\documentclass[review]{elsarticle}

\usepackage{lineno,hyperref}

\journal{arXiv.org}

\graphicspath{{figures/}}
\DeclareGraphicsExtensions{.pdf,.jpeg,.png,.eps}

\usepackage{subfig}

\usepackage{threeparttable}
\usepackage{booktabs}
\usepackage{multirow}
\usepackage{rotating}
\usepackage{adjustbox}

\usepackage[ruled,vlined]{algorithm2e}
\SetAlFnt{\small}

\makeatletter
\newcommand{\algrule}[1][.01pt]{\par\vskip.5\baselineskip\hrule height #1\par\vskip.5\baselineskip}
\makeatother

\makeatletter
\newcommand{\StatexIndent}[1][3]{%
  \setlength\@tempdima{\algorithmicindent}%
  \Statex\hskip\dimexpr#1\@tempdima\relax}
\makeatother

\usepackage{bm} 
\usepackage{amsfonts}
\usepackage{xfrac}

\usepackage{amsmath}
\usepackage{amssymb}
\usepackage{amsthm}

\usepackage{array}

\usepackage{url}

\usepackage{listings}   
\usepackage{calc}  	
\usepackage{multicol}

\usepackage[table]{xcolor}

\usepackage{alphalph}

\usepackage{mathtools}
\DeclarePairedDelimiter{\ceil}{\lceil}{\rceil}
\DeclareMathOperator*{\argmax}{arg\,max}

\DeclareMathOperator*{\median}{median}

\usepackage{booktabs,xcolor,siunitx}
\definecolor{lightgray}{gray}{0.95}

\begin{document}

\begin{frontmatter}

\title{Distributed dual vigilance fuzzy adaptive resonance theory learns online, retrieves arbitrarily-shaped clusters, and mitigates order dependence}

\author[address1,address3]{Leonardo Enzo Brito da Silva\corref{mycorrespondingauthor}}
\cortext[mycorrespondingauthor]{Corresponding author}
\ead{leonardoenzo@ieee.org}

\author[address2]{Islam Elnabarawy}

\author[address1]{Donald C. Wunsch II}

\address[address1]{Applied Computational Intelligence Laboratory, Department of Electrical and Computer Engineering, \\Missouri University of Science and Technology, Rolla, MO 65409 USA.}
\address[address2]{Applied Computational Intelligence Laboratory, Department of Computer Science, \\Missouri University of Science and Technology, Rolla, MO 65409 USA.}
\address[address3]{CAPES Foundation, Ministry of Education of Brazil, Bras\'ilia, DF 70040-020, Brazil.}

\begin{abstract}
This paper presents a novel adaptive resonance theory (ART)-based modular architecture for unsupervised learning, namely the distributed dual vigilance fuzzy ART (DDVFA). DDVFA consists of a global ART system whose nodes are local fuzzy ART modules. It is equipped with the distinctive features of distributed higher-order activation and match functions, using dual vigilance parameters responsible for cluster similarity and data quantization. Together, these allow DDVFA to perform unsupervised modularization, create multi-prototype clustering representations, retrieve arbitrarily-shaped clusters, and control its compactness. Another important contribution is the reduction of order-dependence, an issue that affects any agglomerative clustering method. This paper demonstrates two approaches for mitigating order-dependence: preprocessing using visual assessment of cluster tendency (VAT) or postprocessing using a novel Merge ART module. The former is suitable for batch processing, whereas the latter can be used in online learning. Experimental results in the online learning mode carried out on 30 benchmark data sets show that DDVFA cascaded with Merge ART statistically outperformed the best other ART-based systems when samples were randomly presented. Conversely, they were found to be statistically equivalent in the offline mode when samples were pre-processed using  VAT. Remarkably, performance comparisons to non-ART-based clustering algorithms show that DDVFA (which learns incrementally) was also statistically equivalent to the non-incremental (offline) methods of DBSCAN, single linkage hierarchical agglomerative clustering (HAC), and offline version of k-means, while retaining the appealing properties of ART. Links to the source code and data are provided. Considering the algorithm's simplicity, online learning capability, and performance, it is an ideal choice for many agglomerative clustering applications. 
\end{abstract}


\end{frontmatter}

\section{Introduction} \label{Sec:intro}

There is a rich literature of clustering methods~\cite{xu2005, xu2009, xu2010}, and among the neural network-based ones, adaptive resonance theory (ART)~\cite{Carpenter1987} is of great interest due to its many useful properties~\cite{Wunsch2009}, particularly the fact that it addresses the \textit{stability-plasticity dilemma}. After sufficient exposure to the environment, a competitive learning neural network eventually learns prototypical representations or archetypes that reflect groups of samples~\cite{Bartfai1994}; i.e., it learns a succinct or compressed representation of the data.

Numerous ART-based architectures have been conceived, such as fusion ART~\cite{tan2007}, whose variants have been effectively used for semi-supervised~\cite{Meng2014}, supervised~\cite{Tan1995}, and reinforcement learning applications~\cite{tan2004, tan2006, tan2008}; BARTMAP~\cite{xu2011,bartmap2016} for biclustering applications, such as unsupervised gene expression analysis, as well as architectures with distinct internal category representations such as hyperboxes~\cite{Carpenter1991}; gaussians~\cite{williamson1996, vigdor2007}; hyperspheres~\cite{anagnostopoulos2000}; hyperellipsoids~\cite{anagnostopoulos2001}; and others.

Particularly, ART has been used as the basis for several hierarchical clustering methods, which can be classified into bottom-up (agglomerative or merging methods) and top-down (divisive or splitting methods)~\cite{xu2009}. Hierarchical ART architectures generally follow two main designs~\cite{Massey2009}: (a)~a series/cascade of ART modules where the output of one ART (i.e., a prototype) is the input of the next~\cite{Carpenter1990c, Ishihara1995, Hung1996b, Hung1996, Bartfai1996, Bartfai1997, Bartfai1997b, Chen1999, Chen2001, Yavas2012, Benites2017} or (b)~parallel ART modules sharing the same inputs and using different vigilance values~\cite{Wunsch1991, Wunsch1993, Bartfai1994, Sejun2011, Tscherepanow2010, Tscherepanow2011b, Tscherepanow2012b, Tscherepanow2012a, Svaco2015}. Generally, the hierarchical relationships between ART modules are defined implicitly by the input signal flow, explicitly by enforcing constraints or connections, and/or by the setting of multiple vigilance parameters to define hierarchies. Alternatively, hierarchies within the same ART can be created by designing custom ART activation functions~\cite{Lavoie1997, Lavoie1999b} or by analyzing its distributed activation patterns~\cite{Davenport2004}. ART-based hierarchical approaches have been successfully applied, for instance, in text mining~\cite{Bouchachia2003, Massey2009} and robotics~\cite{Yavas2012, Svaco2015}.

Another branch of clustering includes multi-prototype-based methods. These allow multiple prototypes to represent a single cluster and more accurately capture the data topology, thereby typically handling clusters with arbitrary shapes. Multi-prototype representations have been successfully used for clustering~\cite{Guha1998, tyree1999, tasdemir2009, araujo20131, araujo20132}, visualization~\cite{ultsch1990, tasdemir2009, leonardo2018a}, and validation purposes~\cite{halkidi2008, tasdemir2011}. In the context of ART, examples include the combination of an ART-like system using quadratic neurons~\cite{ChunSu2001} and hierarchical clustering~\cite{ChunSu2002, ChunSu2005} and the related approach~\cite{leonardo2015} using fuzzy ART~\cite{Carpenter1991}. Other methods have augmented ART-based systems by employing dual vigilance parameters~\cite{leonardo2018b}, connecting the first and second resonating categories~\cite{Isawa2007, Isawa2008, Isawa2008b, Isawa2009, Tscherepanow2010, Tscherepanow2011b, Tscherepanow2012b, Tscherepanow2012a}, or replacing fuzzy ART's nodes with growing cell structures~\cite{Fritzke1994} in a hybrid architecture~\cite{Kim2011}. 

Although they are based on multi-prototype representation, many of the previously mentioned approaches do not adopt  distributed activation, match or learning, which improves a network's noise robustness and compactness~\cite{Carpenter1997b, Carpenter1998a}. 
The distributed ART model~\cite{Carpenter1997b} is endowed with all of these distributed features, however it does not possess a mechanism to build, in an unsupervised manner, a permanent and binary many-to-one mapping (i.e., a multi-prototype cluster representation). Thus, it is still limited by its nested hyperbox cluster abstractions. Distributed learning is also featured in the ART variants introduced in~\cite{Kondadadi2002, Yousuf2010}. In the ART literature, the power of distributed activation has been harnessed to perform, for instance, (a)~unsupervised feature extraction~\cite{lam2015}; (b)~hierarchical clustering~\cite{Carpenter1990c, Chen2001} -- although featuring distributed representation, the latter approaches are cascade architectures not designed to model arbitrarily-shaped clusters since they are limited by their category representations at each hierarchical level; and (c)~supervised learning systems such as the distributed ARTMAP~\cite{Carpenter1998a}, which is a generalization of a variety of ART models~\cite{Carpenter2003} such as~\cite{Carpenter1991, carpenter1992, Carpenter2003, Amis2007, Carpenter1998b} and uses distributed ART as its building block, some topoART variants~\cite{Tscherepanow2011a, Tscherepanow2012c}, default ARTMAPs~\cite{Carpenter2003, Amis2007}, and adaptive resonance associative map~\cite{Tan1995} variants~\cite{Sapozhnikova2009, Benites2017}. 

The distributed dual vigilance fuzzy ART (DDVFA) introduced here belongs to the class of modular neural networks~\cite{Auda1996, Auda1998, Auda1999}. Specifically, it is designed for the unsupervised learning task of clustering. This class of network architectures employs a divide-and-conquer approach and shares the following main features~\cite{Auda1996, Auda1998, Auda1999}: task decomposition (breaking down a complex problem) and multi-module decision making (combining local decisions in a single global consensus). Commonly, unsupervised learning methods are used as a pre-processing stage to partition the data to be handled by simple, fast, and efficient supervised modules. ART-based systems have been used for such purposes in supervised modular networks~\citep{Auda1996, Auda1998, Auda1999}. 
A current challenge for incremental learners, such as ART-based systems, is the order of sample presentation. Thus, suitable pre- and post-processing strategies are usually employed when applicable (see references in~\cite{leonardo2018}). Specifically, post-processing merging strategies are commonly used in conjunction with incremental learners (e.g.,~\cite{Lei2006, lughofer2008, Isawa2008, Isawa2008b, Isawa2009, Zhang2006, Swope2012, Benites2017}); here, a novel ART-based network provides such functionality. Additionally, visualization and assessment are valuable assets when performing cluster analysis~\cite{xu2009, Bezdek2017, leonardo2018a}; here, the visual assessment of cluster tendency (VAT)~\cite{bezdek2002,Bezdek2017} technique is used for its sample ordering properties to emulate scenarios in which such data pre-processing is practical, as per~\cite{leonardo2018}.

This paper presents the following main contributions: 
\begin{enumerate}
    \item A novel modular fuzzy ART-based architecture (DDVFA). Unsupervised dynamic modularization (creation of new local modules as needed) and multi-prototype representation are accomplished by employing dual vigilance parameters associated with global and local fuzzy ART modules.
    \item Novel higher order distributed activation and normalized match functions based on hierarchical agglomerative clustering (HAC) methods embedded in the incremental learning process. Suitably setting the HAC-based activation/match functions allows DDVFA to retrieve arbitrarily shaped clusters, and higher order match functions have the potential to generate more compact DDVFA networks~(as per~\cite{Carpenter1997b, Carpenter1998a}) and extend the regions of successful dual vigilance parameter combinations.
    \item A novel Merge ART module compatible with DDVFA for post-processing in online learning applications. This procedure compensates for the errors caused by the random order of input presentations thus enabling improved performance.
    \item An analysis of the behavior of the DDVFA with and without pre-processing (VAT) and post-processing (Merge ART) strategies, as well as with respect to its kernel width parameter. 
\end{enumerate}

The results show that together, these features enable DDVFA to yield an improved performance compared to other current state-of-the-art fuzzy ART-based technologies.

The remainder of this paper is divided as follows: Section~\ref{Sec:theory} provides a brief review of ART, fuzzy ART, fuzzy topoART, and dual vigilance fuzzy ART; Section~\ref{Sec:DDVFA} introduces distributed dual vigilance fuzzy ART; Section~\ref{Sec:experiments} describes the experimental set-up; Section~\ref{Sec:Results} reports and discusses the results; and Section~\ref{Sec:conclusion} is the conclusion.

\section{Adaptive Resonance Theory} \label{Sec:theory}

Adaptive resonance theory (ART)~\cite{Grossberg1976a} is the theory that learning is often mediated by resonant feedback in neural circuits. It inspired the development of many neural network architectures, each with its own internal categorical representation, while sharing the same design principles~(Fig.~\ref{Fig:gen_ART}). The ART \textit{matching rule}~\cite{Carpenter1987} is a key property of these ART systems~\cite{Carpenter2003,Amis2007}; it regulates the interaction between top-down expectations (represented by the internal categories or templates) and the bottom-up inputs. This process is guided by an orienting subsystem, which performs a hypothesis test, called the \textit{vigilance check}, that either shuts down or enables an ART category to learn. ART templates have specific properties and governing equations based on their internal representation. They allow for a discretization of the data space, thus summarizing it as clusters. The vigilance parameter (see Eq.~(3)) controls category size and thus the granularity of this discretization.

\subsection{Fuzzy ART} \label{Sec:FA}

Fuzzy ART~\cite{Carpenter1991} is an ART architecture designed to work with real-valued data. Concisely, when a sample $\bm{x} \in \mathbb{R}^d$ is presented at the feature representation field $F_1$, it activates the category~$j$ at the category representation field $F_2$ whose weight vector~$\bm{w}_j$ maximizes the following \textit{activation function}:

\begin{figure}[!t]
\centerline{
\includegraphics[width=\columnwidth]{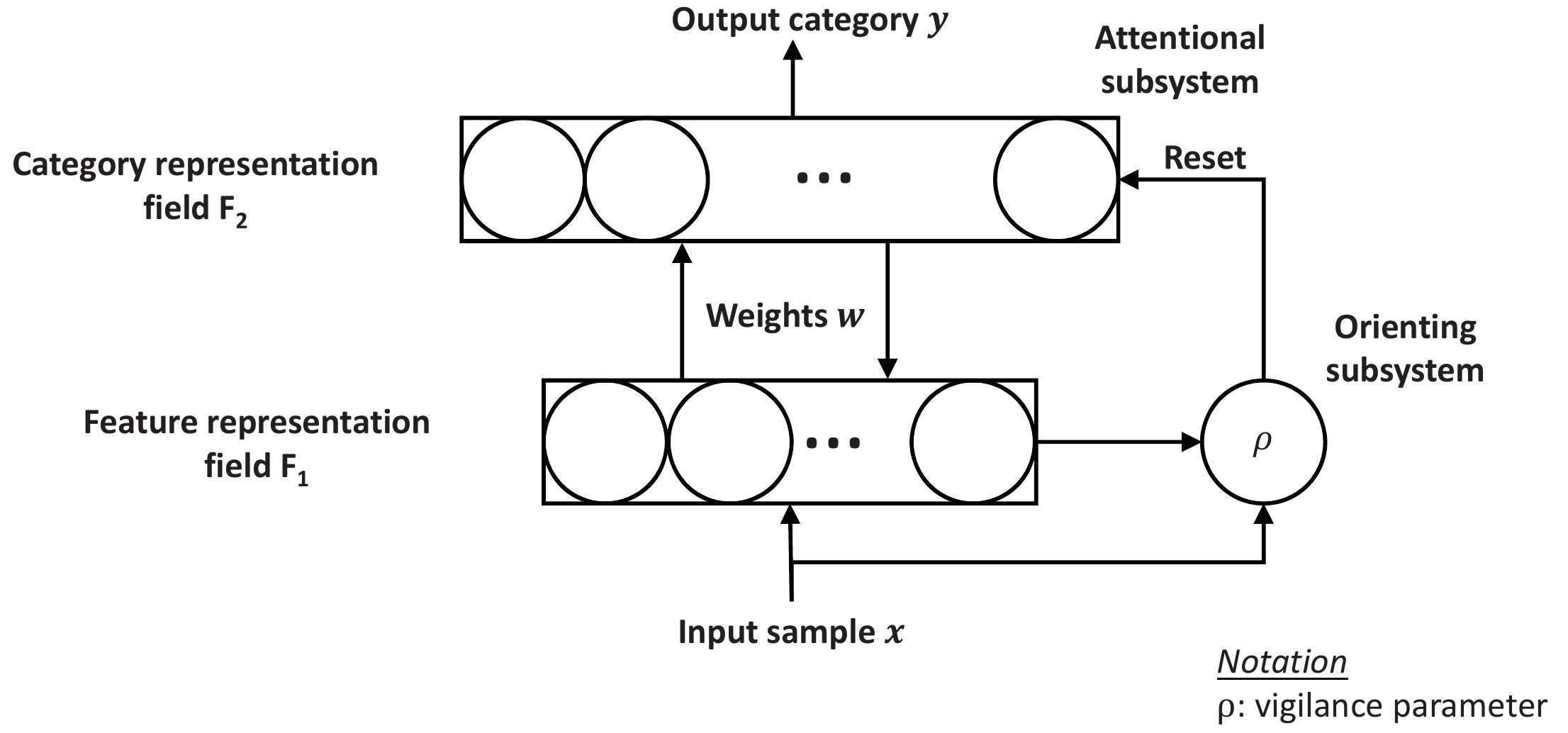}}
\caption{Generic ART architecture, underlying various ART designs.}
\label{Fig:gen_ART}
\end{figure}

\begin{equation}
T_j = \frac{|\bm{x} \wedge \bm{w}_j|}{\alpha +|\bm{w}_j|},
\end{equation}

\noindent where $|\cdot|$ is the $L_1$ norm and $\alpha>0$ is the choice parameter, which is usually set to a small value. A comprehensive study on its behavior can be found in~\cite{Georgiopoulos1996}. 

Next, a \textit{match function} evaluates the best matching category as:

\begin{equation}
M_j = \frac{|\bm{x} \wedge \bm{w}_j|}{|\bm{x}|},
\end{equation}

and a \textit{vigilance check} $\nu$ is performed using the computed match value:

\begin{equation}
\nu: M_j \geq \rho,
\label{Eq:nu}
\end{equation}

\noindent where $0 \leq \rho \leq 1$ is the vigilance parameter. If $\nu$ is satisfied, then the winning category's weight vector is updated as:

\begin{equation}
\bm{w}_j^{new} = (1-\beta)\bm{w}_j^{old} + \beta(\bm{x} \wedge \bm{w}_j^{old}),
\label{Eq:update}
\end{equation}

\noindent where $ 0 < \beta \leq 1$ is the learning rate parameter. Otherwise, this category is deactivated, and the search continues by activating the next highest ranked category. If none of them satisfies this constraint, then a new category is created to encode sample $\bm{x}$. Thus, the problem of selecting the number of clusters is traded for the one of selecting the vigilance value $\rho$. 

Fuzzy ART features many appealing properties such as scalability, speed, stability, plasticity, online (one pass) and offline incremental learning modes, as well as simple implementation, transparency, and novelty detection (rare/unusual events)~\cite{Carpenter2003, Amis2007, Wunsch2009, xu2009, mulder2003}.

\subsection{Fuzzy topoART} \label{Subsec:TopoART}

Fuzzy topoART~\cite{Tscherepanow2010} incorporates topology-based learning~\cite{Furao2006} into ART. Briefly, it consists of multiple independent fuzzy ART modules where the preceding modules filter the shared inputs to subsequent ones. Standard topoART consists of two identical modules: A and B. During training, which is processed in parallel for all modules, an ``instance counting'' feature accounts for the number of samples $n$ learned by a given category. Every $\tau$ learning cycles/iterations (number of sample presentations), a noise thresholding procedure is performed to remove categories with less than $\phi$ samples. Once the threshold is surpassed, ``candidate'' categories become ``permanent'' categories. A sample is propagated to module~B if it has resonated with a ``permanent'' category of module~A.

The granularity of the solutions is defined by the modules' different vigilance parameter values. Module~B's vigilance parameter is~\cite{Tscherepanow2010,Tscherepanow2011b,Tscherepanow2012b}:

\begin{equation}
\rho_b = \frac{1}{2}\left( \rho_a + 1 \right),
\end{equation}

\noindent where $\rho_a$ is module~A's vigilance parameter. Since $\rho_b \geq \rho_a$, modules A and B yield increasingly finer partitions of a given data set. Categories are laterally connected by edges between the first and second resonating categories (i.e., the two highest ranked categories that simultaneously satisfy the vigilance test~(Eq.~(\ref{Eq:nu}))) to mirror the input distribution. This multi-prototype method enables topoART modules to learn topologies and capture clusters with arbitrary geometries. Besides competitive learning, it also uses cooperative learning by allowing the second winner ($sbm$) to learn with a smaller learning rate than the first ($bm$): $\beta_{sbm}<\beta_{bm}=1$. Finally, to compensate for fuzzy ART's bias toward small categories, topoART uses a particular activation function for prediction, which is independent of category size~\cite{Tscherepanow2010,Tscherepanow2011b,Tscherepanow2012b}:

\begin{equation}
T_j = 1 - \frac{|\left( \bm{x} \wedge \bm{w}_j \right) - \bm{w}_j|}{|\bm{x}|}.
\end{equation}

TopoART has spawned several variants for unsupervised~\cite{Tscherepanow2011b, Tscherepanow2012b, Tscherepanow2012a}, supervised~\cite{Tscherepanow2012c, Tscherepanow2011a}, and semi-supervised~\cite{nooralishahi2018} learning paradigms.

\subsection{Dual vigilance fuzzy ART} \label{Subsec:DVFA}

Dual vigilance fuzzy ART (DVFA)~\cite{leonardo2018b} consists of a single ART module equipped with two layered vigilance parameters. The larger vigilance value is referred to as the ``upper bound'' ($\rho_{UB}$) and is responsible for the data compression/quantization, whereas the lower vigilance value is referred to as the ``lower bound'' ($\rho_{LB}$) and is responsible for the cluster similarity. Briefly, when a category is activated after a winner-takes-all competition, then a vigilance check with a large value is performed (using $\rho_{UB}$ in Eq.~(\ref{Eq:nu})); if it is satisfied, then it behaves identically to fuzzy ART. However, if this test fails, then a second test is performed with a slightly smaller vigilance value (using $\rho_{LB}$ in Eq.~(\ref{Eq:nu})). If the same category satisfies this looser constraint, then a new category is created and assigned to the same cluster as the tested category in an output mapping matrix like fuzzy ARTMAP's~\cite{carpenter1992}. Therefore, a many-to-one mapping of categories to clusters is created (this is a multi-prototype approach). In this manner, the data distribution can be more faithfully mirrored, and clusters of arbitrary geometries may be retrieved. 

\section{Distributed dual vigilance fuzzy ART} \label{Sec:DDVFA}

The distributed dual vigilance fuzzy ART (DDVFA) neural network architecture described in Section~\ref{Subsec:DDVFA} can be viewed as an ``\textit{ART of ARTs}'', in which each node in the category representation field $F_2$ of a global ART is itself a local ART, where the latter represents a given data cluster. Equivalently, it can be seen as an unsupervised modular neural network consisting of local ARTs whose multi-module decision making system is a global ART. Since ART-based systems are sensitive to the order of input presentation, Section~\ref{Subsec:MergeART} presents an approach to compensate for this dependency: the output of a DDVFA module (layer~1) is cascaded into a compatible Merge ART module (layer~2). 

\begin{table}[!b]
\centering
\caption{Notation for DDVFA}
\resizebox{\columnwidth}{!}{
\begin{threeparttable}
\begingroup\setlength{\fboxsep}{0pt}
\colorbox{lightgray}{
\begin{tabular}{p{0.31\columnwidth}p{0.6\columnwidth}}
\toprule
Notation 
& Description 
\\
\midrule
\midrule
$\bm{X}$ 
& 
a data set $\bm{X}=\{\bm{x}_l\}_{l=1}^N \in \mathbb{R}^d$. \\
$ART^{(i)}_j$ 
& 
global ART's $F_2$ node $j$ (layer $i$). \\
$T^{ART^{(i)}_j}$, $M^{ART^{(i)}_j}$
& 
activation and match functions of local $ART^{(i)}_j$, respectively. \\
$\bm{w}^{ART^{(i)}_j}_k$ 
& 
$k^{th}$ category weight vector of local $ART^{(i)}_j$. \\
$T^{ART^{(i)}_j}_k$, $M^{ART^{(i)}_j}_k$
& 
activation and match functions of $\bm{w}^{ART^{(i)}_j}_k$, respectively. \\
$\gamma \ge 1$
& 
kernel width. \\
$0 \leq \gamma^{*} \leq \gamma $
&
reference kernel width. \\
$\rho^{(i)}_{UB} \ge \rho^{(i)}_{LB}$  
&
lower and upper bound vigilance parameters (layer $i$). \\
$\bm{T}_{p,q}$, $\bm{M}_{p,q}$
& 
activation and match matrices between local $ART^{(1)}_p$ and local $ART^{(2)}_q$. \\
$n_k^{ART^{(i)}_j}$ 
&
number of samples encoded by category $k$ of local $ART^{(i)}_j$ (instance counting). \\
${n^{ART^{(i)}_j}}$ 
&
total number of samples encoded by local $ART^{(i)}_j$ (instance counting). \\
\bottomrule
\end{tabular}
}\endgroup
\end{threeparttable}
}
\label{tab:notations}
\end{table}

\subsection{DDVFA architecture} \label{Subsec:DDVFA} 

\begin{figure}[!ht]
\centerline{\includegraphics[width=\textwidth]{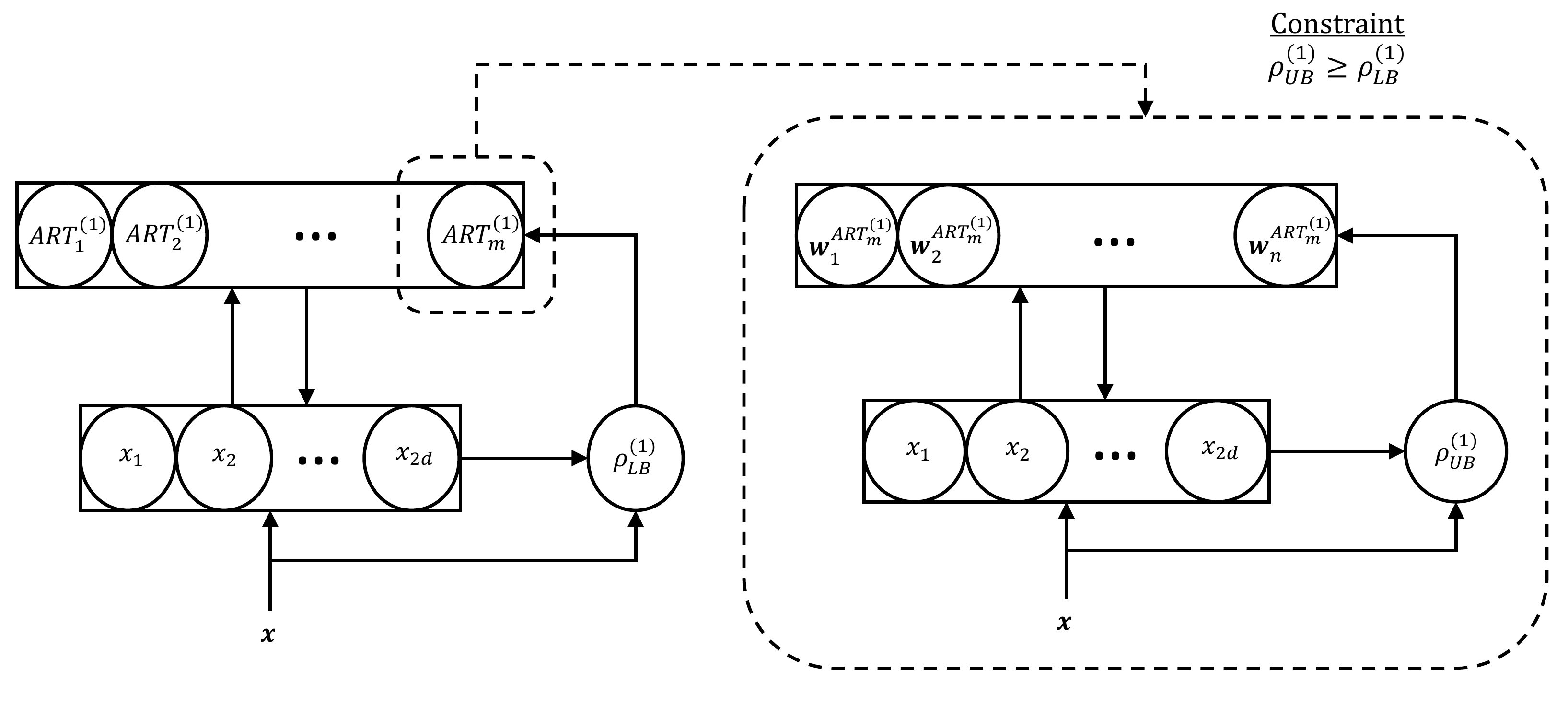}}
\caption{DDVFA architecture. Each global ART's $F_2$ node is a local fuzzy ART (as portrayed in Fig.~\ref{Fig:gen_ART}) with shared input $\bm{x}$ and vigilance $\rho=\rho_{UB}^{(1)} \geq \rho_{LB}^{(1)}$.}
\label{Fig:DDpART}
\end{figure}

Table~\ref{tab:notations} lists the notation used in this section, and Fig.~\ref{Fig:DDpART} depicts a generic DDVFA. It is a modular structure in which a global ART controls local parallel ARTs via a vigilance feedback between these modules -- cf. ART tree~\cite{Wunsch1991, Wunsch1993}, in which $F_2$ nodes are also ART modules, but these are not controlled by a global ART module. The global ART acts as a mapping mechanism analogous to the inter-ART module in fuzzy ARTMAP architectures~\cite{carpenter1992, Asfour1993}, thus maintaining hierarchical consistency. This relates to self-consistent modular ART~\cite{Bartfai1994}; however, DDVFA uses a bottom-up agglomerative approach, whereas the former uses a top-down divisive approach limited to hyperrectangular cluster representations. Concretely, DDVFA is a multi-prototype hierarchical agglomerative clustering (HAC) method that builds a self-consistent two-level hierarchy of categories.

Similar to DVFA, the vigilance parameters of the global and local ARTs are denoted as $\rho_{LB}$ and $\rho_{UB}$, respectively, where the constraint $\rho_{LB} \leq \rho_{UB}$ is enforced. Setting $\rho_{UB} = \rho_{LB}$ reduces the DDVFA to a generic fuzzy ART framework, which ensures that each global ART's $F_2$ node (i.e., each local ART) encodes one category. Alternately, setting $\rho_{UB}$ strictly greater than $\rho_{LB}$ builds a multiple category representation for each cluster, thus enabling an approximation of that cluster's geometry over the data space according to the underlying assumption of the activation and match functions, which are to be set a priori. The vigilance parameters $\rho_{LB}$ and $\rho_{UB}$ reflect the minimum similarity of a cluster and the granularity level of the data quantization (i.e., the categories' sizes), respectively. In other words, the rationale is to restrict the maximum internal category size of each local ART while maintaining a smaller similarity constraint for the cluster represented by each global ART $F_2$ node. Thus, local ART modules (or clusters) can be added as needed.

The inner workings of DDVFA are the same as a generic ART architecture, as reviewed in Section~\ref{Sec:theory}. However, the activation $T^{ART_i}(\cdot)$ and match $M^{ART_i}(\cdot)$ functions of the global ART's $F_2$ node $i$ are a distributed version of the local $ART_i$ categories' activation $T^{ART_i}_j$ and match $M^{ART_i}_j$ functions based on HAC, where $j=\{1,...,k\}$ represents the categories. Specifically, the activation and match functions of global ART's $F_2$ node~$i$ in layer~(1) are given by a function of local $ART^{(1)}_i$'s $k$ nodes: 

\begin{equation}
T^{ART^{(1)}_i} = f\left( T^{ART^{(1)}_i}_1, T^{ART^{(1)}_i}_2,~...~, T^{ART^{(1)}_i}_k\right),
\label{Eq:T1}
\end{equation}

\noindent where

\begin{equation}
T^{ART^{(1)}_i}_j  = \left( \frac{|\bm{x} \wedge \bm{w}^{ART^{(1)}_i}_j|}{\alpha +|\bm{w}^{ART^{(1)}_i}_j|} \right)^\gamma,~j \in \{1,~...~,k\},
\label{Eq:T2}
\end{equation}

\noindent and 

\begin{equation}
M^{ART^{(1)}_i} = g\left( M^{ART^{(1)}_i}_1, M^{ART^{(1)}_i}_2,~...~, M^{ART^{(1)}_i}_k\right),
\label{Eq:M1}
\end{equation}

\noindent where

\begin{equation}
M^{ART^{(1)}_i}_j = \left( \frac{|\bm{x} \wedge \bm{w}^{ART^{(1)}_i}_j|}{|\bm{x}|}\right)^\gamma,~j \in \{1,~...~,k\}. 
\label{Eq:M2}
\end{equation}

In this study, for simplicity, $f\left( \cdot \right) = g\left( \cdot \right)$ in~(\ref{Eq:T1}) and~(\ref{Eq:M1}), i.e., the same functional relationship is used for the activation and match functions. These are listed in Table~\ref{tab:DDpART_T_M} and are based on HAC methods~\cite{xu2009}. A power parameter $\gamma \geq 1$ is employed here in both the activation and match functions. Like the power parameter used in~\cite{Carpenter1997b, Carpenter1998a}, $\gamma$~assumes the role of a kernel width, facilitates the dual vigilance parameters selection, and reduces category proliferation (Section~\ref{Sec:Gamma}). Setting $\gamma=1$ corresponds to a standard fuzzy ART module, in which a moderately far sample would still have a reasonably large value for the match function. 

\begin{figure}[!ht]
\centerline{
\subfloat[$\gamma=1$]{\includegraphics[width=.5\columnwidth]{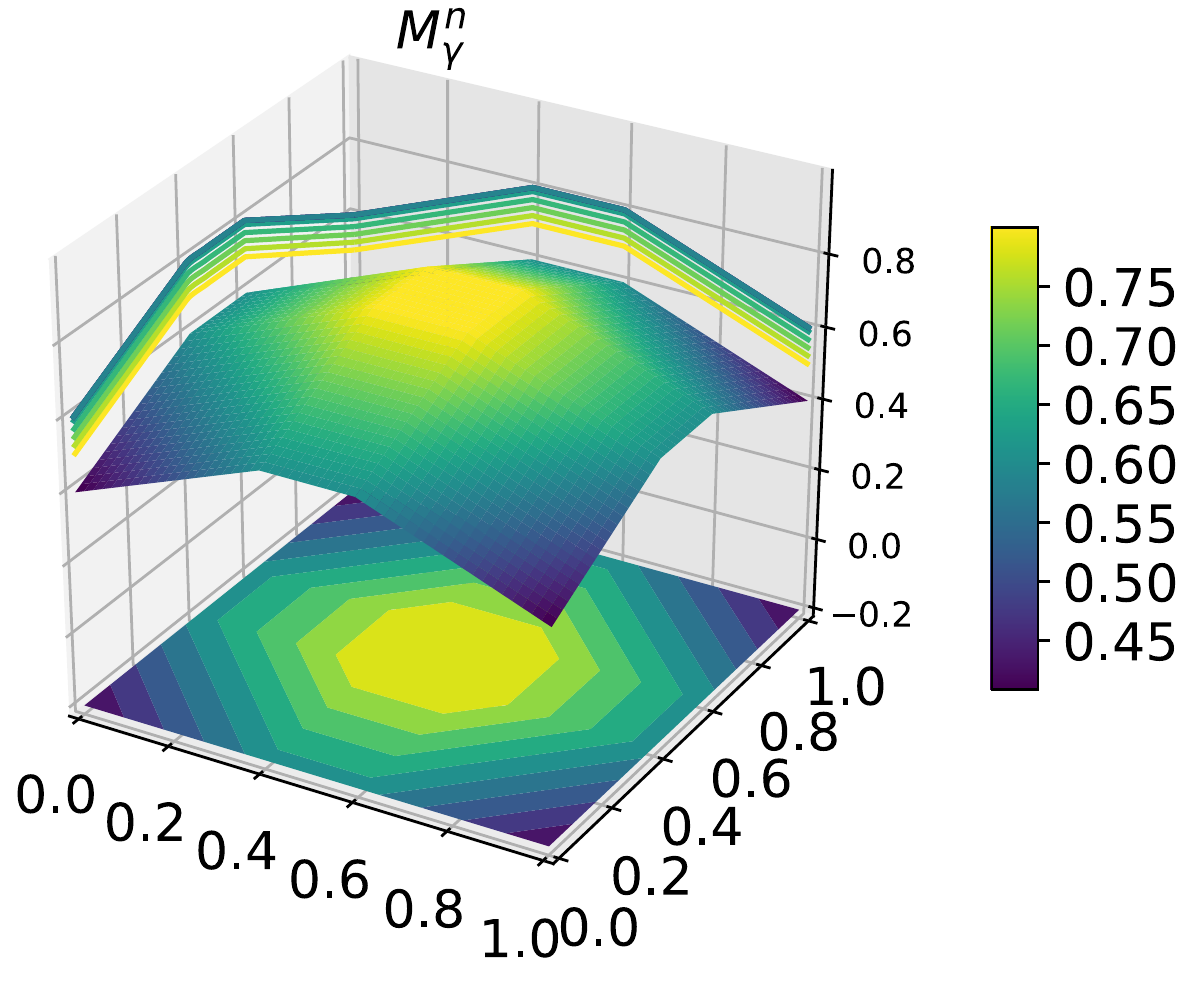}}
\hfil
\subfloat[$\gamma=10$]{\includegraphics[width=.5\columnwidth]{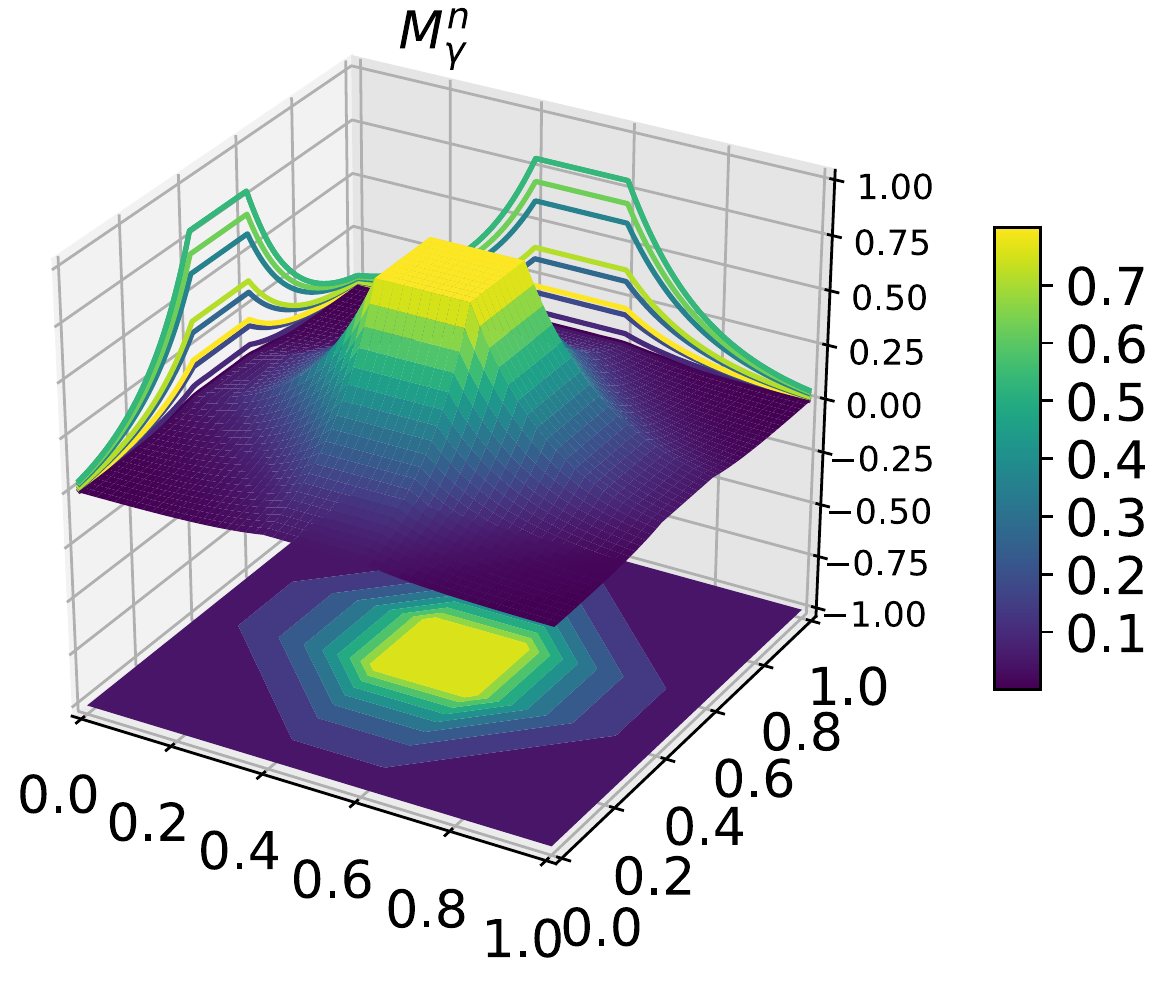}
\label{Fig:perf_R_seeds}}
}
\caption{3D surfaces, contours, and cross-section cuts representing the normalized match functions ($M_{\gamma}^n$) using $\gamma^{*}=1$ and different values of $\gamma$.}
\label{Fig:gammas}
\end{figure}

This extension of successful dual vigilance parameters occurs because the match and activation functions (when $\gamma=1$) decay linearly and slowly for samples outside a category's hyperrectangular boundaries and thus, by increasing~$\gamma$, steeper decays are created (Fig.~\ref{Fig:gammas}). A similar behavior is exhibited by fuzzy min-max neural networks~\cite{Simpson1992, Simpson1993, Gabrys2000} and the variant~\cite{Gabrys2000} addresses it by devising a custom fuzzy membership function, whose sensitivity parameter performs the same role of controlling the membership value decays. Furthermore, the higher order membership class of functions has been shown to enhance fuzzy ART performance~\cite{elnabarawy2017}.  

The property exploited here is the fact that the activation and match functions become more ``selective'' (as expected from a power rule as a contrast-enhancement procedure~\cite{Carpenter1997b,Carpenter1998a}); e.g., in Fig.~\ref{Fig:gammas} their trapezoidal form approaches a rectangular membership function. Therefore, regarding the match function, increasing $\gamma$ makes far samples less similar and a category's vigilance region~\cite{Meng2015} smaller (Fig.~\ref{Fig:gammas}). Naturally, when applying a power rule to a scalar in the range $[0,1]$, such as the case of the match and activation functions, its value decreases with~$\gamma$. Therefore, to account for the scaling effect, instead of using~(\ref{Eq:M2}), the match function is normalized in practice as:

\begin{equation}
M^{ART^{(1)}_i}_j = \left( \frac{|\bm{w}^{ART^{(1)}_i}_j|}{|\bm{x}|}\right)^{\gamma^{*}}T^{ART^{(1)}_i}_j, j \in \{1,...,k\} 
\label{Eq:M3}
\end{equation}

\begin{table}[!t]
\centering
\caption{DDVFA's activation and match functions.}
\begin{threeparttable}
\begingroup\setlength{\fboxsep}{0pt}
\colorbox{lightgray}{
\begin{tabular*}{\columnwidth}{@{\extracolsep{\fill}}lll@{}}
\toprule
Method 
& $T^{ART^{(1)}_i}=f(\cdot)$ 
& $M^{ART^{(1)}_i}=g(\cdot)$ \\
\midrule
\midrule
single 	  
& $\max\limits_{j}\left( T^{ART^{(1)}_i}_j \right)$	
& $\max\limits_{j}\left( M^{ART^{(1)}_i}_j \right)$ \\
complete  
& $\min\limits_{j}\left( T^{ART^{(1)}_i}_j \right)$	
& $\min\limits_{j}\left( M^{ART^{(1)}_i}_j \right)$ \\
median    
& $\median\limits_{j}\left( T^{ART^{(1)}_i}_j \right)$	
& $\median\limits_{j}\left( M^{ART^{(1)}_i}_j \right)$ \\
average\tnote{a} 
& $\frac{1}{k_i}\sum\limits_{j=1}^{k_i} T^{ART^{(1)}_i}_j$	
& $\frac{1}{k_i}\sum\limits_{j=1}^{k_i} M^{ART^{(1)}_i}_j$ \\
weighted\tnote{b} 
& $\sum\limits\limits_{j=1}^{k_i} p_jT^{ART^{(1)}_i}_j$	
& $\sum\limits\limits_{j=1}^{k_i} p_jM^{ART^{(1)}_i}_j$ \\
centroid\tnote{c}  
& $\left(\frac{|\bm{x} \wedge \bm{w}_c|}{\alpha +|\bm{w}_c|}\right)^\gamma$    
& $\left( \frac{|\bm{x} \wedge \bm{w}_c|}{|\bm{x}|} \right)^\gamma$  \\
\bottomrule
\end{tabular*}
}\endgroup
\begin{tablenotes}[normal,flushleft]
\item[a,b]$k_i$ represents the number of categories in $ART^{(1)}_i$.
\item[b]$p_j=\frac{n_j^{ART^{(1)}_i}}{n^{ART^{(1)}_i}}$ and $n^{ART^{(1)}_i}=\sum\limits_j{n_j^{ART^{(1)}_i}}$. This represents an a priori probability of $ART^{(1)}_i$'s category $j$  analogous to~\cite{williamson1996,vigdor2007}.
\item[c]$\bm{w}_c$ is the centroid representing all categories of $ART^{(1)}_i$, where its $l$ component is given by $w_{c,l}=\min\limits_{j}\left( w_{j,l}\right)$ for \mbox{$l=\{1,...,2d\}$}.
\end{tablenotes}
\end{threeparttable}
\label{tab:DDpART_T_M}
\end{table}

\noindent where $0 \leq \gamma^{*} \leq \gamma $ is the reference kernel width with respect to which the match function is normalized (see~\ref{AppendixA}). In this paper's experiments, such normalization was performed with respect to the match function values of a standard fuzzy ART (i.e., $\gamma^{*}=1$). Note that the higher order HAC-based activation functions in Eq.~(\ref{Eq:T2}) do not change the search order for global ART when varying $\gamma$ for single, complete, and centroid methods; but it may for weighted and average. Additionally, it also does not affect the search order within the local fuzzy ART module using the higher order activation and match functions.

\textbf{Remark 1.} A power law was introduced in distributed ART/ARTMAP~\cite{Carpenter1997b} for the increased gradient content-addressable memory rule as a contrast enhancement procedure, and it has been used in other ART variants such as distributed ARTMAP~\cite{Carpenter1998a} and default ARTMAPs~\cite{Carpenter2003, Amis2007}. As opposed to the latter ART systems, where the activation functions are normalized to $1$ with respect to a subset of highly active nodes, DDVFA's activation functions are not normalized, but rather its match functions. Specifically, the latter are normalized using a reference parameter $\gamma^{*}$ and with respect to an individual category; additionally, DDVFA's match-reset-search mechanism itself is distinct and uses winner-takes-all learning, as opposed to distributed ART's distributed learning. 

\textbf{Remark 2.} There are subtle, yet fundamental, differences between DVFA and DDVFA besides the architecture itself and the distributed HAC-based higher order nature of the activation and normalized match functions. The first one relates to the search mechanism. In DVFA, it is theoretically possible for categories mapped to the same cluster to be brought up during the search process. Conversely, in DDVFA, if a global ART node does not satisfy the vigilance test, then its local ART and the cluster it represents (which includes all its categories) is shut down and will not appear again during global ART's search. Another difference is that, according to Eq.~(\ref{Eq:M1}) and Table~\ref{tab:DDpART_T_M}, the match functions are distributed, and, in the case of single and complete variants, the category selected by winner-takes-all competition and the category subjected to the vigilance test are not required to be the same.

Naturally, DDVFA integrates a winner-take-all mechanism to select among global ART's $F_2$ nodes (i.e., local FAs) with a variety of distributed HAC-based activation/match functions, which are computed using local fuzzy ART's weight vectors. According to their definitions (Table~\ref{tab:DDpART_T_M}), they range from winner-take-all (single) and loser-take-all (complete) to completely distributed (average, centroid, and weighted). DDVFA can be viewed as an ART-based online incremental approximate (prototype-based) HAC method. If $\rho_{UB}^{(1)}=1$, then the approach reduces to an ART-based HAC, since each local fuzzy ART's category encodes a single sample, and the dendrogram cut-level is defined by the global ART module's vigilance parameter $\rho_{LB}^{(1)}$. Algorithm~\ref{Alg:DDVFA} summarizes the DDVFA's pseudocode.

\begin{algorithm}[!t]
\DontPrintSemicolon
\SetKwInOut{Input}{Input}\SetKwInOut{Output}{Output}
\BlankLine
\Input{$\bm{x}$, $\alpha$, $\beta$, $\rho_{UB}^{(1)}$, $\rho_{LB}^{(1)}$, $\gamma$, $\gamma^{*}$, method.} 
\Output{DDVFA clusters.}
\algrule
\nl Present input sample $\bm{x} \in \bm{X}$. \;
\nl Compute $T^{ART^{(1)}_i}_j,~\forall~i,j$ (Eq.~(\ref{Eq:T2})). \;
\nl Compute $T^{ART^{(1)}_i},~\forall~i$ (Eq.~(\ref{Eq:T1}), Table~\ref{tab:DDpART_T_M}'s method). \;
\nl \label{DDVFA_search_global}Find the winning node $I \leftarrow \argmax\limits_{i}\{T^{ART^{(1)}_i}\}$. \;
\nl Compute $M^{ART^{(1)}_I}_j,~\forall~j$ (Eq.~(\ref{Eq:M3})). \;
\nl Compute $M^{ART^{(1)}_I}$ (Eq.~(\ref{Eq:M1}), Table~\ref{tab:DDpART_T_M}'s method). \;
\nl Evaluate vigilance test $\nu_1 : M^{ART^{(1)}_I} \geq \rho_{LB}^{(1)}$. \;
\nl \uIf{$\nu_1$ is satisfied (resonance)}{
\nl \label{DDVFA_search_local}Find winning category $J \leftarrow \argmax\limits_{j}\{T^{ART^{(1)}_I}_j\}$. \;
\nl Evaluate vigilance test $\nu_2 : M^{{ART^{(1)}_I}}_J \geq \rho_{UB}^{(1)}$. \;
\nl \uIf{$\nu_2$ is satisfied (resonance)}{
\nl Update category $J$ weight vector $\bm{w}^{ART^{(1)}_I}_{J}$ (Eq.~\ref{Eq:update}).\;
}
\nl \Else{
\nl Reset category $J$. If there are still active categories in local ART $I$ then go to step~\ref{DDVFA_search_local}; otherwise create a new category using fast commit ($\bm{w}^{ART^{(1)}_I}_{new} \leftarrow \bm{x}$).\;
}
  }
\nl \Else{
\nl Reset $F_2$ node $I$. If there are still active nodes in global ART then go to step~\ref{DDVFA_search_global}; otherwise create a new ART node and apply fast commit ($\bm{w}^{ART^{(1)}_{new}}_{new} \leftarrow \bm{x}$).\;
  }
\caption{DDVFA}\label{Alg:DDVFA}  
\end{algorithm}

\subsection{Merge ART module}  \label{Subsec:MergeART}

\begin{figure}[!t]
\centerline{\includegraphics[width=0.7\columnwidth]{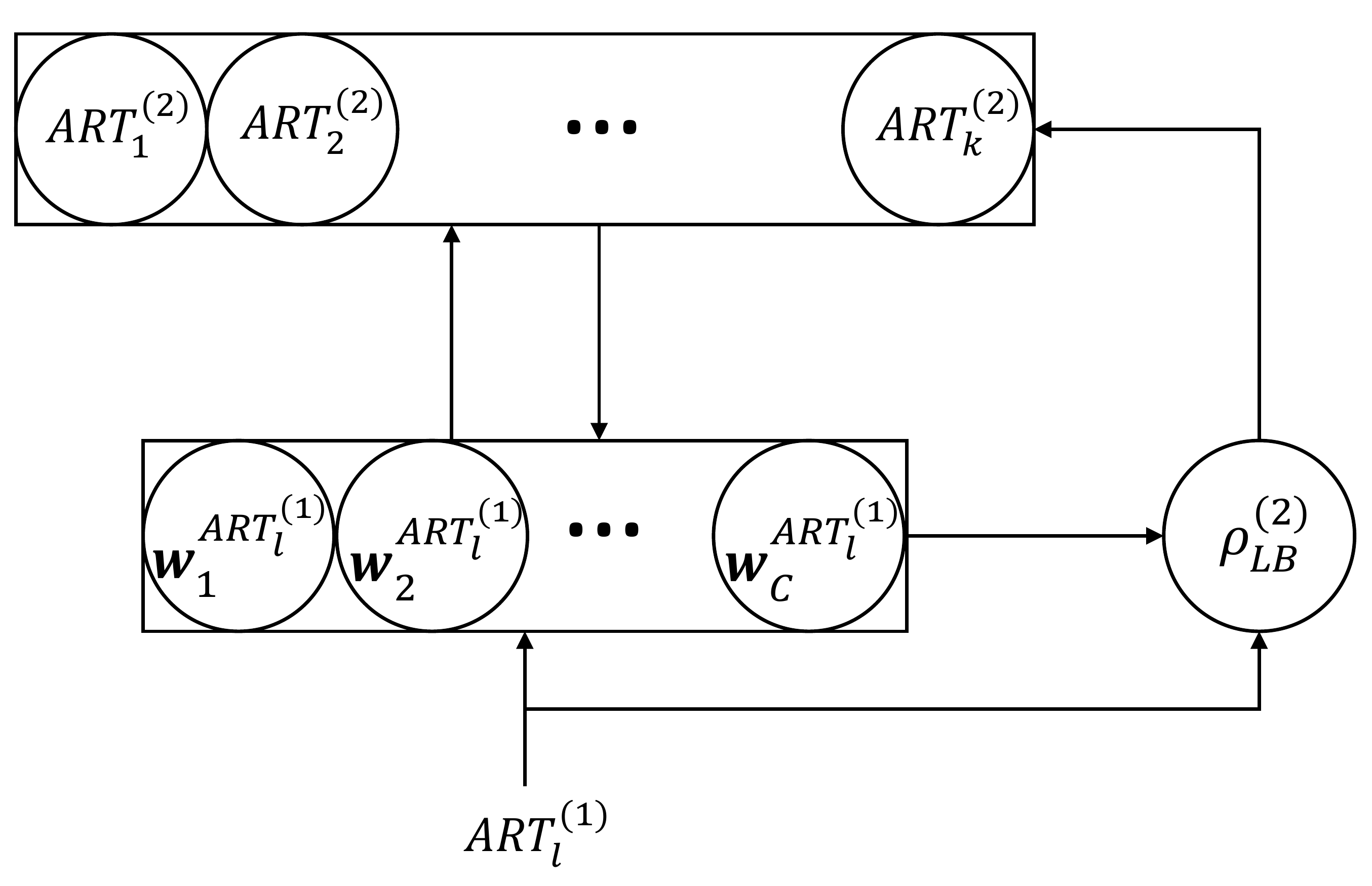}}
\caption{Merge ART module. Each $ART^{(2)}$ is a fuzzy ART with $\rho=\rho^{(2)}_{UB}$.}
\label{Fig:MergeART}
\end{figure}

The order of input presentation is a challenge for incremental learners as it plays a significant role in such systems' performance (see references in~\cite{leonardo2018}). For this reason, a Merge ART module (Fig.~\ref{Fig:MergeART}) is introduced here to be placed at layer 2, i.e., on top of the DDVFA in a cascade design. It acts as another ART module with dual vigilance parameters in which the inputs are ART nodes from DDVFA. It has its own set of parameters that are independent of DDVFA. However, for simplicity, DDVFA's activation and match functions functional forms were kept to maintain the same underlying cluster assumptions, and $(\rho^{(2)}_{LB}, \rho^{(2)}_{UB})$ were set to $(\rho^{(1)}_{LB} , \rho^{(1)}_{UB})$. 

The merging process consists of unions or concatenation of local fuzzy ARTs followed by compressions within each set of local fuzzy ARTs. Let $\bm{T}_{k,l}=[t_{ij}]_{R \times C}$ and $\bm{M}_{k,l}=[m_{ij}]_{R \times C}$ be the activation and match matrices of Merge ART's $F_2$ node $ART^{(2)}_k$ when the input $ART^{(1)}_l$ (from DDVFA) is presented, where $R$ and $C$ are the number of categories of Merge ART's $ART^{(2)}_k$ and DDVFA's $ART^{(1)}_l$, respectively. The entries of matrices $\bm{T}_{k,l}$ and $\bm{M}_{k,l}$ are computed as:

\begin{equation}
t_{i,j} = \left( \frac{|\bm{w}_j^{ART^{(1)}_l} \wedge \bm{w}_i^{ART^{(2)}_k}|}{\alpha +|\bm{w}_i^{ART^{(2)}_k}|} \right)^\gamma,
\label{Eq:tij}
\end{equation}

\begin{equation}
m_{i,j} = \left( \frac{|\bm{w}_i^{ART^{(2)}_k}|}{|\bm{w}_j^{ART^{(1)}_l}|} \right)^{\gamma^{*}}t_{i,j}.
\label{Eq:wij}
\end{equation}

The activation and match functions of the Merge ART module are listed in Table~\ref{tab:MergeART_T_M}. When resonance is triggered, i.e., when the condition \mbox{$M^{ART^{(2)}_K} \geq \rho_{LB}^{(2)}$} is satisfied, then $ART^{(2)}_K(new) \leftarrow ART^{(2)}_K(old) \cup ART^{(1)}_l$. Finally, to compress the representation, i.e., to reduce the number of categories, in the last step of the Merge ART procedure, the category weight vectors $\bm{w}^{ART^{(2)}_k}$ and instance countings $n^{ART^{(2)}_k}$ of each local ART module are fed to a fuzzy ART with higher order activation and match functions, using the parameters~$\rho=\rho_{UB}^{(2)}$, $\gamma^{*}=1$, and $\gamma$; in this case, when a category learns using Eq.~(\ref{Eq:update}) then its instance counting is updated as $n^{new}=n^{old} + n^{w}$, where $n^{w}$ is the instance counting of the category presented as an input.
 
\begin{table}[!t]
\centering
\caption{Merge ART's activation and match functions.}
\begin{threeparttable}
\begingroup\setlength{\fboxsep}{0pt}
\colorbox{lightgray}{
\begin{tabular*}{\columnwidth}{@{\extracolsep{\fill}}lll@{}}
\toprule
Method 
& $T^{ART^{(2)}_k}=f(\cdot)$ 
& $M^{ART^{(2)}_k}=g(\cdot)$ \\
\midrule
\midrule
single 	  
& $\max\limits_{i,j}\left( [t_{ij}] \right)$	
& $\max\limits_{i,j}\left( [m_{ij}] \right)$ \\
complete  
& $\min\limits_{i,j}\left( [t_{ij}] \right)$	
& $\min\limits_{i,j}\left( [m_{ij}] \right)$ \\
median    
& $\median\limits_{i,j}\left( [t_{ij}] \right)$	
& $\median\limits_{i,j}\left( [m_{ij}] \right)$ \\
average 
& $\frac{1}{RC}\sum\limits_{i=1}^{R}\sum\limits_{j=1}^{C} t_{ij}$	
& $\frac{1}{RC}\sum\limits_{i=1}^{R}\sum\limits_{j=1}^{C} m_{ij}$ \\
weighted\tnote{a}  
& $\sum\limits_{i=1}^{R}\sum\limits_{j=1}^{C} p_ip_jt_{ij}$		
& $\sum\limits_{i=1}^{R}\sum\limits_{j=1}^{C} p_ip_jm_{ij}$ \\
centroid\tnote{b}   
& $\left(\frac{|\bm{w}^{ART^{(2)}_k}_c \wedge \bm{w}^{ART^{(1)}_l}_c|}{\alpha +|\bm{w}^{ART^{(2)}_k}_c|}\right)^\gamma$    
& $\left(\frac{|\bm{w}^{ART^{(2)}_k}_c \wedge \bm{w}^{ART^{(1)}_l}_c|}{|\bm{w}^{ART^{(1)}_l}_c|}\right)^\gamma$  \\
\bottomrule
\end{tabular*}
}\endgroup
\begin{tablenotes}[normal,flushleft]
\item[a]$p_i=\frac{n_i^{ART^{(2)}_k}}{n^{ART^{(2)}_k}}$ and $p_j=\frac{n_j^{ART^{(1)}_l}}{n^{ART^{(1)}_l}}$.
This represents an a priori probability of categories $i$ and $j$ analogous to~\cite{williamson1996,vigdor2007}. Statistical independence is assumed.
\item[b]$\bm{w}^{ART^{(2)}_k}_c$ and $\bm{w}^{ART^{(1)}_l}_c$ are the centroids representing all categories of $ART^{(2)}_k$ and $ART^{(1)}_l$, respectively. Each of their $n$ components is given by $w^{ART^{(2)}_k}_{c,n}=\min\limits_{j}\left( w^{ART^{(2)}_k}_{j,n}\right)$ and $w^{ART^{(1)}_l}_{c,n}=\min\limits_{j}\left( w^{ART^{(1)}_l}_{j,n}\right)$, where $n=\{1,...,2d\}$.
\end{tablenotes}
\end{threeparttable}
\label{tab:MergeART_T_M}
\end{table}

The Merge ART module can be triggered at any stage during incremental learning. For convenience, in this study it is activated by the end of one epoch (a full pass through the data, similar to~\cite{Swope2012}), i.e., after $N$ samples are presented to the learning system, where $N$ is made equal to the data cardinality. Therefore, this framework may perform online incremental approximate HAC without computing a distance matrix with the entire data or requiring full recomputations when new samples are presented. Again, as the vigilance parameter $\rho_{UB}$ approaches~1, there is little to no data compression. Merge ART relates to traditional HAC approaches using ART's activation function as the similarity measure and the match function as the dendrogram threshold level, i.e., the activation and match functions of the Merge ART module perform an ART-based HAC using the ART weights created by DDVFA. Algorithm~\ref{Alg:MergeART} summarizes the Merge ART module's pseudocode.

\begin{algorithm}[!b]
\DontPrintSemicolon
\SetKwInOut{Input}{Input}\SetKwInOut{Output}{Output}
\BlankLine
\Input{DDVFA, $\left\{\alpha,\beta,\rho_{UB}^{(2)},\rho_{LB}^{(2)},\gamma,\gamma^{*}, method \right\}$ inherited from DDVFA, number of iterations.} 
\Output{Merge ART clusters.}
\algrule
\nl \Repeat{stopping criteria: reaching a predefined number of iterations or there is no change in Merge ART nodes}{
\nl \For{$l=\{1,~...~,No.~global~ART~F_2~nodes\}$}{
\nl Present input node $ART^{(1)}_l \in$ DDVFA. \;
\nl Compute $T^{ART^{(2)}_k},~\forall~k$ (Table~\ref{tab:MergeART_T_M}'s method). \;
\nl \label{MergeART_search}Find the winning node $K \leftarrow \argmax\limits_{k}\{T^{ART^{(1)}_k}\}$. \;
\nl Compute $M^{ART^{(2)}_K}$ (Table~\ref{tab:MergeART_T_M}'s method). \;
\nl Evaluate vigilance test $\nu_1 : M^{ART^{(2)}_K} \geq \rho_{LB}^{(2)}$. \;
\nl \uIf{$\nu_1$ is satisfied (resonance)}{
\nl $ART^{(2)}_K \leftarrow ART^{(2)}_K \cup ART^{(1)}_l$.  \;
}
\nl \Else{
\nl Reset node $K$. If there are still active nodes in Merge ART then go to step~\ref{MergeART_search}; otherwise create a new ART node and apply fast commit ($ART^{(2)}_{new} \leftarrow ART^{(1)}_l$).\;}
}
\nl DDVFA $\leftarrow$ Merge ART.\;
}
\nl \For{each $ART^{(2)}_k \in$ Merge ART}{
\nl $ART^{(2)}_k \leftarrow FA\left(\{\bm{w},n\} \in ART^{(2)}_k, \rho_{UB}^{(2)}, \gamma, \gamma^{*}, \alpha, \beta \right)$. \;
\tcc{FA: Fuzzy ART algorithm.} }
\caption{Merge ART module}\label{Alg:MergeART}
\end{algorithm}

\textbf{Remark 3.} Merging strategies are commonly employed in ART-based systems. The Merge ART module presented here is closely related to the ART category merging methods discussed in~\cite{Lei2006,Isawa2008,Isawa2008b,Isawa2009,Zhang2006,Swope2012,Benites2017} and especially the frameworks in~\cite{Swope2012,Benites2017}. In the latter, fuzzy ART weights are merged via a fuzzy ART module with its own set of parameters. Although both the DDVFA + Merge ART and the strategy in~\cite{Swope2012,Benites2017} use a fuzzy ART framework for merging, they have the following fundamental differences: (a)~Merge ART's inputs are local fuzzy ART modules from DDVFA (i.e., subsets of categories) to be merged using a fuzzy ART framework augmented with HAC-based distributed higher order activation and match functions; (b)~the output of the merging procedure includes not only categories but also ART modules; (c)~Merge ART's compression step does not use an activation threshold (as in~\cite{Swope2012}), but instead it uses higher order activation/match functions (in contrast to~\cite{Swope2012,Benites2017}); (d)~the weight update is not based on an overlap/gap between weights (as in~\cite{Swope2012}), but instead it follows standard fuzzy ART rules (Eq.~(\ref{Eq:update})) which correspond to the weight merging in~\cite{Benites2017} (and~\cite{Zhang2006} in fast learning mode); and (e)~the vigilance parameter used to cluster samples is also used to merge weights during the compression step (in contrast to~\cite{Swope2012}).

The Merge ART module was designed such that its output can be used to replace DDVFA when the merging procedure is done. The fact that $\rho^{(2)}_{LB}$ used to concatenate DDVFA's local FAs is smaller than $ \rho^{(1)}_{UB}$ used to cluster the samples, ($\rho^{(2)}_{LB} = \rho^{(1)}_{LB} \leq \rho^{(1)}_{UB}=\rho^{(2)}_{UB}$), conforms with the findings reported in~\cite{Swope2012} that this setting yields a good performance for merging fuzzy ART weights. This is expected, since the overall architecture (DDVFA + Merge ART) is multi-layered and related to ART-based serial structures (e.g., \cite{Ishihara1995,Bartfai1996}), which in turn typically follow similar parameterization.

\section{Experimental Setup} \label{Sec:experiments}

\subsection{Data sets}

A mix of $30$ real world and artificial benchmark data sets comprising diverse characteristics were used in the experiments. They are available at the UCI Machine Learning Repository~\cite{uci}, Fundamental Clustering Problem Suite~\cite{fcps}, Clustering data sets~\cite{shape}, and Data package~\cite{datapkg}. Fig.~\ref{Fig:datasets} illustrates these data sets, and Table~\ref{Tab:datasets} summarizes their characteristics. Linear normalization was applied to all data sets to scale their features to the range $[0,1]$, as well as complement coding, which is a useful data representation technique to mitigate category proliferation in fuzzy ART.

\newcommand{\y}{0.16}
\begin{figure}[!ht]
\centerline{
\subfloat[Aggreg.]{\includegraphics[height=\y\textwidth]{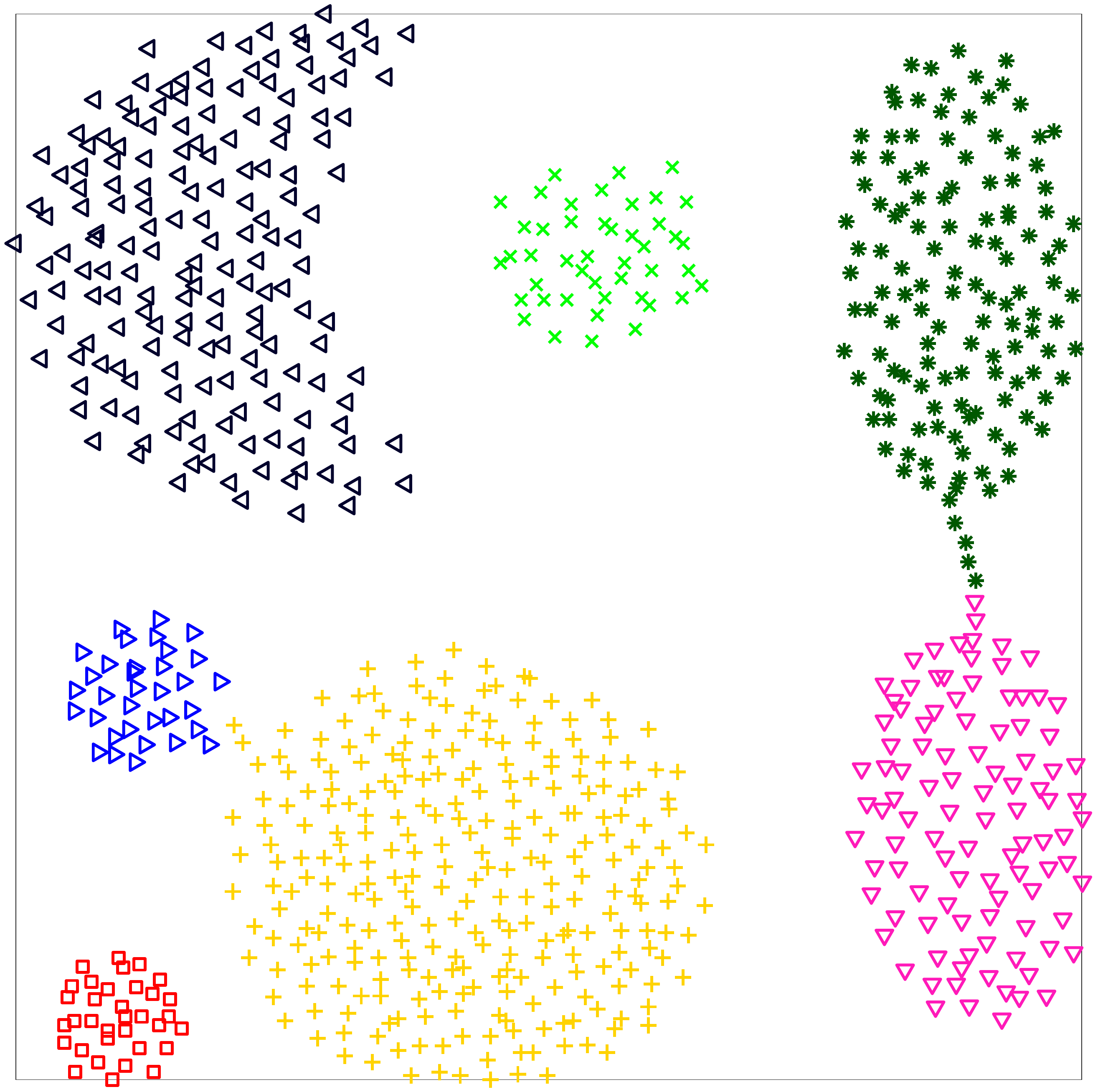}
\label{Fig:datasetsA}}
\subfloat[Atom]{\includegraphics[height=\y\textwidth]{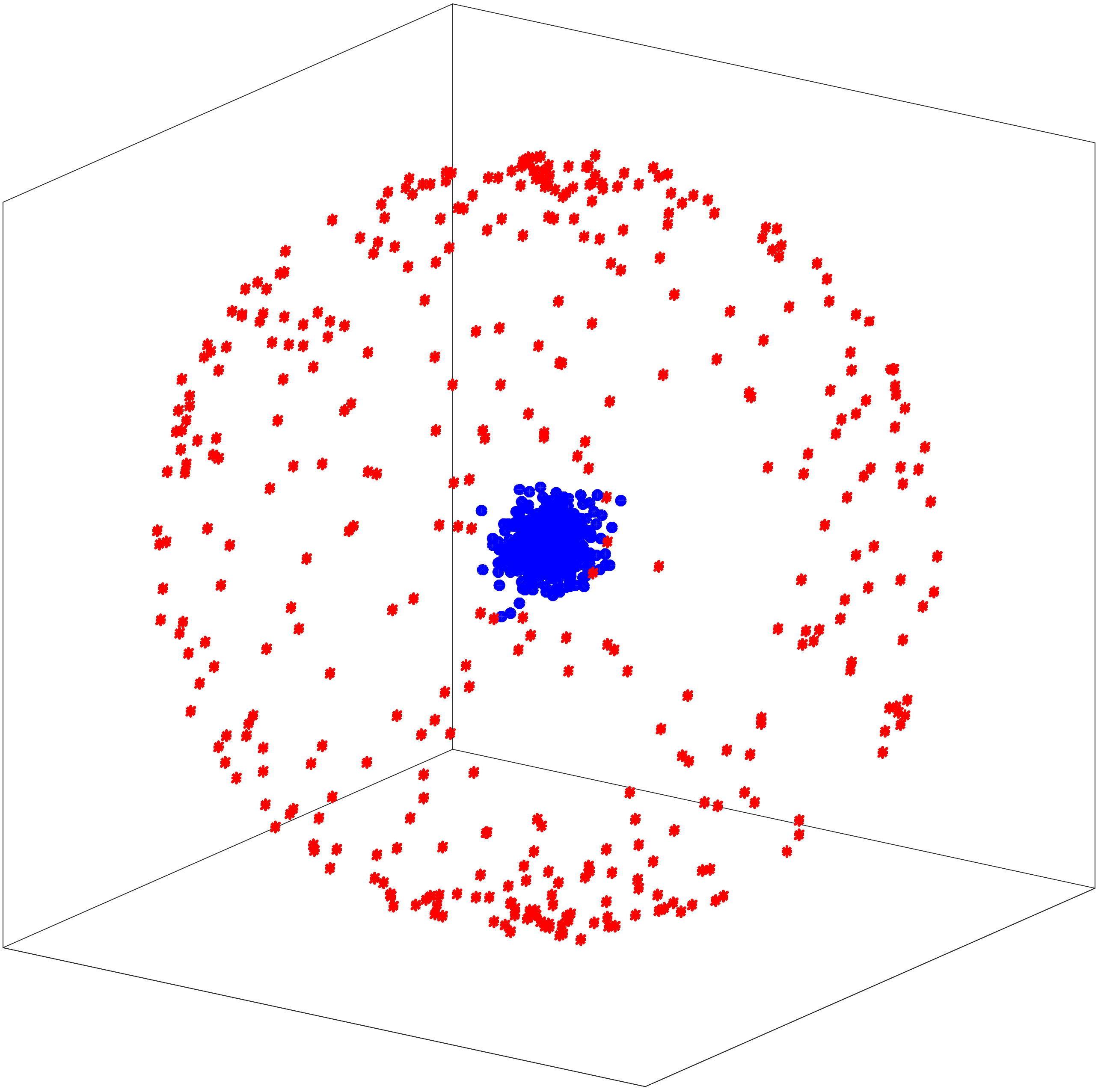}
\label{Fig:datasetsB}}
\subfloat[Chainlink]{\includegraphics[height=\y\textwidth]{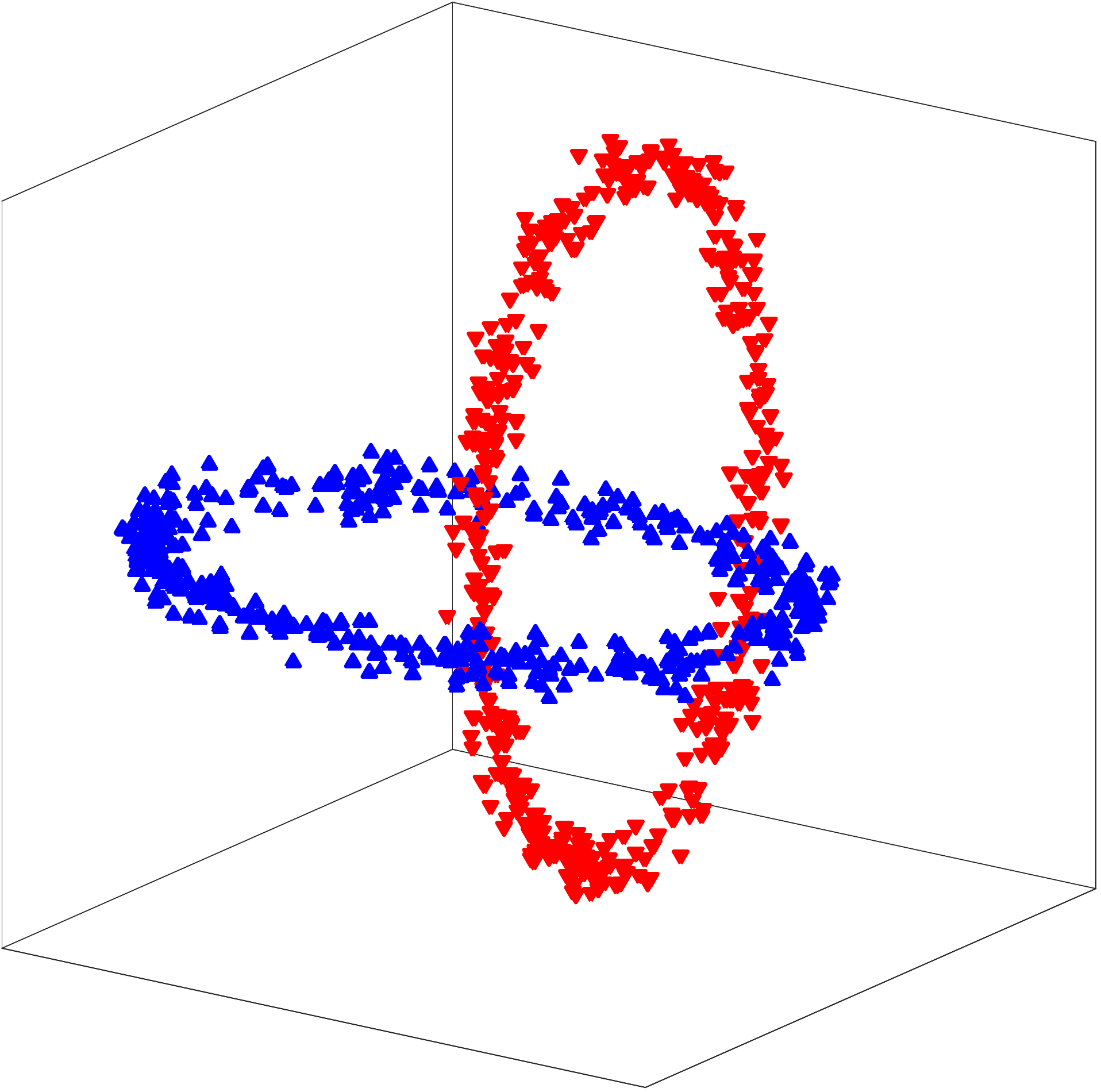}
\label{Fig:datasetsC}}
\subfloat[Compound]{\includegraphics[height=\y\textwidth]{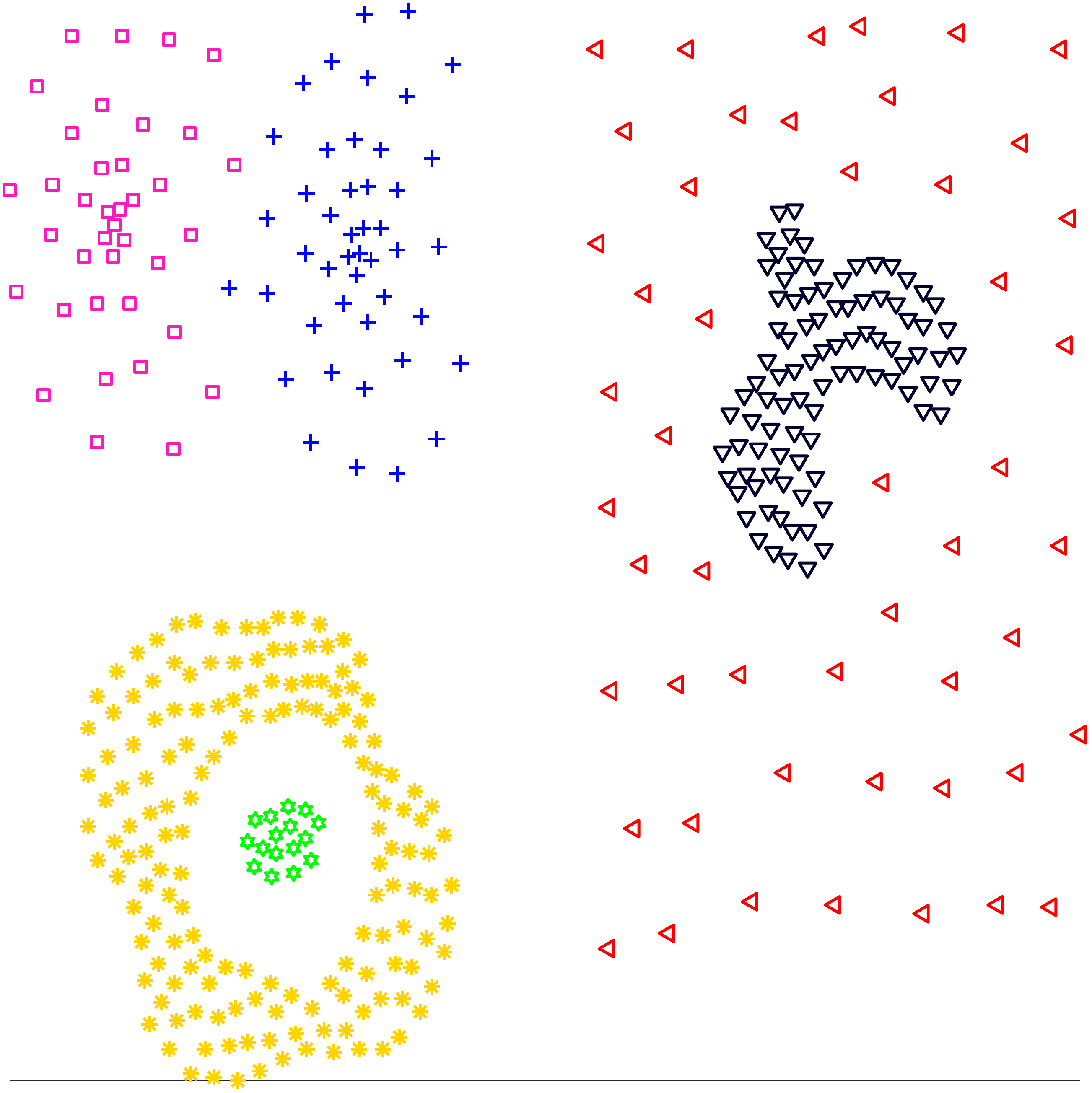}
\label{Fig:datasetsD}}
\subfloat[Dermat.]{\includegraphics[height=\y\textwidth]{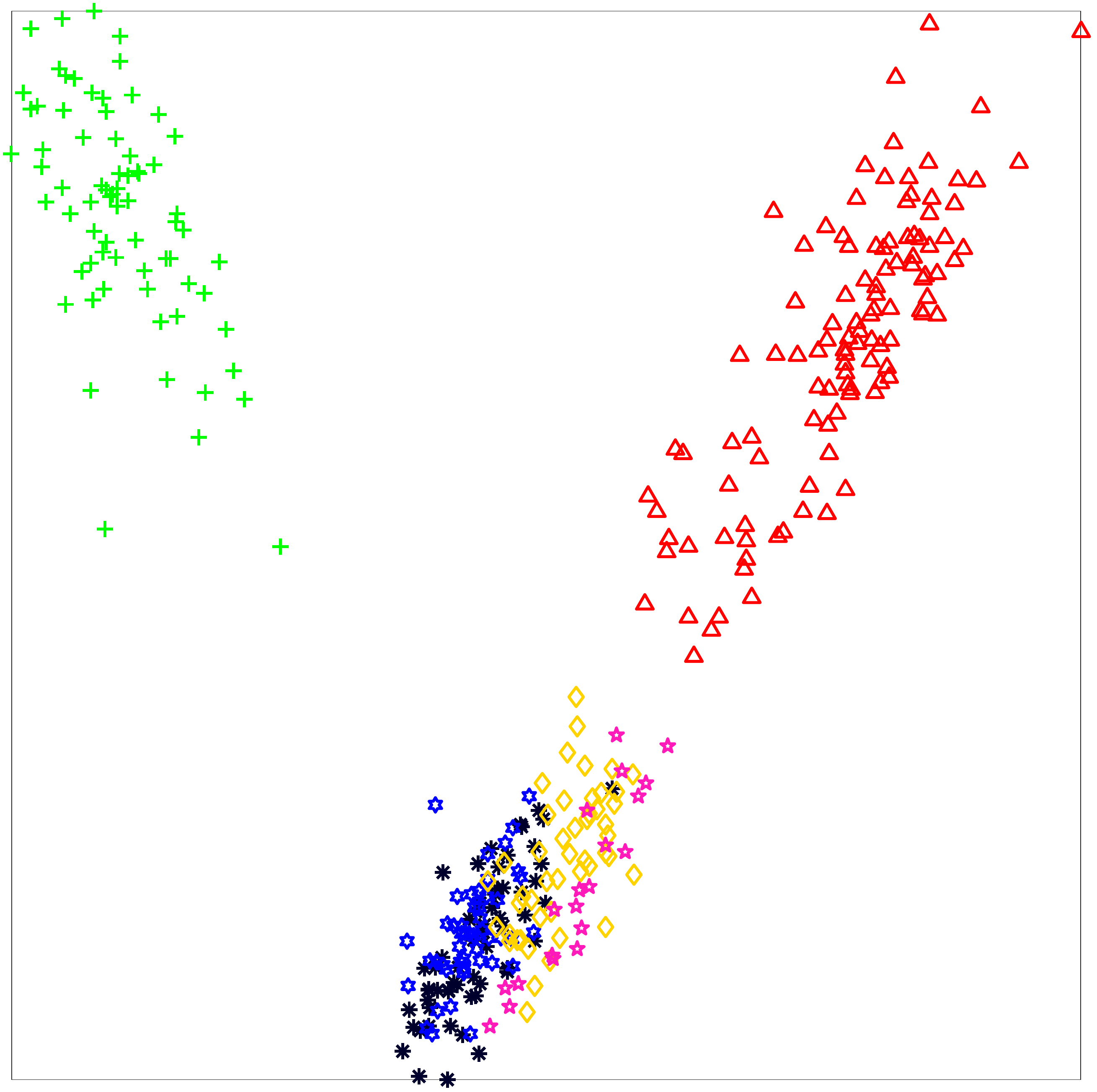}
\label{Fig:datasetsE}}
}
\vspace{-0.5\baselineskip}
\centerline{
\subfloat[Ecoli]{\includegraphics[height=\y\textwidth]{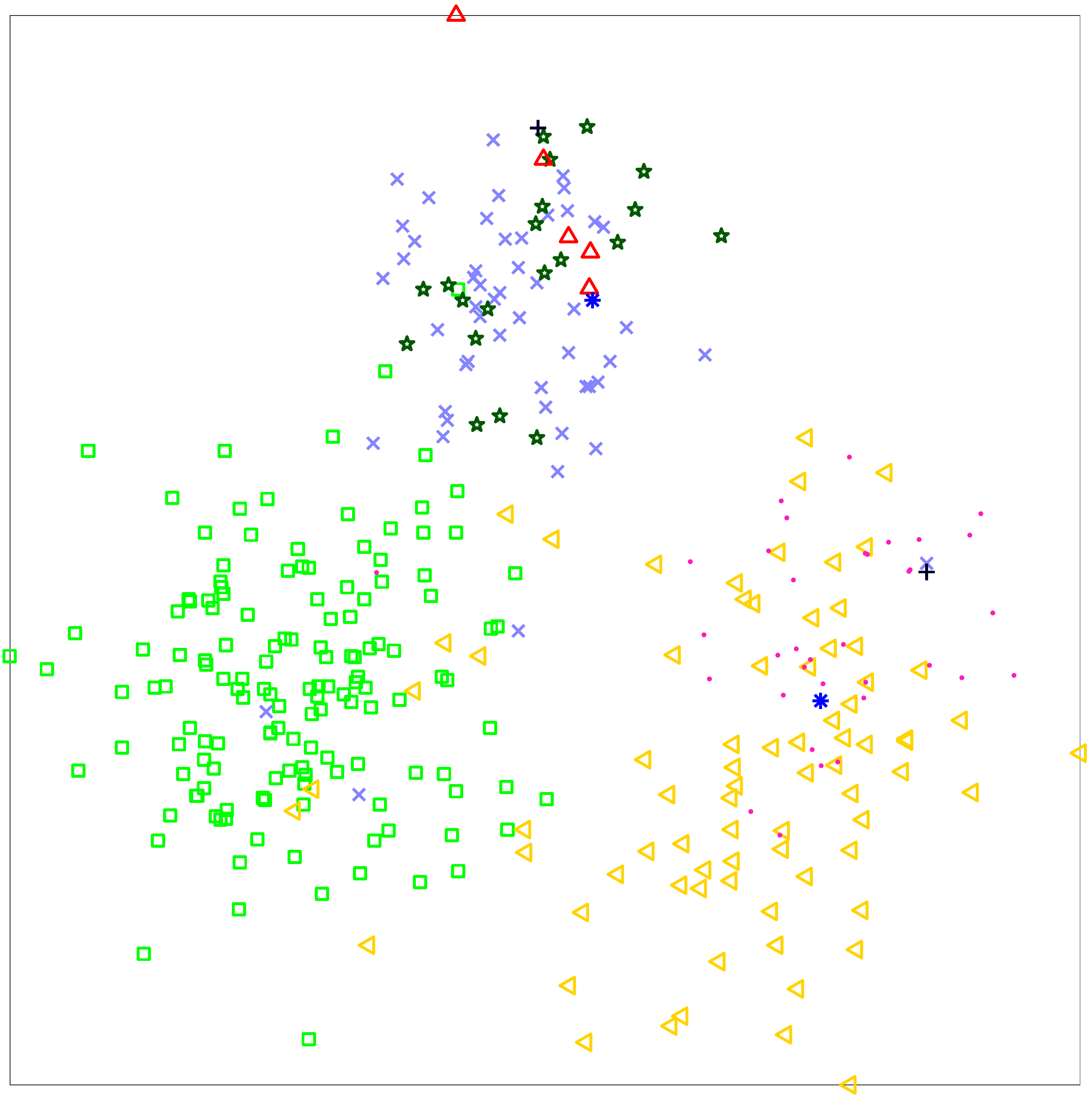}
\label{Fig:datasetsF}}
\subfloat[Face]{\includegraphics[height=\y\textwidth]{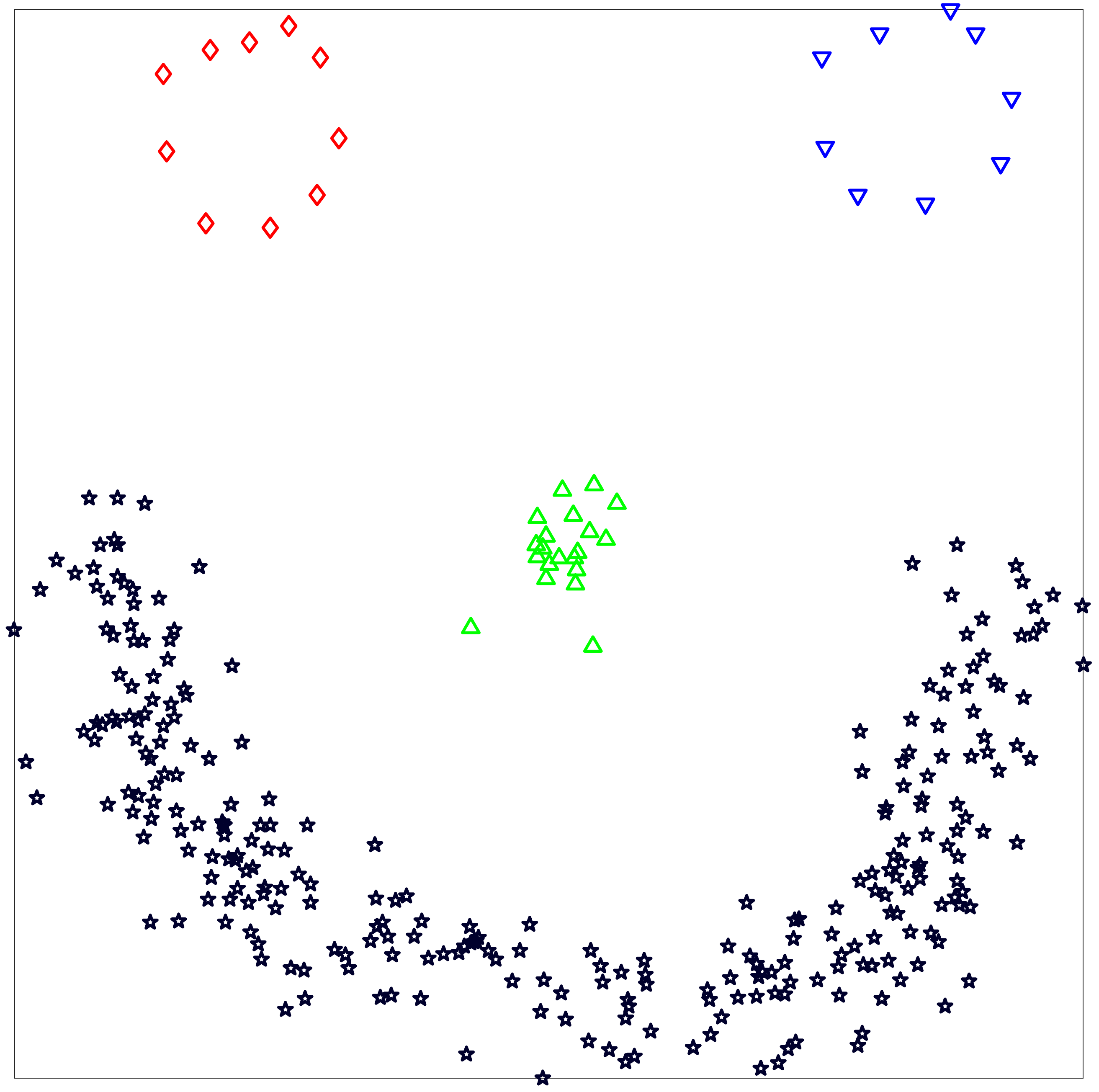}
\label{Fig:datasetsG}}
\subfloat[Flag]{\includegraphics[height=\y\textwidth]{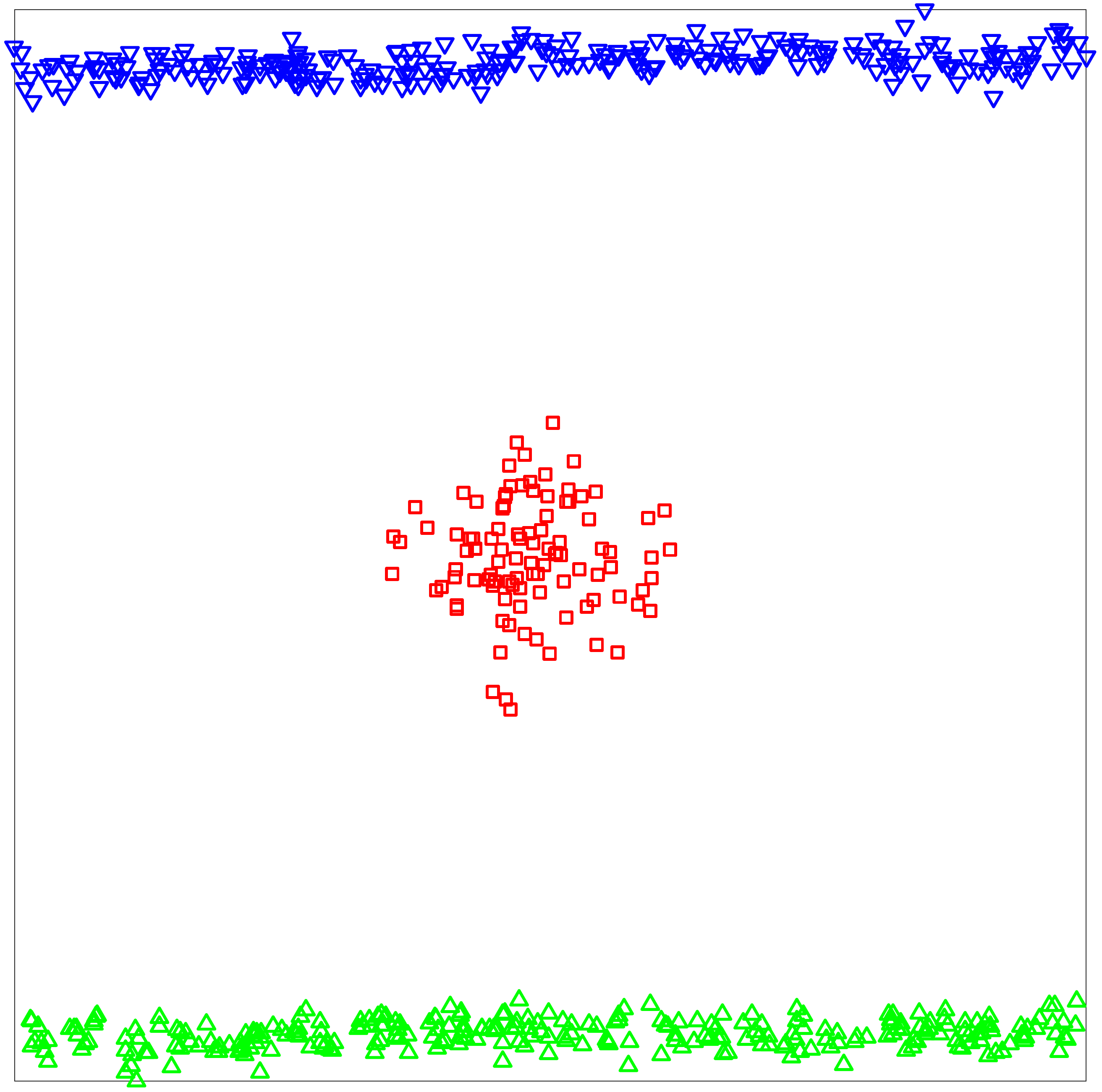}
\label{Fig:datasetsH}}
\subfloat[Flame]{\includegraphics[height=\y\textwidth]{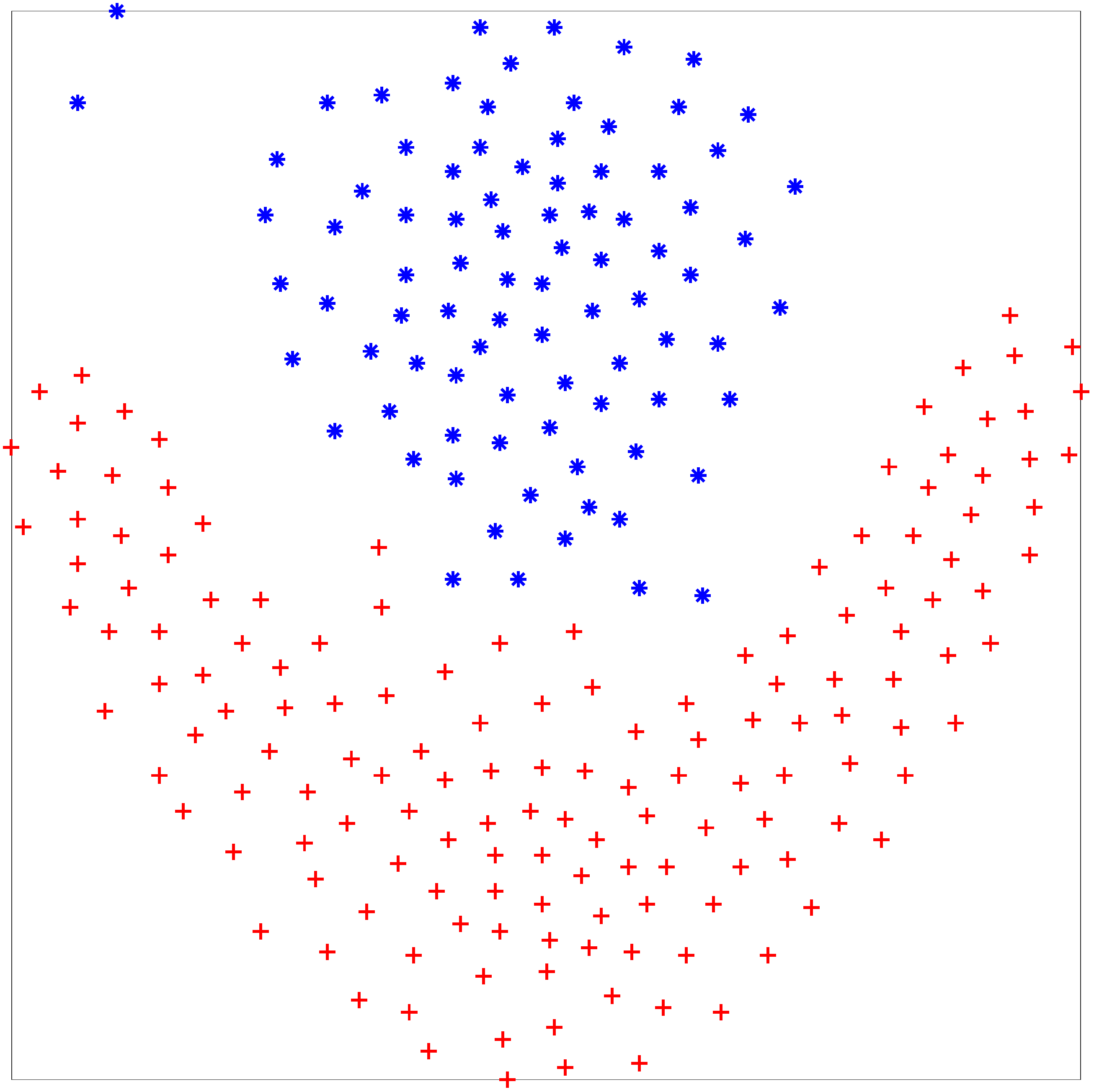}
\label{Fig:datasetsI}}
\subfloat[Giant]{\includegraphics[height=\y\textwidth]{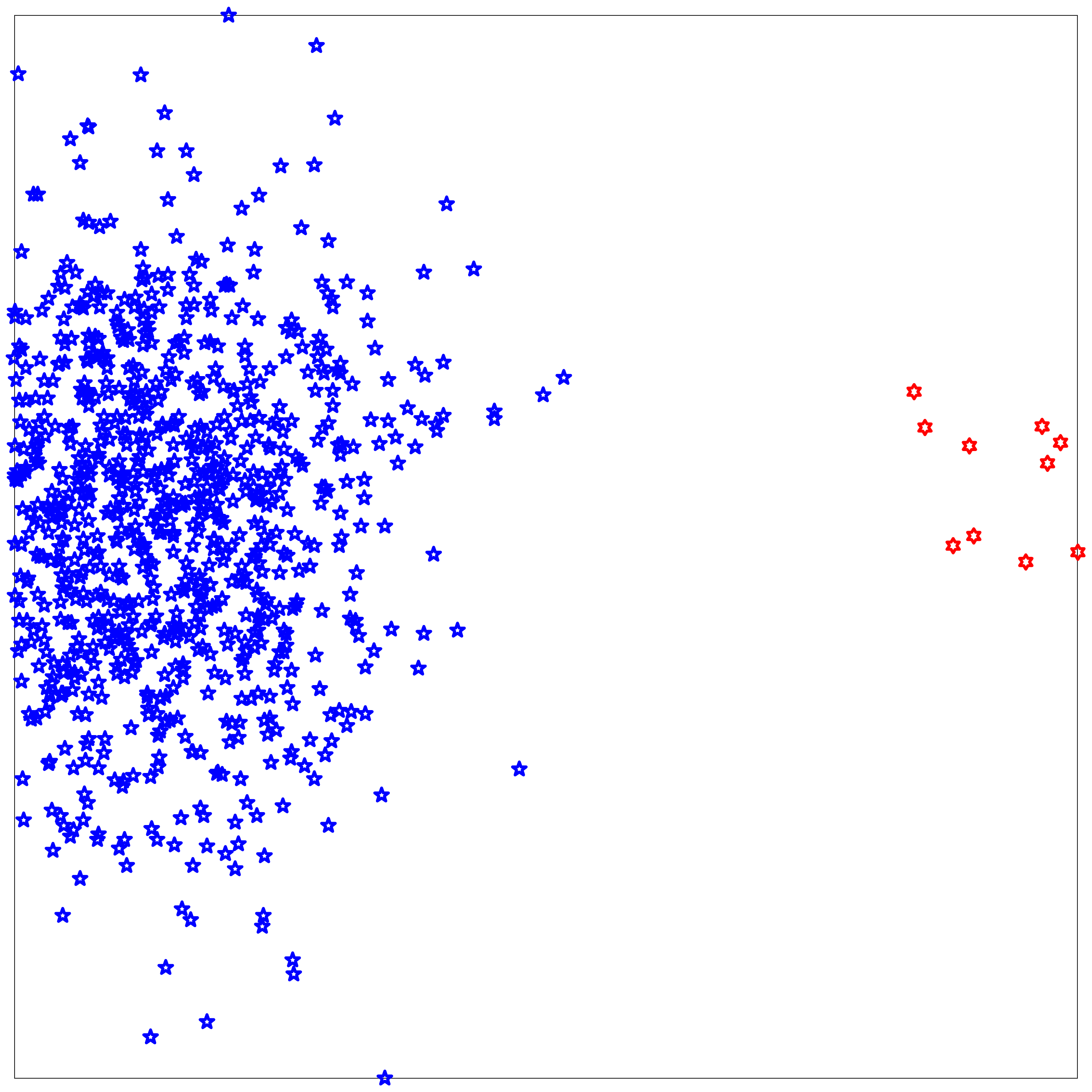}
\label{Fig:datasetsJ}}
}
\vspace{-0.5\baselineskip}
\centerline{
\subfloat[Glass]{\includegraphics[height=\y\textwidth]{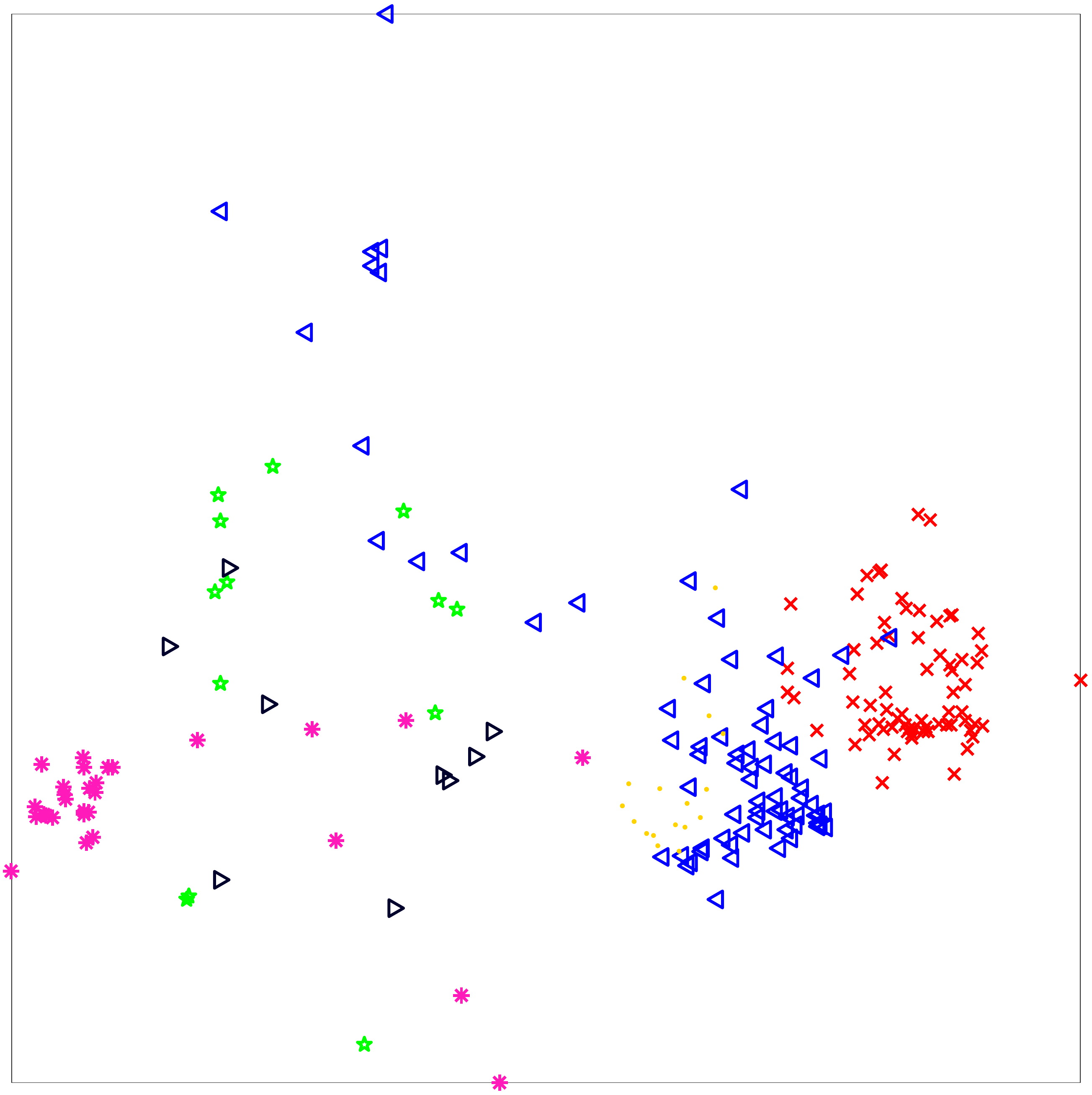}
\label{Fig:datasetsK}}
\subfloat[Hepta]{\includegraphics[height=\y\textwidth]{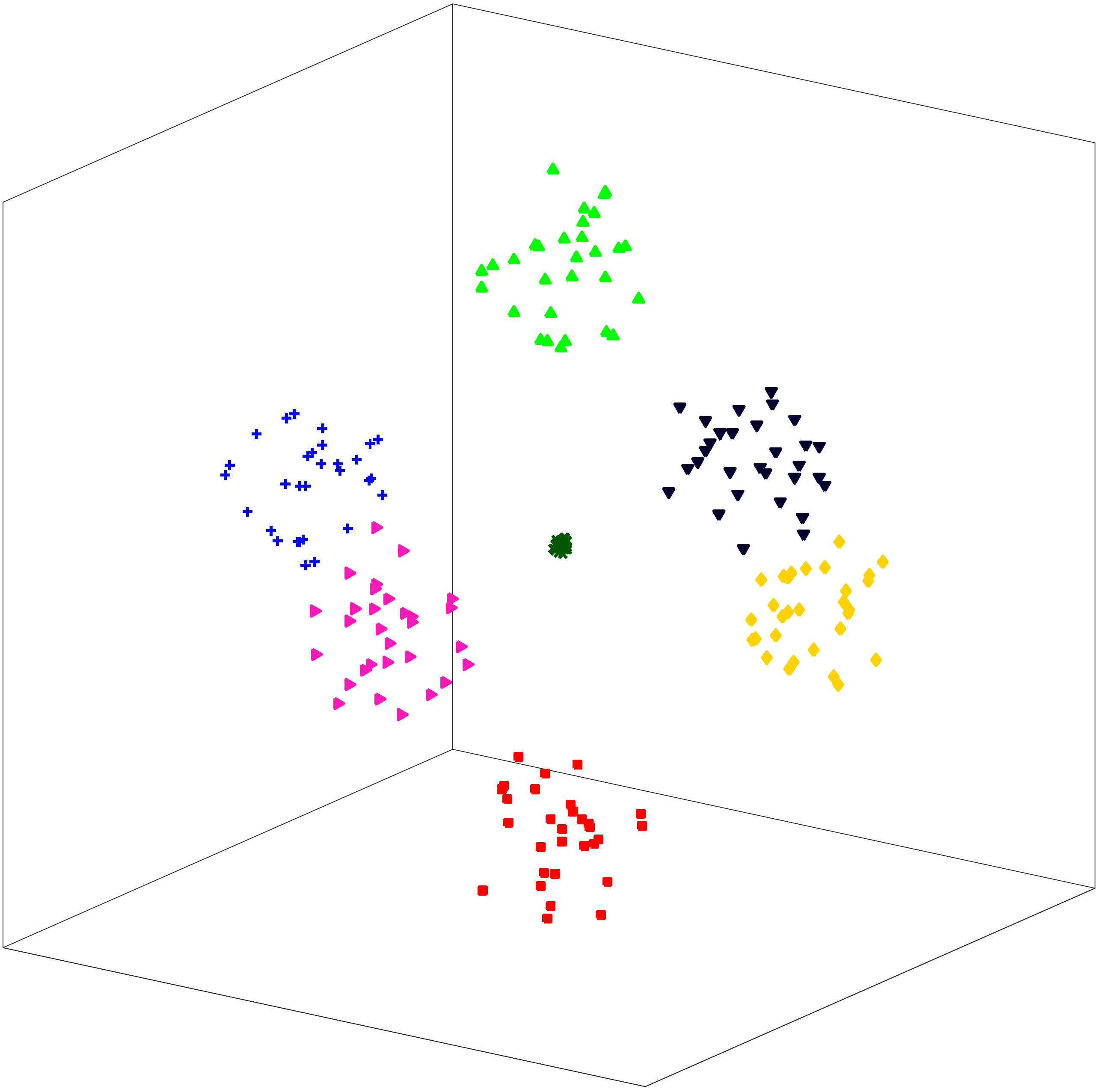}
\label{Fig:datasetsL}}
\subfloat[Iris]{\includegraphics[height=\y\textwidth]{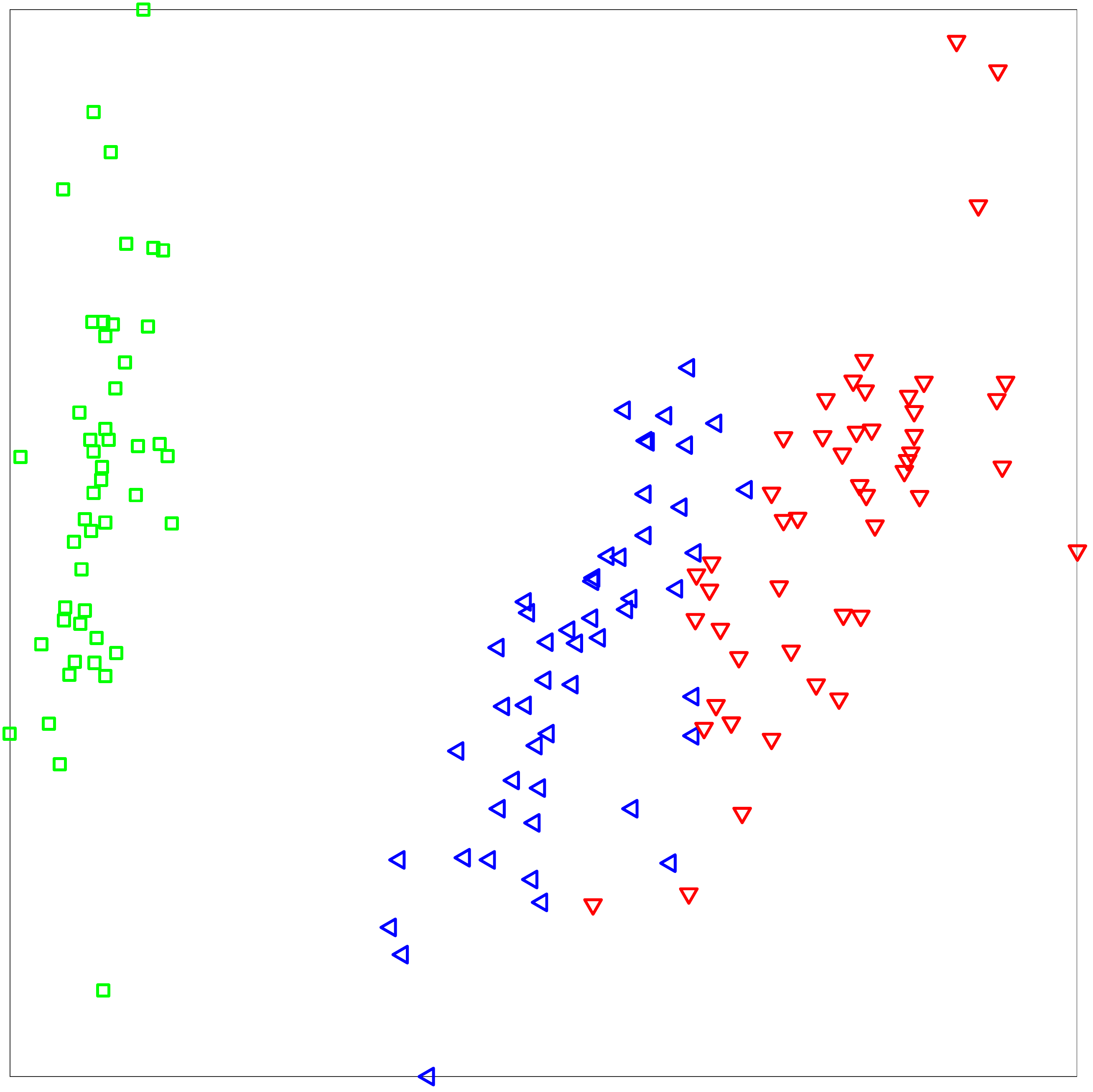}
\label{Fig:datasetsM}}
\subfloat[Jain]{\includegraphics[height=\y\textwidth]{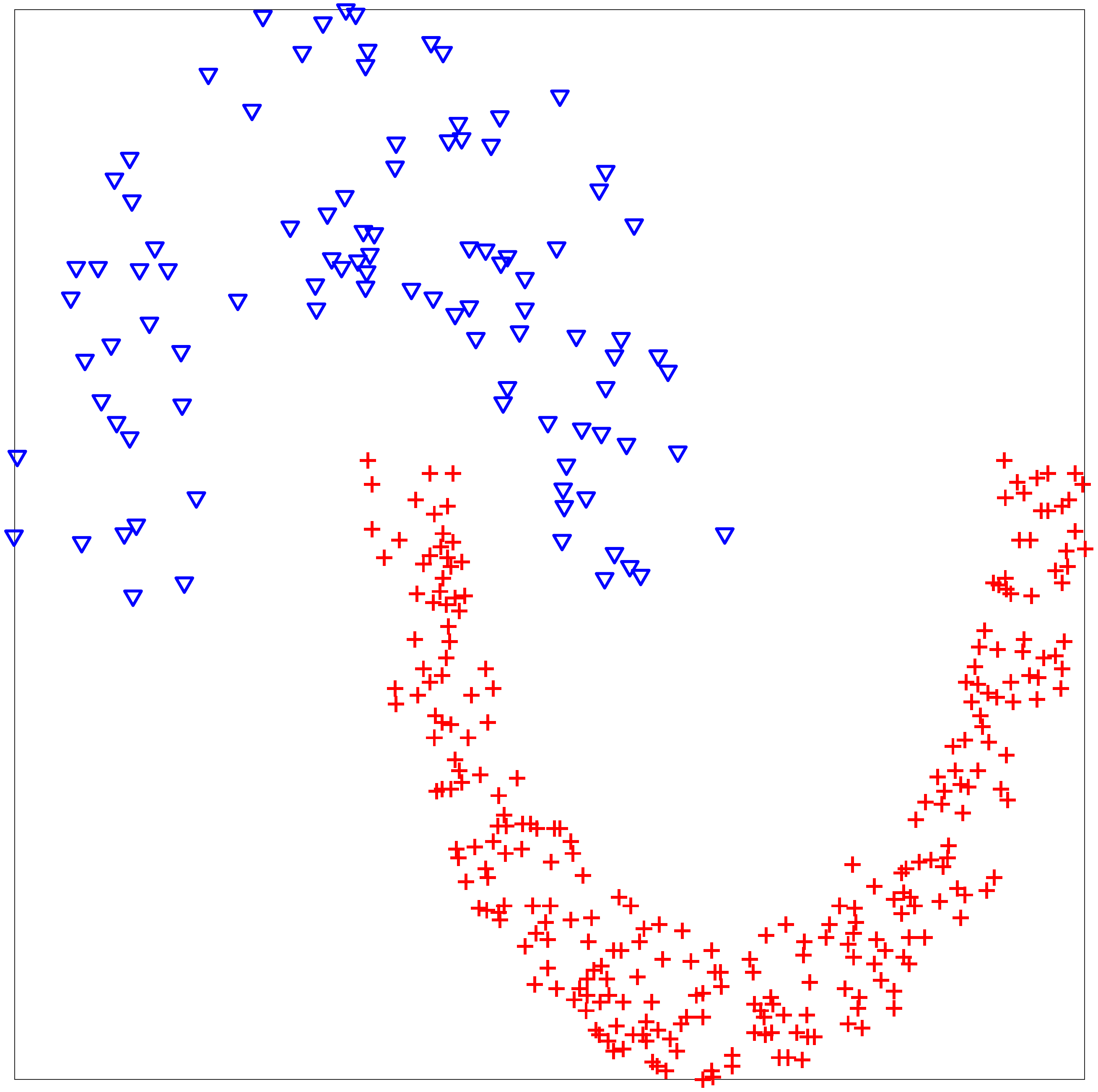}
\label{Fig:datasetsN}}
\subfloat[Lsun]{\includegraphics[height=\y\textwidth]{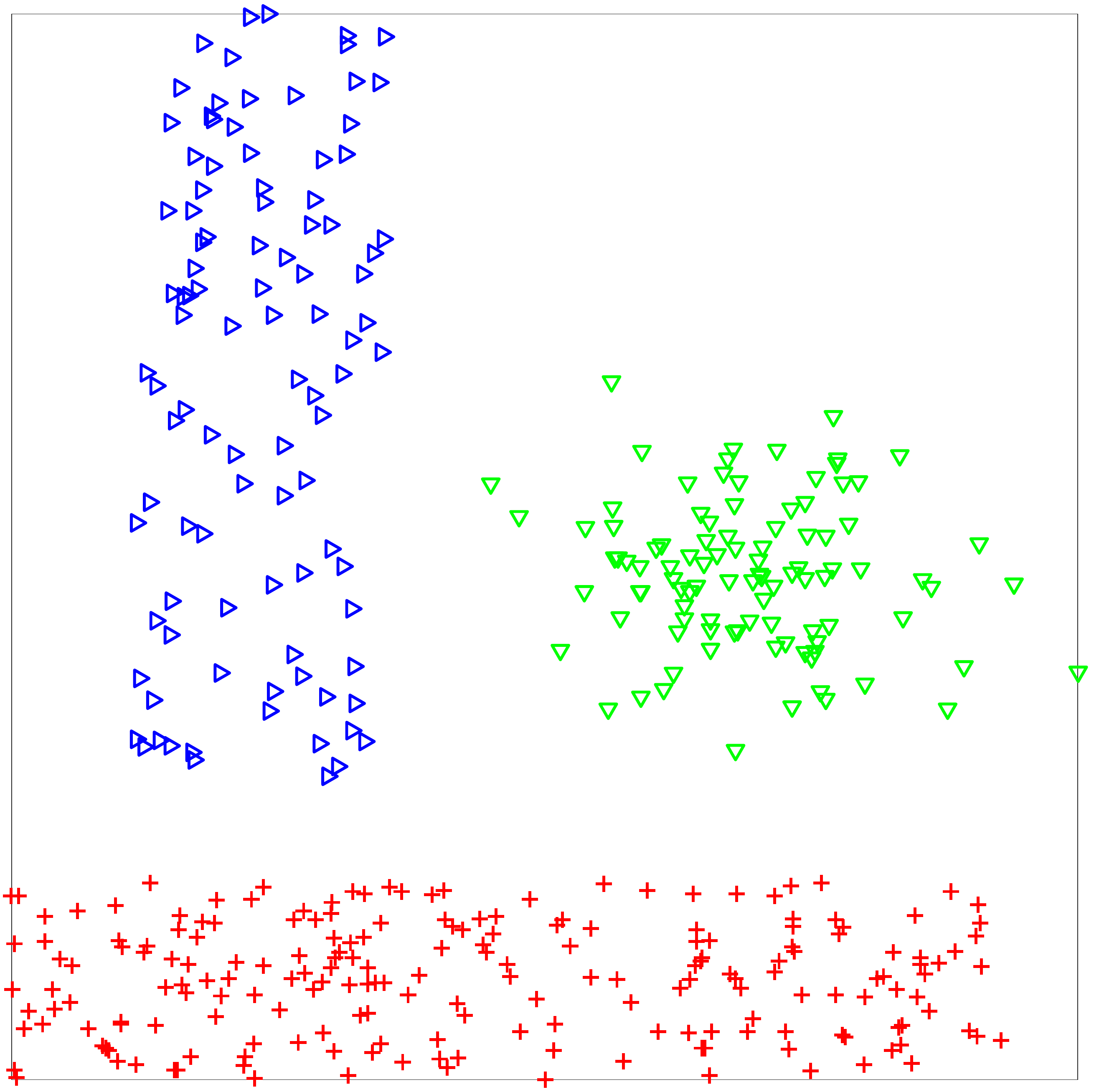}
\label{Fig:datasetsO}}
}
\vspace{-0.5\baselineskip}
\centerline{
\subfloat[Moon]{\includegraphics[height=\y\textwidth]{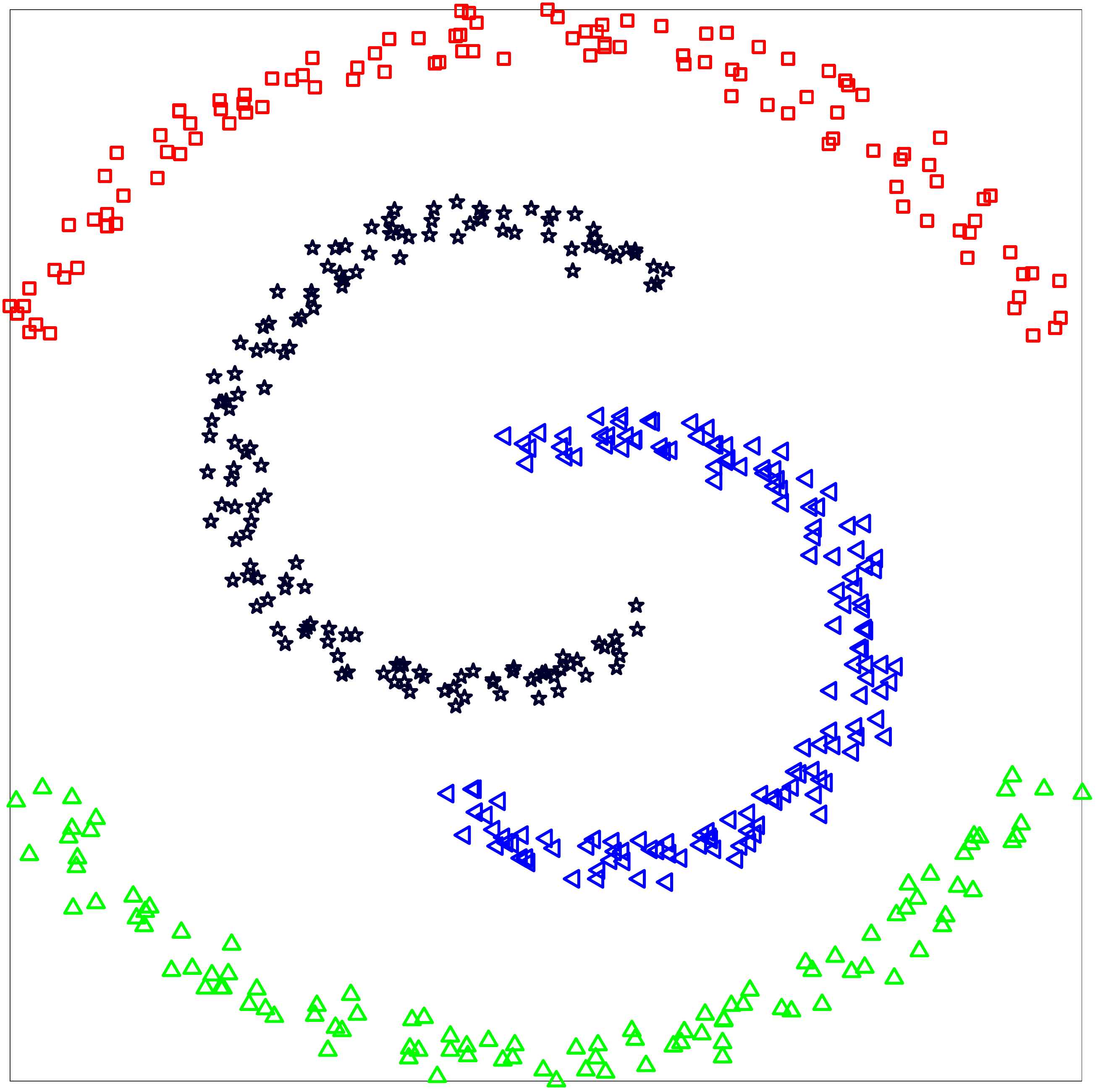}
\label{Fig:datasetsP}}
\subfloat[Path Based]{\includegraphics[height=\y\textwidth]{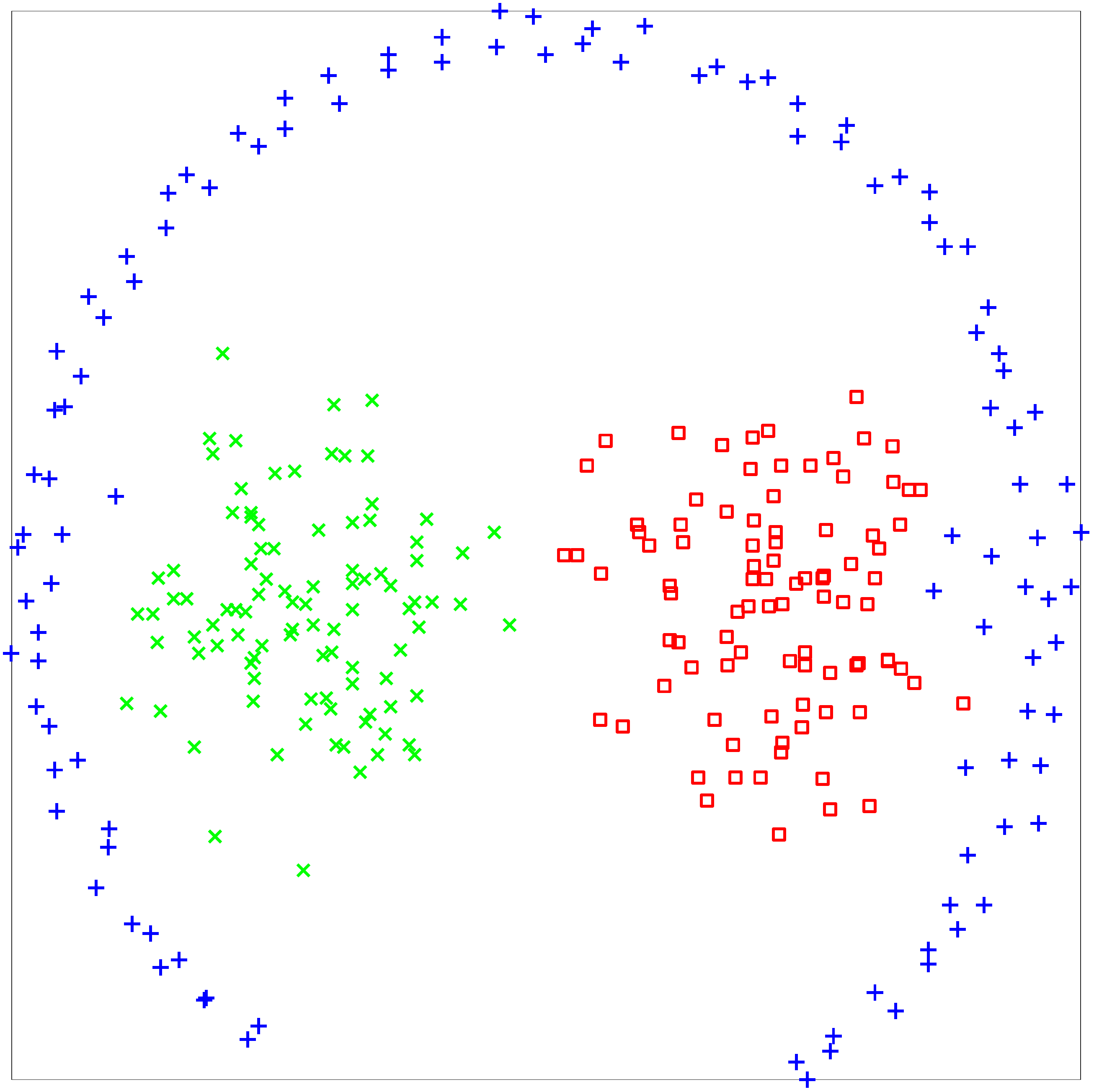}
\label{Fig:datasetsQ}}
\subfloat[R15]{\includegraphics[height=\y\textwidth]{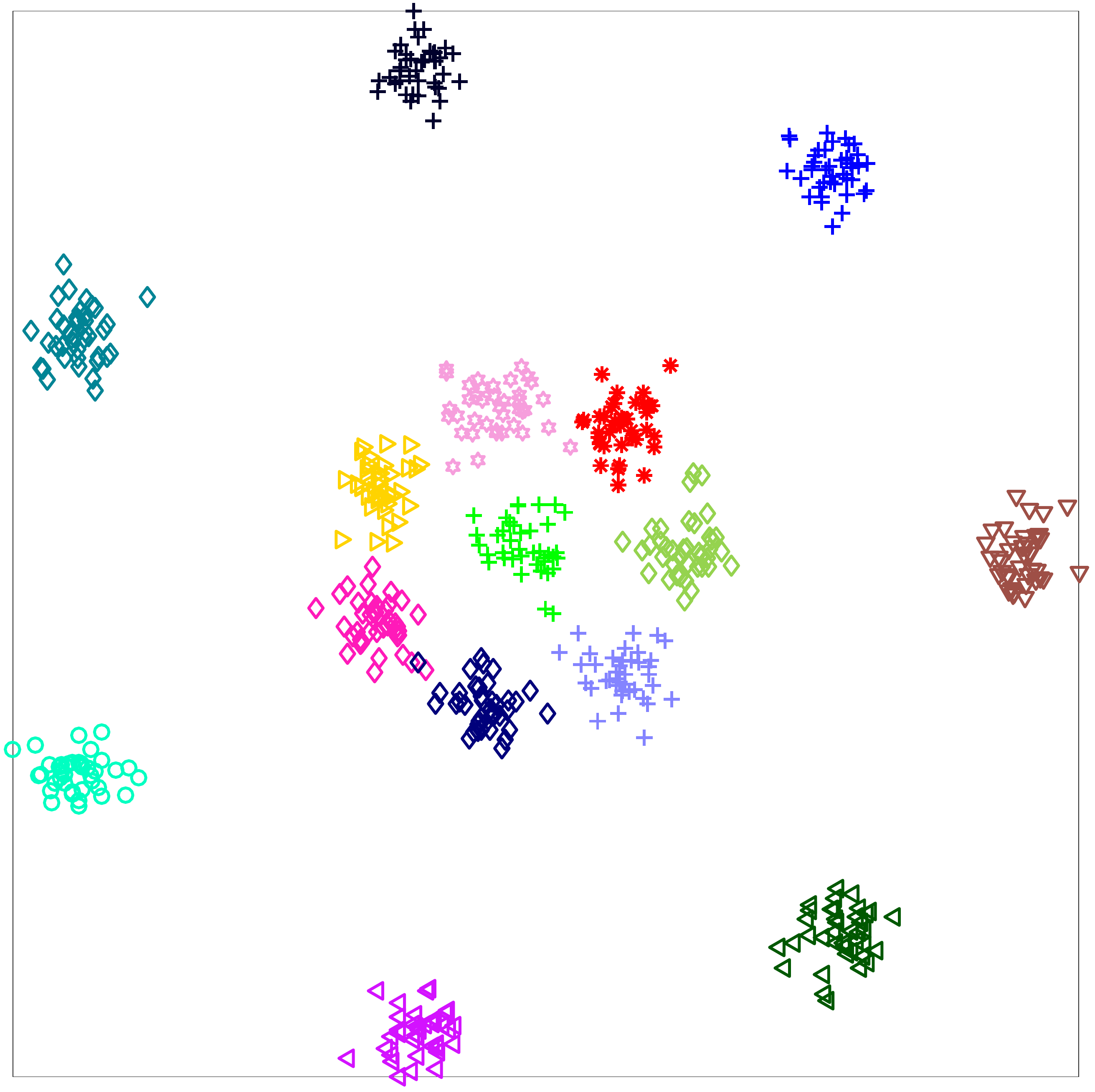}
\label{Fig:datasetsR}}
\subfloat[Ring]{\includegraphics[height=\y\textwidth]{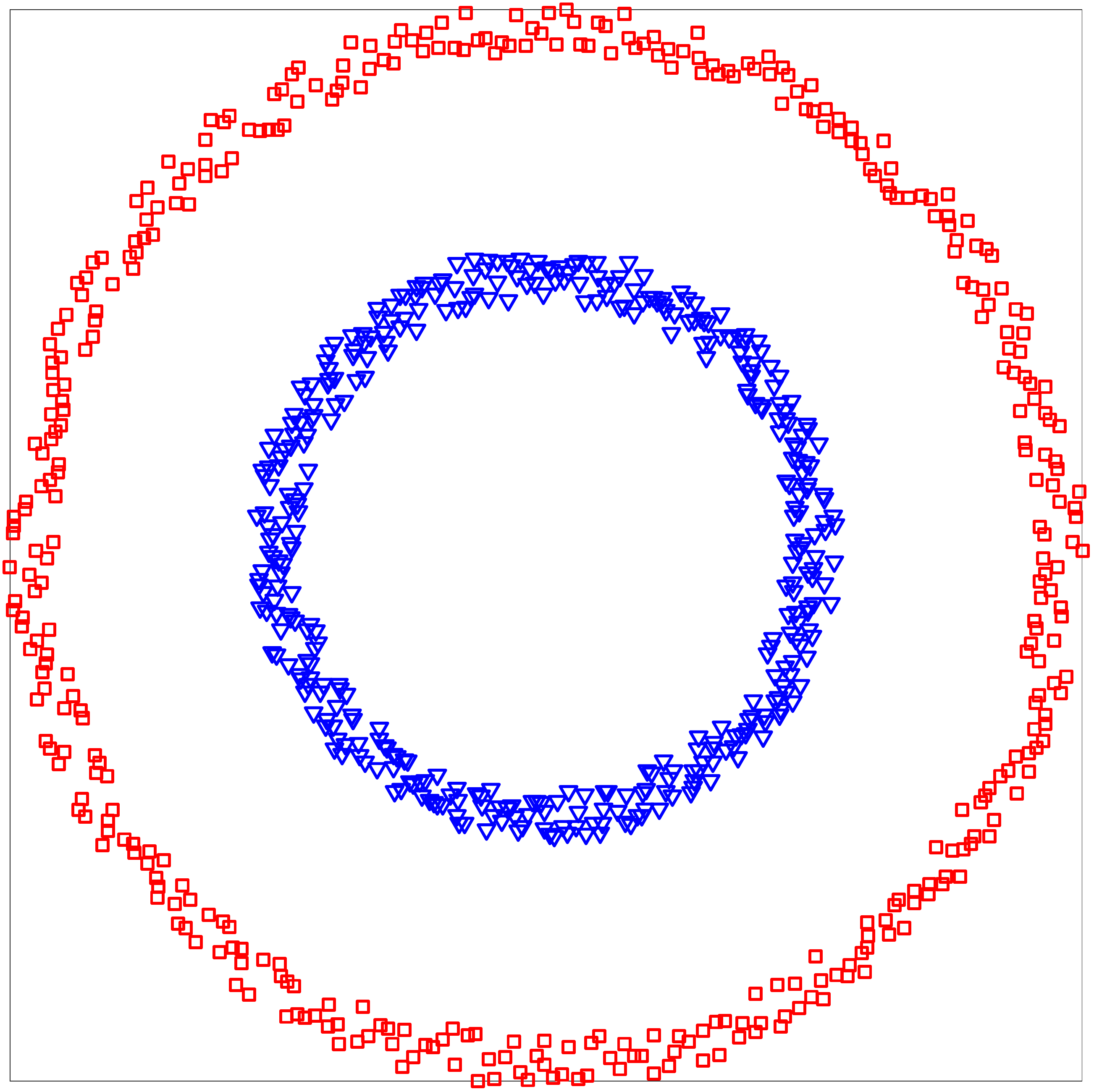}
\label{Fig:datasetsS}}
\subfloat[Seeds]{\includegraphics[height=\y\textwidth]{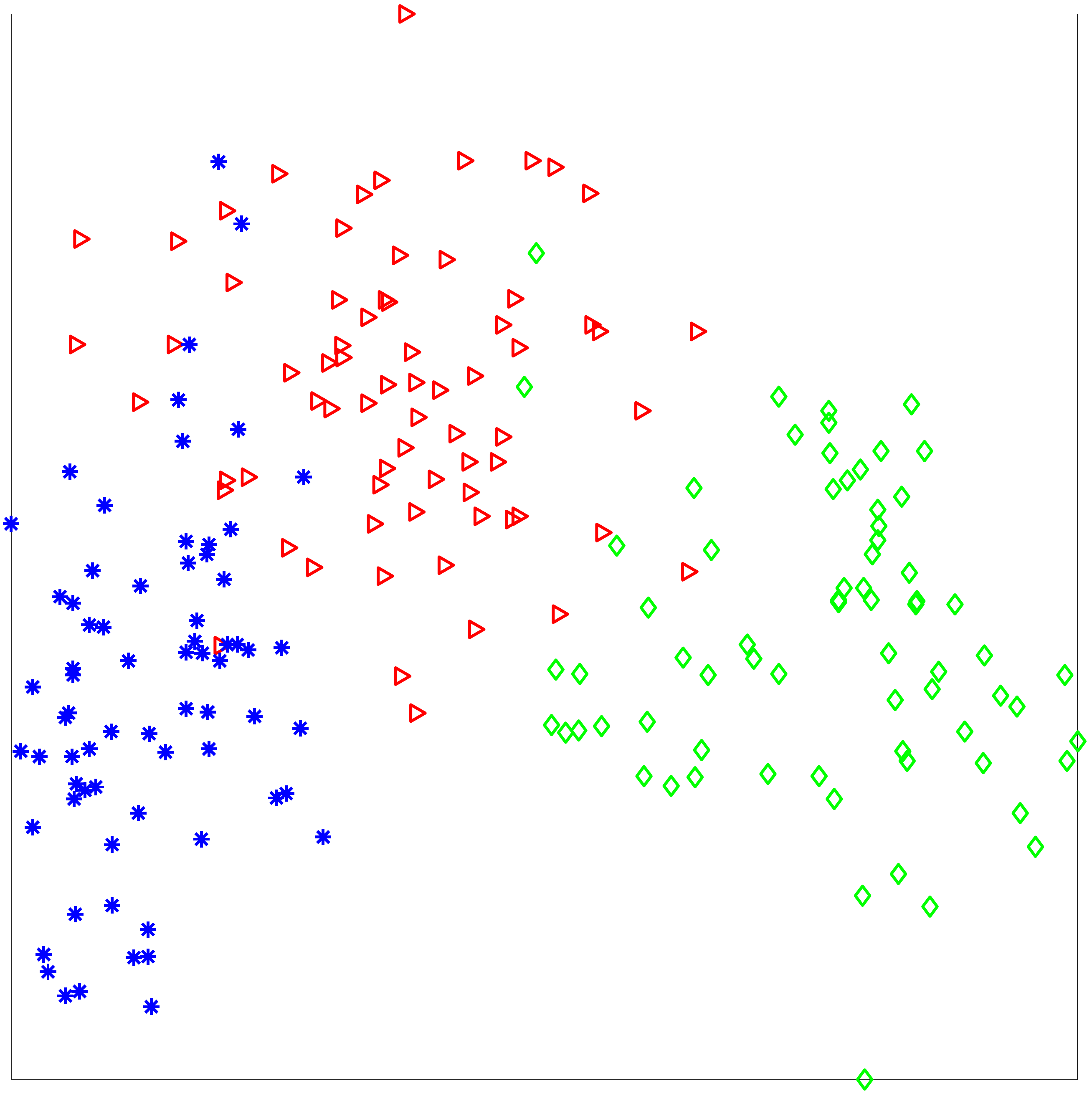}
\label{Fig:datasetsT}}
}
\vspace{-0.5\baselineskip}
\centerline{
\subfloat[Spiral]{\includegraphics[height=\y\textwidth]{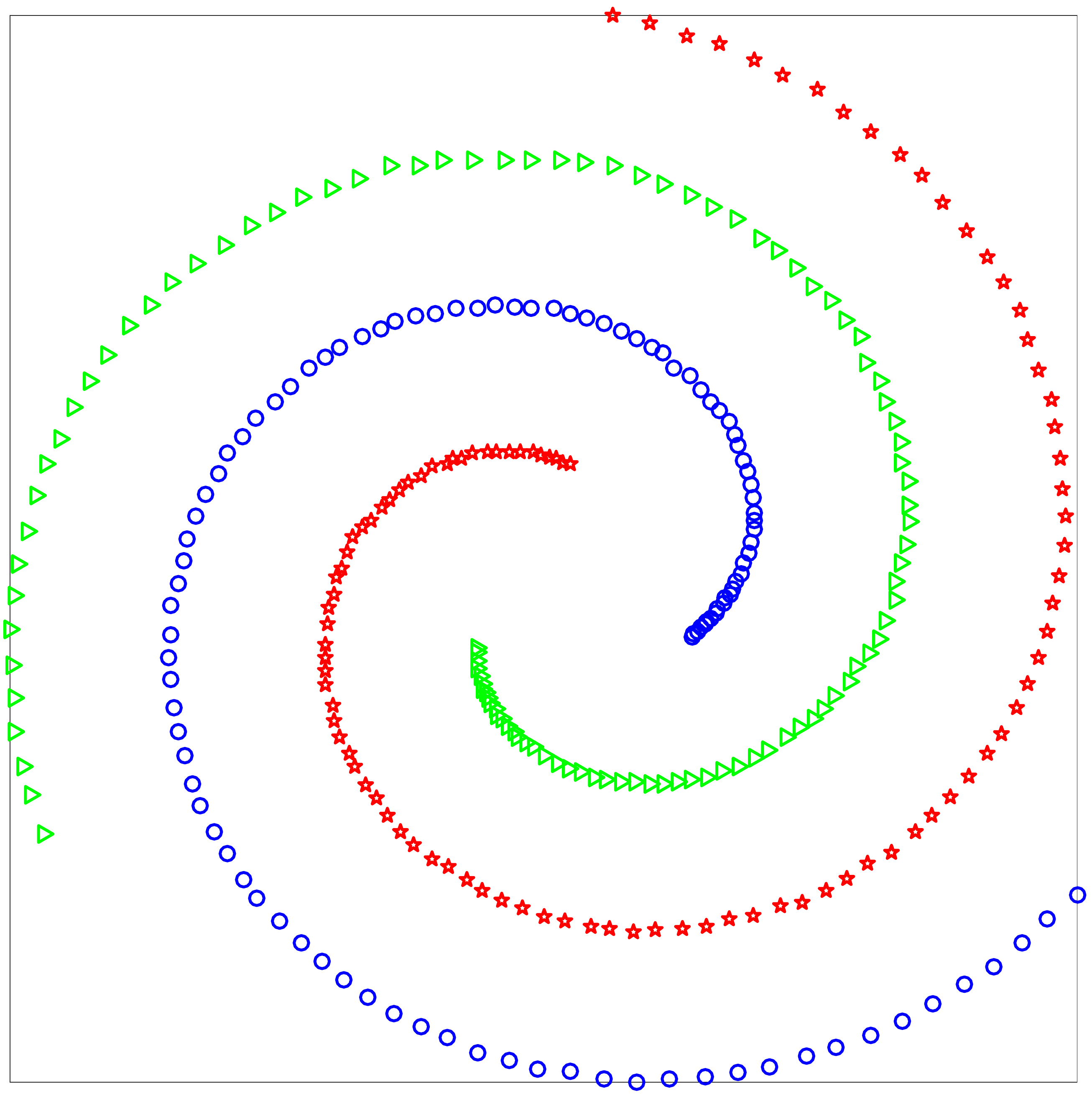}
\label{Fig:datasetsU}}
\subfloat[S. Control]{\includegraphics[height=\y\textwidth]{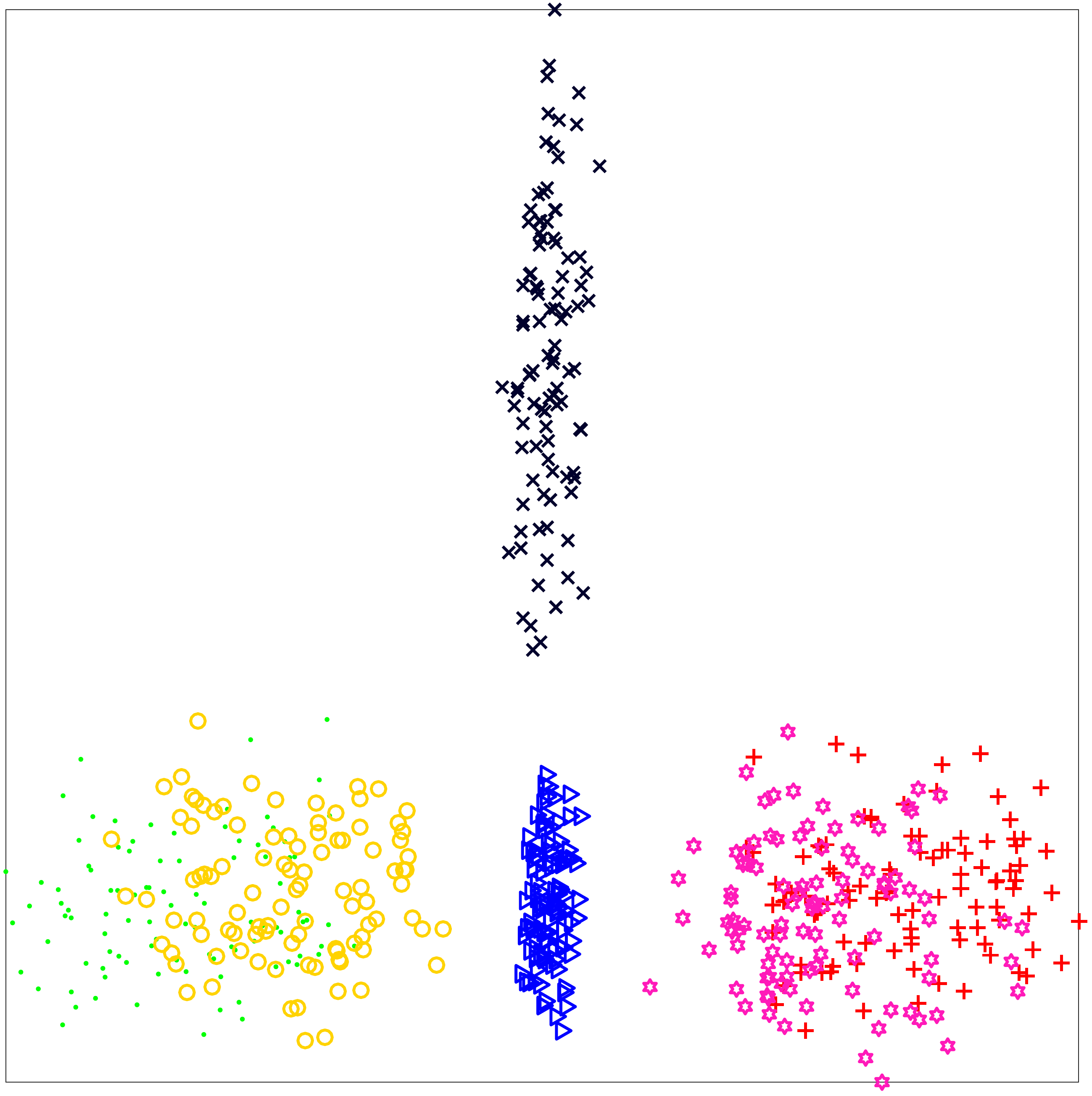}
\label{Fig:datasetsV}}
\subfloat[Target]{\includegraphics[height=\y\textwidth]{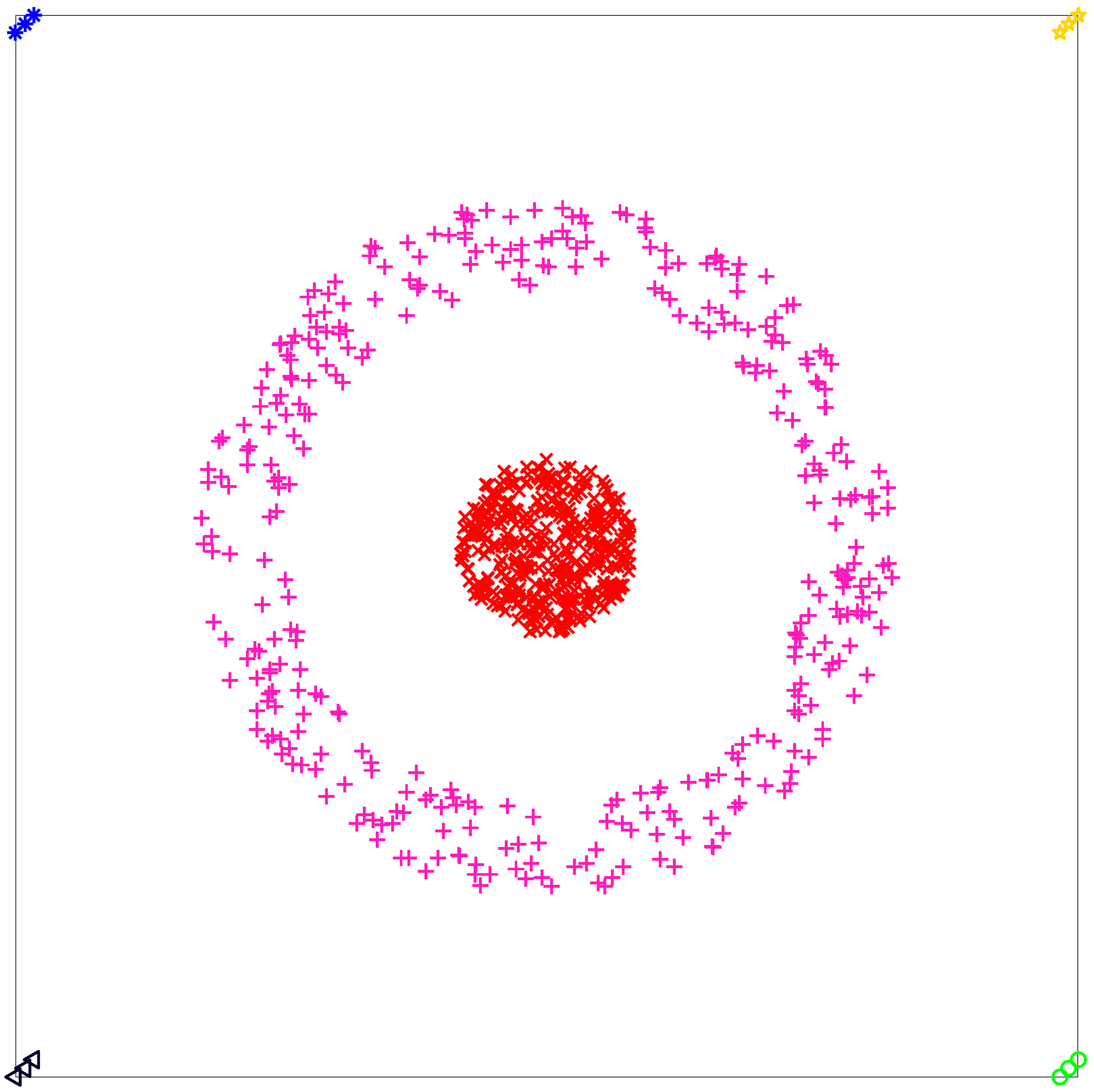}
\label{Fig:datasetsW}}
\subfloat[Tetra]{\includegraphics[height=\y\textwidth]{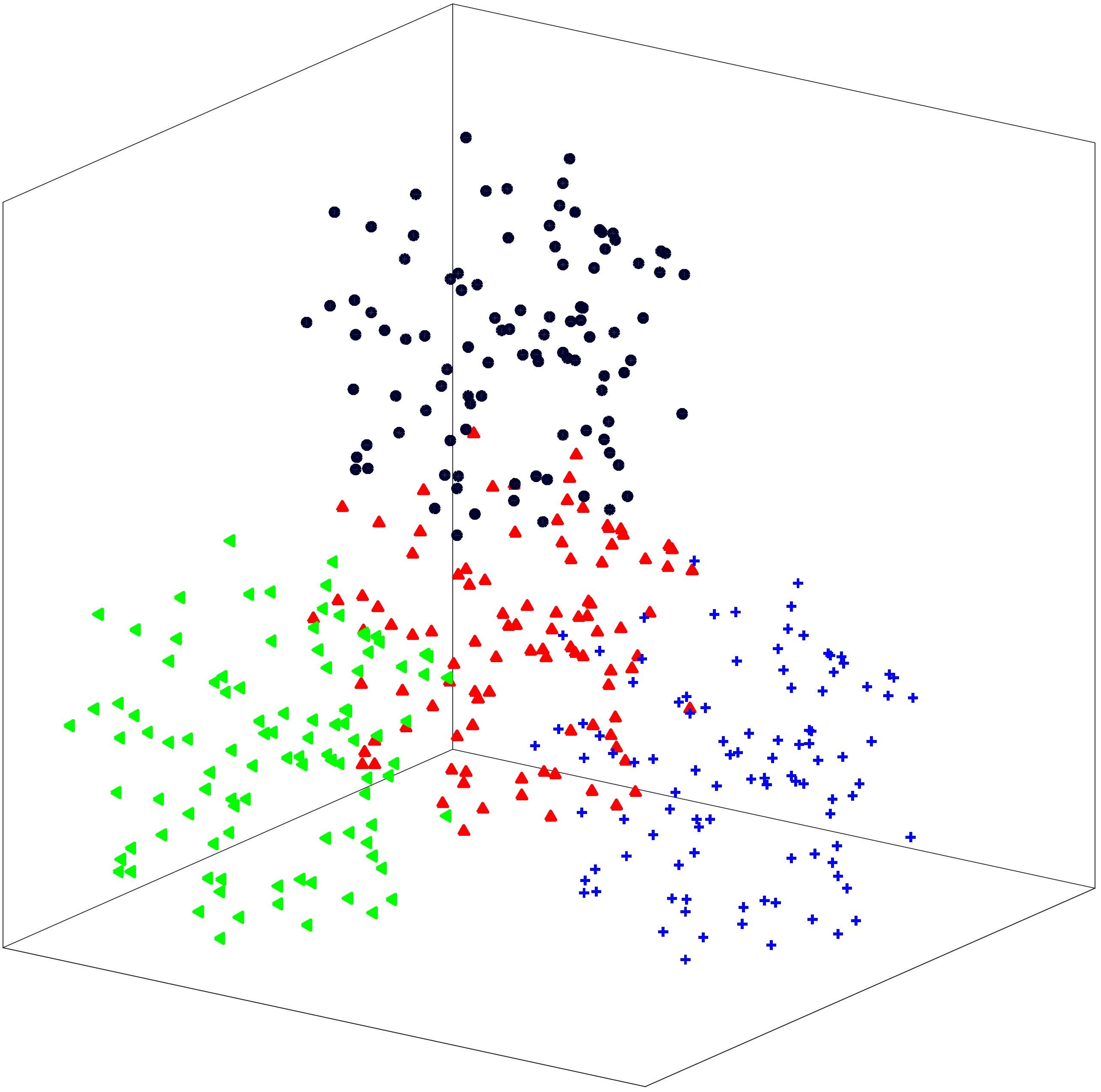}
\label{Fig:datasetsX}}
\subfloat[Twodiam.]{\includegraphics[height=\y\textwidth]{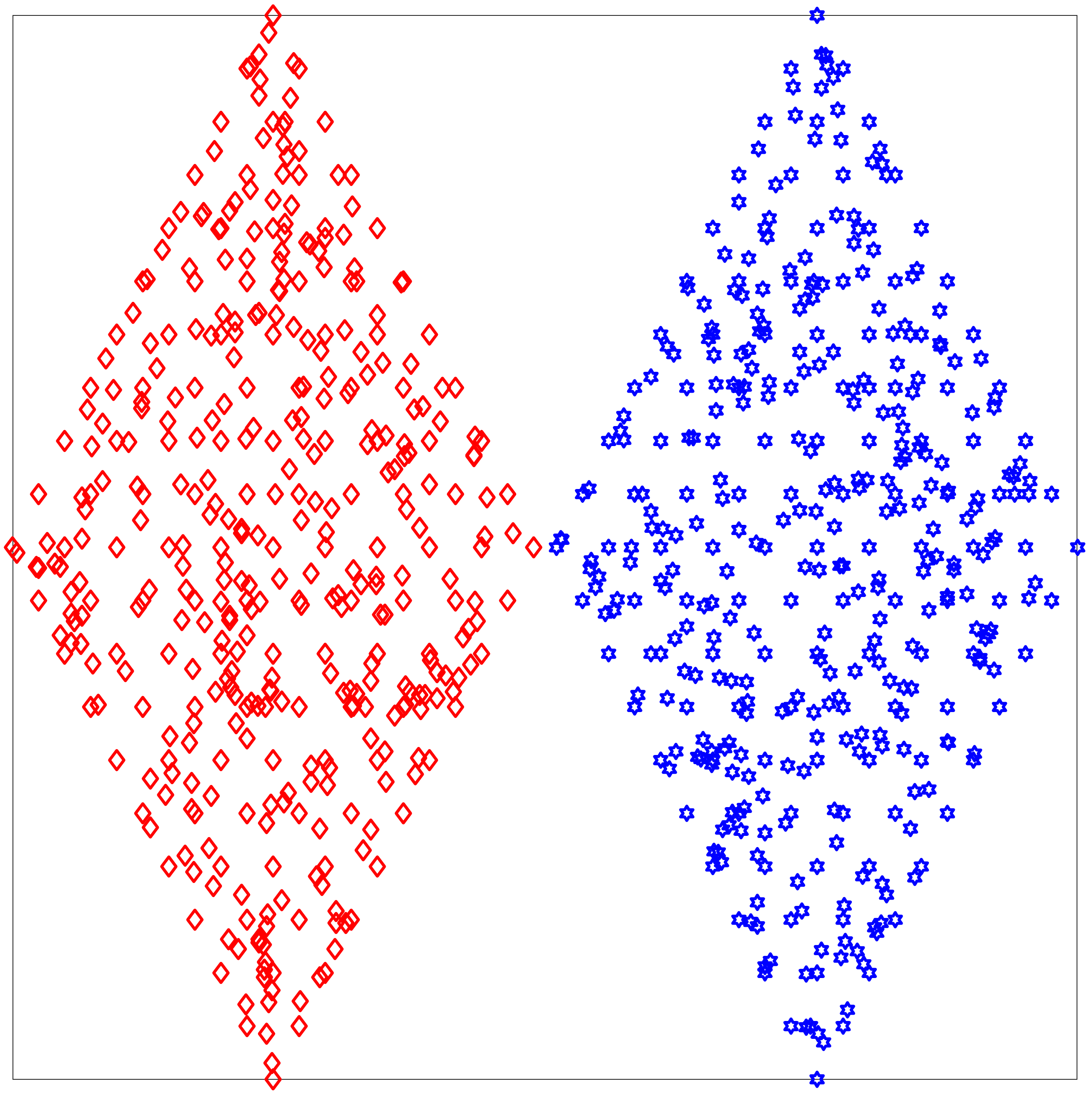}
\label{Fig:datasetsY}}
}
\vspace{-0.5\baselineskip}
\centerline{
\subfloat[Wave]{\includegraphics[height=\y\textwidth]{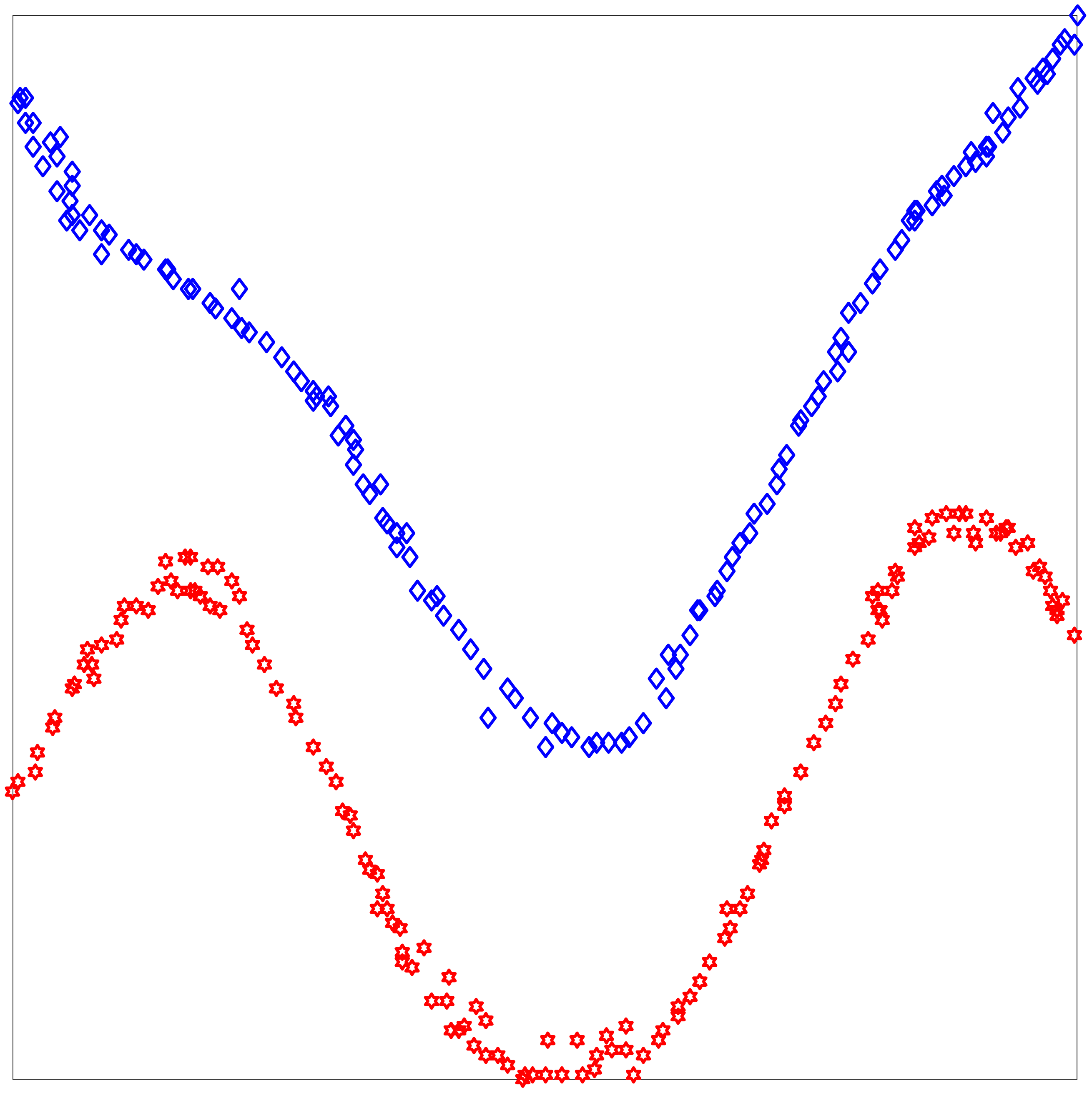}
\label{Fig:datasetsZ}}
\subfloat[Wine]{\includegraphics[height=\y\textwidth]{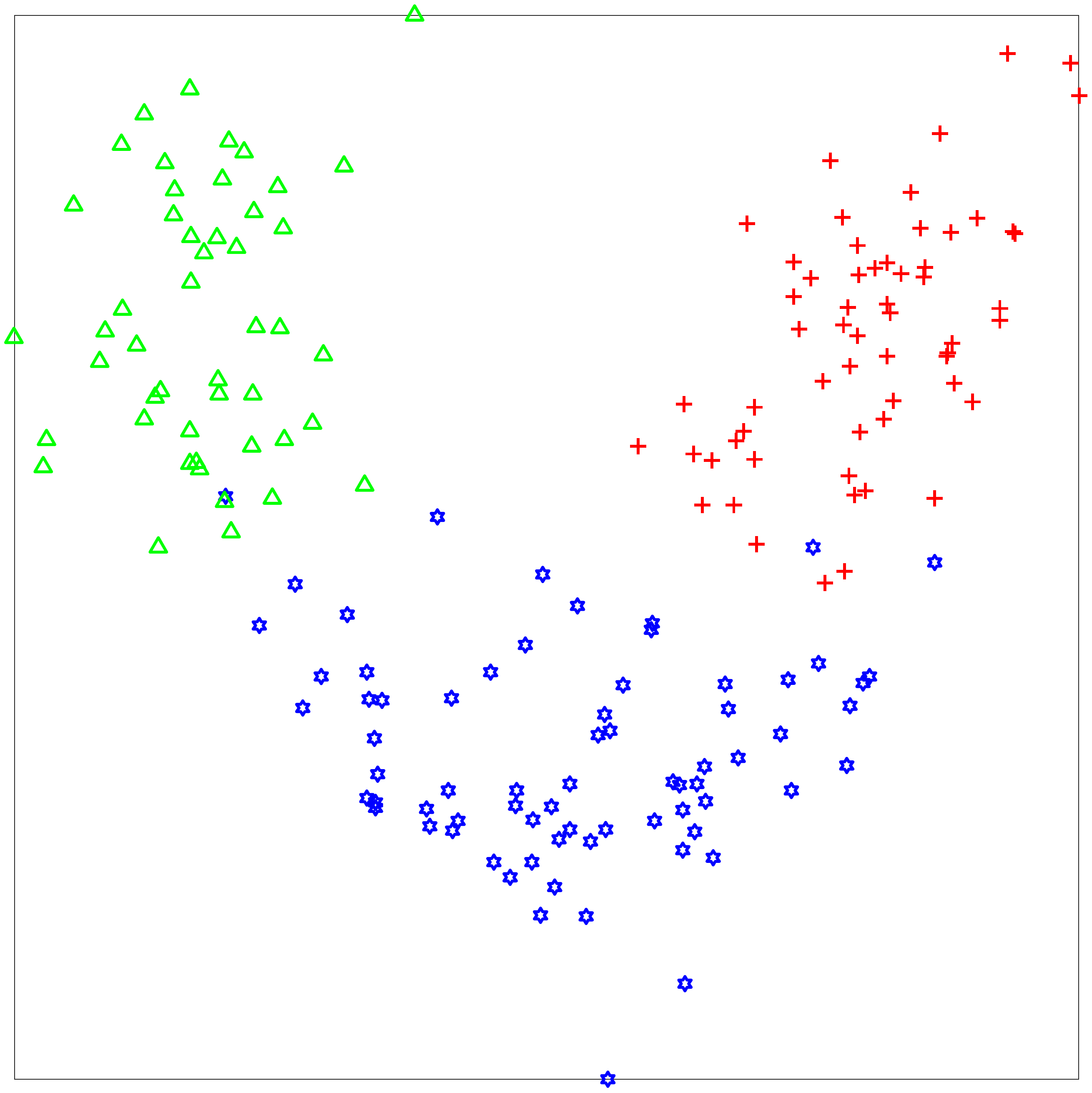}
\label{Fig:datasetsAA}}
\subfloat[Wingnut]{\includegraphics[height=\y\textwidth]{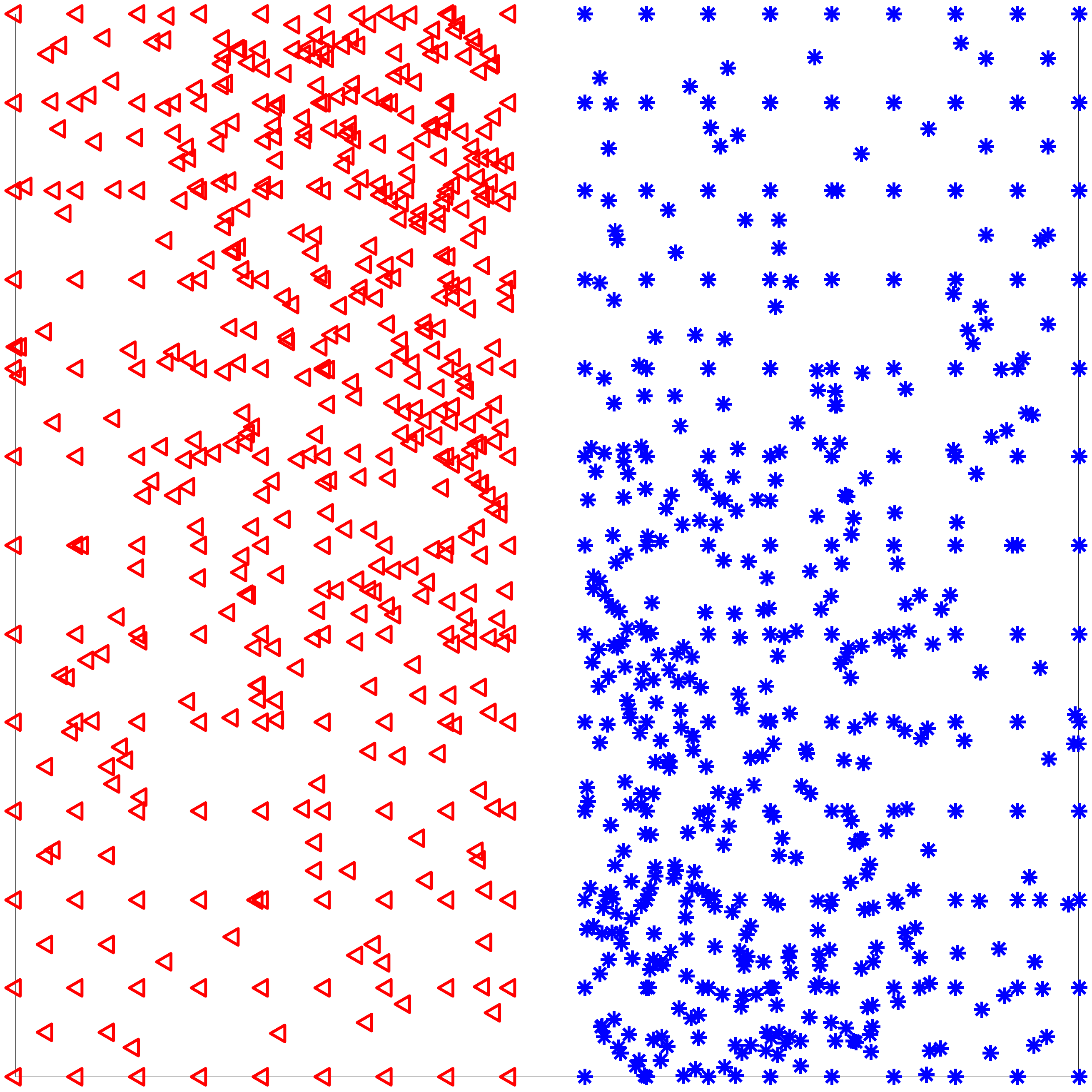}
\label{Fig:datasetsAB}}
\subfloat[Wisconsin ]{\includegraphics[height=\y\textwidth]{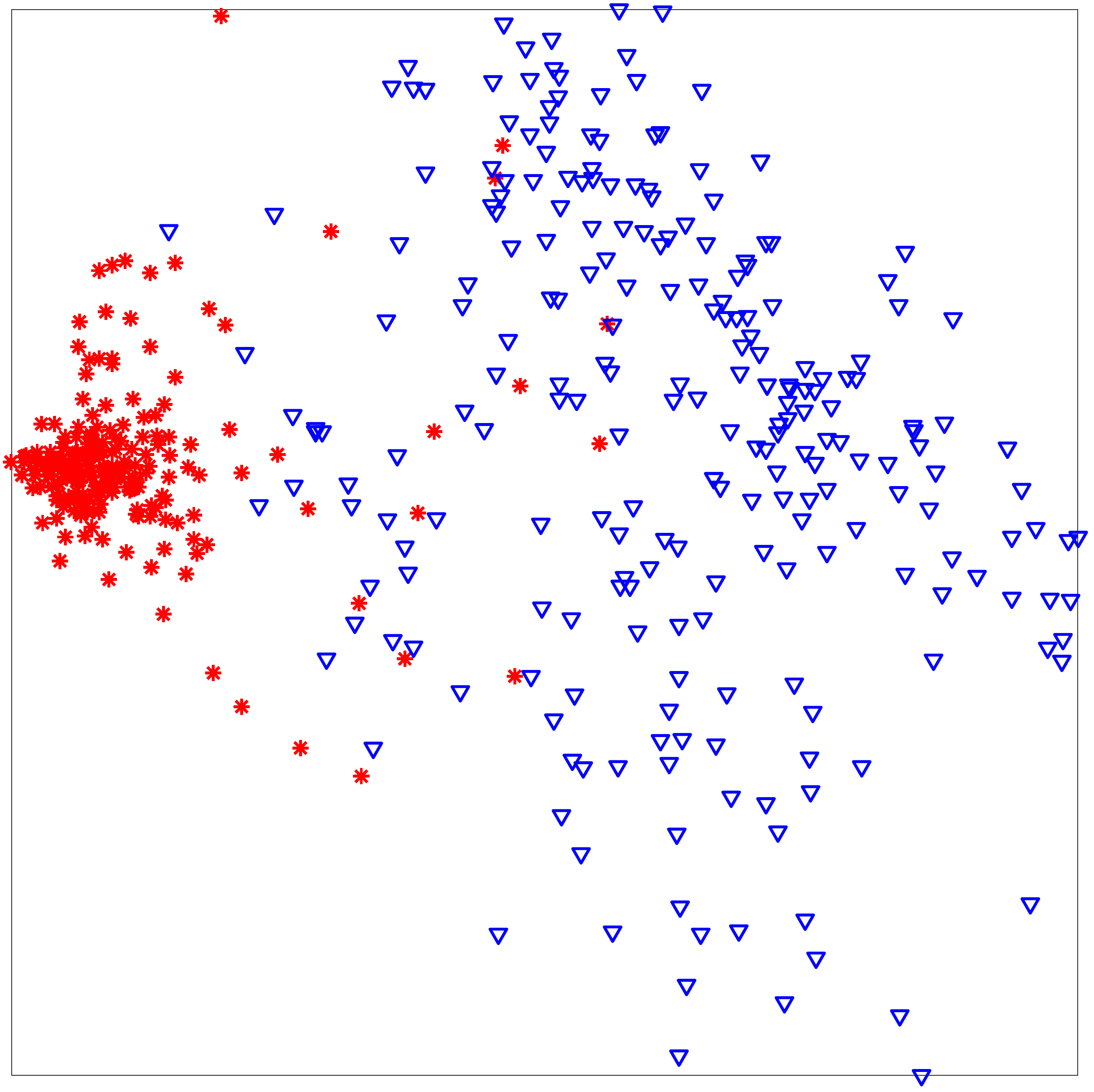}
\label{Fig:datasetsAC}}
\subfloat[WDBC]{\includegraphics[height=\y\textwidth]{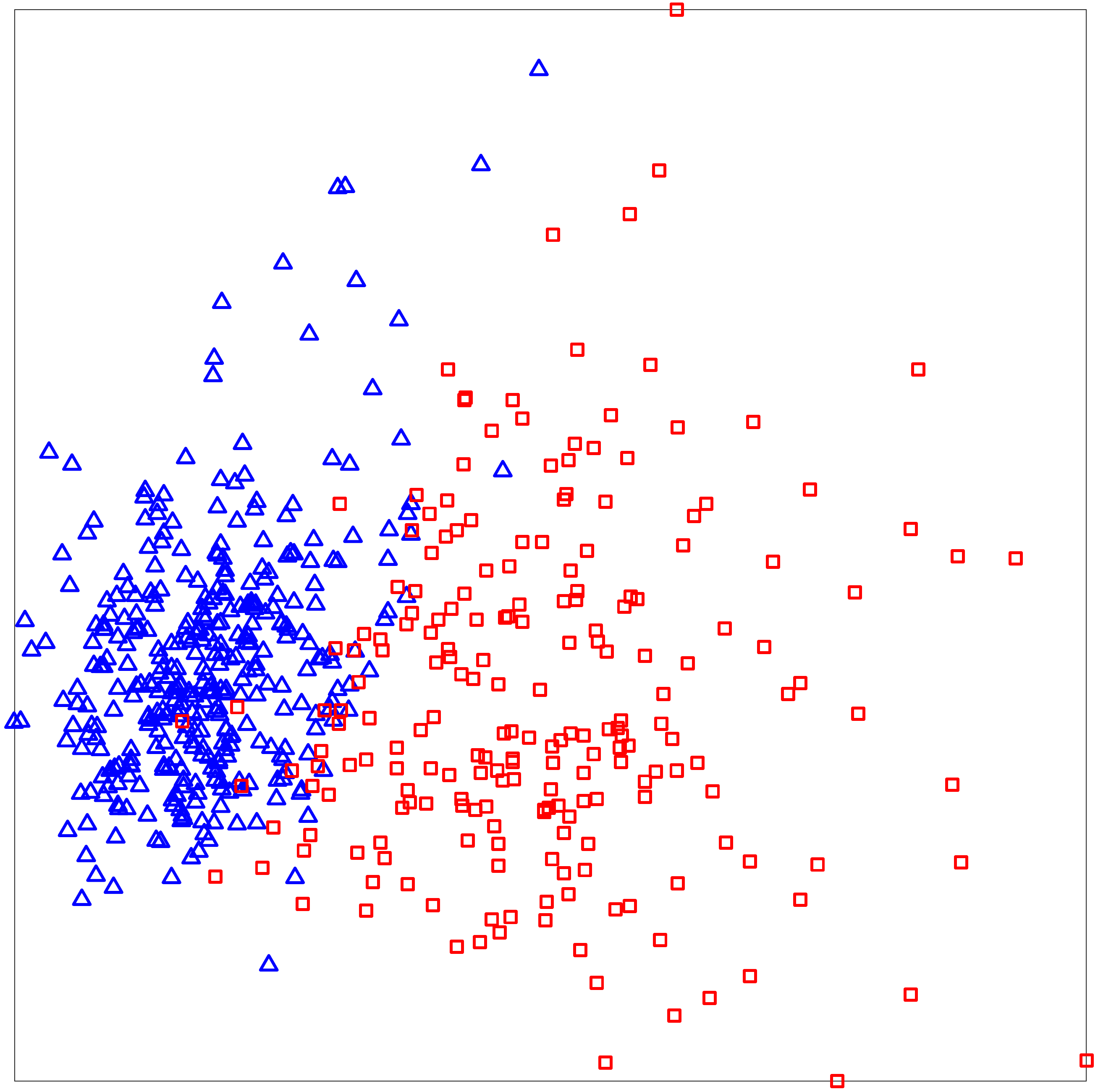}
\label{Fig:datasetsAD}}
}
\caption{Data sets used in the experiments. Solely for visualization purposes, the data sets \textit{Iris}, \textit{Wine}, \textit{Seeds}, \textit{WDBC}, \textit{Synthetic Control}, \textit{Glass}, and \textit{Ecoli} are depicted using principal component analysis projection. The data sets' features and projections are scaled to the range $[0,1]$.}
\label{Fig:datasets}
\end{figure}

\begin{table}[!t]\centering
\caption{Summary of the data sets' characteristics.}
\resizebox{.91\columnwidth}{!}{
\begin{threeparttable}
\begingroup\setlength{\fboxsep}{0pt}
\colorbox{lightgray}{
\begin{tabular}{lllllll}
\toprule
Data set 	&	\# samples &	\# features	& \# clusters & type & reference(s)  \\
\midrule
\midrule
Aggregation & 788   & 2     & 7 & Artificial & \cite{Gionis2007,shape} \\
Atom  & 800   & 3     & 2 & Artificial & \cite{fcps}\\
Chainlink & 1000  & 3     & 2 & Artificial & \cite{fcps}\\
Compound & 399   & 2     & 6 & Artificial & \cite{zahn1971,shape}\\
Dermatology & 358   & 34    & 6 & Real World & \cite{uci}\\
Ecoli & 336   & 7     & 8 & Real World & \cite{uci}\\
Face  & 320   & 2     & 4 & Artificial & \cite{datapkg,Ilc2011} \\
Flag  & 640   & 2     & 3 & Artificial & \cite{datapkg,Ilc2011,Ilc2012}\\
Flame & 240   & 2     & 2 & Artificial & \cite{fu2007,shape}\\
Giant & 862   & 2     & 2 & Artificial & \cite{datapkg,Ilc2011,Ilc2012}\\
Glass & 214   & 10    & 6 & Real World & \cite{uci}\\
Hepta & 212   & 3     & 7 & Artificial & \cite{fcps} \\
Iris  & 150   & 4     & 3 & Real World & \cite{fisher1936,uci}\\
Jain  & 373   & 2     & 2 & Artificial & \cite{shape,jain2005}\\
Lsun  & 400   & 2     & 3 & Artificial & \cite{fcps} \\
Moon  & 514   & 2     & 4 & Artificial & \cite{datapkg,Ilc2011,Ilc2012}\\
Path based & 300   & 2     & 3 & Artificial & \cite{chang2008,shape}\\
R15   & 600   & 2     & 15 & Artificial & \cite{veenman2002,shape}\\
Ring  & 800   & 2     & 2 & Artificial & \cite{datapkg,Ilc2011,Ilc2012}\\
Seeds\tnote{a} & 210   & 7     & 3 & Real World & \cite{charytanowicz2010,uci} \\
Spiral & 312   & 2     & 3 & Artificial & \cite{chang2008,shape}\\
Synthetic Control\tnote{b} & 600   & 60    & 6 & Real World & \cite{uci}\\
Target & 770   & 2     & 6 & Artificial & \cite{fcps}\\
Tetra & 400   & 3     & 4 & Artificial & \cite{fcps}\\
Two Diamonds & 800   & 2     & 2 & Artificial & \cite{fcps}\\
Wave  & 287   & 2     & 2 & Artificial & \cite{datapkg,Ilc2011,Ilc2012}\\
Wine  & 178   & 13    & 3 & Real World & \cite{uci}\\
Wingnut & 1016  & 2     & 2 & Artificial & \cite{fcps}\\
Wisconsin & 683   & 9     & 2 & Real World & \cite{uci}\\
WDBC\tnote{c} & 569   & 30    & 2 & Real World & \cite{uci}  \\
\bottomrule
\end{tabular}
}\endgroup
\begin{tablenotes}[normal,flushleft]
\item[a]The contributors gratefully acknowledge support of their work by the Institute of Agrophysics of the Polish Academy of Sciences in Lublin.
\item[b]Image courtesy of Eamonn Keogh.
\item[c]Wisconsin Diagnostic Breast Cancer.
\end{tablenotes}
\end{threeparttable}
}
\label{Tab:datasets}
\end{table}

\subsection{Clustering algorithms and parameter tuning} \label{Subsec:tuning}

To set the parameters of the clustering algorithms employed in the experiments, grid searches were performed through their parameter spaces. For all algorithms, the best solution was selected according to the parameter combination that yielded the peak average performance. 

\subsubsection{ART-based clustering methods} Fuzzy ART, fuzzy topoART, and DVFA were compared to DDVFA. In the experiments performed, fuzzy ART's, DVFA's and DDVFA's vigilance parameters were scanned in the range $[0,1]$ with identical step sizes equal to $0.01$ (DVFA's and DDVFA's vigilances were also subjected to the constraint $\rho^{UB} \ge \rho^{LB}$). For all fuzzy ART modules, the maximum number of epochs was set to $1$ (online mode),  the choice parameter~($\alpha$) was set to~$0.001$, and the learning rate~($\beta$) was set to 1 (fast learning).
DDVFA's parameters $\gamma^{*}$ and $\gamma$ were set to $1$ and $3$, respectively; and, for simplicity, $\rho^{(1)}_{UB}=\rho^{(2)}_{UB}$ and $\rho^{(1)}_{LB}=\rho^{(2)}_{LB}$. Moreover, in all the fuzzy ART implementations, no uncommitted category participated in the winner-take-all competitive process. If none of the current committed categories satisfy the vigilance criteria, then a new one is created and set to the current sample (fast commit). Regarding topoART, the parameters $\rho_a$, $\beta_{sbm}$, $\phi$ and $\tau$ were scanned in the ranges $[0,1]$ with a step size of $0.008$, $[0,0.75]$ with a step size of $0.25$, $[1,4]$ with a step size of $1$, and $[10\%,30\%]$ of the data cardinality with a step size of $10\%$, respectively. These ranges and step sizes generated approximately the same number of parameter combinations for topoART, DVFA, and DDVFA. Module B's clusters were taken as topoART's output. Finally, for all these methods, $30$ runs were performed for each data set in both random and VAT ordered presentation scenarios.

\subsubsection{Non-ART-based clustering methods} DBSCAN~\cite{Ester1996}, affinity propagation (AP)~\cite{Frey2007}, k-means~\cite{kmeans}, and single linkage (SL-HAC)~\cite{xu2009} were compared to DDVFA. In the experiments performed, DBSCAN's~$MinPts$ parameter was varied in the range $[1,4]$ with a step size of $1$, while~$eps$ was scanned in the range $[0,\sqrt{d}]$ with a step size of $0.005$, where $d$ is the dimensionality of the data (thus encompassing the full range of possible distance values in the $d$-dimensional unit cube). The number of clusters~$k$ in k-means was varied in the range $\left[1, \ceil*{\sqrt{N}}\right]$, where $N$ is the cardinality of the data set (this upper bound is usually taken as a rule of thumb~\cite{Bezdek1995,pbm}). Additionally, k-means was repeated $10$ times, and the best solution, according to the cost function being minimized, was selected for each value of $k$. The AP's damping factor~$\lambda$ was varied in the range $[0.5,1]$ with a step size of $0.005$, and the preference parameter was set as the median of the data samples' similarities. SL-HAC used Euclidean distance, and its dendrogram was cut at all merging levels. Finally, for all these methods, a single run was performed for each randomized data set, since they are global approaches that are either not (or almost not) order dependent.

\subsection{Clustering performance assessment}

The adjusted rand index ($AR$)~\cite{hubert1985} is an external cluster validity index commonly used in the unsupervised learning literature to measure the level of agreement between a data sets' reference partition (i.e., ground truth structure) and a discovered partition~\cite{xu2009}. It was used in this work to evaluate the quality of the solutions returned by all clustering algorithms. The ($AR$) is defined as:

\begin{equation}
AR = \frac{{N \choose 2}(tp + tn)-\left[(tp+fp)(tp+fn)+(fn+tn)(fp+tn)\right]}{{N \choose 2}^2-\left[(tp+fp)(tp+fn)+(fn+tn)(fp+tn)\right]},
\label{Eq:ari}
\end{equation}

\noindent where $tp$, $tn$, $fp$ and $fn$ stand for true positive, true negative, false positive, and false negative, respectively.

\subsection{Statistical analysis methodology}
The clustering algorithms were compared following the procedures discussed in~\cite{Demsar2006}:
\begin{enumerate}
    \item The quantities of interest (i.e., performance in terms of AR and network compactness) were tested for equality using Iman-Davenport's correction~\cite{Iman1980} of Friedman's non-parametric rank sum test~\cite{friedman1937,friedman1940}.
    \item If there was sufficient evidence to reject the null hypothesis, then a critical difference (CD) diagram~\cite{Demsar2006} was generated using Nemenyi's post-hoc test~\cite{Nemenyi1963}.
\end{enumerate}

\subsection{Software and code}

The experiments were conducted using \mbox{MATLAB}, scikit-learn~\cite{scikit-learn}, Orange~\cite{Demsar2013}, and Cluster Validity Analysis Platform~\cite{cvap}. The MATLAB code for fuzzy ART, DVFA, and DDVFA is available at the Applied Computational Intelligence Laboratory group GitHub repositories\footnote[1]{https://github.com/ACIL-Group/DVFA.}\textsuperscript{,}\footnote[2]{https://github.com/ACIL-Group/DDVFA.}. The topoART experiments were carried out using Lib\-TopoART\footnote[3]{LibTopoART v0.74, available at https://www.libtopoart.eu.}~\cite{Tscherepanow2010}, whereas the other clustering algorithms' implementations were from scikit-learn\footnote[4]{http://scikit-learn.org/}.

\section{Results and discussion} \label{Sec:Results}

\subsection{DDVFA results with pre- and post-processing} \label{Sec:ResultsA}

This study investigates DDVFA's order of presentation dependency by analyzing two frameworks: an offline approach that consists of pre-ordering the shuffled samples using VAT~\cite{bezdek2002}, as per~\cite{leonardo2018}, and an online approach in which the samples are solely randomized prior to presentation. The latter is a more realistic scenario when an online incremental learner is required, i.e., a learning system is confronted with a data stream. That is why all the experiments were conducted with one epoch (single pass), so each data sample is only presented once. 

\begin{figure}[!b]
\centerline{
\includegraphics[width=.5\textwidth]{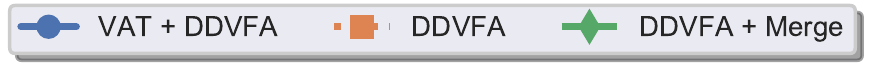}
}
\centerline{
\subfloat[average]{\includegraphics[width=.32\textwidth]{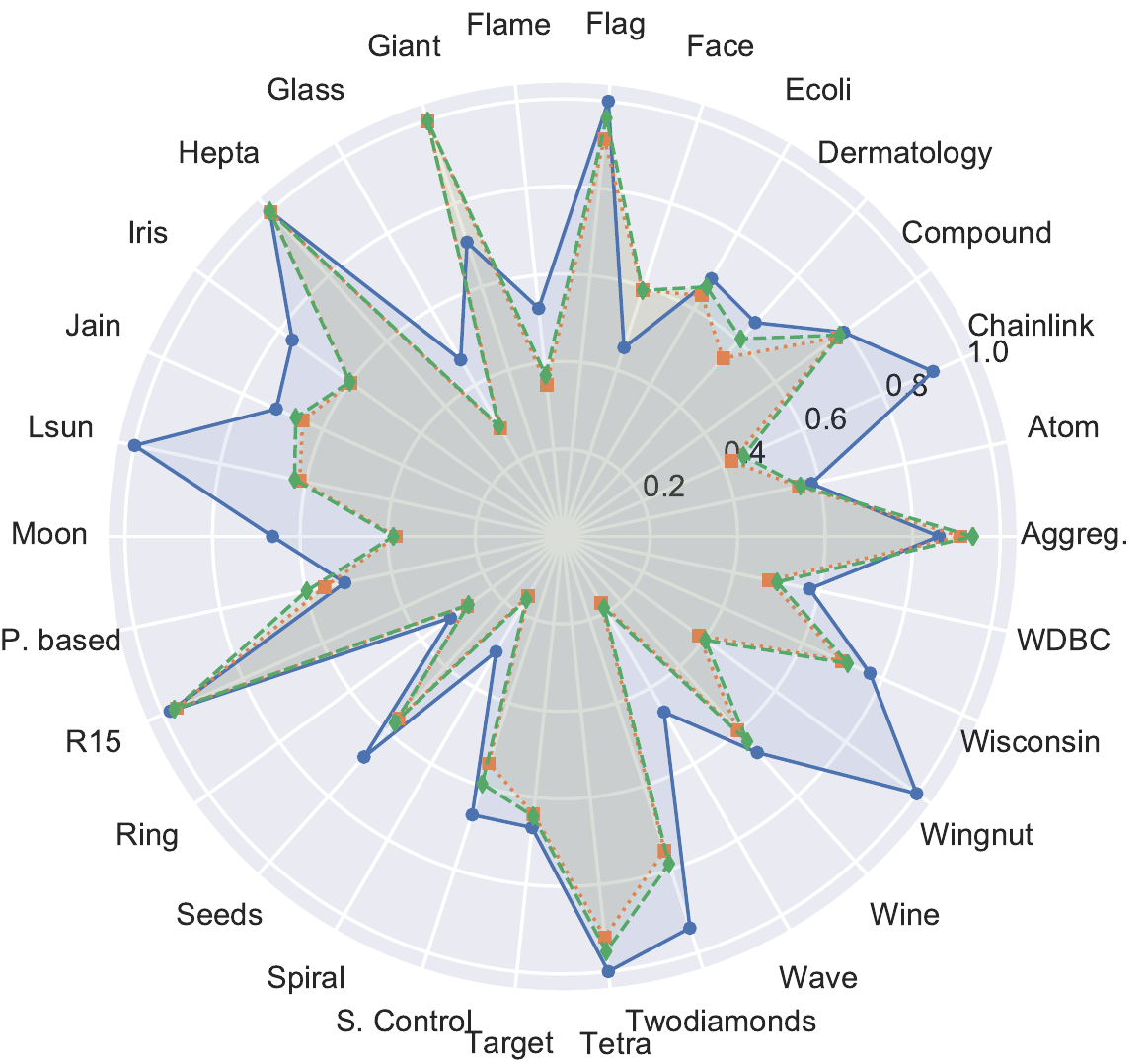}
\label{Fig:average1}} 
\hfil
\subfloat[centroid]{\includegraphics[width=.32\textwidth]{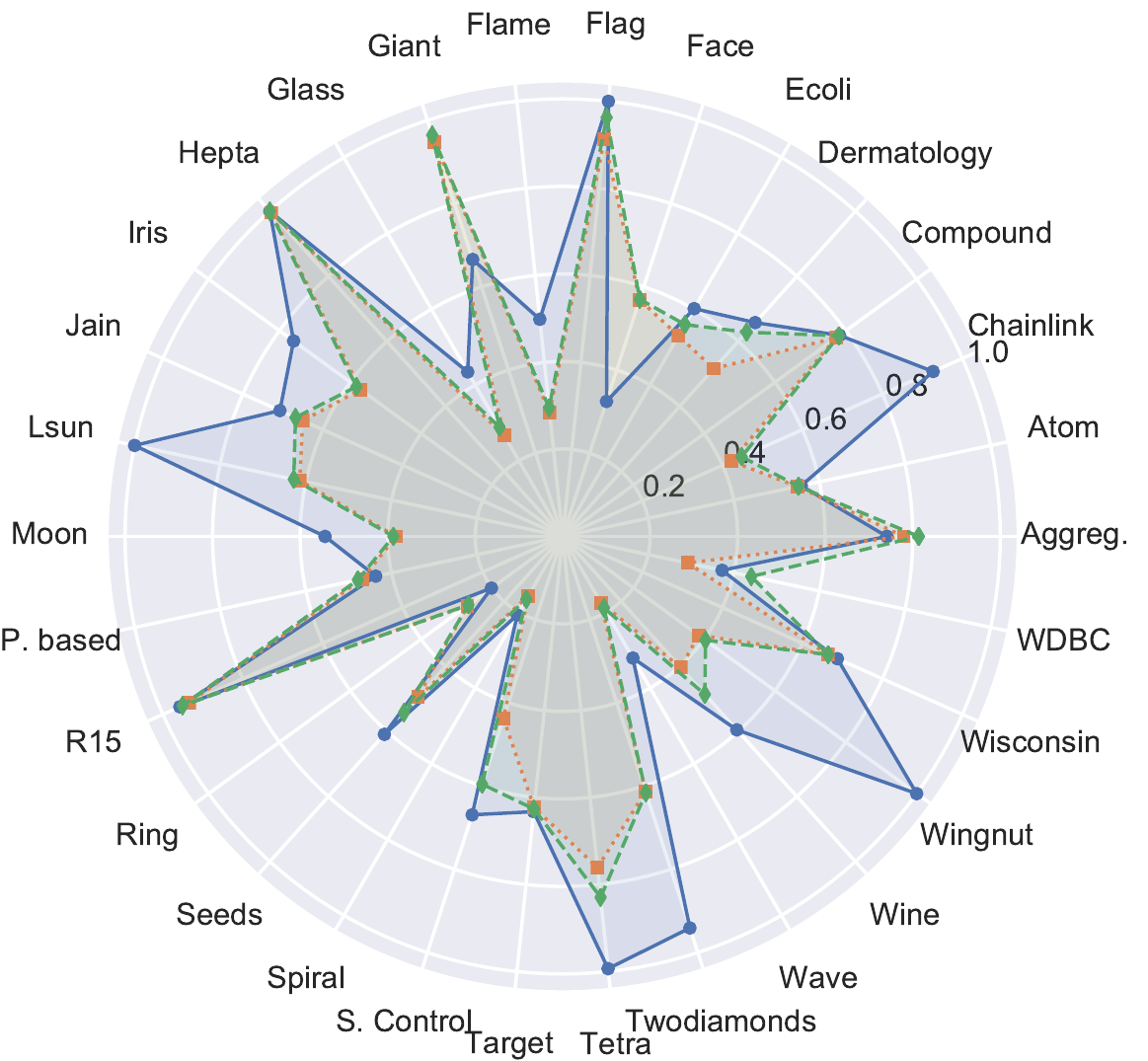}
\label{Fig:centroid1}} 
\hfil
\subfloat[complete]{\includegraphics[width=.32\textwidth]{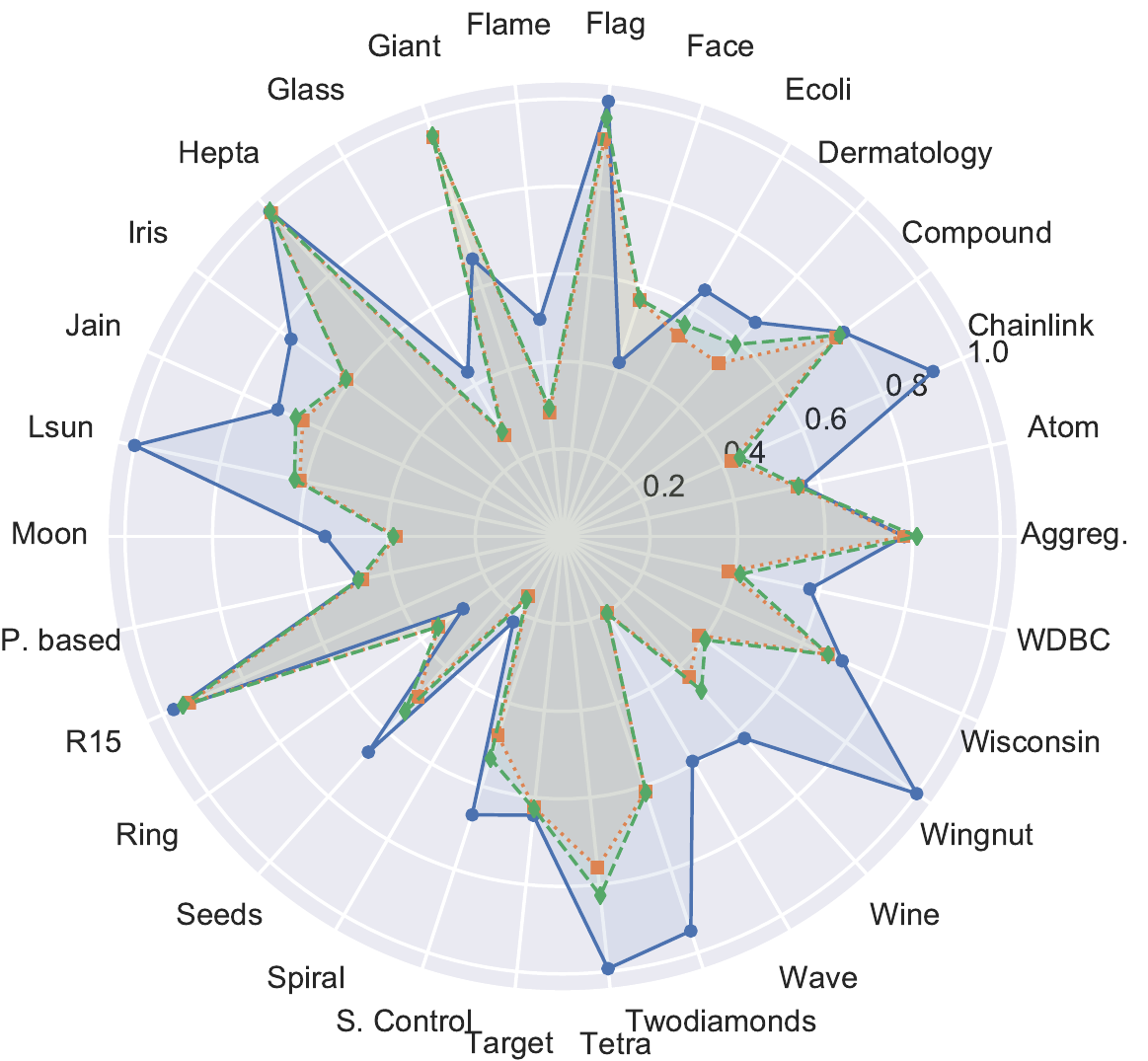}
\label{Fig:complete1}}
}
\centerline{
\subfloat[median]{\includegraphics[width=.32\textwidth]{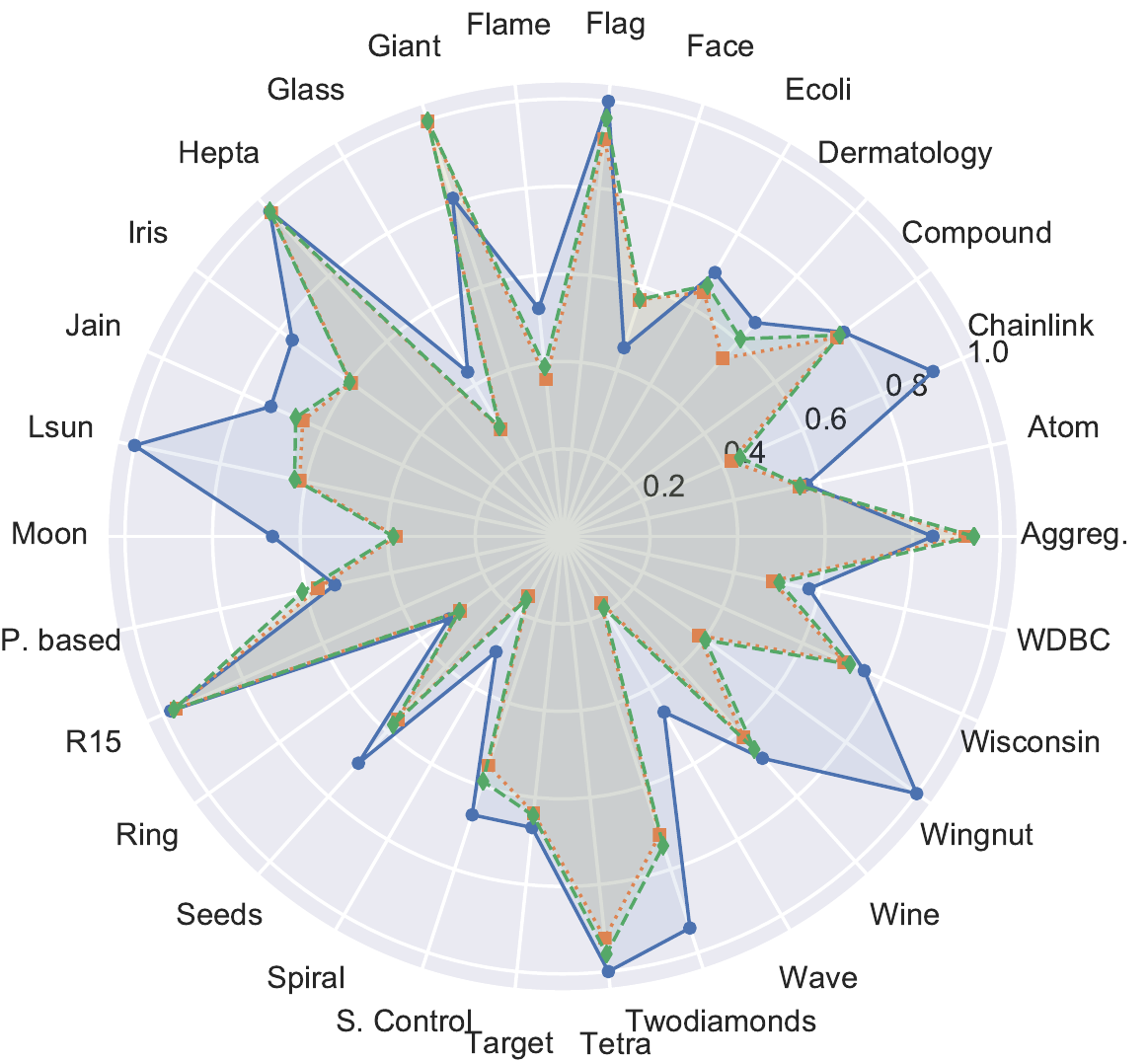}
\label{Fig:median1}} \\
\hfil
\subfloat[single]{\includegraphics[width=.32\textwidth]{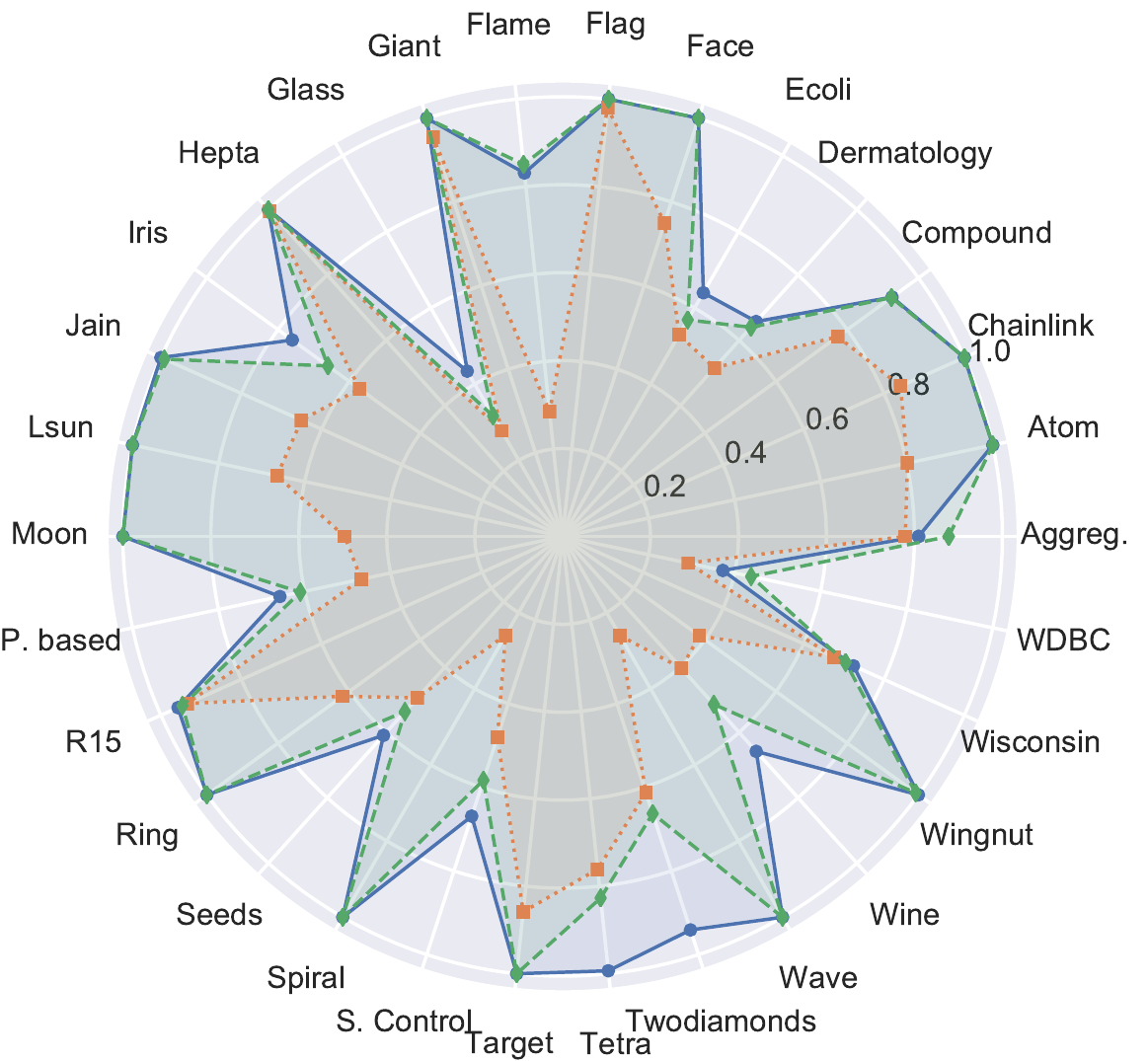}
\label{Fig:single1}}
\hfil
\subfloat[weighted]{\includegraphics[width=.32\textwidth]{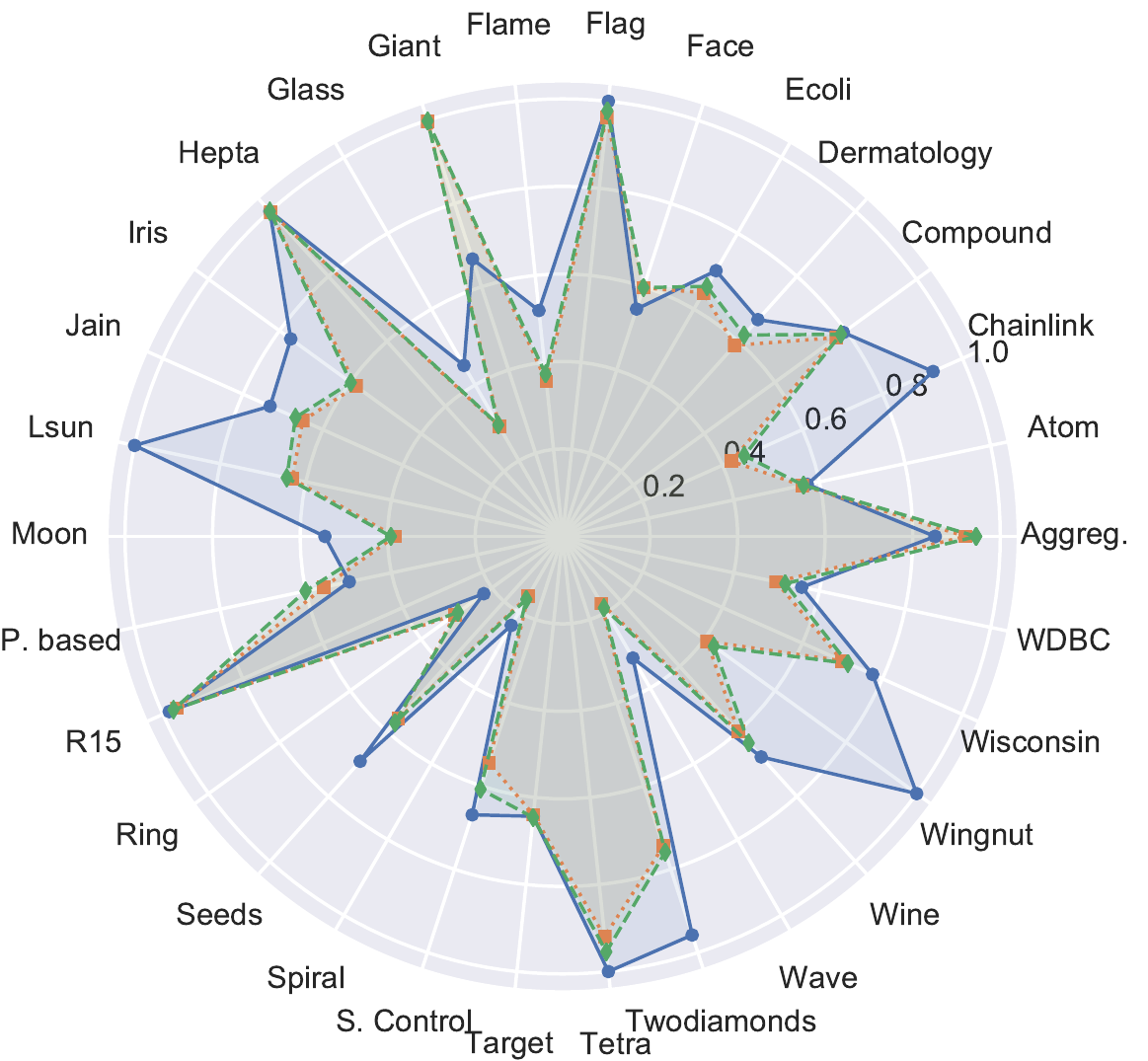}
\label{Fig:weighted1}}
}
\caption{Radar charts of the peak average performances ($AR$) of all three different DDVFA systems, which are grouped by the type of activation/match functions (a)-(e). The results are based on $30$ runs per data set using $\gamma^{*}=1$ and $\gamma=3$. Typically, VAT pre-ordering yielded the best performance, while DDVFA and DDVFA + Merge ART appear to yield a similar performance, with the exception of the single-linkage-based DDVFA, in which using Merge ART makes a noticeable difference when compared to DDVFA by itself.}
\label{Fig:DDpART2}
\end{figure}

Employing the methodology described in subsection~\ref{Subsec:tuning}, the experiments were performed with the following three systems: (1) DDVFA, (2) VAT + DDVFA, and (3) DDVFA + Merge ART. The results are summarized in Fig.~\ref{Fig:DDpART2}, which depicts radar charts of the peak average performance of all the mentioned systems grouped by the type of HAC-based activation/match functions 
(i.e., per Tables~\ref{tab:DDpART_T_M} and \ref{tab:MergeART_T_M}'s method): (\ref{Fig:average1}) average, (\ref{Fig:centroid1}) centroid, (\ref{Fig:complete1}) complete, (\ref{Fig:median1}) median, (\ref{Fig:single1}) single, and (\ref{Fig:weighted1}) weighted. It shows that, in general, VAT pre-ordering yields a better performance than pure DDVFA or post-processing with Merge ART. The latter approaches yielded a similar performance across all types of activation/match functions, except for the single-linkage based DDVFA, in which using Merge ART makes a significant difference compared to DDVFA by itself. For instance, Fig.~\ref{Fig:Merge_examples} illustrates the outputs of DDVFA before and after cascading it with Merge ART for the \textit{Spiral}, \textit{Face}, \textit{Atom} and \textit{Chainlink} data sets.

\newcommand{\egMerge}{0.24}
\begin{figure}[!b]
\centerline{
\subfloat[]{\includegraphics[width=\egMerge\columnwidth]{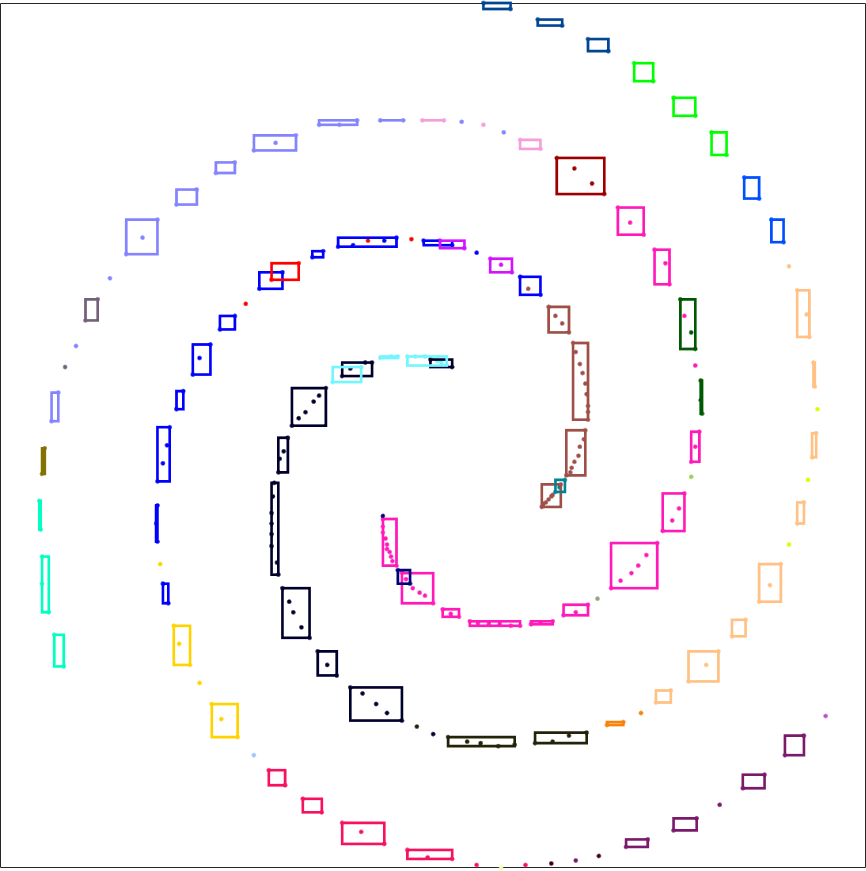}}
\hfil
\subfloat[]{\includegraphics[width=\egMerge\columnwidth]{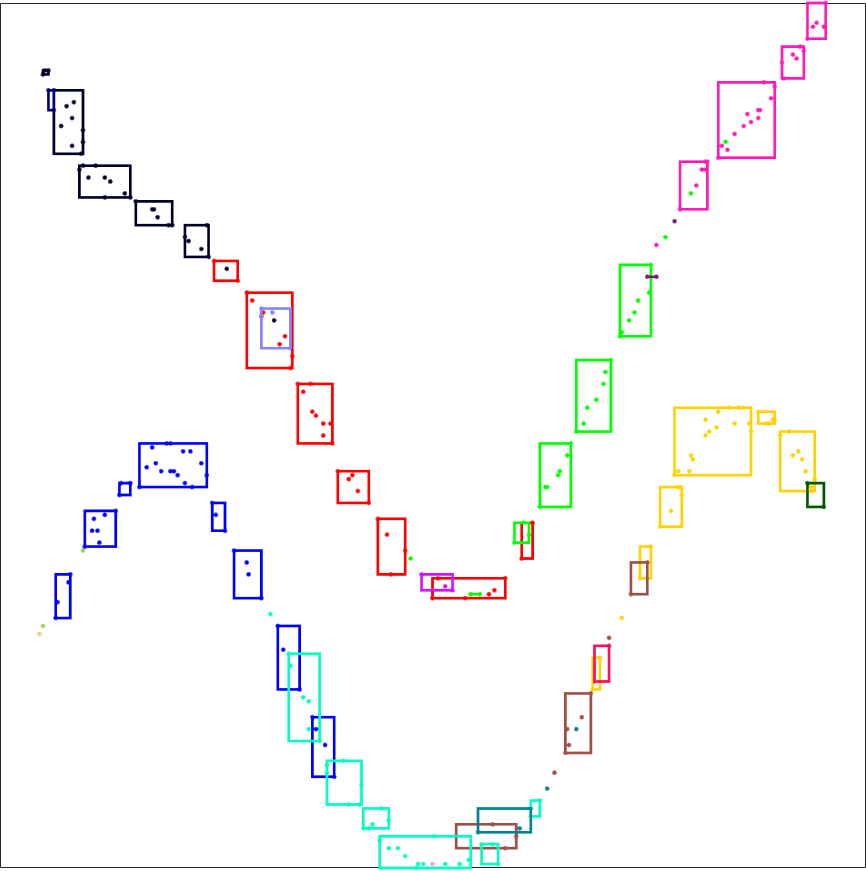}}
\hfil
\subfloat[]{\includegraphics[width=\egMerge\columnwidth]{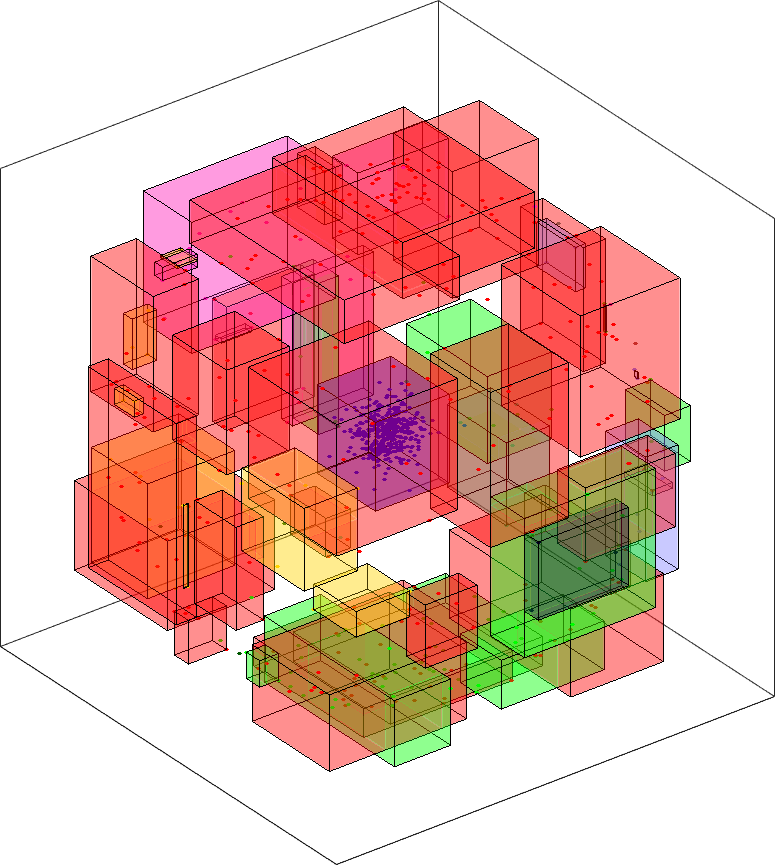}}
\hfil
\subfloat[]{\includegraphics[width=\egMerge\columnwidth]{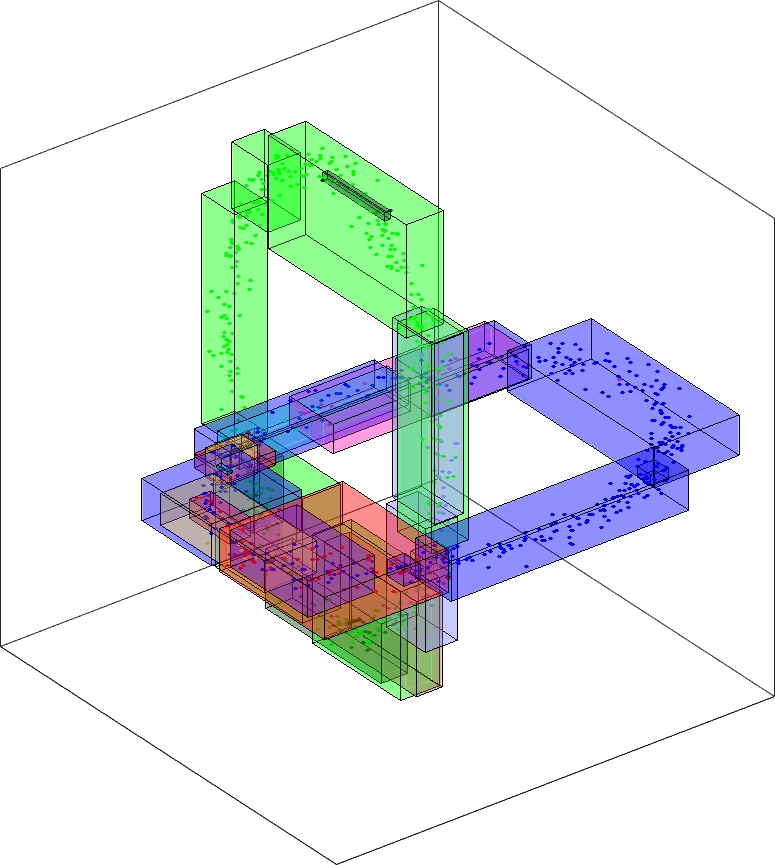}}
}
\centerline{
\subfloat[]{\includegraphics[width=\egMerge\columnwidth]{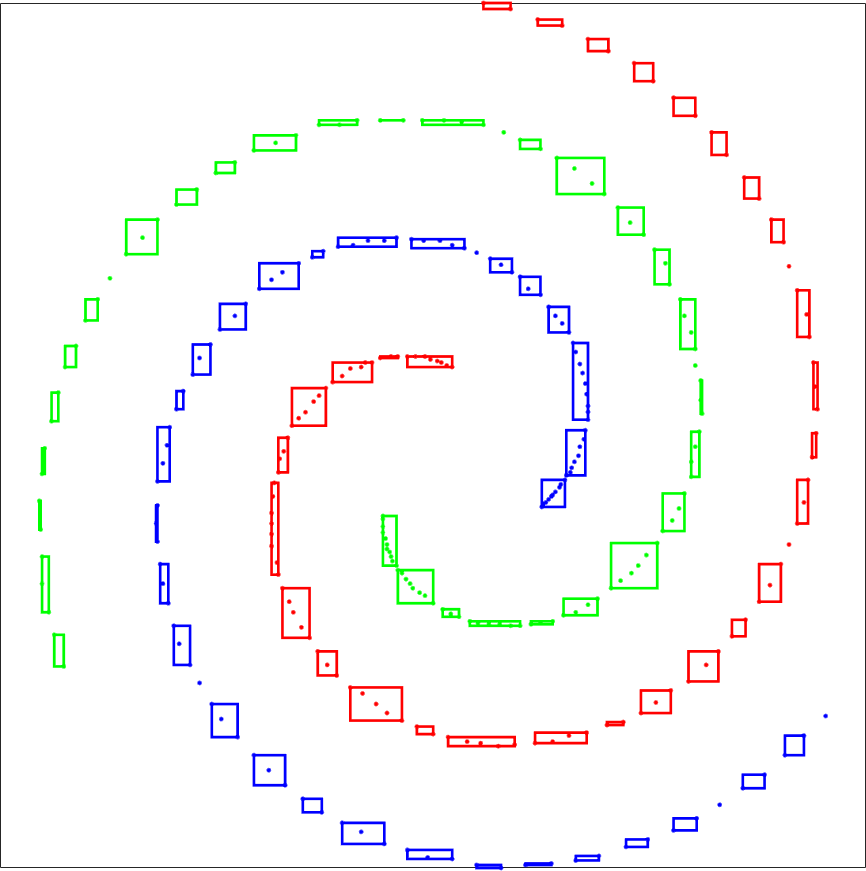}}
\hfil
\subfloat[]{\includegraphics[width=\egMerge\columnwidth]{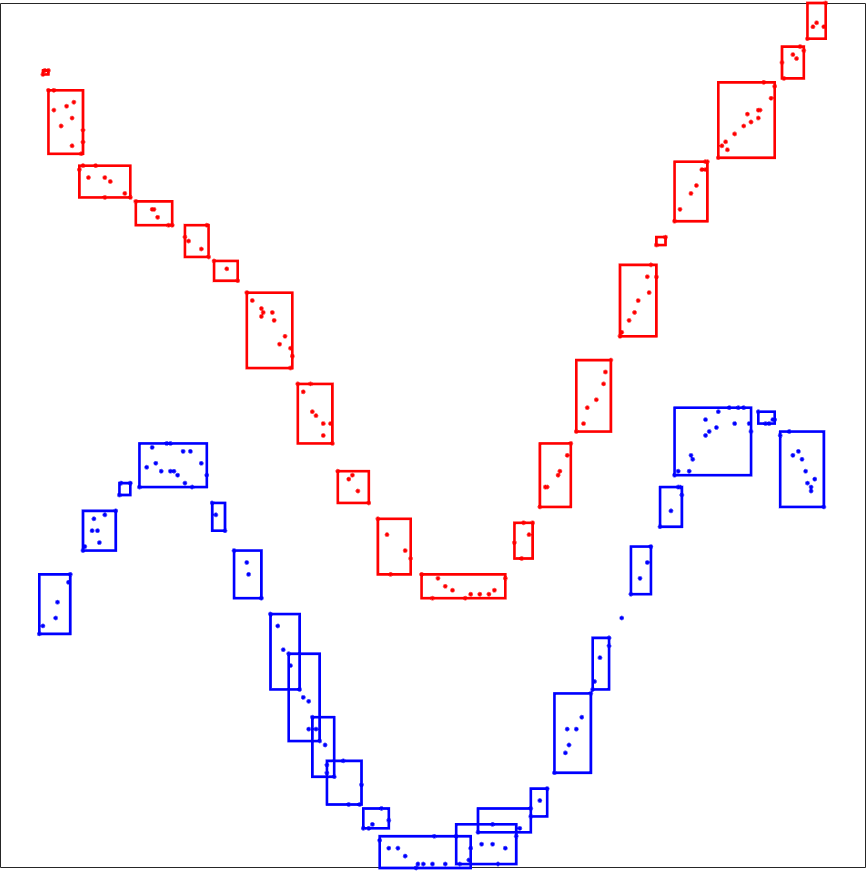}}
\hfil
\subfloat[]{\includegraphics[width=\egMerge\columnwidth]{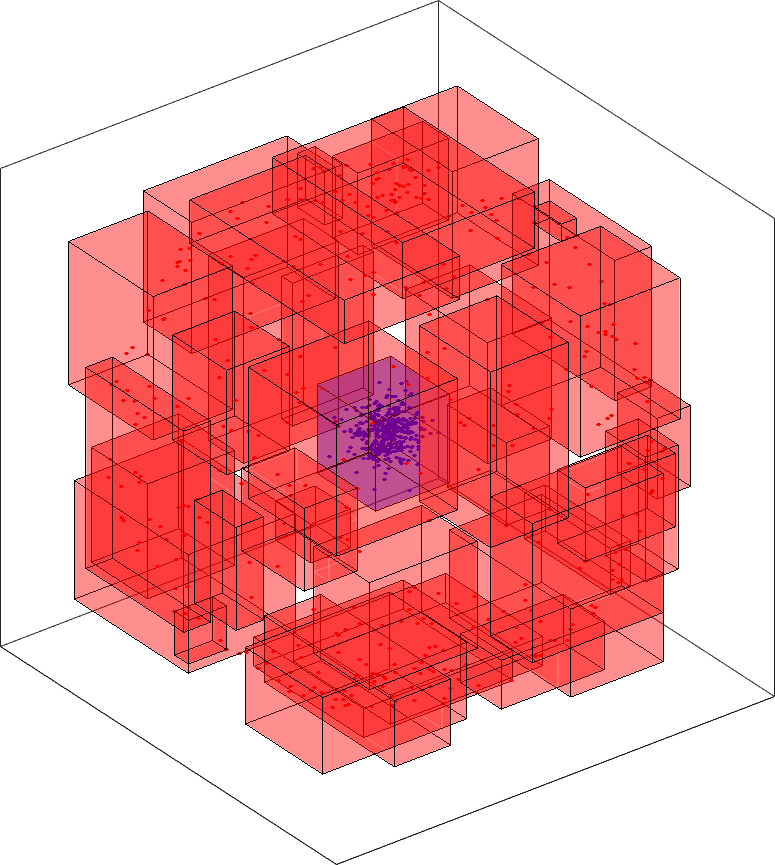}}
\hfil
\subfloat[]{\includegraphics[width=\egMerge\columnwidth]{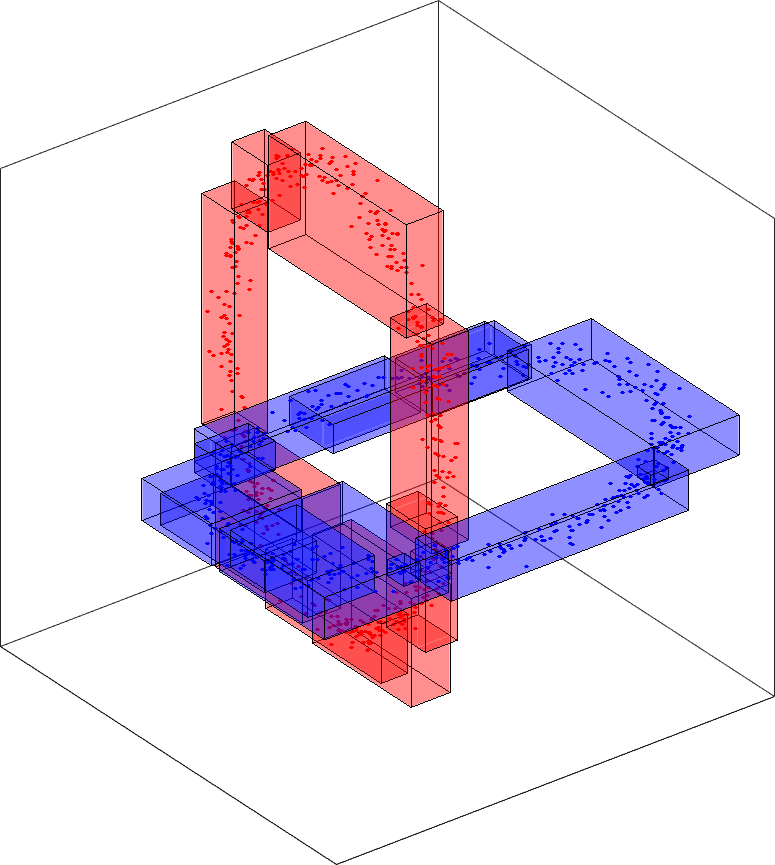}}
}
\caption{Output partitions of the DDVFA system (a)-(d) before, and (e)-(f) after cascading the Merge ART module for the (a,e) \textit{Spiral}, (b,f) \textit{Face}, (c,g) \textit{Atom}, and (d,h) \textit{Chainlink} data sets.}
\label{Fig:Merge_examples}
\end{figure}

\subsubsection{Statistical analysis of performance} 

Using the Iman-Davenport test, a statistical analysis was conducted to quantitatively assess if the performances of the different types of HAC-based activation/match functions (average vs. centroid vs. complete vs. median vs. single vs. weighted) were equivalent when fixing the type of DDVFA system. All these performance equivalency hypotheses were rejected at a $0.05$ significance level (Table~\ref{Tab:pvalues1}). Therefore, Nemenyi's test was performed, and Fig.~\ref{Fig:CD_methods_1} depicts the resulting CD diagrams. They indicate that the best performing groups seem to be: (Fig.~\ref{Fig:systemA}) \{average, single, weighted, median\}, (Fig.~\ref{Fig:systemB}) \{weighted, median\}, and (Fig.~\ref{Fig:systemC}) \{single, weighted\}; and the worst performing groups seem to be: (Fig.~\ref{Fig:systemA}) \{centroid\}, (Fig.~\ref{Fig:systemB}) \{centroid, complete\}, and (Fig.~\ref{Fig:systemC}) \{centroid, complete\}, respectively. The fact that the best average rank for DDVFA is achieved by the weighted variant is expected since it considers additional information in the form of local prior probabilities.

\begin{table}[!b]
\centering
\caption{A statistical comparison of the different HAC activation/match functions' performances per DDVFA system: Friedman-Iman-Davenport p-values.}
\begin{threeparttable}
\begingroup\setlength{\fboxsep}{0pt}
\colorbox{lightgray}{
\begin{tabular*}{\columnwidth}{@{\extracolsep{\fill}}llll@{}}
\toprule
System & DDVFA  & VAT + DDVFA  & DDVFA + Merge ART \\
\midrule
\midrule
p-value\tnote{a} & 1.1056e-09 & 4.2657e-08 & 6.8745e-13 \\
\bottomrule
\end{tabular*}
}\endgroup
\begin{tablenotes}[normal,flushleft]
\item[a]Considering a given system, all HAC activation/match function types are statistically compared.
\end{tablenotes}
\end{threeparttable}
\label{Tab:pvalues1}
\end{table}

\newcommand{\mem}{1}
\begin{figure}[!ht]
\centerline{
\subfloat[VAT + DDVFA]{\includegraphics[width=\mem\columnwidth]{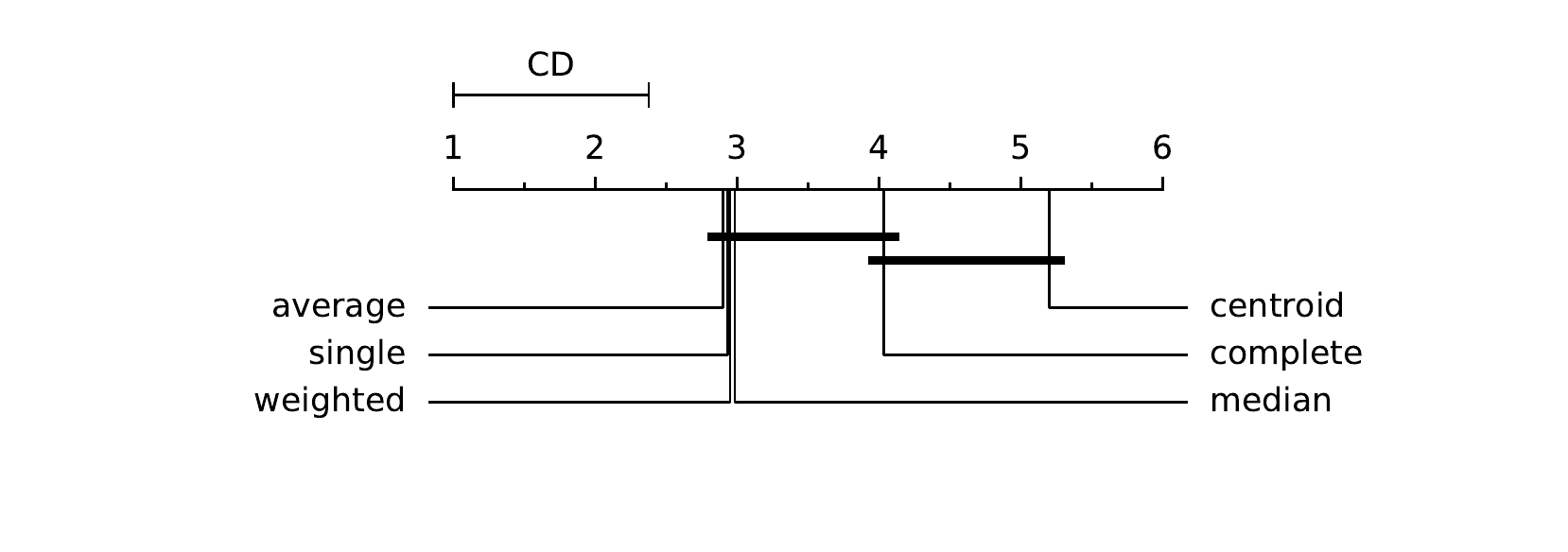} \label{Fig:systemA}}
}
\centerline{
\subfloat[DDVFA]{\includegraphics[width=\mem\columnwidth]{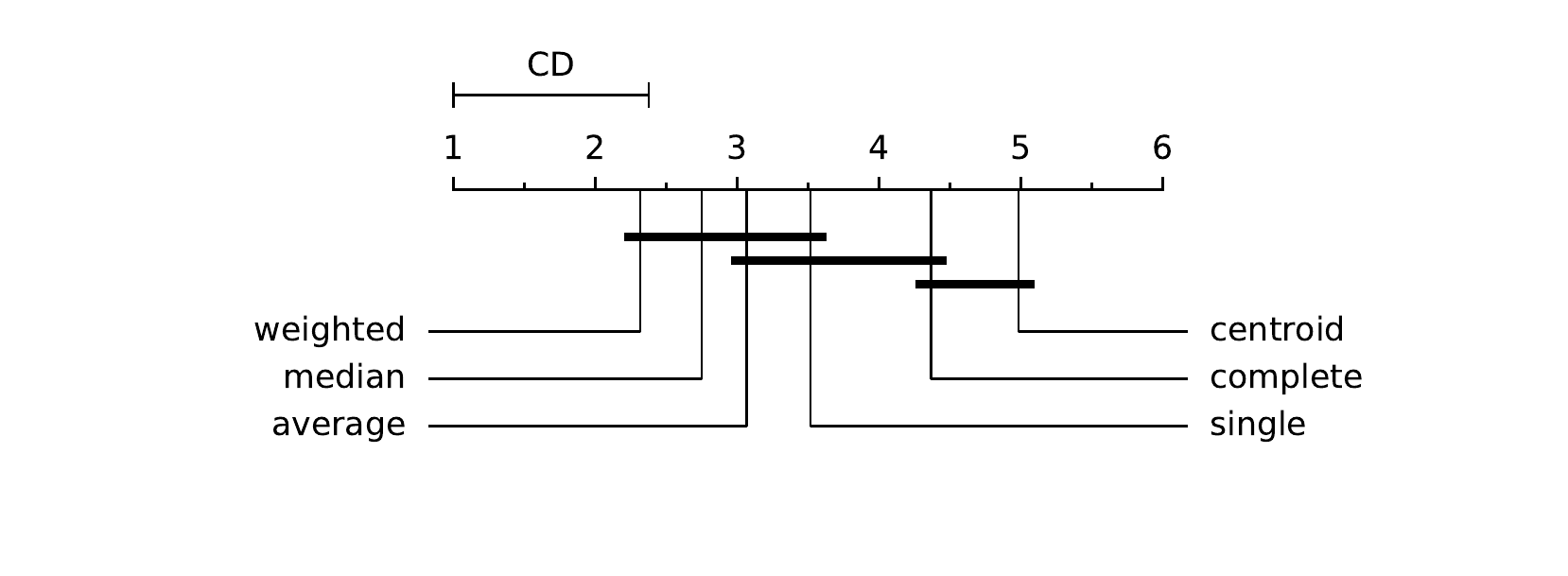} \label{Fig:systemB}}
}
\centerline{
\subfloat[DDVFA + Merge ART]{\includegraphics[width=\mem\columnwidth]{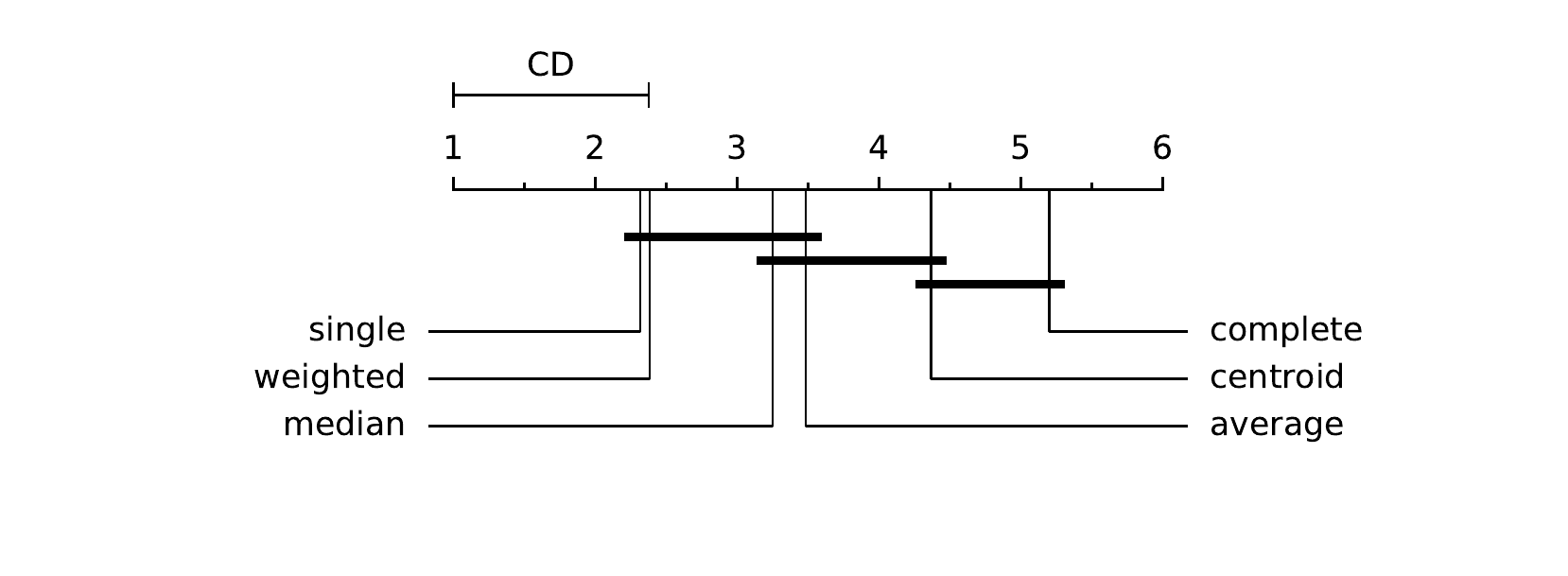} \label{Fig:systemC}}
}
\caption{CD diagrams for all three DDVFA systems considering all HAC-based distributed activation/match functions.}
\label{Fig:CD_methods_1}
\end{figure}

A similar statistical analysis was conducted to determine if the performances of the systems (DDVFA vs. VAT + DDVFA vs. DDVFA + Merge ART) were equivalent when fixing the type of activation/match functions. All these null hypotheses were rejected at a $0.05$ significance level (Table~\ref{Tab:pvalues2}). Therefore, Nemenyi's test was performed, and, for clarity, Fig.~\ref{Fig:CD_methods_2} solely depicts the resulting CD diagrams of selected HAC-based activation/match functions. Typically, pre-processing with VAT or post-processing with the Merge ART module are statistically equivalent, and, as expected, they are statistically better than just feeding the shuffled data directly to DDVFA.  

\begin{table}[!ht]
\centering
\caption{A statistical comparison of the different systems' performances per HAC activation/match function type: Friedman-Iman-Davenport p-values.}
\resizebox{\columnwidth}{!}{
\begin{threeparttable}
\begingroup\setlength{\fboxsep}{0pt}
\colorbox{lightgray}{
\begin{tabular}{lllllll}
\toprule
Method & average  & centroid  & complete  & median  & single  & weighted  \\
\midrule
\midrule
p-value\tnote{a} & 3.1048e-11 & 3.7364e-10 & 2.4092e-14 & 3.8147e-13 & 1.1102e-16 & 9.8684e-10 \\
\bottomrule
\end{tabular}
}\endgroup
\begin{tablenotes}[normal,flushleft]
\item[a]Considering a given activation/match function type, all three DDVFA systems are statistically compared.
\end{tablenotes}
\end{threeparttable}
}
\label{Tab:pvalues2}
\end{table}

\newcommand{\me}{1}
\begin{figure}[!ht]
\centerline{
\subfloat[Average]{\includegraphics[width=\me\columnwidth]{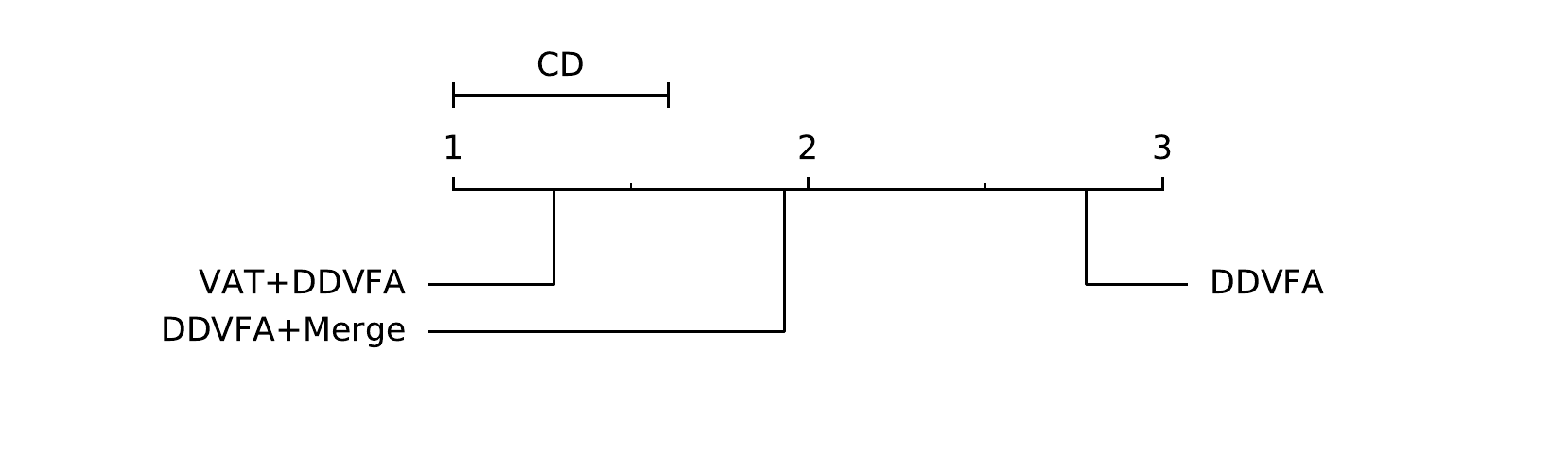} \label{Fig:Average}}
}
\centerline{
\subfloat[Single]{\includegraphics[width=\me\columnwidth]{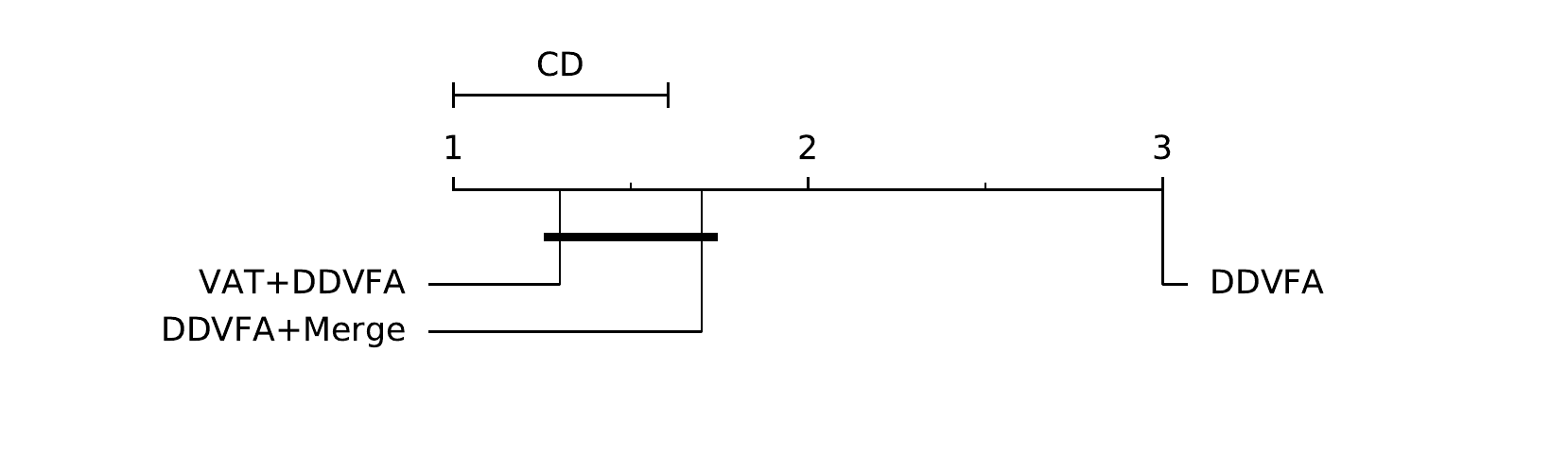} \label{Fig:Single}}
}
\caption{CD diagram for selected distributed HAC-based activation/match functions considering all three DDVFA systems. The CD diagram of the single variant is also representative for centroid, complete, median, and weighted.}
\label{Fig:CD_methods_2}
\end{figure}

\subsubsection{Summary}

The statistical analysis suggests that pre-processing with VAT or post-processing with Merge ART yields better results than just DDVFA. Furthermore, in general, single, median, average and weighted HAC-based activation/match functions appear to be statistically equivalent. Thus, the recommended systems are DDFVA + Merge ART for online learning mode and random presentation, and VAT + DDVFA for offline learning mode and applications where pre-ordering is feasible; for both of these systems the single-linkage variant is recommended since it appeared in the top 2 average rank for both learning modes.

\begin{sidewaystable}[!hp]
\centering
\caption{Experimental results summary: performance in terms of $AR$ ($mean \pm standard~deviation$).}
\resizebox{\textwidth}{!}{
\begin{threeparttable}
\begingroup\setlength{\fboxsep}{0pt}
\colorbox{lightgray}{
\begin{tabular}{l|llll|llll|llll}
\toprule
\multicolumn{1}{c|}{\multirow{3}[6]{*}{Data set}} 
& \multicolumn{8}{c|}{FA-based Clustering Algorithms\tnote{a}}          
& \multicolumn{4}{c}{Non-FA-based Clustering Algorithms\tnote{b}} \\
\cmidrule{2-13}          
& \multicolumn{4}{c|}{Random Order} 
& \multicolumn{4}{c|}{VAT pre-order\tnote{d}} 
& \multicolumn{1}{c}{\multirow{2}[4]{*}{SL-HAC}} 
& \multicolumn{1}{c}{\multirow{2}[4]{*}{DBSCAN}} 
& \multicolumn{1}{c}{\multirow{2}[4]{*}{K-means}} 
& \multicolumn{1}{c}{\multirow{2}[4]{*}{AP}} \\
\cmidrule{2-9}          
& \multicolumn{1}{c}{FA} 
& \multicolumn{1}{c}{DVFA} 
& \multicolumn{1}{c}{TopoART B} 
& \multicolumn{1}{c|}{DDVFA\tnote{c}} 
& \multicolumn{1}{c}{FA} 
& \multicolumn{1}{c}{DVFA} 
& \multicolumn{1}{c}{TopoART B} 
& \multicolumn{1}{c|}{DDVFA} 
&       &       &       &  \\
\midrule
\midrule
Aggregation 
& $0.5085 \pm 0.1014$ 
& $0.7428 \pm 0.0892$ 
& $\bm{0.8828 \pm 0.0610}$ 
& $0.8780 \pm 0.0500$ 
& $0.7502 \pm 0.0093$ 
& $0.8089 \pm 0.0000$ 
& $\bm{0.9810 \pm 0.0005}$ 
& $0.8095 \pm 0.0001$ 
& $0.8186$ & $0.8901$ & $0.7948$ & $0.5937$ \\
Atom  
& $0.5128 \pm 0.0028$ 
& $0.5739 \pm 0.0746$ 
& $0.9033 \pm 0.1777$ 
& $\bm{1.0000 \pm 0.0000}$ 
& $0.8740 \pm 0.0452$ 
& $\bm{1.0000 \pm 0.0000}$ 
& $0.9278 \pm 0.0075$ 
& $\bm{1.0000 \pm 0.0000}$ 
& $1.0000$ & $1.0000$ & $0.5880$ & $0.5232$ \\
Chainlink 
& $0.3714 \pm 0.2440$ 
& $0.4002 \pm 0.1616$ 
& $0.9993 \pm 0.0018$
& $\bm{1.0000 \pm 0.0000}$ 
& $0.9510 \pm 0.0317$ 
& $\bm{1.0000 \pm 0.0000}$ 
& $\bm{1.0000 \pm 0.0000}$ 
& $\bm{1.0000 \pm 0.0000}$ 
& $1.0000$ & $1.0000$ & $0.2528$ & $0.2172$ \\
Compound 
& $0.6026 \pm 0.1338$ 
& $0.6674 \pm 0.1053$ 
& $0.7964 \pm 0.0658$ 
& $\bm{0.9242 \pm 0.0000}$ 
& $0.7860 \pm 0.0040$ 
& $0.9182 \pm 0.0000$ 
& $0.8885 \pm 0.0601$ 
& $\bm{0.9258 \pm 0.0000}$ 
& $0.9270$ & $0.9531$ & $0.7294$ & $0.3831$ \\
Dermatology 
& $0.2073 \pm 0.0664$ 
& $0.6302 \pm 0.1143$ 
& $0.4422 \pm 0.1311$ 
& $\bm{0.6399 \pm 0.0603}$ 
& $0.5994 \pm 0.0658$ 
& $0.5994 \pm 0.0658$ 
& $\bm{0.7224 \pm 0.0277}$ 
& $0.6576 \pm 0.0877$ 
& $0.3740$ & $0.4273$ & $0.8703$ & $0.6328$ \\
Ecoli
& $0.2192 \pm 0.0740$ 
& $\bm{0.6074 \pm 0.1252}$ 
& $0.5170 \pm 0.0956$ 
& $0.5687 \pm 0.0887$ 
& $0.5081 \pm 0.0033$ 
& $0.6102 \pm 0.0041$ 
& $\bm{0.6580 \pm 0.0093}$ 
& $0.6398 \pm 0.0007$ 
& $0.5101$ & $0.4979$ & $0.6966$ & $0.2516$ \\
Face  
& $0.5319 \pm 0.1847$ 
& $0.5319 \pm 0.1847$ 
& $0.9941 \pm 0.0106$ 
& $\bm{1.0000 \pm 0.0000}$ 
& $0.3848 \pm 0.2697$ 
& $\bm{1.0000 \pm 0.0000}$ 
& $\bm{1.0000 \pm 0.0000}$ 
& $\bm{1.0000 \pm 0.0000}$ 
& $1.0000$ & $1.0000$ & $0.2815$ & $0.2879$ \\
Flag  
& $0.8107 \pm 0.1649$ 
& $0.8107 \pm 0.1649$ 
& $0.9998 \pm 0.0009$ 
& $\bm{1.0000 \pm 0.0000}$ 
& $0.7374 \pm 0.0000$ 
& $\bm{1.0000 \pm 0.0000}$ 
& $0.9321 \pm 0.0799$ 
& $\bm{1.0000 \pm 0.0000}$ 
& $1.0000$ & $1.0000$ & $0.7184$ & $0.7147$ \\
Flame 
& $0.2041 \pm 0.0706$ 
& $0.4075 \pm 0.2072$ 
& $0.4440 \pm 0.1429$ 
& $\bm{0.8508 \pm 0.0000}$ 
& $0.4563 \pm 0.0151$ 
& $0.5921 \pm 0.2363$ 
& $\bm{0.9766 \pm 0.0083}$ 
& $0.8310 \pm 0.0000$ 
& $0.9172$ & $0.9081$ & $0.4534$ & $0.2048$ \\
Giant 
& $0.8416 \pm 0.3613$ 
& $0.8416 \pm 0.3613$ 
& $0.9956 \pm 0.0244$ 
& $\bm{1.0000 \pm 0.0000}$ 
& $0.6635 \pm 0.4840$ 
& $\bm{1.0000 \pm 0.0000}$ 
& $\bm{1.0000 \pm 0.0000}$ 
& $\bm{1.0000 \pm 0.0000}$ 
& $1.0000$ & $1.0000$ & $0.0444$ & $0.0210$ \\
Glass 
& $0.1707 \pm 0.0488$ 
& $\bm{0.3818 \pm 0.1170}$ 
& $0.3238 \pm 0.1199$ 
& $0.3162 \pm 0.0000$ 
& $0.3507 \pm 0.0299$ 
& $0.3507 \pm 0.0299$ 
& $0.4191 \pm 0.0388$ 
& $\bm{0.4340 \pm 0.0681}$ 
& $0.3210$ & $0.3210$ & $0.5046$ & $0.3405$ \\
Hepta 
& $0.8923 \pm 0.0399$ 
& $0.9865 \pm 0.0248$ 
& $0.9628 \pm 0.0240$ 
& $\bm{1.0000 \pm 0.0000}$ 
& $0.9345 \pm 0.0573$ 
& $\bm{1.0000 \pm 0.0000}$ 
& $0.9433 \pm 0.0225$ 
& $\bm{1.0000 \pm 0.0000}$ 
& $1.0000$ & $1.0000$ & $1.0000$ & $1.0000$ \\
Iris  
& $0.4863 \pm 0.0748$ 
& $0.6526 \pm 0.1075$ 
& $0.6048 \pm 0.0808$ 
& $\bm{0.6596 \pm 0.0000}$ 
& $0.7236 \pm 0.1714$ 
& $0.7247 \pm 0.1703$ 
& $\bm{0.8227 \pm 0.0707}$ 
& $0.7600 \pm 0.0061$ 
& $0.6141$ & $0.5681$ & $0.7163$ & $0.5490$ \\
Jain  
& $0.5629 \pm 0.2221$ 
& $0.5950 \pm 0.2672$ 
& $0.7578 \pm 0.1537$ 
& $\bm{0.9914 \pm 0.0000}$ 
& $0.7124 \pm 0.1672$ 
& $\bm{1.0000 \pm 0.0000}$ 
& $0.6958 \pm 0.0003$ 
& $\bm{1.0000 \pm 0.0000}$ 
& $0.9758$ & $0.9758$ & $0.5767$ & $0.2425$ \\
Lsun  
& $0.4368 \pm 0.1897$ 
& $0.6415 \pm 0.1498$ 
& $0.7890 \pm 0.1489$ 
& $\bm{1.0000 \pm 0.0000}$ 
& $0.9263 \pm 0.0656$ 
& $\bm{1.0000 \pm 0.0000}$ 
& $0.9613 \pm 0.0137$ 
& $\bm{1.0000 \pm 0.0000}$ 
& $1.0000$ & $1.0000$ & $0.6619$ & $0.311$ \\
Moon  
& $0.2774 \pm 0.0668$ 
& $0.3721 \pm 0.1020$ 
& $0.6829 \pm 0.1741$ 
& $\bm{1.0000 \pm 0.0000}$ 
& $0.5398 \pm 0.0330$ 
& $0.9669 \pm 0.0315$ 
& $0.6651 \pm 0.1230$ 
& $\bm{1.0000 \pm 0.0000}$ 
& $1.0000$ & $1.0000$ & $0.3508$ & $0.2762$ \\
Path based 
& $0.2671 \pm 0.0976$ 
& $0.4780 \pm 0.0900$ 
& $0.5278 \pm 0.0593$ 
& $\bm{0.6097 \pm 0.0145}$ 
& $0.4947 \pm 0.0123$ 
& $\bm{0.8513 \pm 0.1218}$ 
& $0.6236 \pm 0.0355$ 
& $0.6573 \pm 0.0008$ 
& $0.6122$ & $0.6087$ & $0.5501$ & $0.3452$ \\
R15   
& $0.7922 \pm 0.0459$ 
& $0.9347 \pm 0.0360$ 
& $0.9205 \pm 0.0280$ 
& $\bm{0.9465 \pm 0.0227}$ 
& $0.9634 \pm 0.0023$ 
& $0.9634 \pm 0.0023$ 
& $\bm{0.9857 \pm 0.0001}$ 
& $0.9575 \pm 0.0116$ 
& $0.9460$ & $0.8976$ & $0.9928$ & $0.9928$ \\
Ring  
& $0.0924 \pm 0.0124$ 
& $0.2022 \pm 0.0429$ 
& $0.9768 \pm 0.0688$ 
& $\bm{1.0000 \pm 0.0000}$ 
& $0.2333 \pm 0.0446$ 
& $\bm{1.0000 \pm 0.0000}$ 
& $0.8418 \pm 0.1609$ 
& $\bm{1.0000 \pm 0.0000}$ 
& $1.0000$ & $1.0000$ & $0.1696$ & $0.1088$ \\
Seeds 
& $0.3414 \pm 0.0675$ 
& $\bm{0.5373 \pm 0.1289}$
& $0.4579 \pm 0.1604$ 
& $0.5360 \pm 0.0850$ 
& $\bm{0.6432 \pm 0.0197}$ 
& $\bm{0.6432 \pm 0.0197}$ 
& $0.5813 \pm 0.0482$ 
& $0.6087 \pm 0.0210$ 
& $0.4259$ & $0.4143$ & $0.7049$ & $0.2822$ \\
Spiral 
& $0.0870 \pm 0.0058$ 
& $0.1740 \pm 0.0145$ 
& $0.3004 \pm 0.0784$ 
& $\bm{1.0000 \pm 0.0000}$ 
& $0.2443 \pm 0.0301$ 
& $\bm{1.0000 \pm 0.0000}$ 
& $0.2019 \pm 0.0023$ 
& $\bm{1.0000 \pm 0.0000}$ 
& $1.0000$ & $1.0000$ & $0.1302$ & $0.1541$ \\
Synthetic Control 
& $0.0894 \pm 0.0283$ 
& $\bm{0.5841 \pm 0.0524}$ 
& $0.3320 \pm 0.0805$ 
& $0.5831 \pm 0.0358$ 
& $0.6081 \pm 0.0266$ 
& $0.6081 \pm 0.0266$ 
& $0.6640 \pm 0.0197$ 
& $\bm{0.6690 \pm 0.0182}$ 
& $0.5530$ & $0.5495$ & $0.6109$ & $0.5285$ \\
Target 
& $0.5679 \pm 0.0394$ 
& $0.6515 \pm 0.0178$ 
& $0.9989 \pm 0.0020$ 
& $\bm{1.0000 \pm 0.0000}$ 
& $0.6407 \pm 0.0023$ 
& $\bm{1.0000 \pm 0.0000}$ 
& $0.8950 \pm 0.1068$ 
& $\bm{1.0000 \pm 0.0000}$ 
& $1.0000$ & $1.0000$ & $0.6838$ & $0.3051$ \\
Tetra 
& $0.3793 \pm 0.0912$ 
& $0.6928 \pm 0.1472$ 
& $0.6143 \pm 0.1392$ 
& $\bm{0.8269 \pm 0.1306}$ 
& $0.9933 \pm 0.0000$ 
& $0.9933 \pm 0.0000$ 
& $\bm{1.0000 \pm 0.0000}$ 
& $0.9933 \pm 0.0000$ 
& $0.9462$ & $0.9326$ & $1.0000$ & $0.6018$ \\
Twodiamonds 
& $0.2917 \pm 0.0934$ 
& $0.5879 \pm 0.2602$ 
& $0.6245 \pm 0.3250$ 
& $\bm{0.6628 \pm 0.4411}$ 
& $\bm{0.9570 \pm 0.0508}$ 
& $\bm{0.9570 \pm 0.0508}$ 
& $0.9460 \pm 0.0482$ 
& $0.9410 \pm 0.0775$ 
& $0.8980$ & $0.7469$ & $1.0000$ & $0.1839$ \\
Wave  
& $0.1407 \pm 0.0912$ 
& $0.1929 \pm 0.0437$ 
& $0.4466 \pm 0.1656$ 
& $\bm{1.0000 \pm 0.0000}$ 
& $0.3770 \pm 0.1039$ 
& $\bm{1.0000 \pm 0.0000}$ 
& $0.3315 \pm 0.0827$ 
& $\bm{1.0000 \pm 0.0000}$ 
& $1.0000$ & $1.0000$ & $0.2081$ & $0.1889$ \\
Wine  
& $0.0893 \pm 0.0320$ 
& $\bm{0.6359 \pm 0.1353}$ 
& $0.4317 \pm 0.1472$ 
& $0.5138 \pm 0.0863$ 
& $0.5807 \pm 0.0729$ 
& $0.5846 \pm 0.1580$ 
& $0.5851 \pm 0.0338$ 
& $\bm{0.6578 \pm 0.0140}$
& $0.4071$ & $0.4363$ & $0.8685$ & $0.4464$ \\
Wingnut 
& $0.3736 \pm 0.3298$ 
& $0.3736 \pm 0.3298$ 
& $0.3422 \pm 0.2140$ 
& $\bm{0.9921 \pm 0.0000}$ 
& $\bm{1.0000 \pm 0.0000}$ 
& $\bm{1.0000 \pm 0.0000}$ 
& $\bm{1.0000 \pm 0.0000}$ 
& $\bm{1.0000 \pm 0.0000}$ 
& $1.0000$ & $1.0000$ & $0.5047$ & $0.2259$ \\
Wisconsin 
& $0.4267 \pm 0.1263$ 
& $0.6909 \pm 0.0084$ 
& $0.6689 \pm 0.0194$ 
& $\bm{0.7044 \pm 0.0011}$ 
& $0.6909 \pm 0.0079$ 
& $\bm{0.7291 \pm 0.0011}$ 
& $0.7171 \pm 0.0635$ 
& $0.7240 \pm 0.0060$ 
& $0.7053$ & $0.8257$ & $0.8465$ & $0.3827$ \\
WDBC  
& $0.0722 \pm 0.0318$ 
& $\bm{0.4772 \pm 0.1568}$ 
& $0.3278 \pm 0.1150$ 
& $0.4381 \pm 0.1596$ 
& $0.3257 \pm 0.0026$ 
& $0.4209 \pm 0.0130$
& $\bm{0.4472 \pm 0.0468}$ 
& $0.3724 \pm 0.0041$ 
& $0.3021$ & $0.3200$ & $0.7302$ & $0.4821$ \\
\bottomrule
\end{tabular}
}
\endgroup
\begin{tablenotes}[normal,flushleft]
\item[] Bold values indicate best average performance across shuffled or VAT pre-ordered data (non-ART based methods are not included).
\item[a] Mean and standard deviations of $AR$ over $30$ runs with respect to the best parameter combination are reported.
\item[b] Since these clustering algorithms are either completely (or almost) insensitive to order presentation, data samples were randomized, a single run was performed, and the peak $AR$ corresponding to the best parameters are reported (except for k-means where $10$ repetitions were performed in such run).
\item[c] DDVFA + Merge ART system.
\item[d] All four ART-based algorithms were fed the VAT pre-ordered data.
\end{tablenotes}
\end{threeparttable}
}
\label{Tab:Comparisons}
\end{sidewaystable}

\subsection{Performance comparison 1: ART-based clustering algorithms} \label{Results-FA}

Table~\ref{Tab:Comparisons} lists the $AR$ peak average performance of fuzzy ART, DVFA, topoART~B, and DDVFA for both random and VAT ordered presentation scenarios. Given the results of Section~\ref{Sec:ResultsA}'s statistical analyses, the VAT + DDVFA and DDVFA + Merge ART systems were selected, and the performance was recorded with respect to single linkage-based activation and match functions variant. 

\subsubsection{Statistical analysis of performance} 
\begin{figure}[!t]
\centerline{
\subfloat[Random order]{\includegraphics[width=\columnwidth]{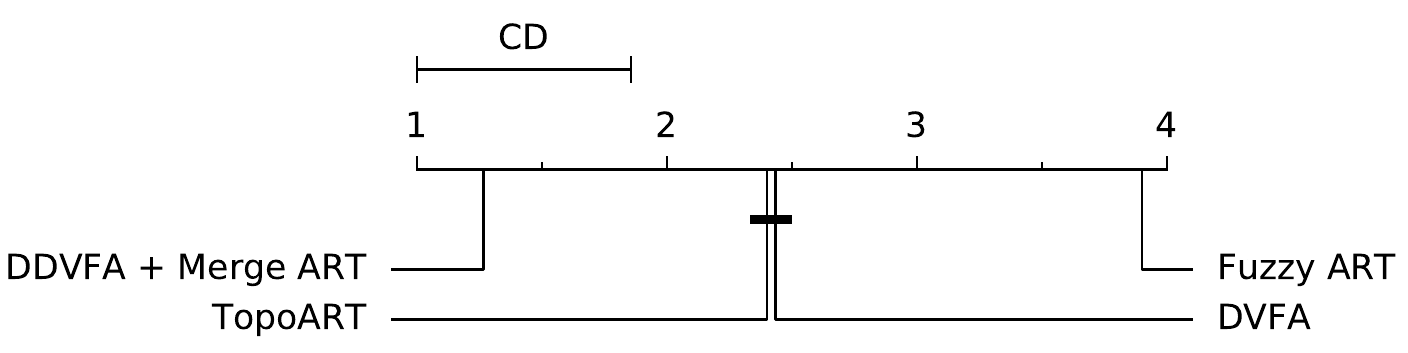} }
}
\centerline{
\subfloat[VAT pre-order]{\includegraphics[width=\columnwidth]{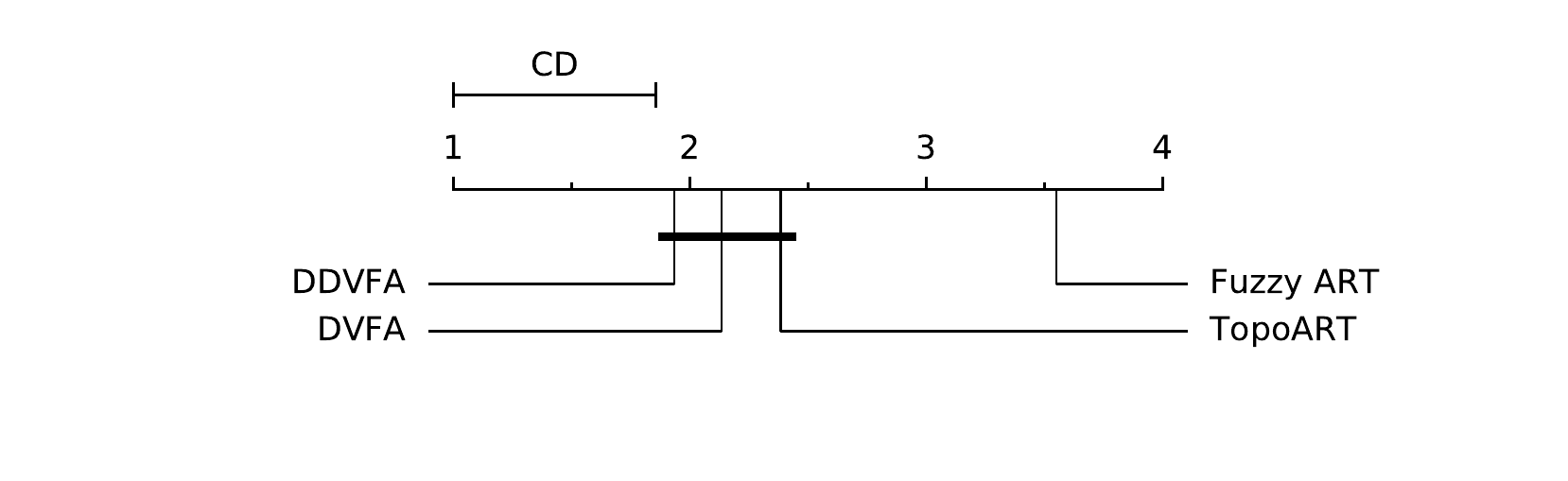} }
}
\caption{CD diagrams comparing the performance of ART-based clustering methods.}
\label{Fig:CD_FA_algs}
\end{figure}

The hypothesis that these algorithms perform equally was tested using the Iman-Davenport statistic and rejected at a $0.05$ significance level for both random (p-value=1.1102E-16) and VAT orderings (p-value=3.2012E-07). Therefore, the CD diagrams were further computed, as shown in Fig.~\ref{Fig:CD_FA_algs}, using Nemenyi's test. As shown, VAT pre-processing (offline incremental mode) equalizes performance, such that all multi-prototype ART-based algorithms become statistically similar, while also outperforming fuzzy ART. Alternately, when data is presented randomly in an online incremental mode DDVFA + Merge ART yields a statistically better performance than all the other ART-based algorithms at a $0.05$ significance level. DVFA and topoART~B were observed to be statistically equivalent (as expected per~\cite{leonardo2018b}) while also surpassing standard fuzzy ART. In the vast majority of the remaining comparisons among TopoART, DVFA, DDVFA systems, and fuzzy ART, no significant statistical difference was observed among the first three, while all of them outperformed fuzzy ART. 

\subsubsection{Statistical analysis of compactness} 

\begin{figure}[!t]
\centerline{
\subfloat[Random order]{\includegraphics[width=\columnwidth]{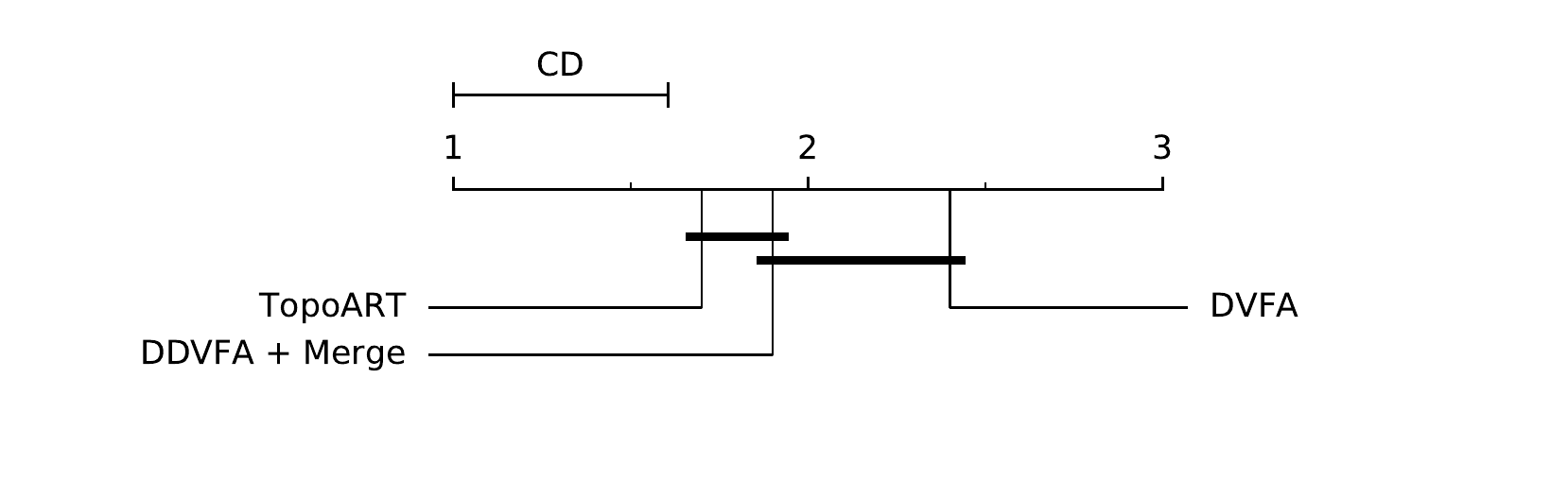} }
}
\centerline{
\subfloat[VAT pre-order]{\includegraphics[width=\columnwidth]{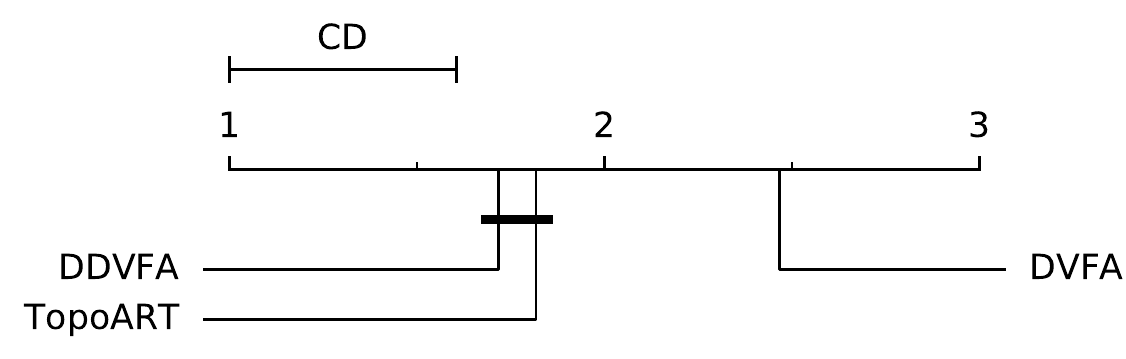} }
}
\caption{CD diagrams comparing the compactness of the multi-prototype  ART-based architectures.}
\label{Fig:CD_FA_algs_compactness}
\end{figure}

The compactness of the multi-prototype ART-based networks were also compared, i.e., the number of categories that were created to represent the data sets' clusters. The hypothesis of equivalence (using Iman-Davenport's test) was rejected at a $0.05$ significance level, with p-values equal to (a) 5.2039E-03 for VAT pre-ordering and (b) 1.7622E-02 for random presentation. Given this outcome, the corresponding CD diagrams were generated as shown in Fig.~\ref{Fig:CD_FA_algs_compactness} using Nemenyi's test. In online learning mode, in which samples are presented randomly, topoART has the best average ranking for compactness. Yet, in offline learning mode, in which order-dependence can be managed via pre-processing strategies, DDVFA + Merge ART has a better average compactness ranking than topoART. However, their observed compactness were similar and with no statistically significant difference. As expected, topoART creates more compact networks than DVFA in all scenarios~\cite{leonardo2018b}. Note that improved compactness may be obtained by carefully tuning parameter $\gamma$.

\subsubsection{Summary} 

The statistical analysis suggests that if pre-processing with VAT, then topoART, DVFA, and DDVFA seem to perform equally; whereas for random presentation DDVFA + Merge ART's performance was observed to be statistically better than the remaining ART-based systems. Moreover, no statistical differences were found between the compactness of topoART and DDVFA systems using single linkage functions for neither randomly or VAT ordered presentations, and both achieved a better average rank than DVFA. 

\begin{figure}[!ht]
\centerline{
\subfloat[DDVFA + Merge ART vs. non-ART-based methods.]{\includegraphics[width=\columnwidth]{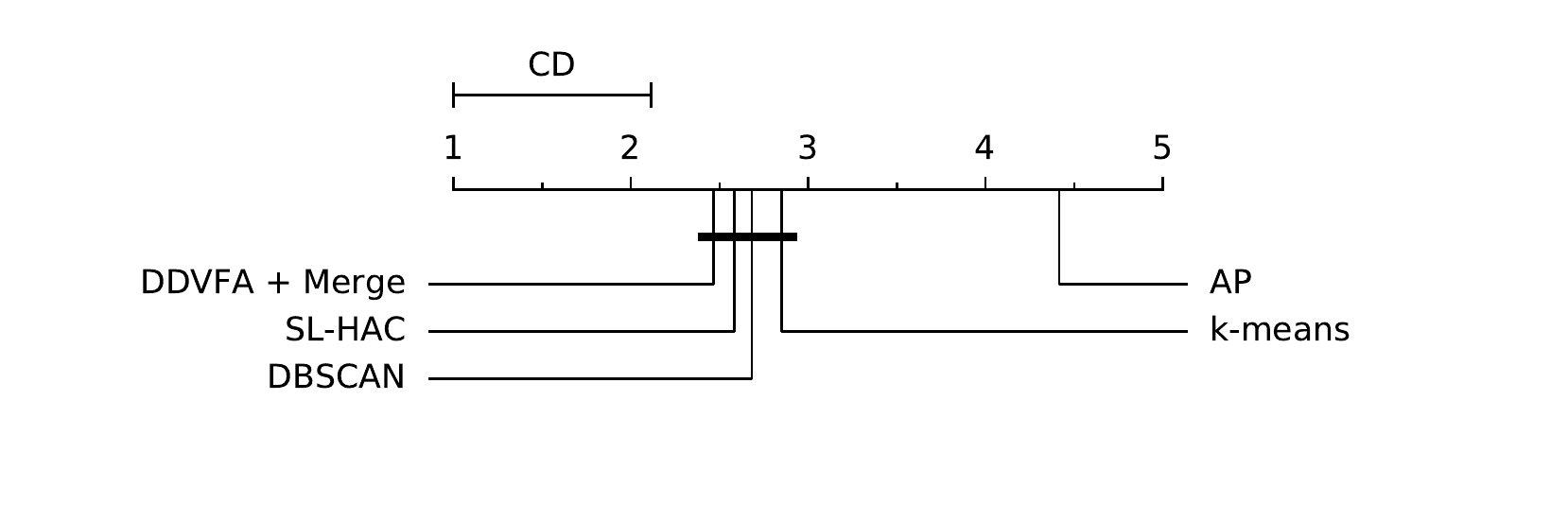} \label{Fig:RNG2}}
}
\centerline{
\subfloat[VAT + DDVFA vs. non-ART-based methods.]{\includegraphics[width=\columnwidth]{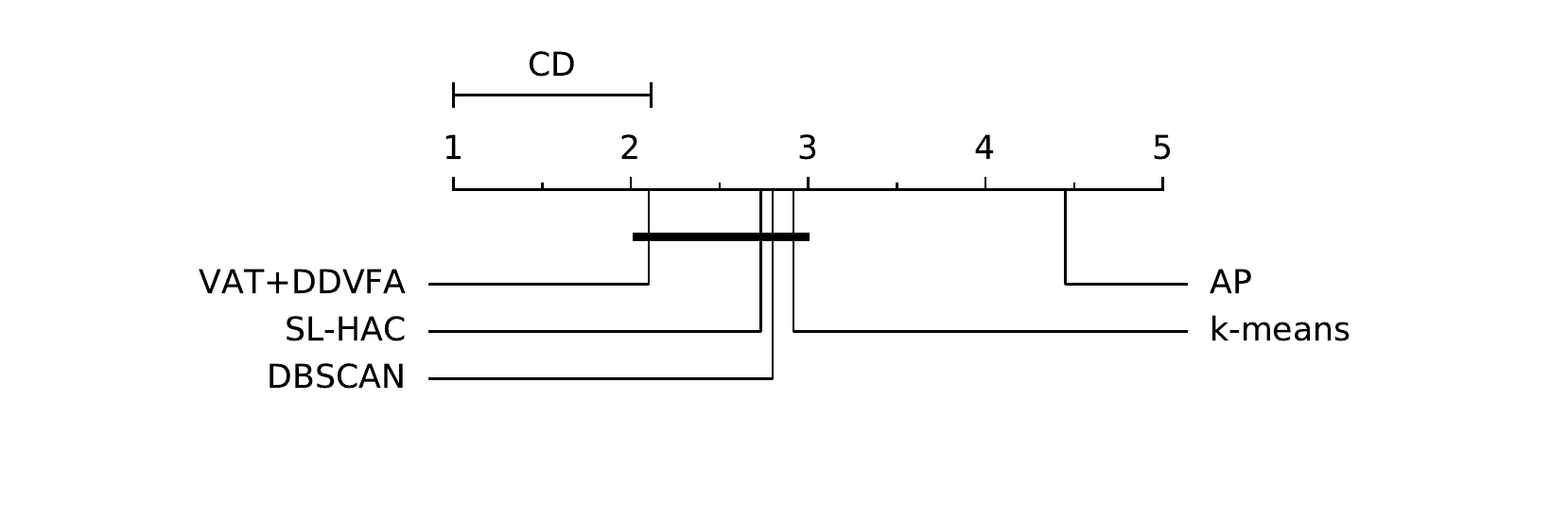} \label{Fig:VAT2}}
}
\caption{CD diagrams comparing the performance of two DDVFA systems to SL-HAC, DBSCAN, k-means, and AP clustering algorithms.}
\label{Fig:CD_clust_algs}
\end{figure}

\subsection{Performance comparison 2: non-ART-based clustering algorithms}

Table~\ref{Tab:Comparisons} also reports the performance of k-means, DBSCAN, affinity propagation (AP), and single linkage (SL-HAC). Again, the Iman-Davenport test was used to compare these algorithms to (a) VAT + DDVFA, and (b) DDVFA + Merge ART. These null hypotheses were rejected at a $0.05$ significance level with p-values equal to (a) 1.4944E-08, and (b) 4.5854E-07. Next, the CD diagrams were generated using Nemenyi's test, as shown in Fig.~\ref{Fig:CD_clust_algs}. It was observed that for these data sets all clustering algorithms seem to be statistically equivalent at a $0.05$ significance level, except for AP. Nevertheless, both DDVFA systems (VAT + DDVFA and DDVFA + Merge ART) have a smaller average rank value (particularly when using the VAT pre-processor). This on par performance is remarkable, especially regarding the comparison with the DDVFA + Merge ART system, since in this case clustering is performed both incrementally and online, as opposed to the other global clustering methods. Re-performing the computations using the entire data set is not required if a new sample is presented (c.f., SL-HAC). Therefore, it is possible to extend the current knowledge base. Moreover, the weights do not cycle, and previously acquired knowledge is not forgotten (c.f., k-means). These important advantages of DDVFA are inherited from ART.

\subsection{Sensitivity to kernel width parameter} \label{Sec:Gamma}

\newcommand{\gammaScale}{0.155}
\begin{figure*}[!ht]
\centerline{
\subfloat[Seeds]{\includegraphics[width=\gammaScale\textwidth]{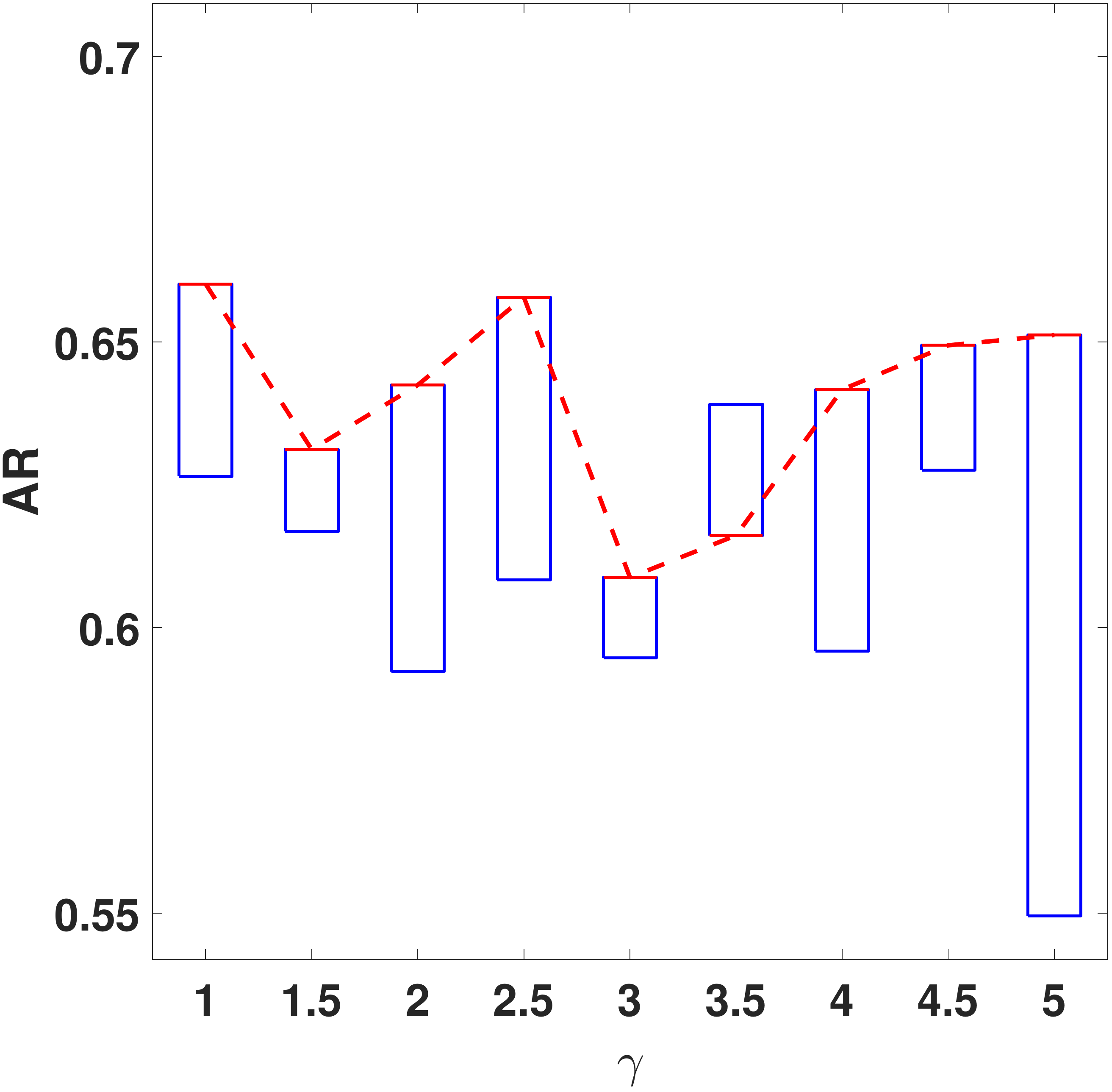} }
\subfloat[Wine]{\includegraphics[width=\gammaScale\textwidth]{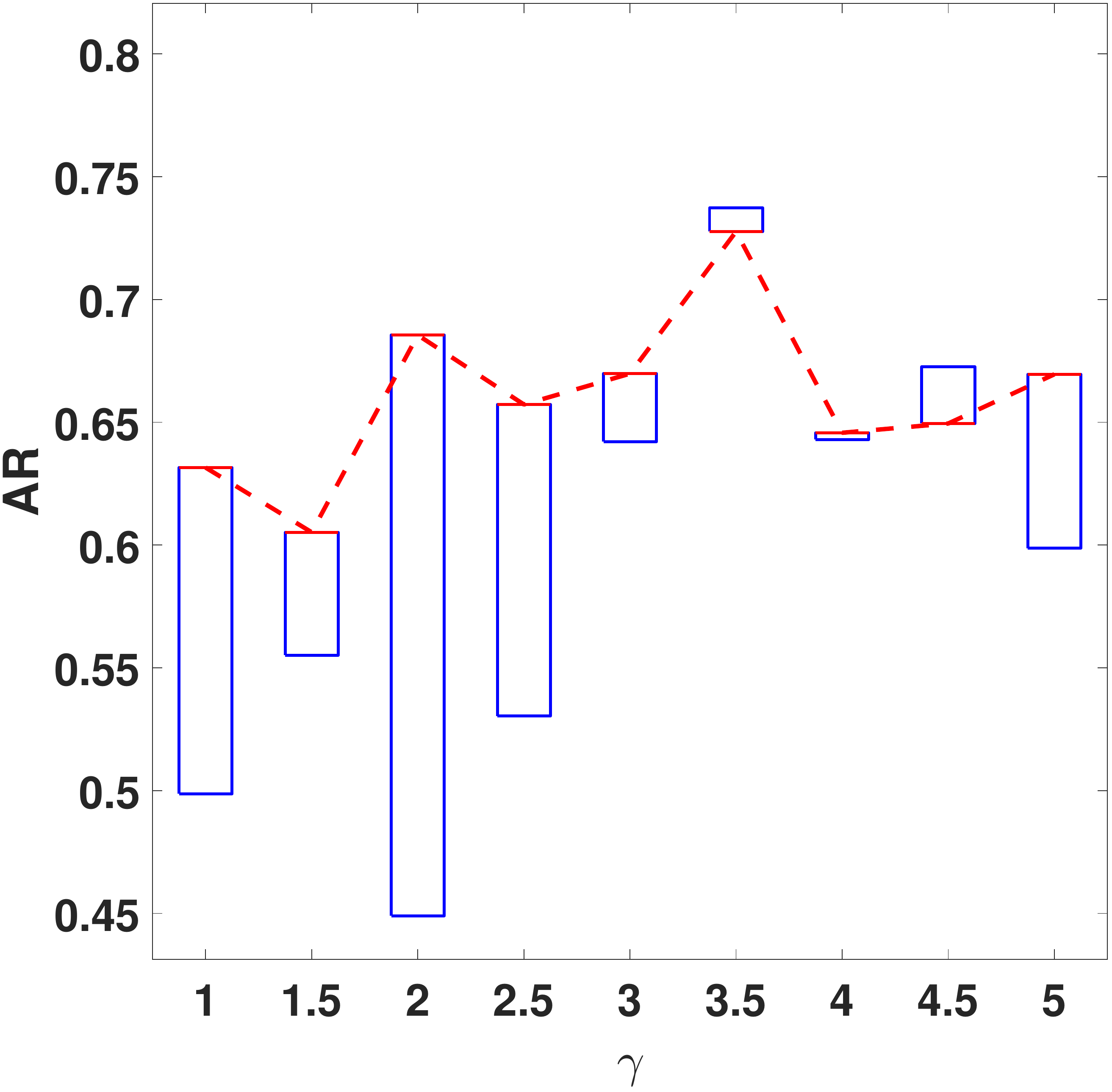} }
\subfloat[Target]{\includegraphics[width=\gammaScale\textwidth]{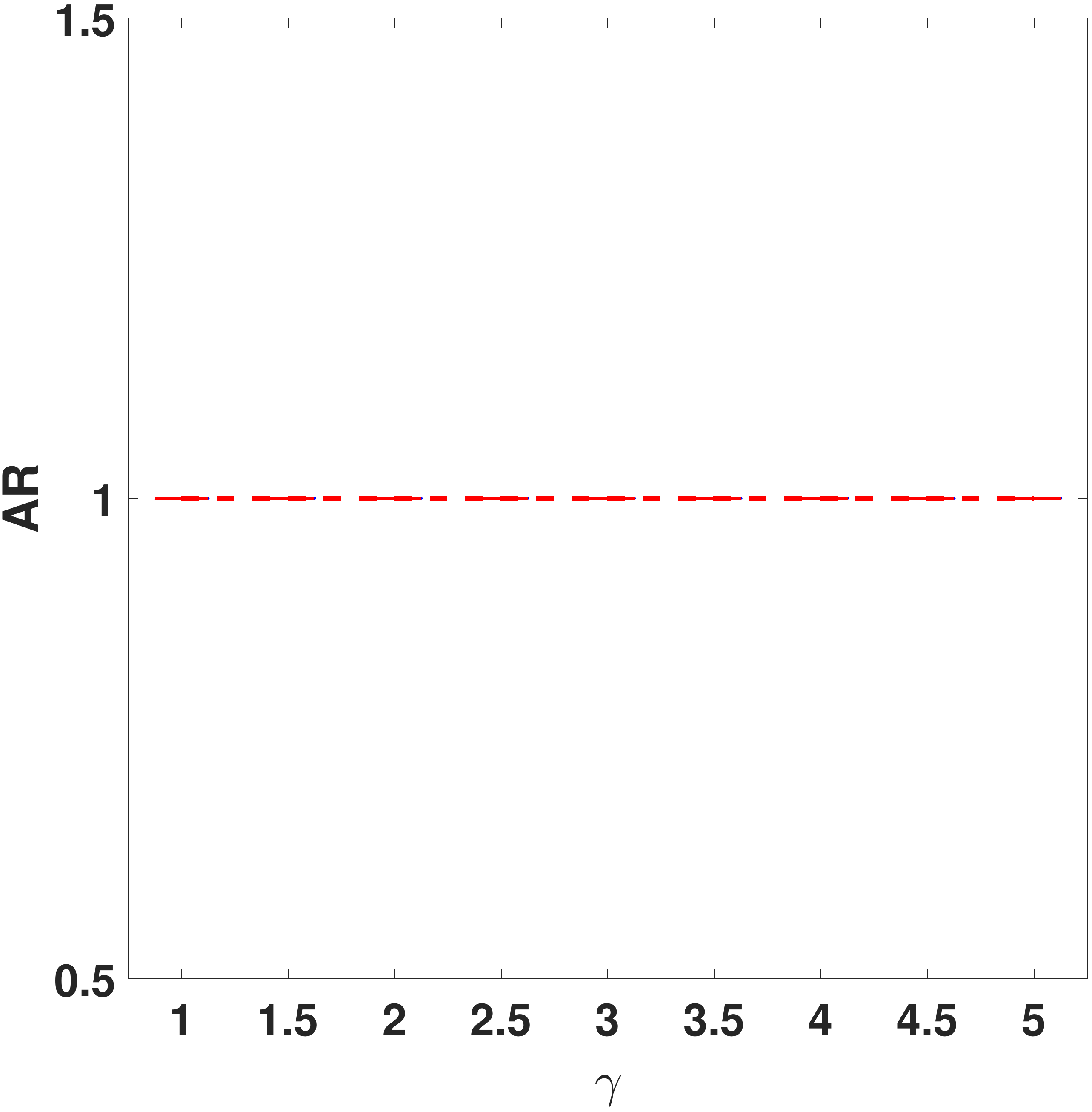} }
\subfloat[Tetra]{\includegraphics[width=\gammaScale\textwidth]{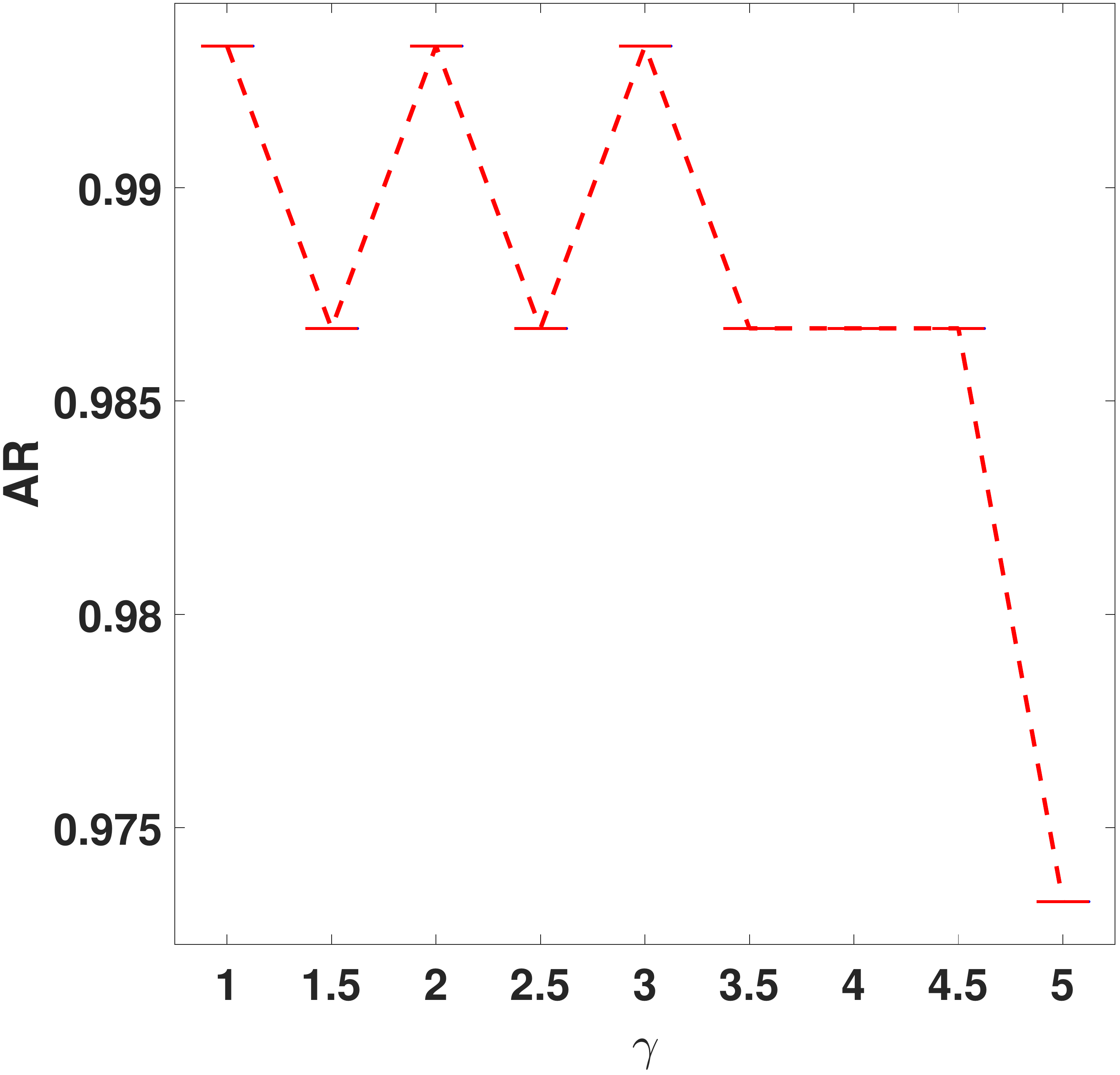} }
\subfloat[Lsun]{\includegraphics[width=\gammaScale\textwidth]{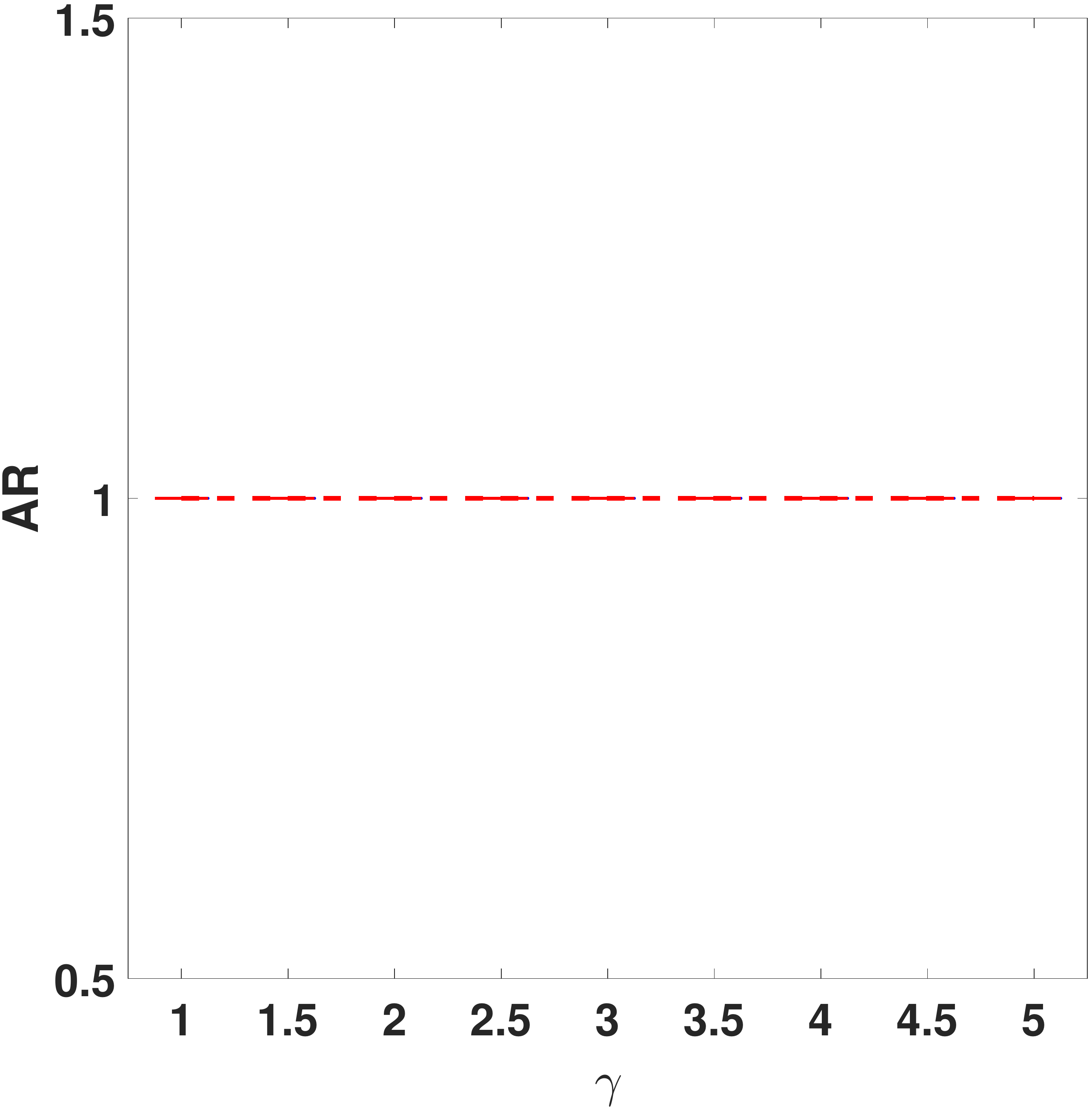} }
\subfloat[Moon]{\includegraphics[width=\gammaScale\textwidth]{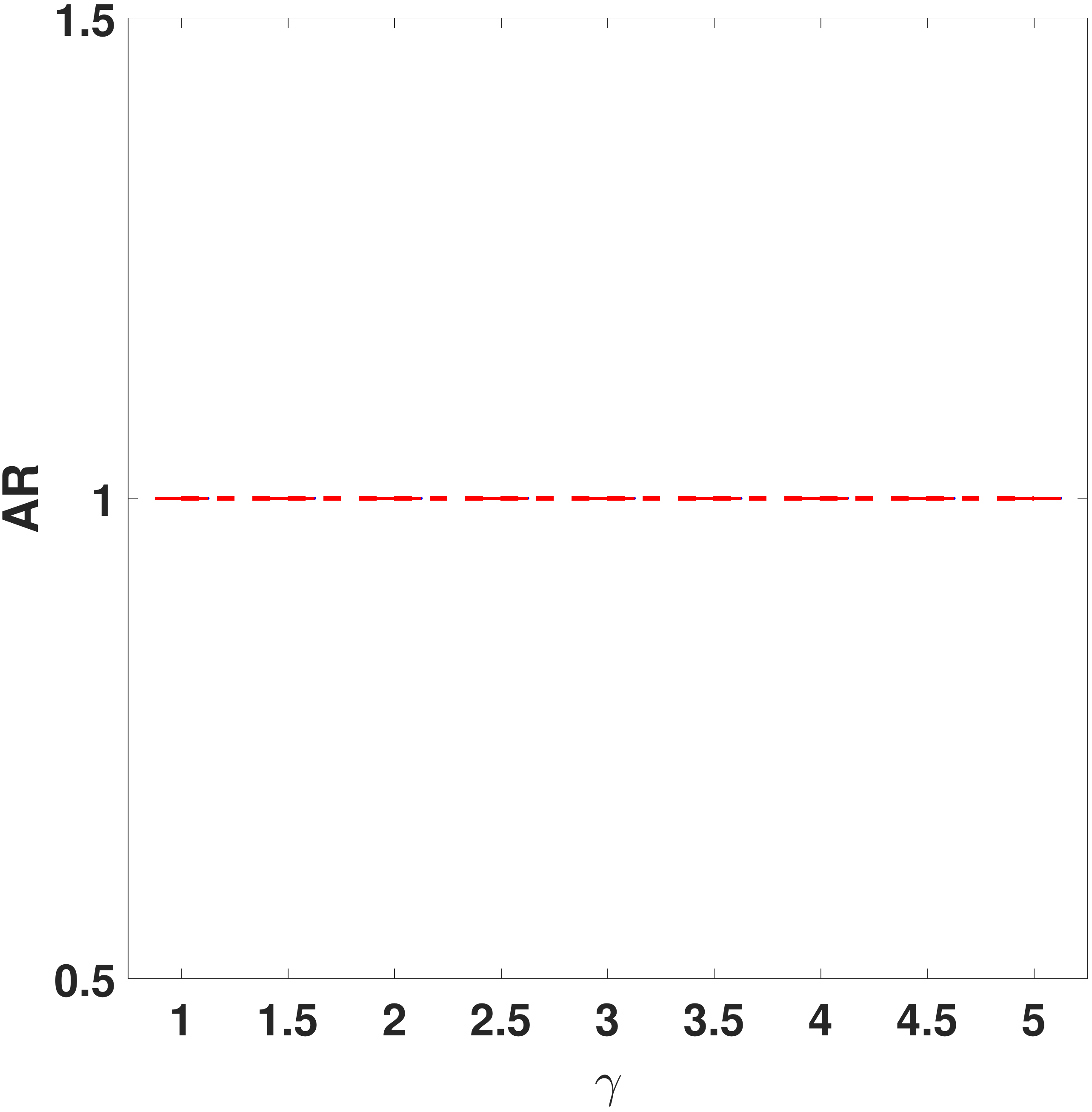} }
}
\vspace{-0.5\baselineskip}
\centerline{
\subfloat[Seeds]{\includegraphics[width=\gammaScale\textwidth]{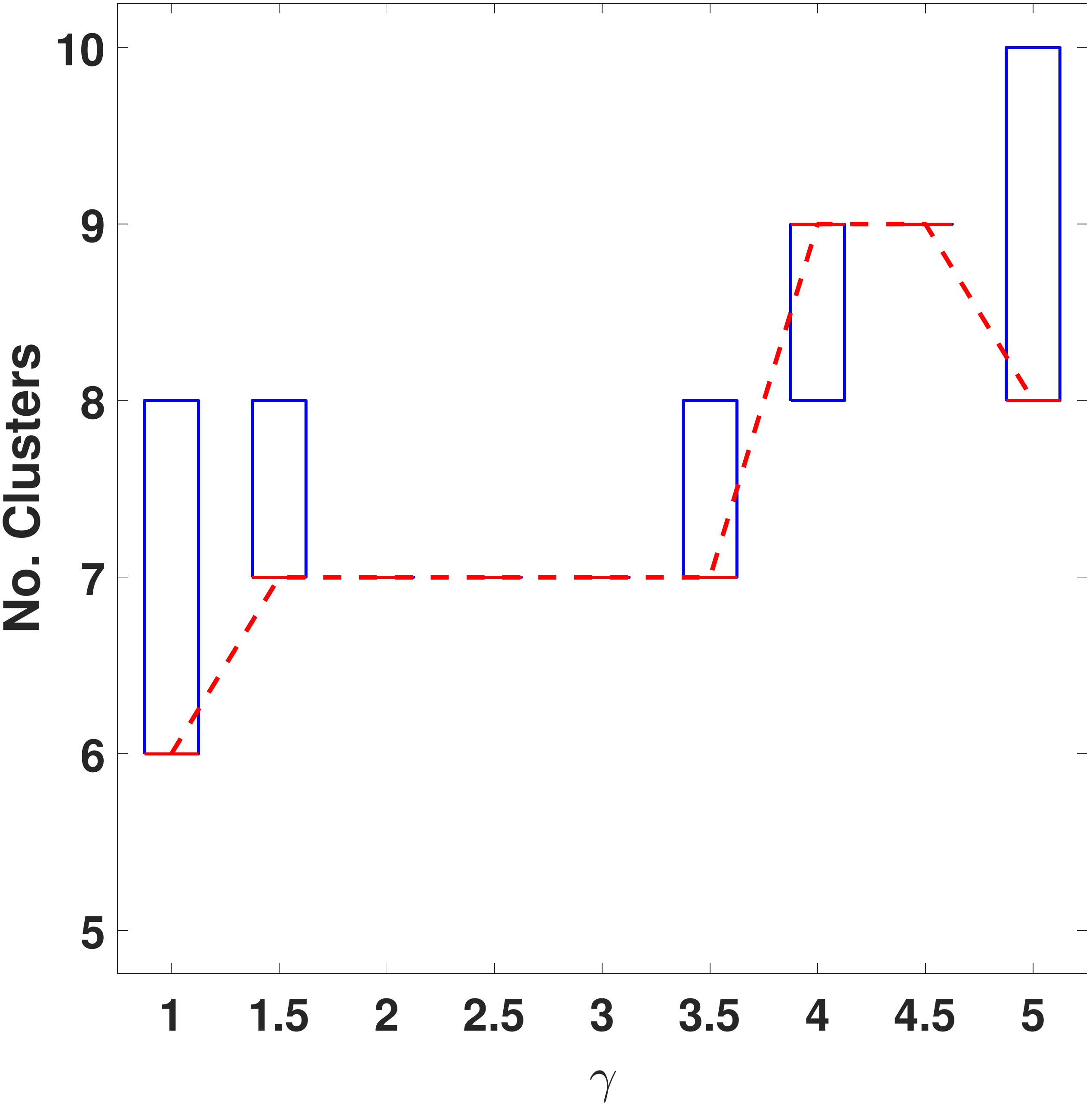} }
\subfloat[Wine]{\includegraphics[width=\gammaScale\textwidth]{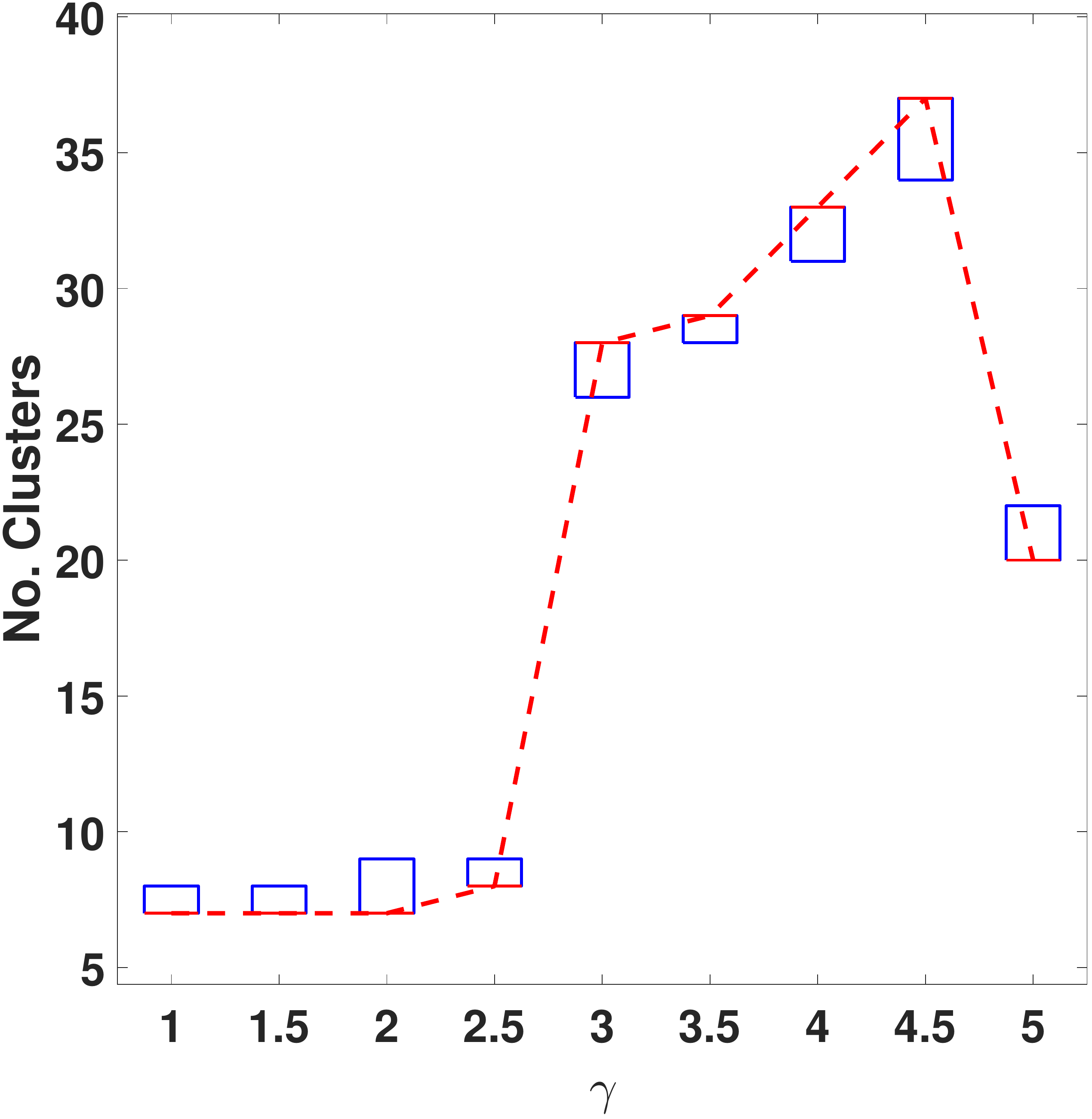} }
\subfloat[Target]{\includegraphics[width=\gammaScale\textwidth]{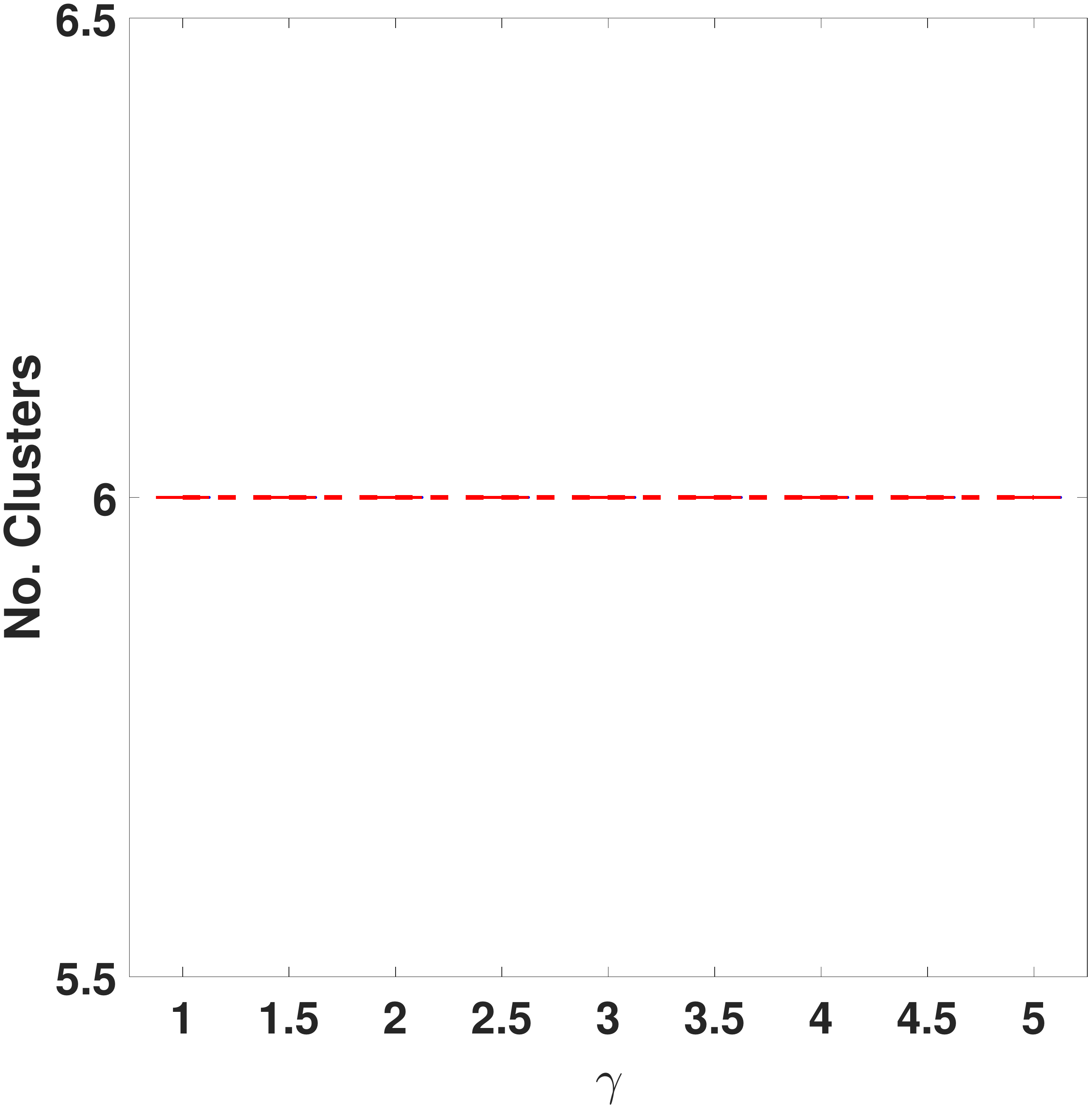} }
\subfloat[Tetra]{\includegraphics[width=\gammaScale\textwidth]{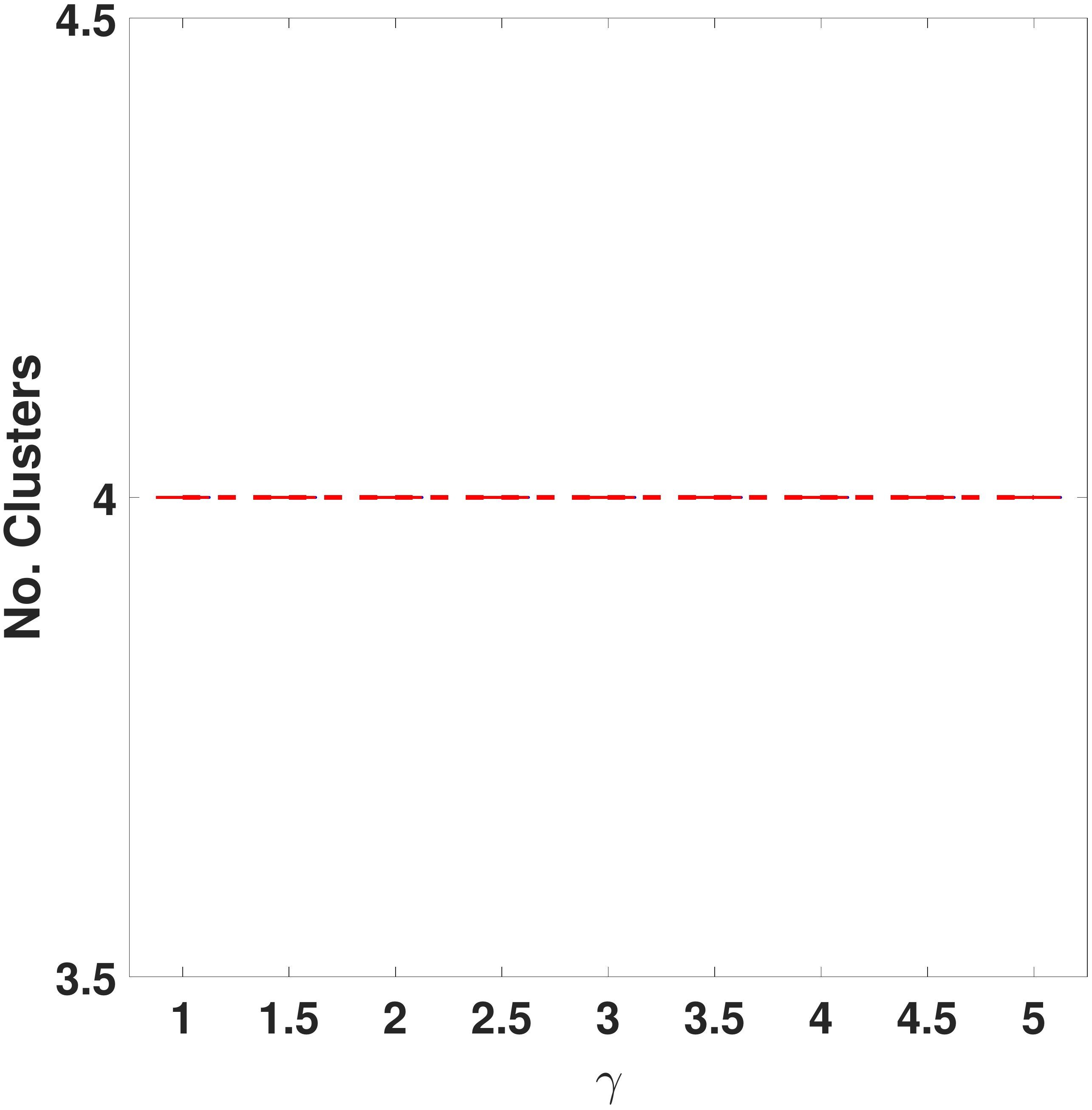} }
\subfloat[Lsun]{\includegraphics[width=\gammaScale\textwidth]{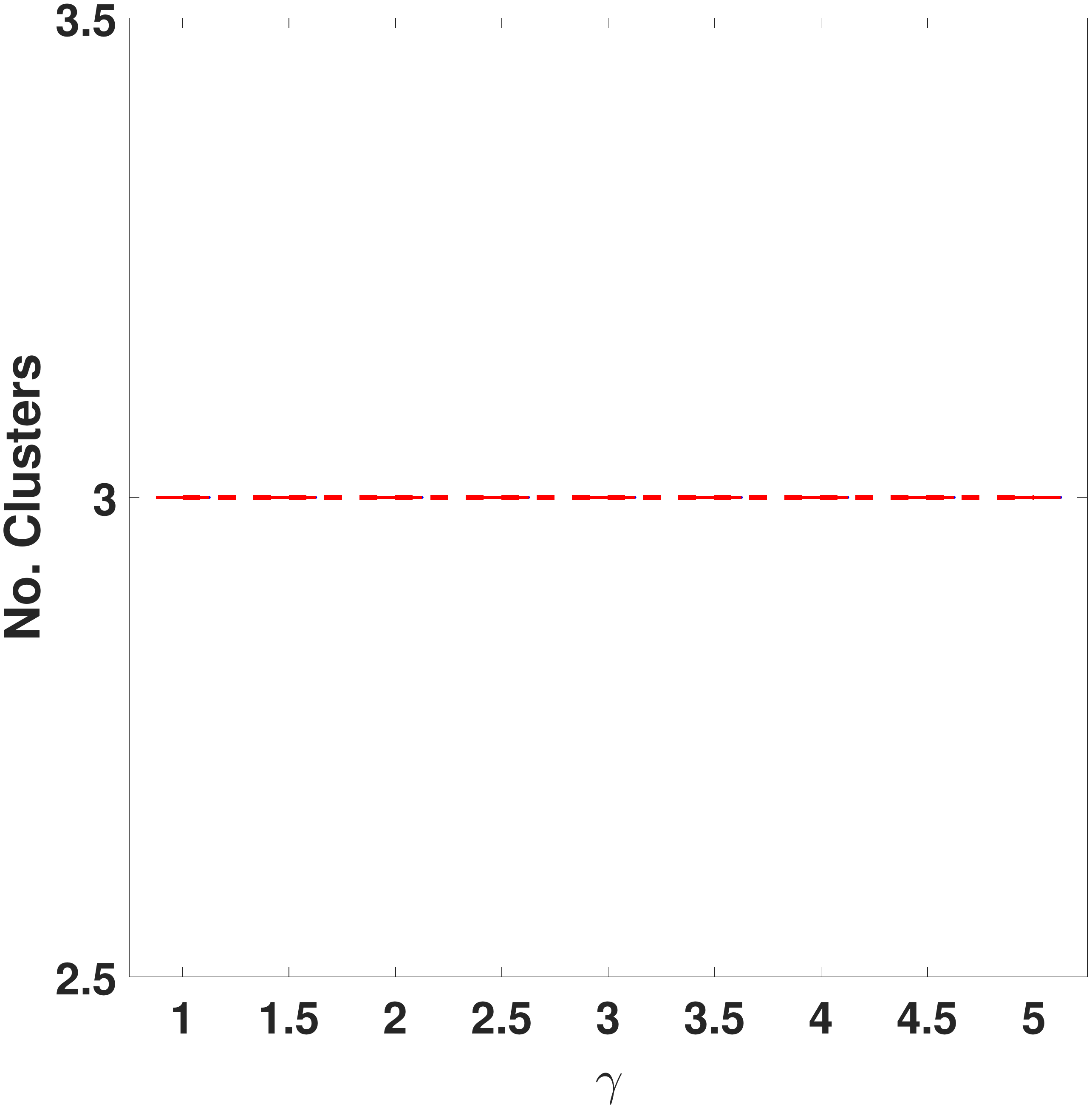} }
\subfloat[Moon]{\includegraphics[width=\gammaScale\textwidth]{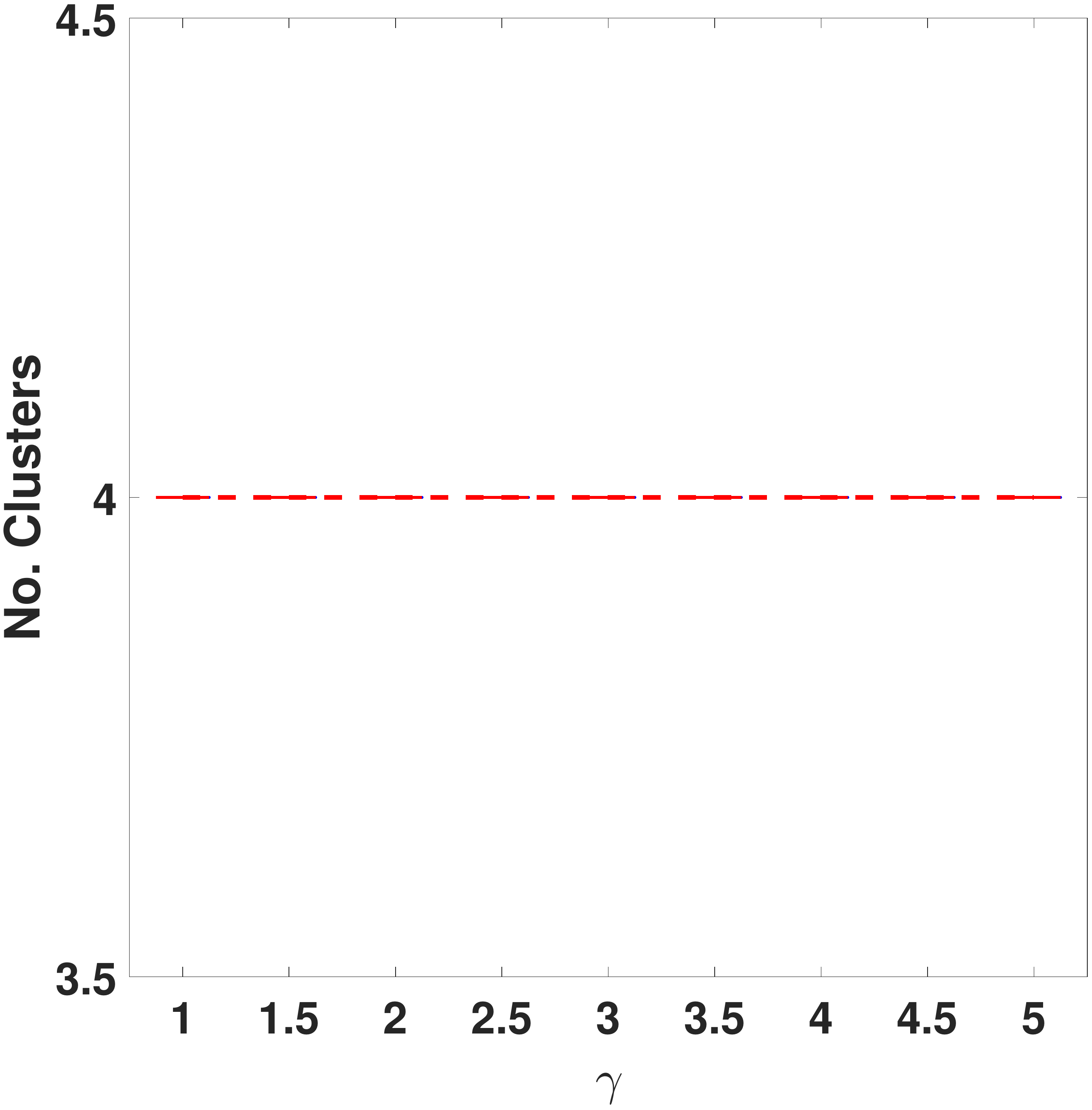} }
}
\vspace{-0.5\baselineskip}
\centerline{
\subfloat[Seeds]{\includegraphics[width=\gammaScale\textwidth]{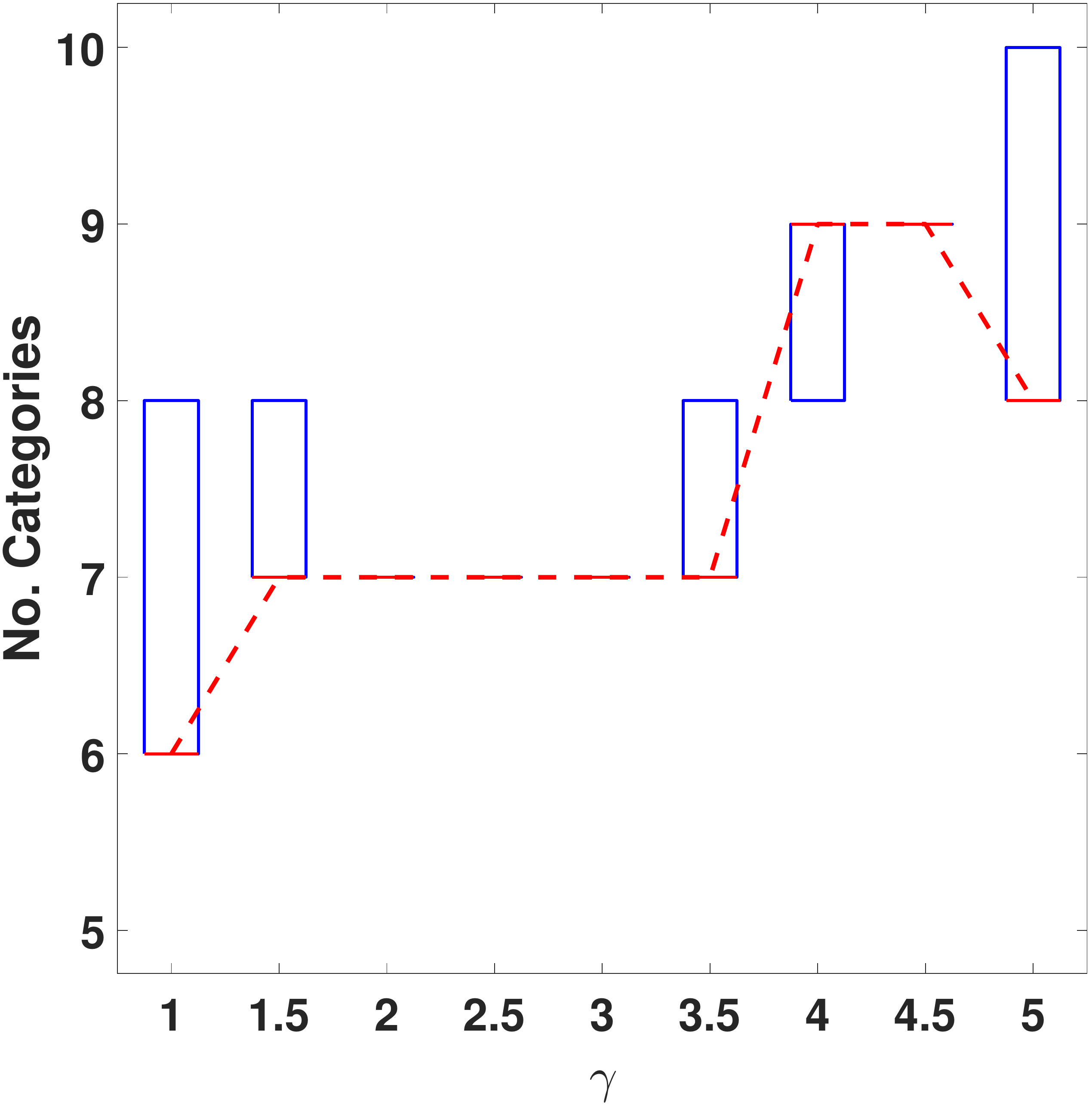} }
\subfloat[Wine]{\includegraphics[width=\gammaScale\textwidth]{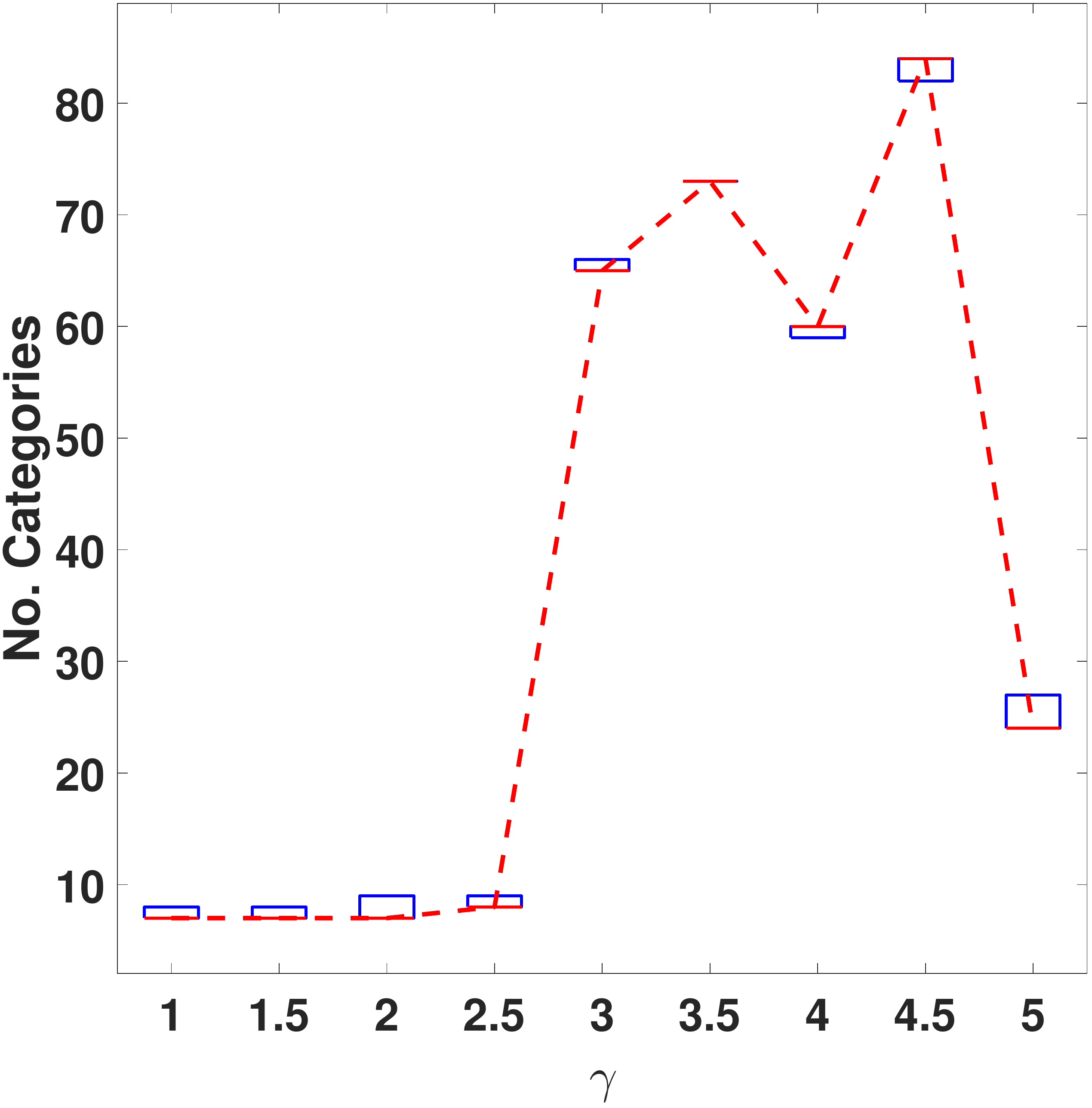} }
\subfloat[Target]{\includegraphics[width=\gammaScale\textwidth]{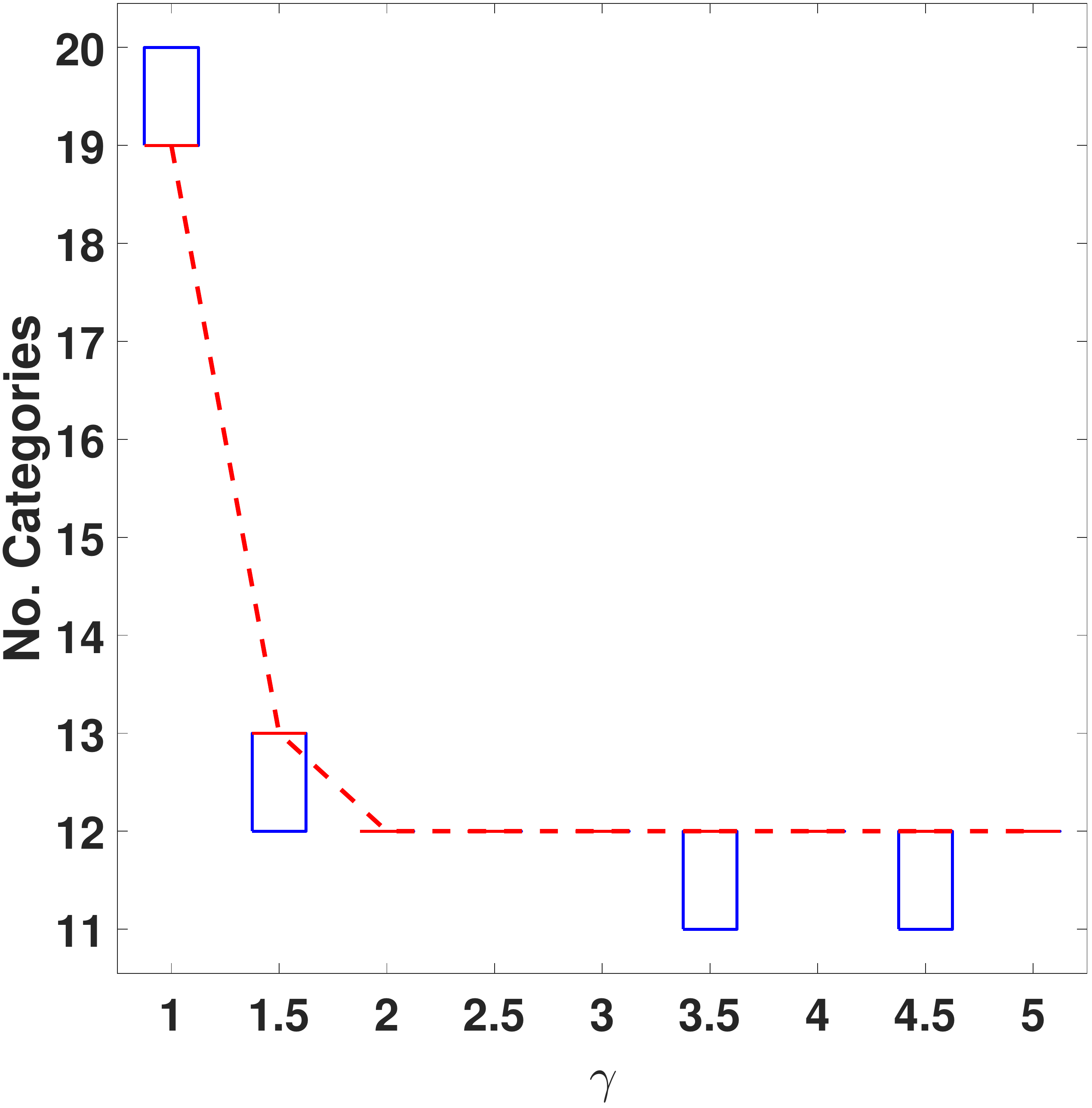} }
\subfloat[Tetra]{\includegraphics[width=\gammaScale\textwidth]{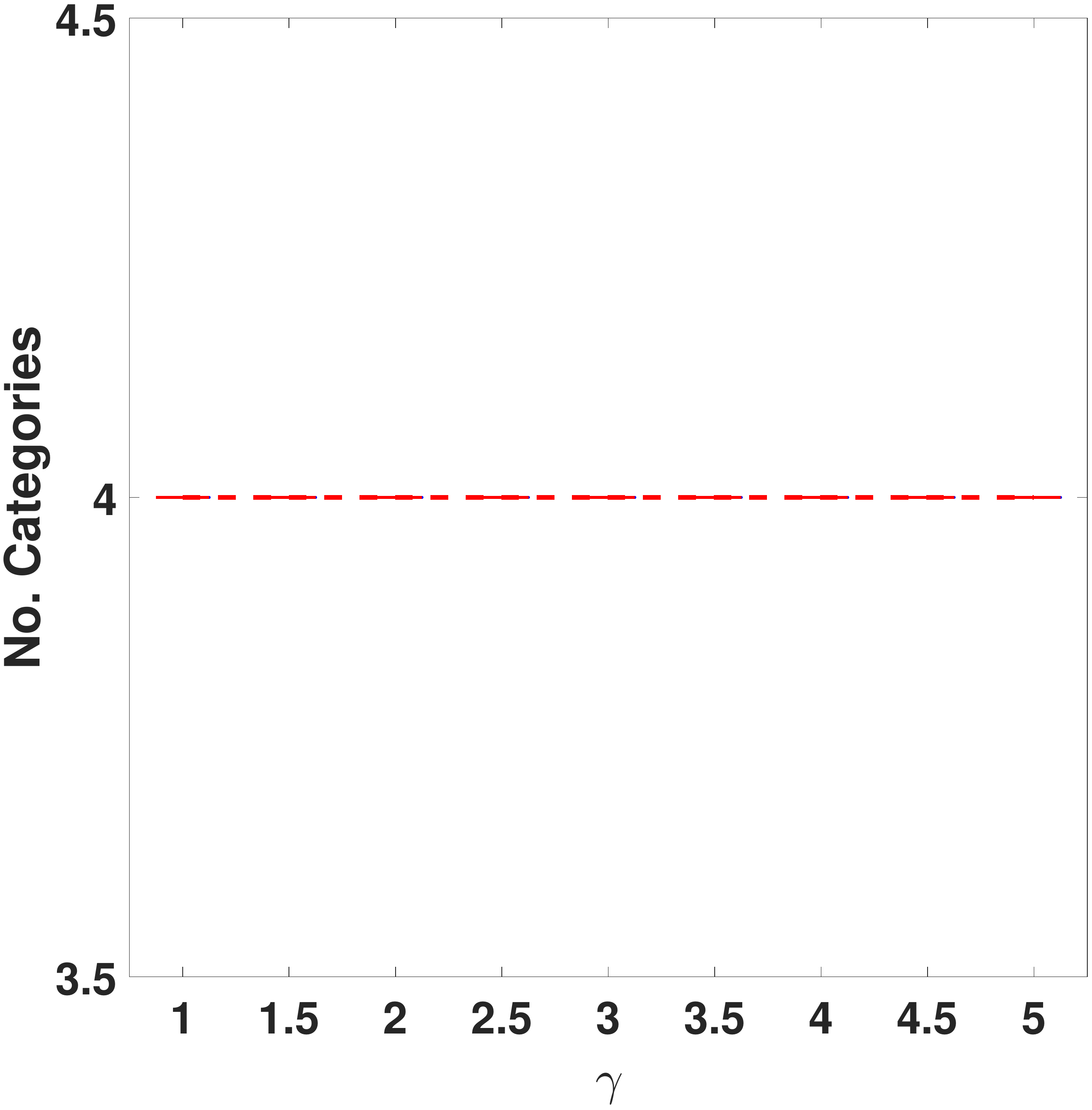} }
\subfloat[Lsun]{\includegraphics[width=\gammaScale\textwidth]{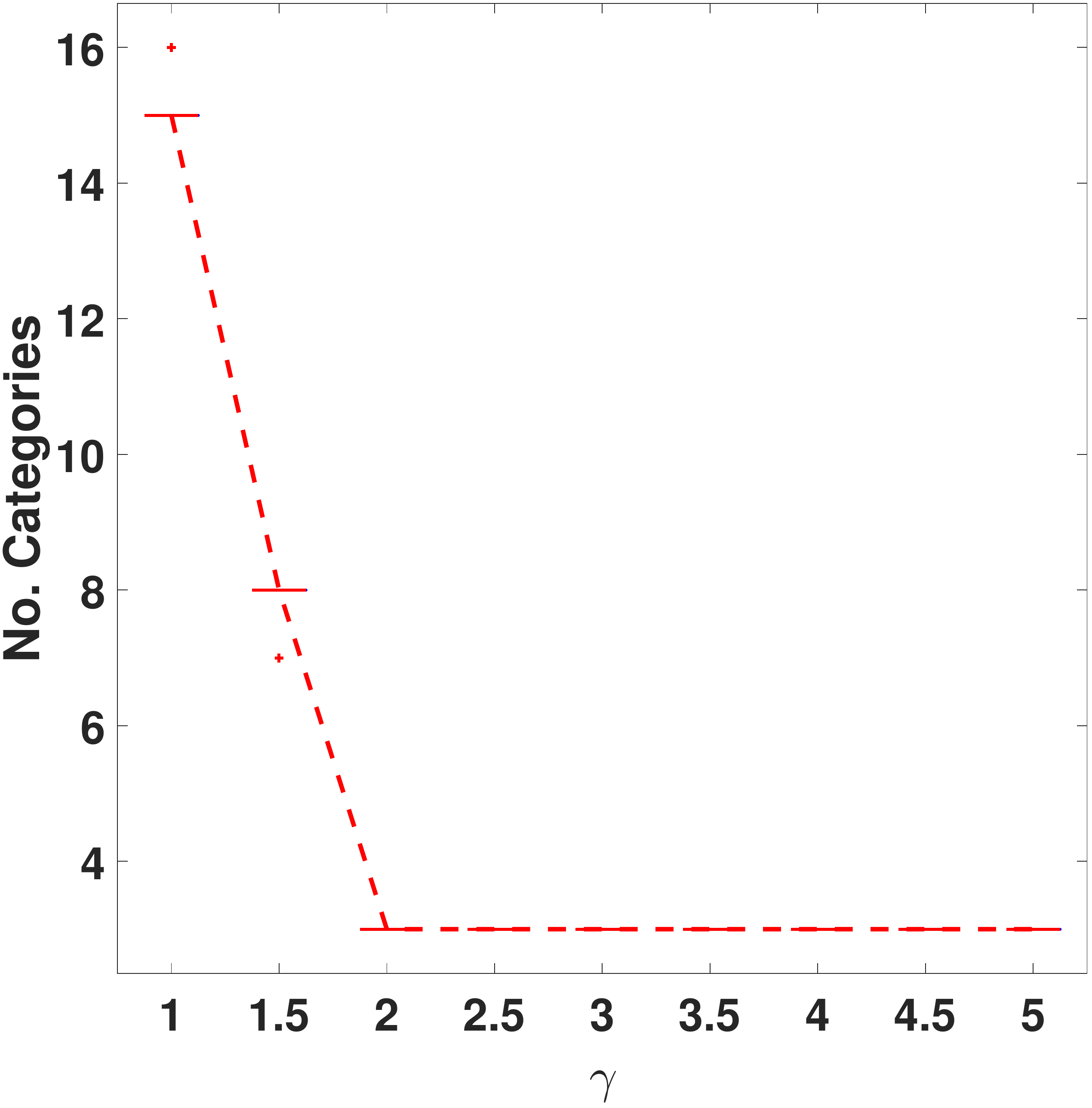} }
\subfloat[Moon]{\includegraphics[width=\gammaScale\textwidth]{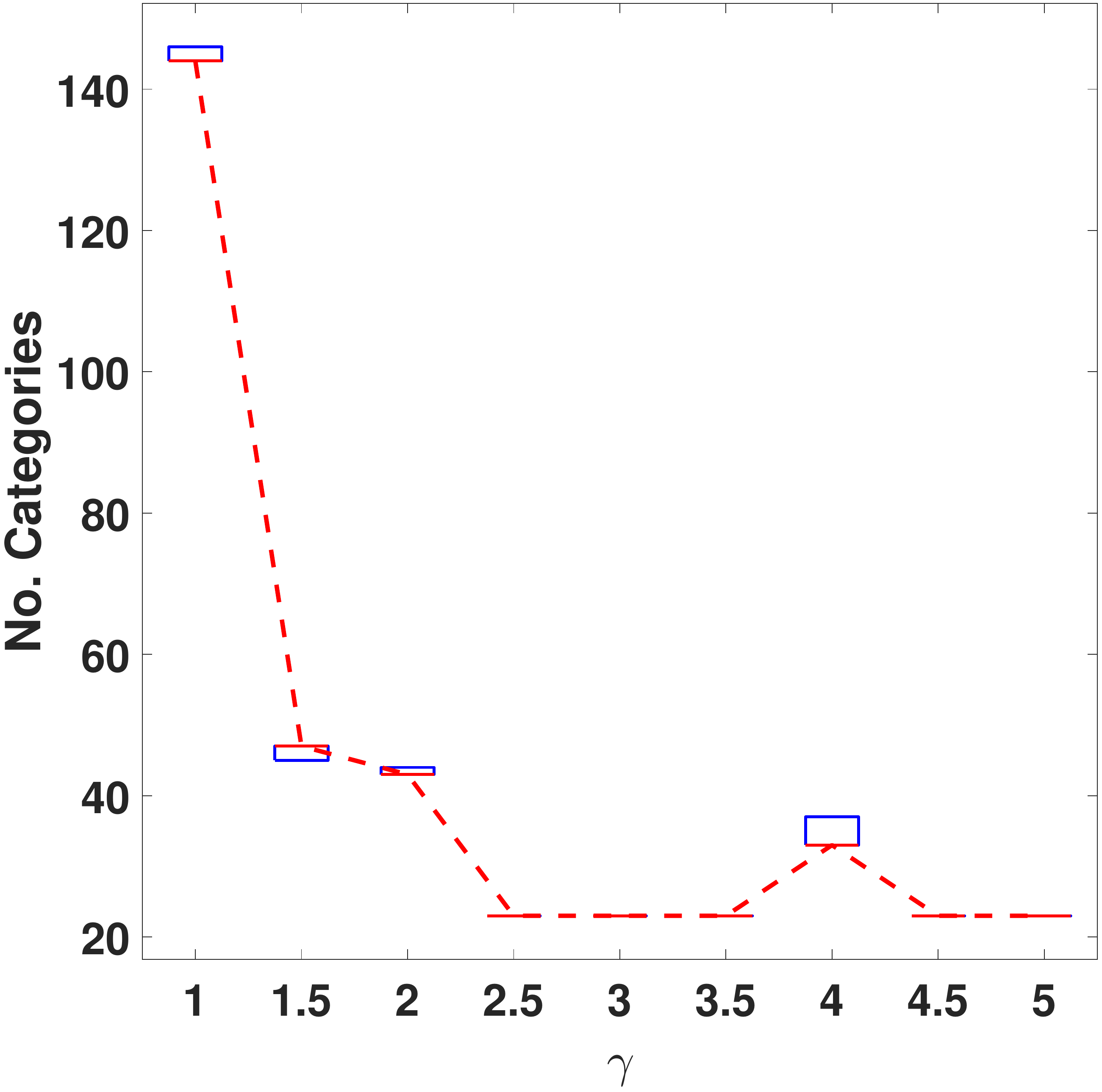} }
}
\caption{The behavior of the VAT + DDVFA system with respect to parameter $\gamma$ using the \textit{Seeds}, \textit{Wine}, \textit{Target}, \textit{Tetra}, \textit{Lsun}, and \textit{Moon} data sets: (a)-(f) peak average performance ($AR$), (g)-(l) number of clusters, and (m)-(r) total number of categories created. Both the number of clusters and categories are taken with respect to the most compact model that yields the depicted peak average performance (i.e., dual vigilance parameterization is \textit{not} held constant while varying parameter $\gamma$).}
\label{Fig:gamma_behavior_01}
\end{figure*}

\begin{figure*}[!ht]
\centerline{
\subfloat[Seeds]{\includegraphics[width=\gammaScale\textwidth]{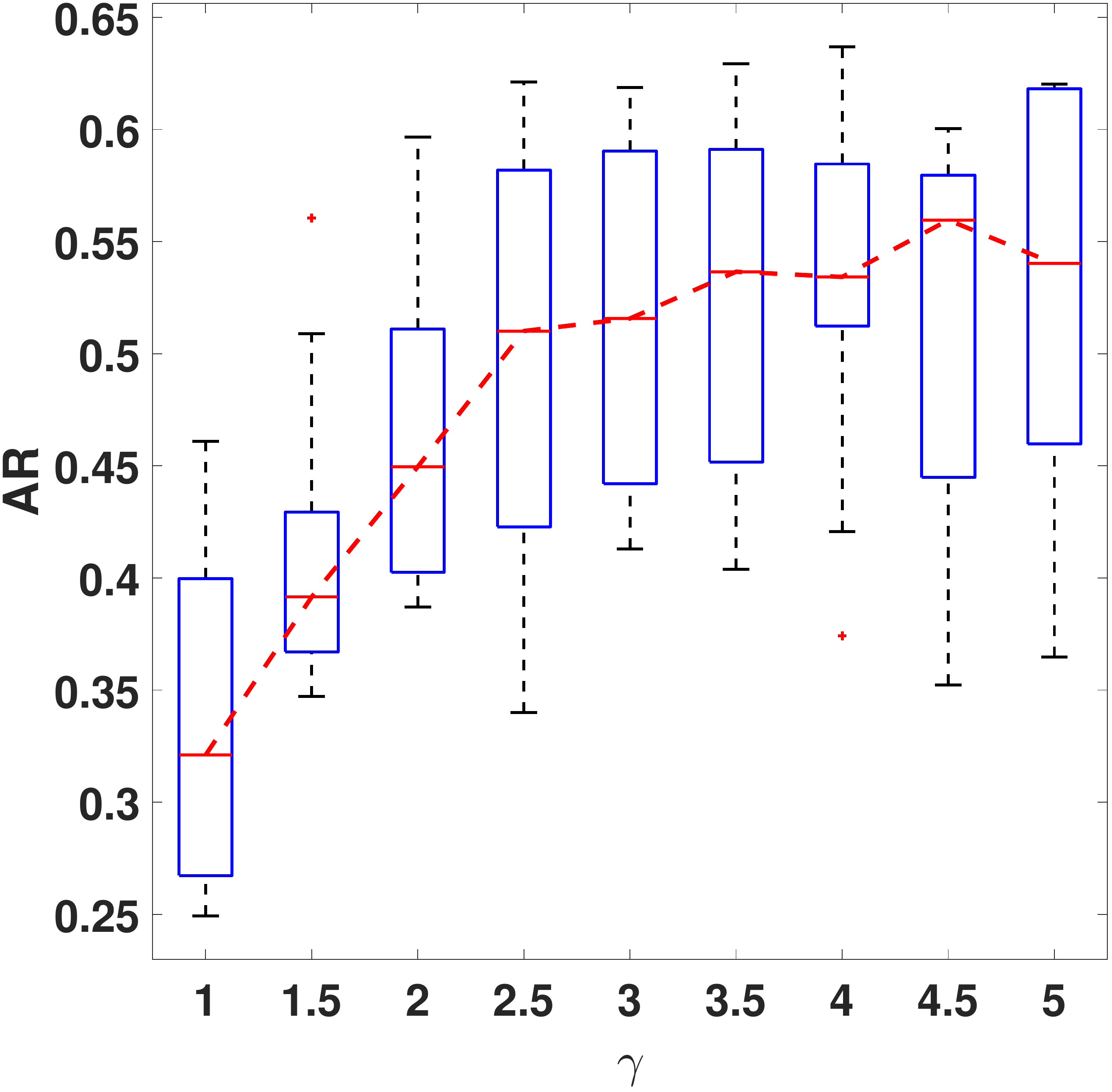} }
\subfloat[Wine]{\includegraphics[width=\gammaScale\textwidth]{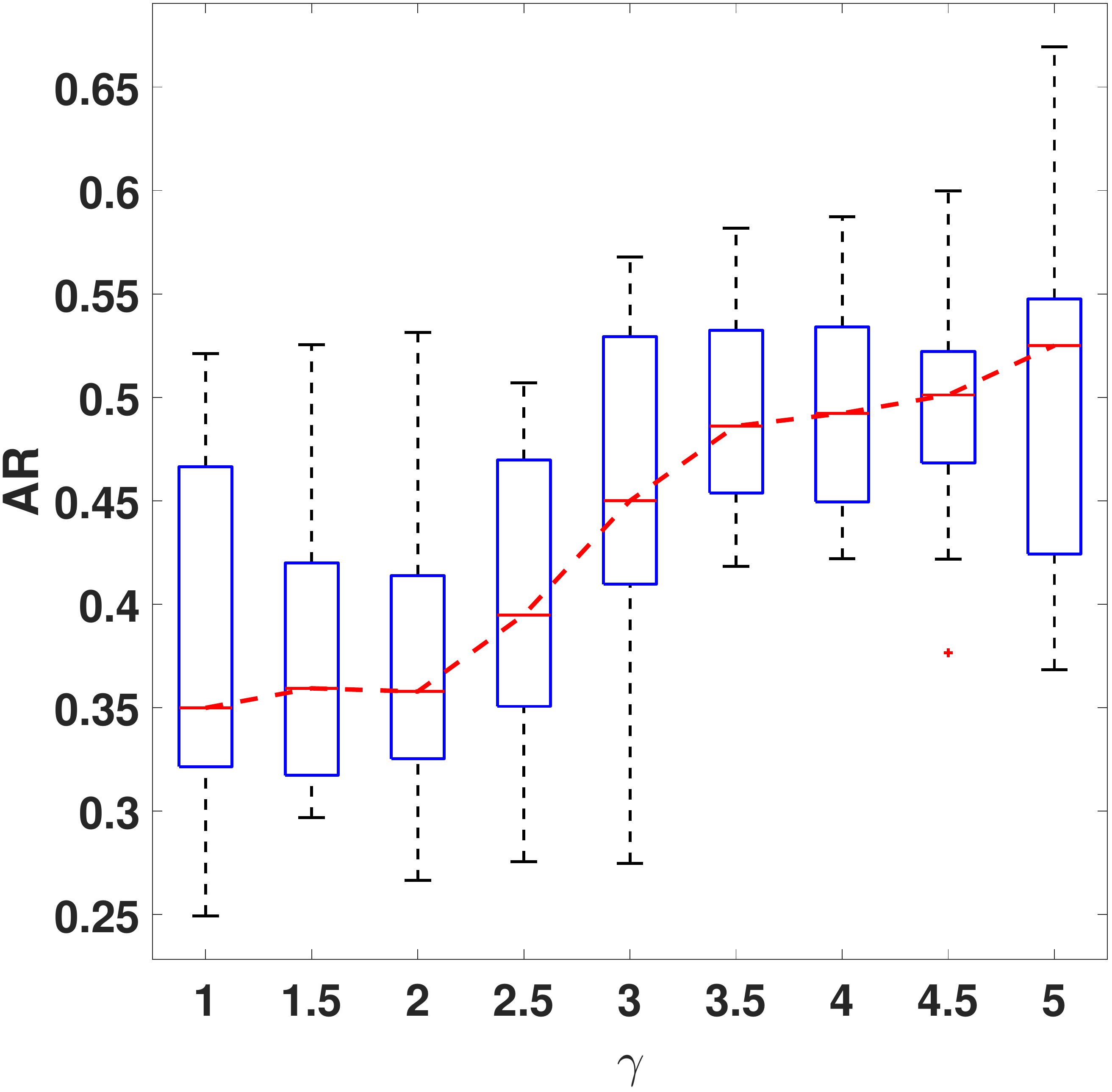} }
\subfloat[Target]{\includegraphics[width=\gammaScale\textwidth]{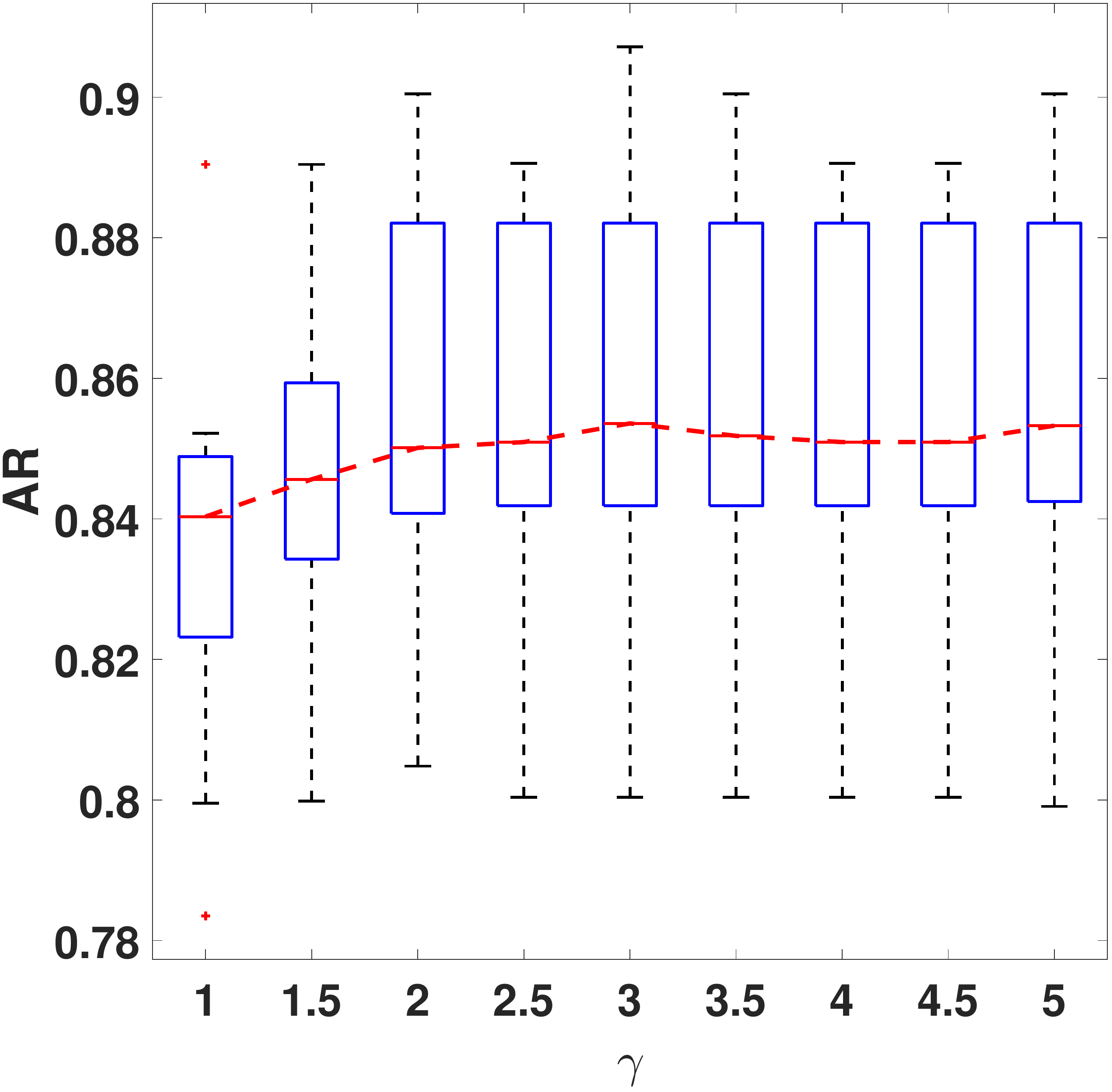} }
\subfloat[Tetra]{\includegraphics[width=\gammaScale\textwidth]{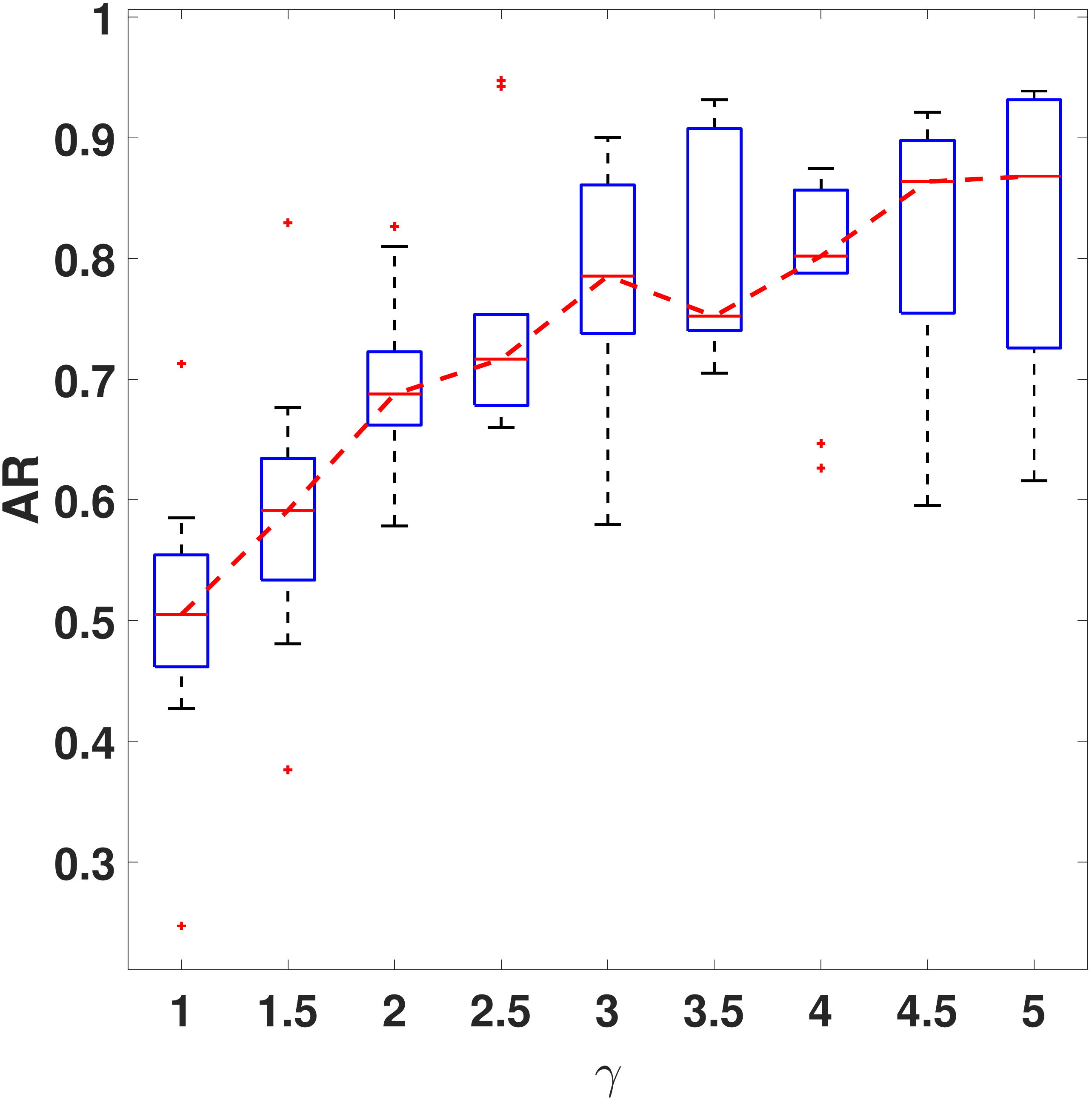} }
\subfloat[Lsun]{\includegraphics[width=\gammaScale\textwidth]{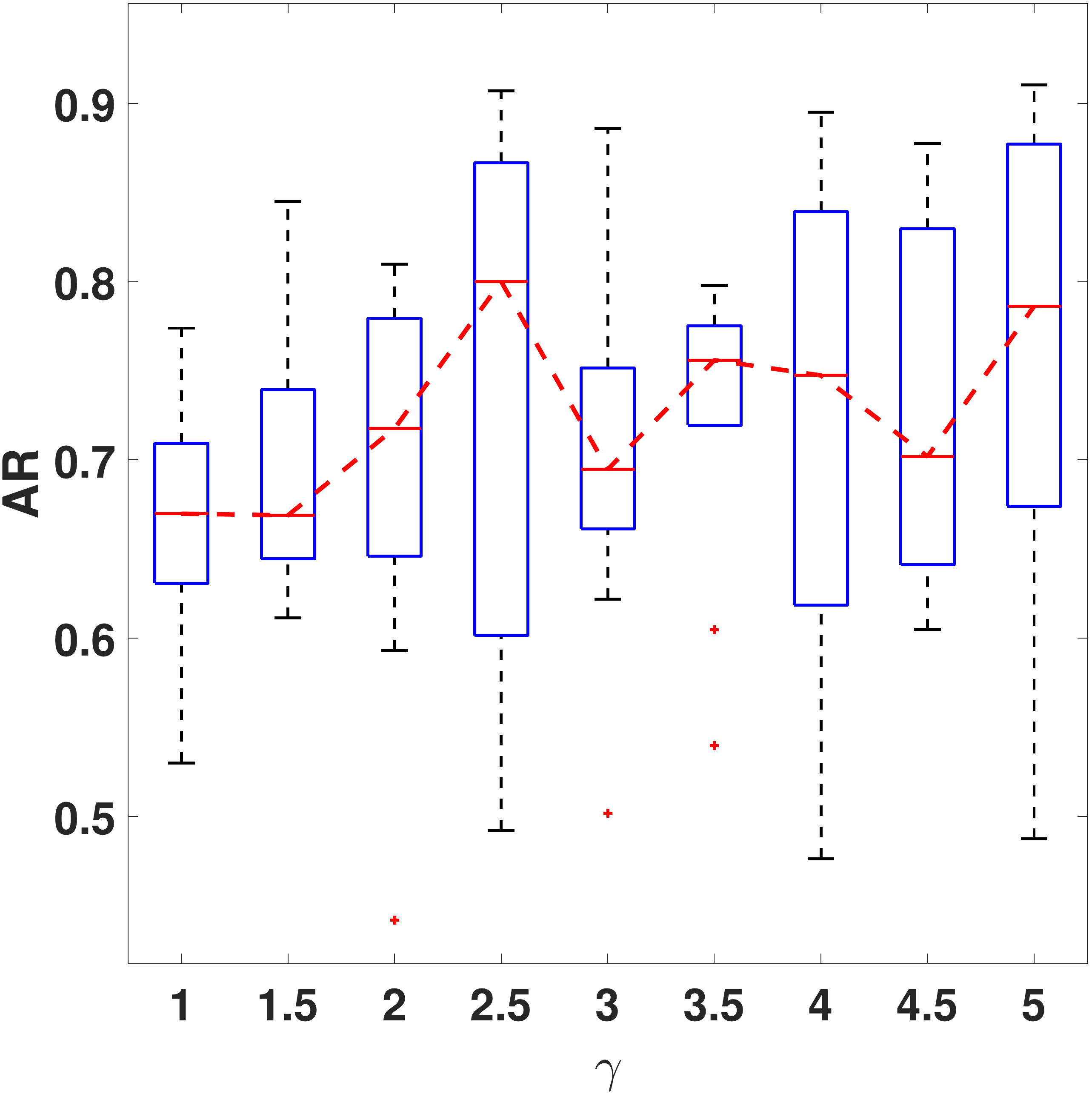} }
\subfloat[Moon]{\includegraphics[width=\gammaScale\textwidth]{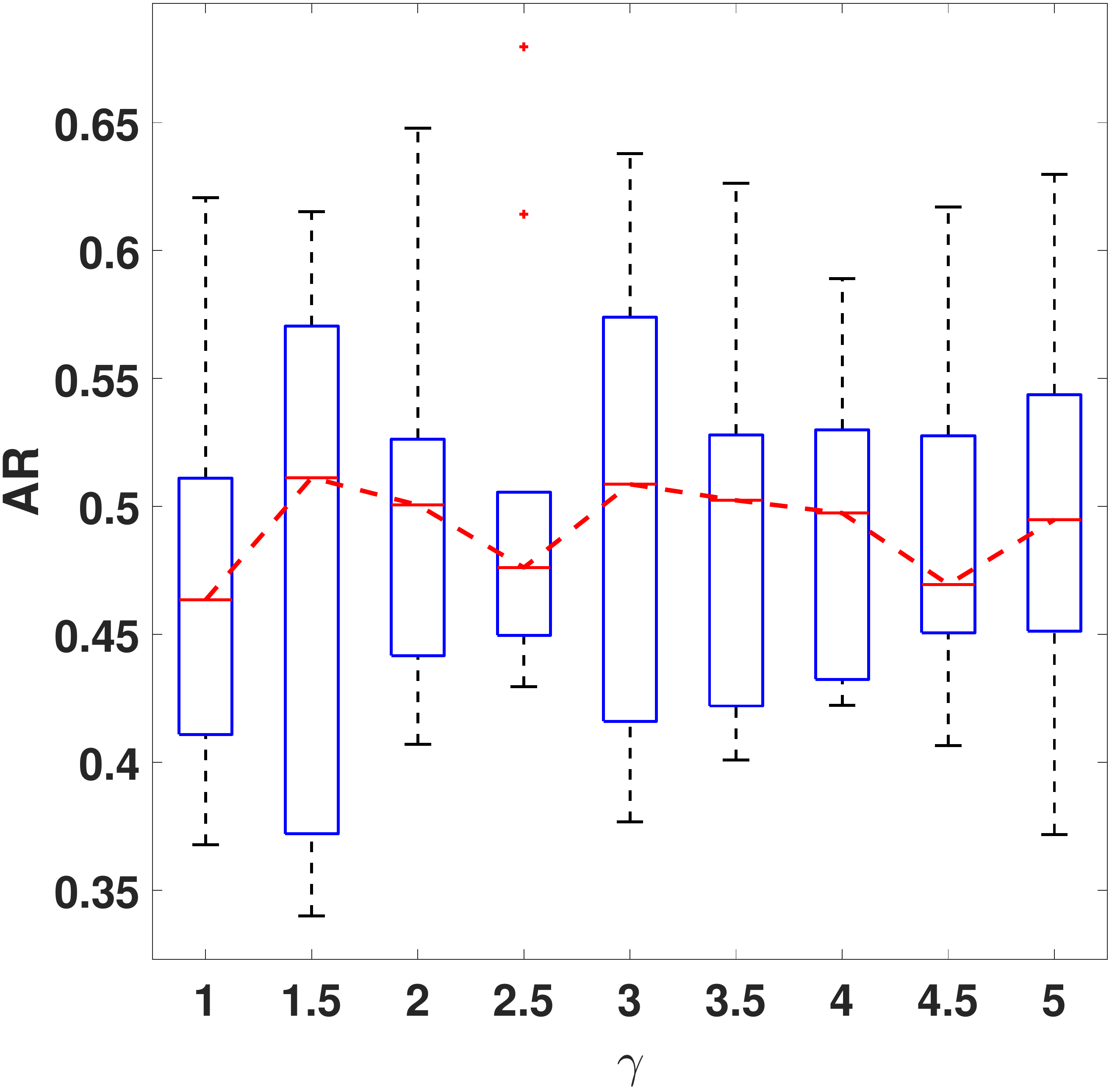} }
}
\vspace{-0.5\baselineskip}
\centerline{
\subfloat[Seeds]{\includegraphics[width=\gammaScale\textwidth]{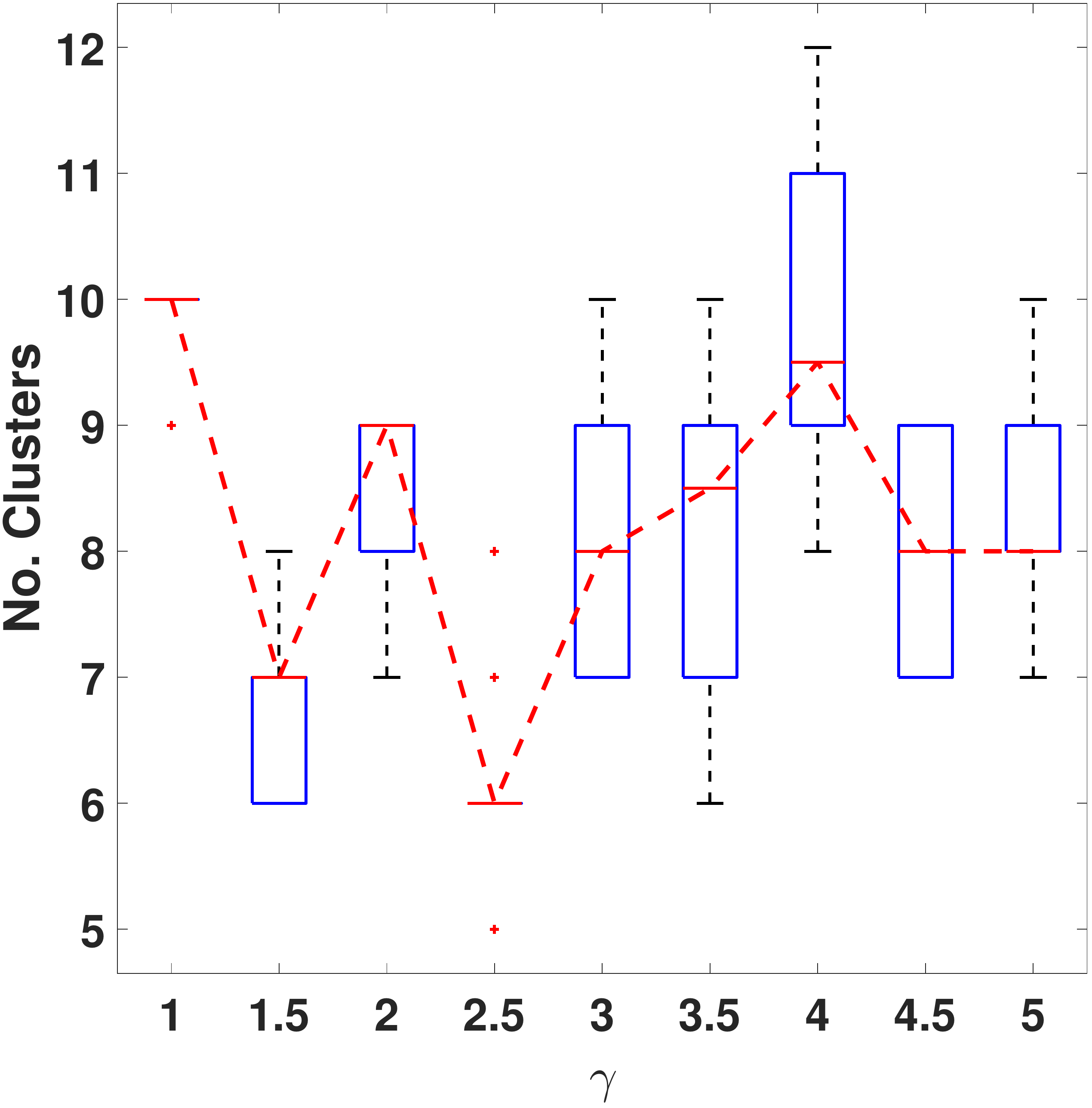} }
\subfloat[Wine]{\includegraphics[width=\gammaScale\textwidth]{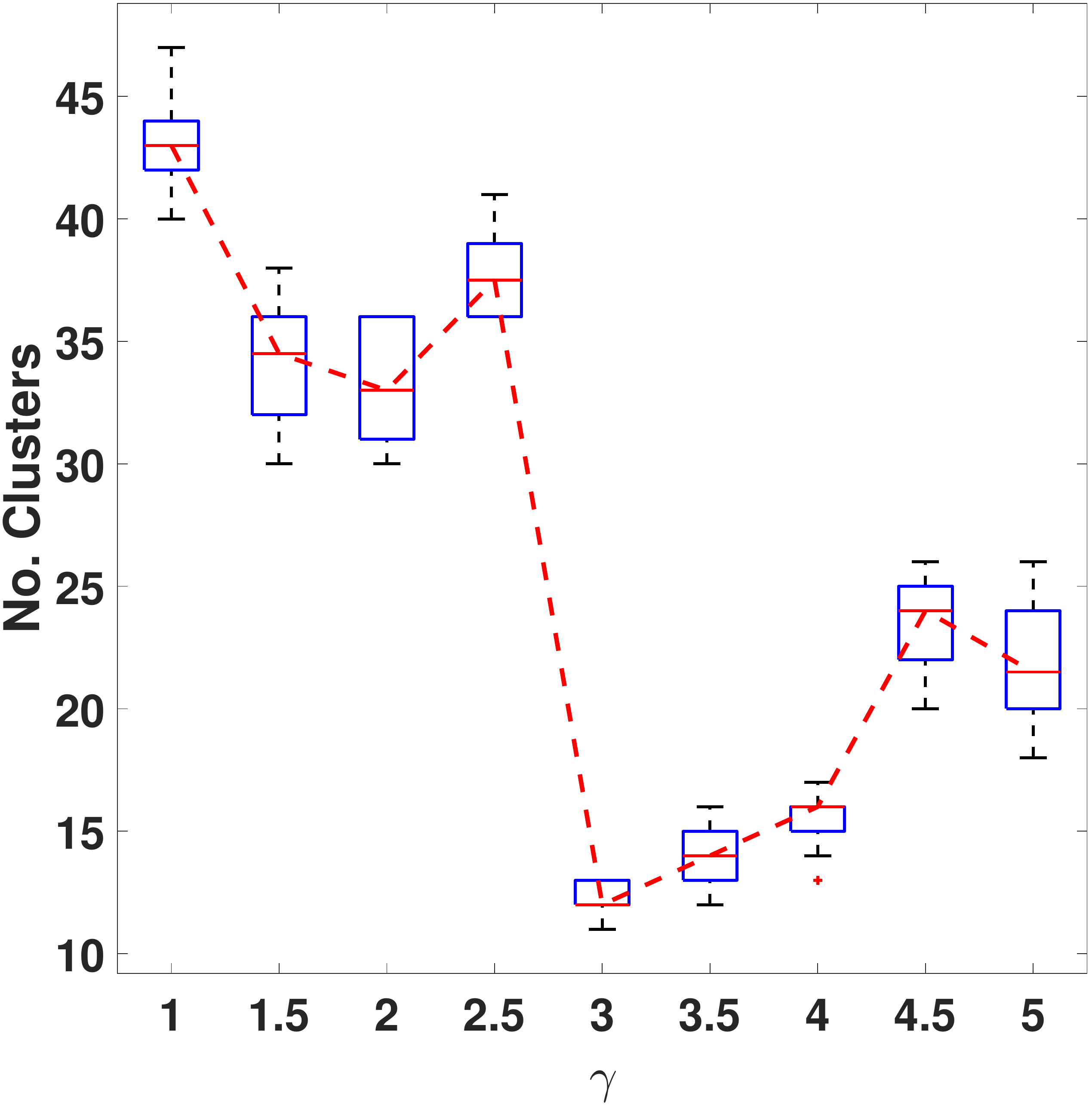} }
\subfloat[Target]{\includegraphics[width=\gammaScale\textwidth]{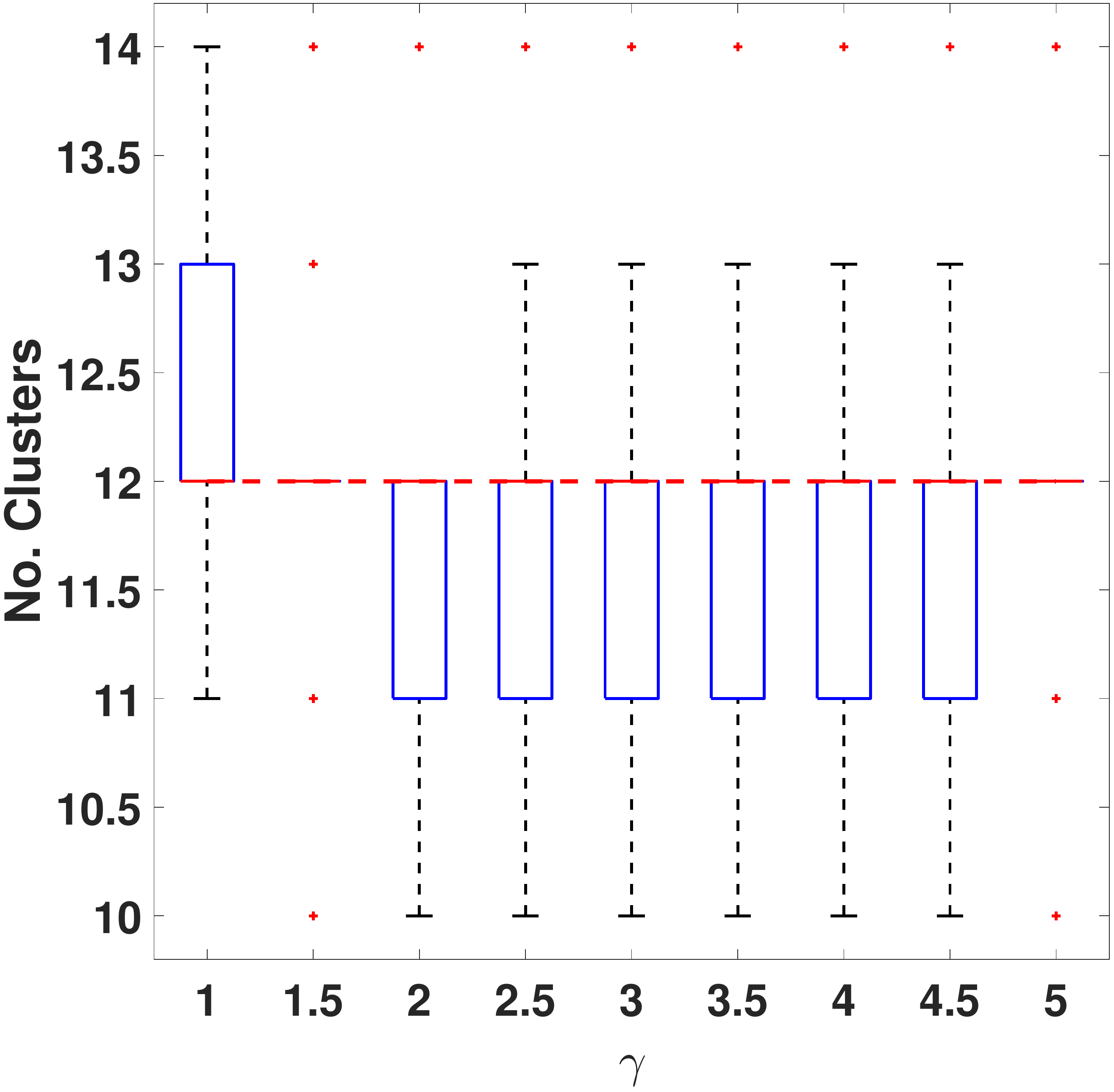} }
\subfloat[Tetra]{\includegraphics[width=\gammaScale\textwidth]{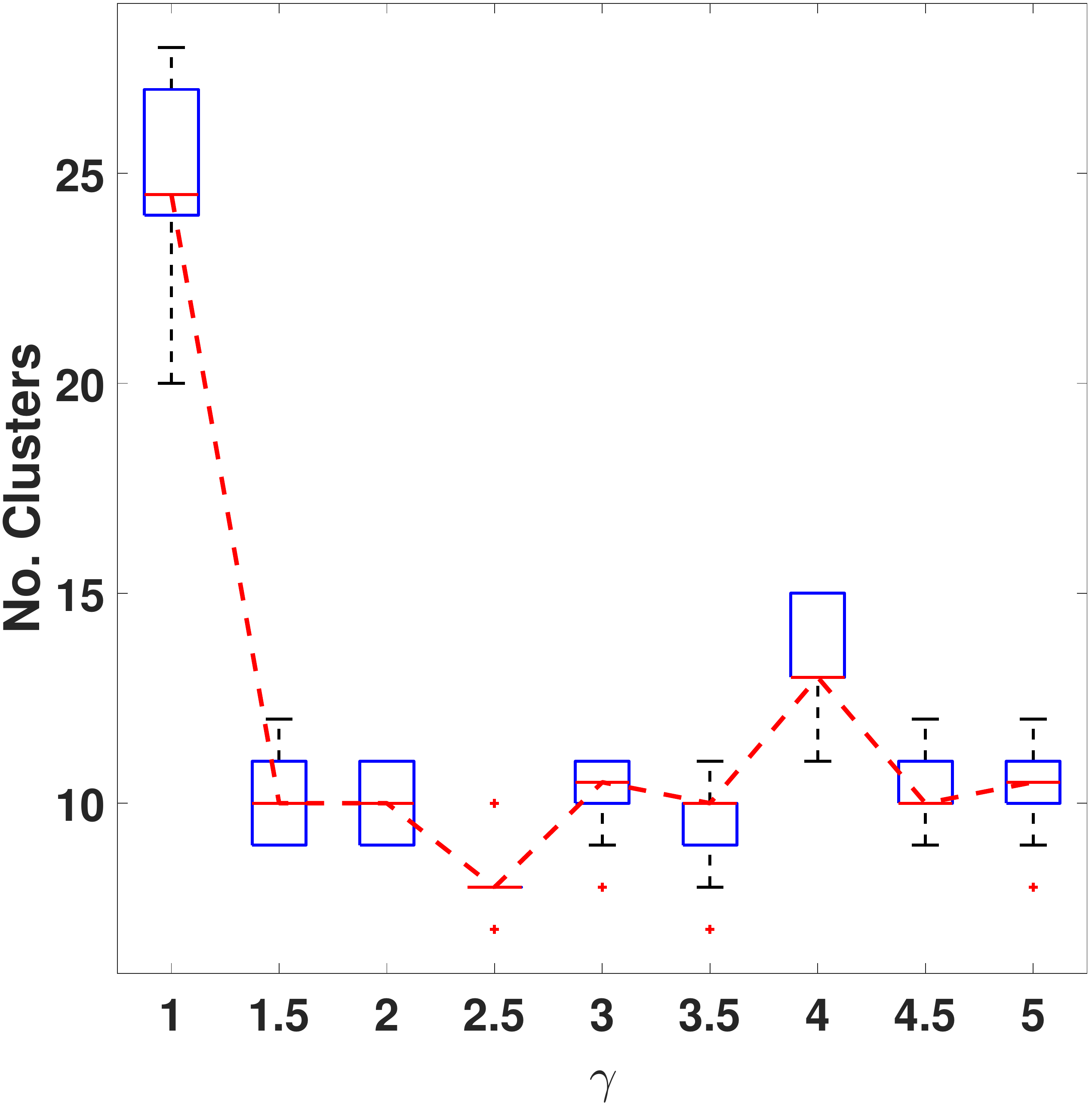} }
\subfloat[Lsun]{\includegraphics[width=\gammaScale\textwidth]{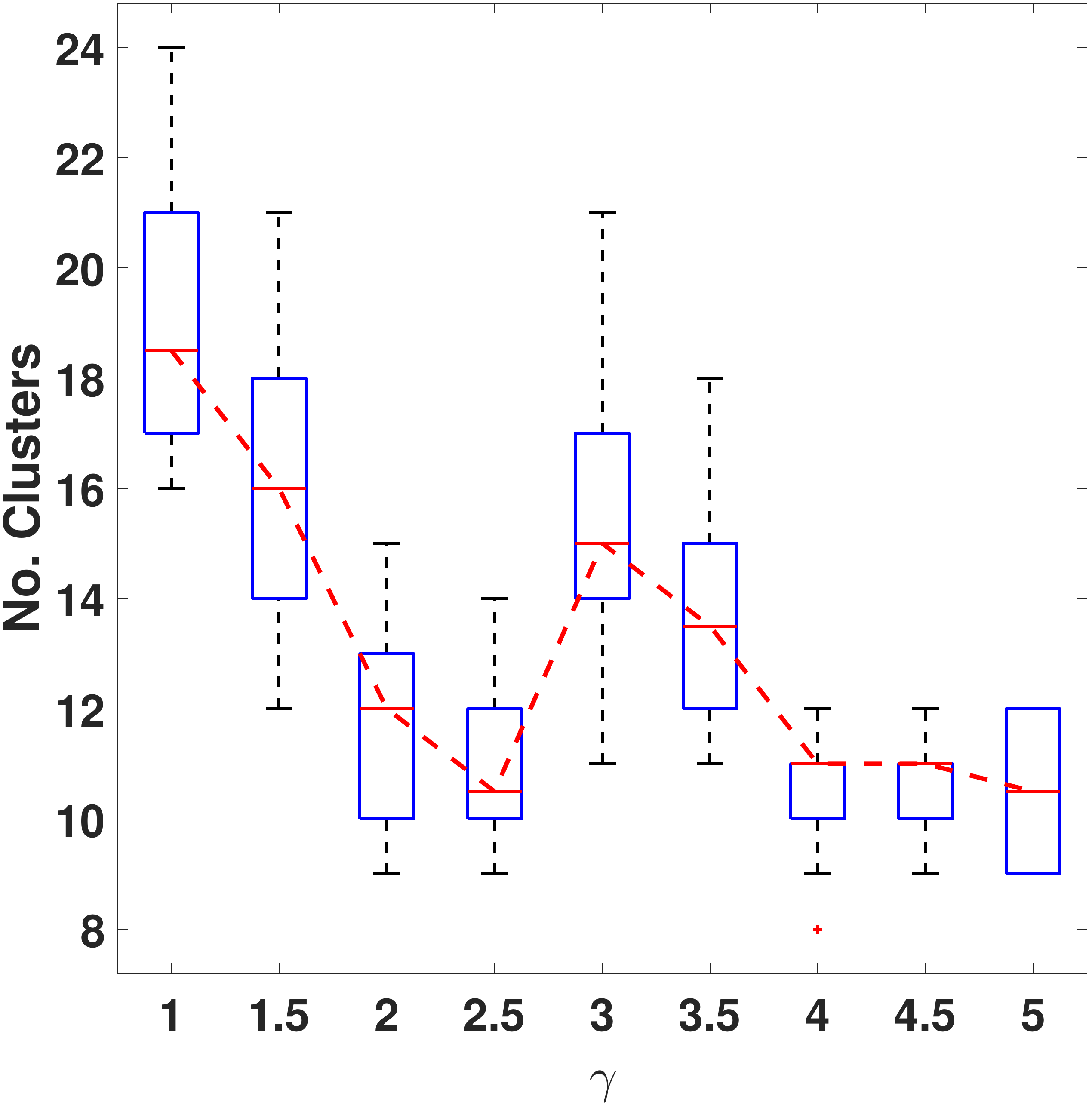} }
\subfloat[Moon]{\includegraphics[width=\gammaScale\textwidth]{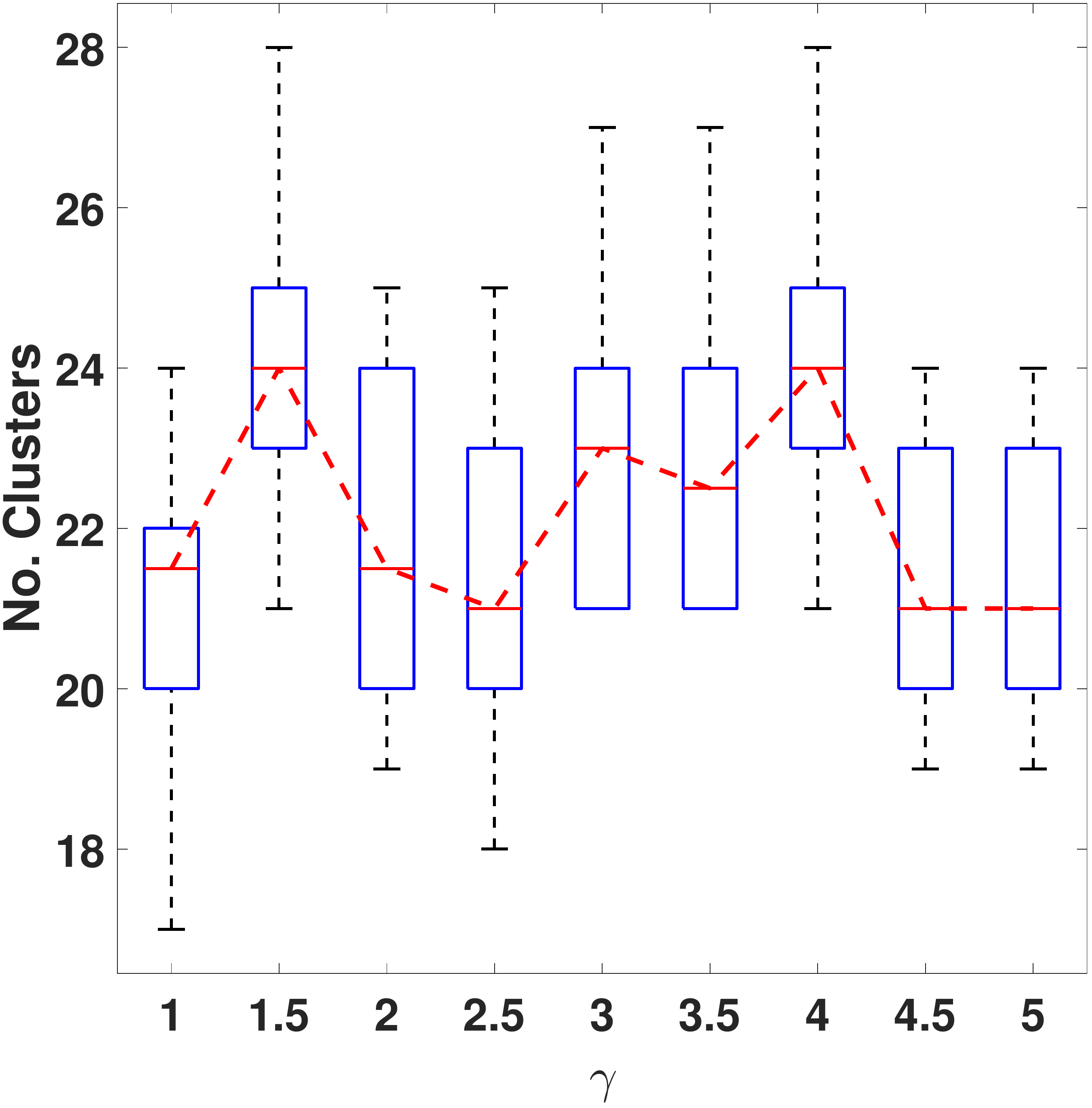} }
}
\vspace{-0.5\baselineskip}
\centerline{
\subfloat[Seeds]{\includegraphics[width=\gammaScale\textwidth]{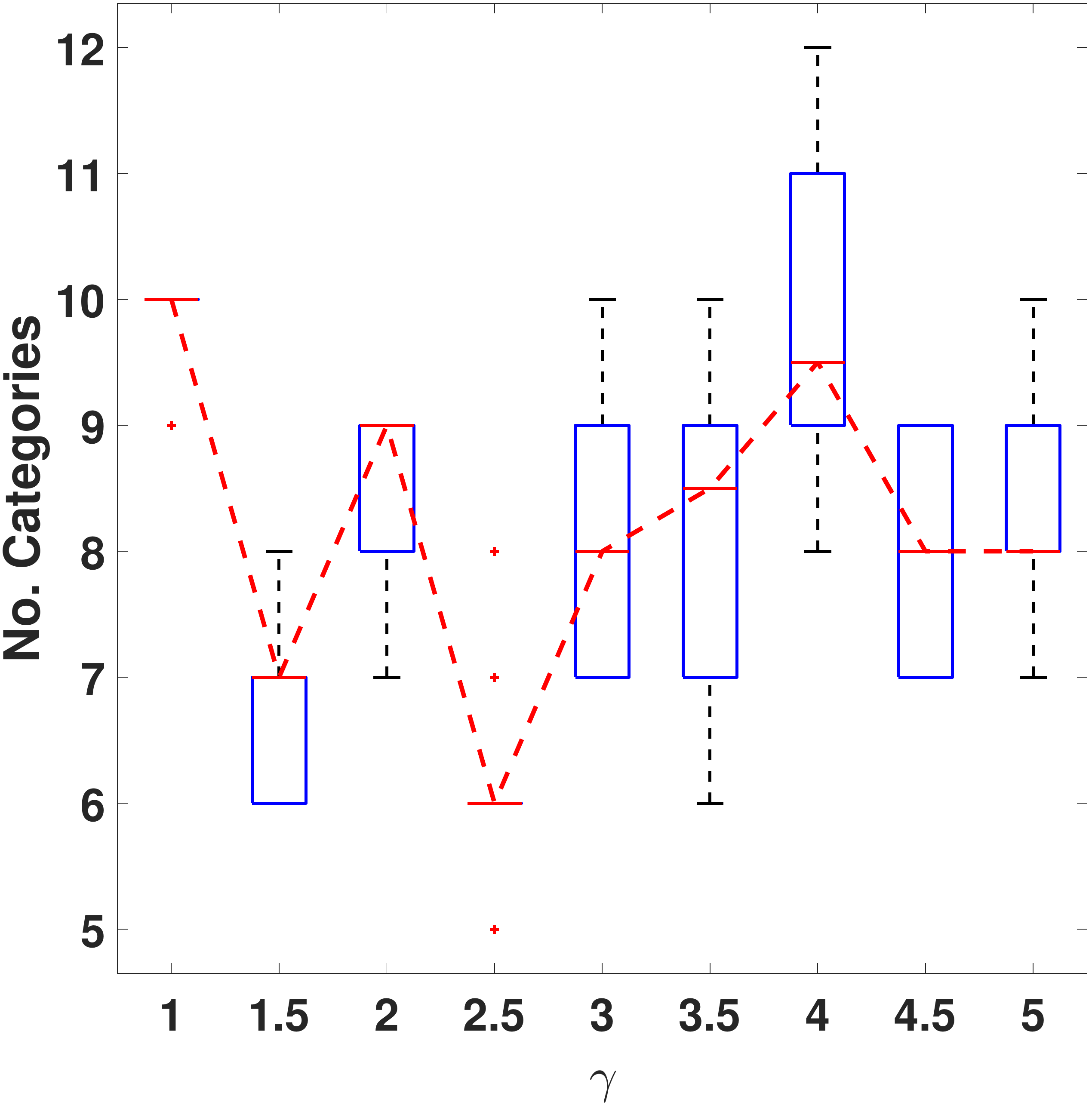} }
\subfloat[Wine]{\includegraphics[width=\gammaScale\textwidth]{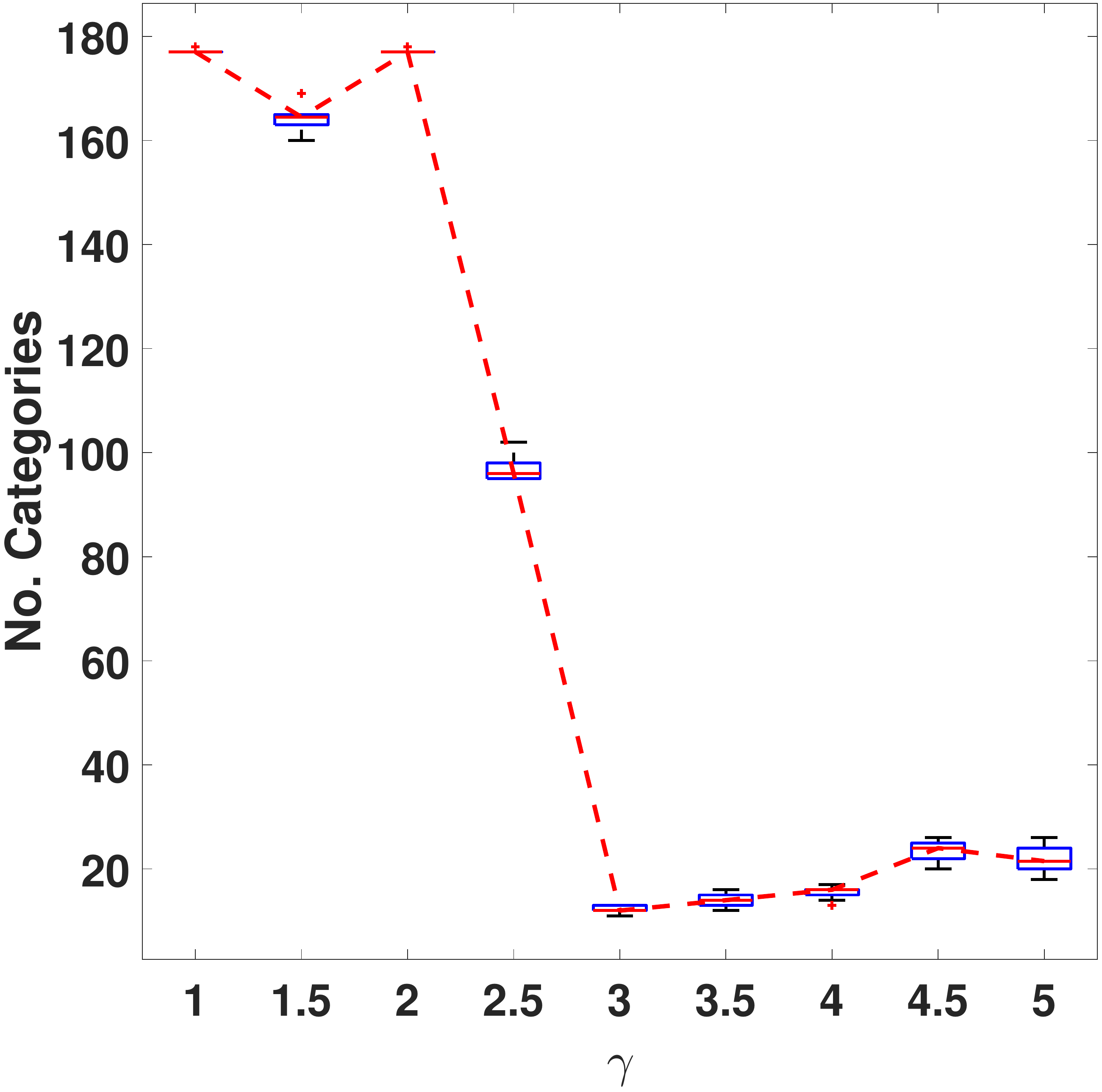} }
\subfloat[Target]{\includegraphics[width=\gammaScale\textwidth]{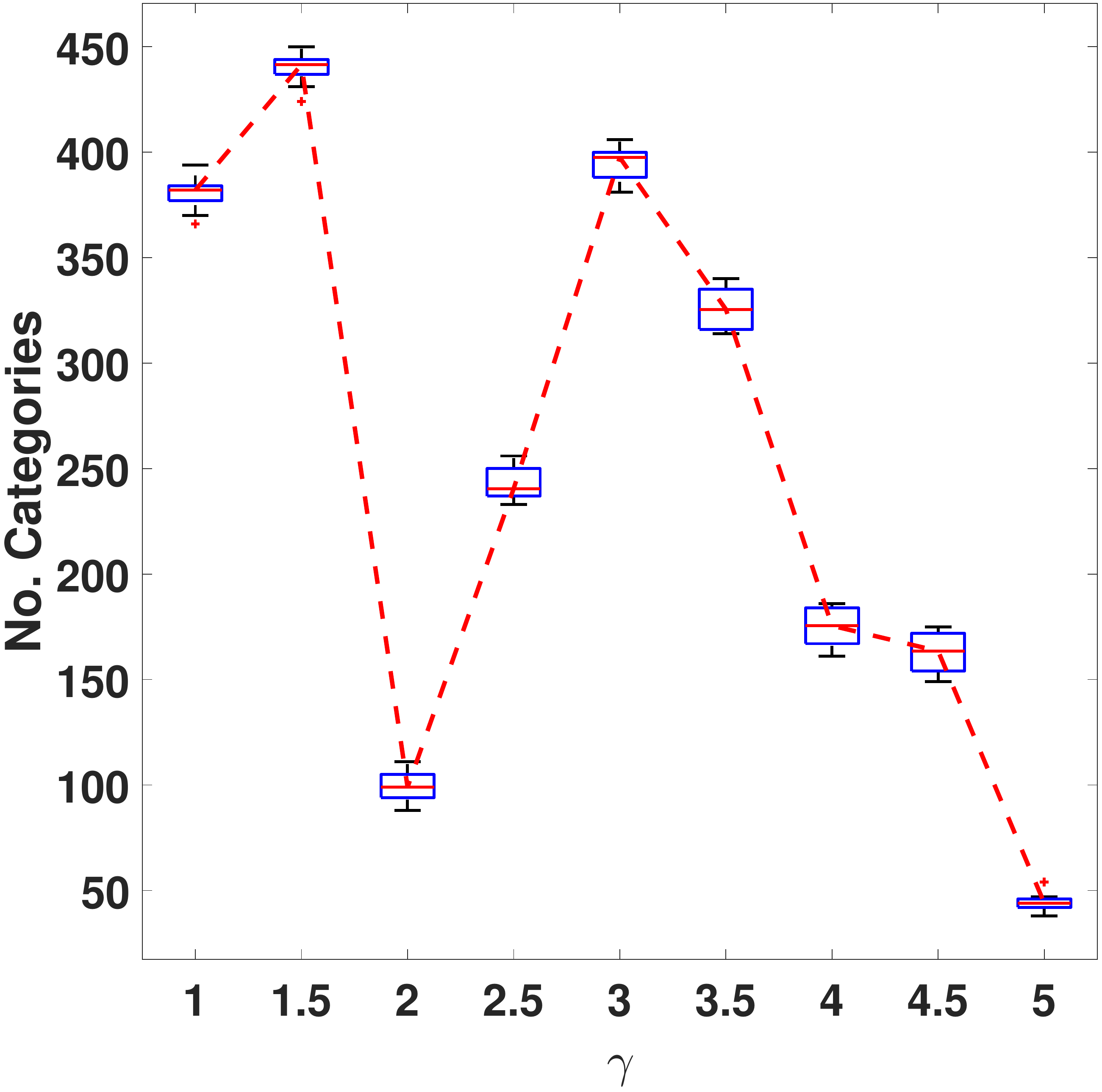} }
\subfloat[Tetra]{\includegraphics[width=\gammaScale\textwidth]{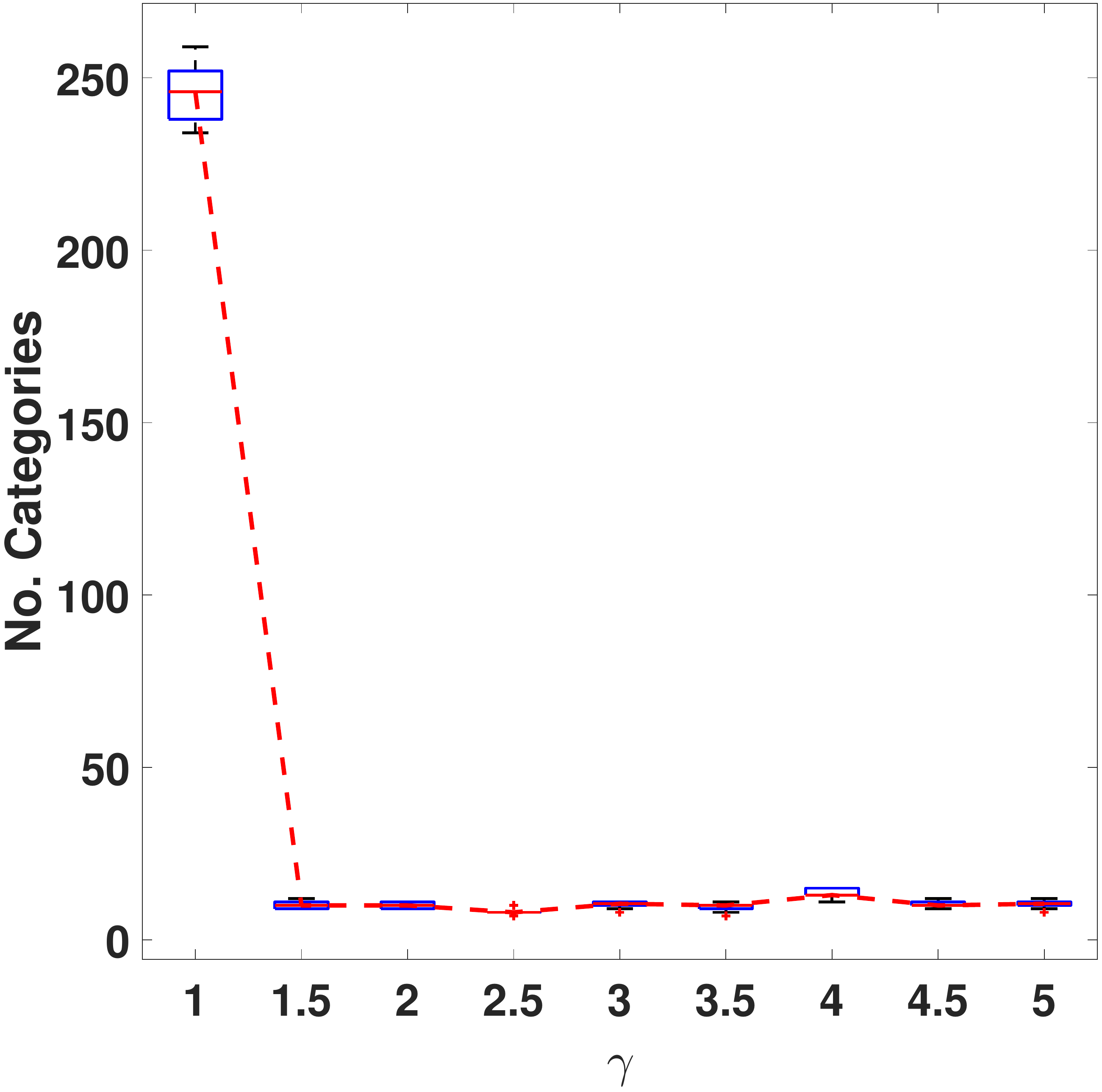} }
\subfloat[Lsun]{\includegraphics[width=\gammaScale\textwidth]{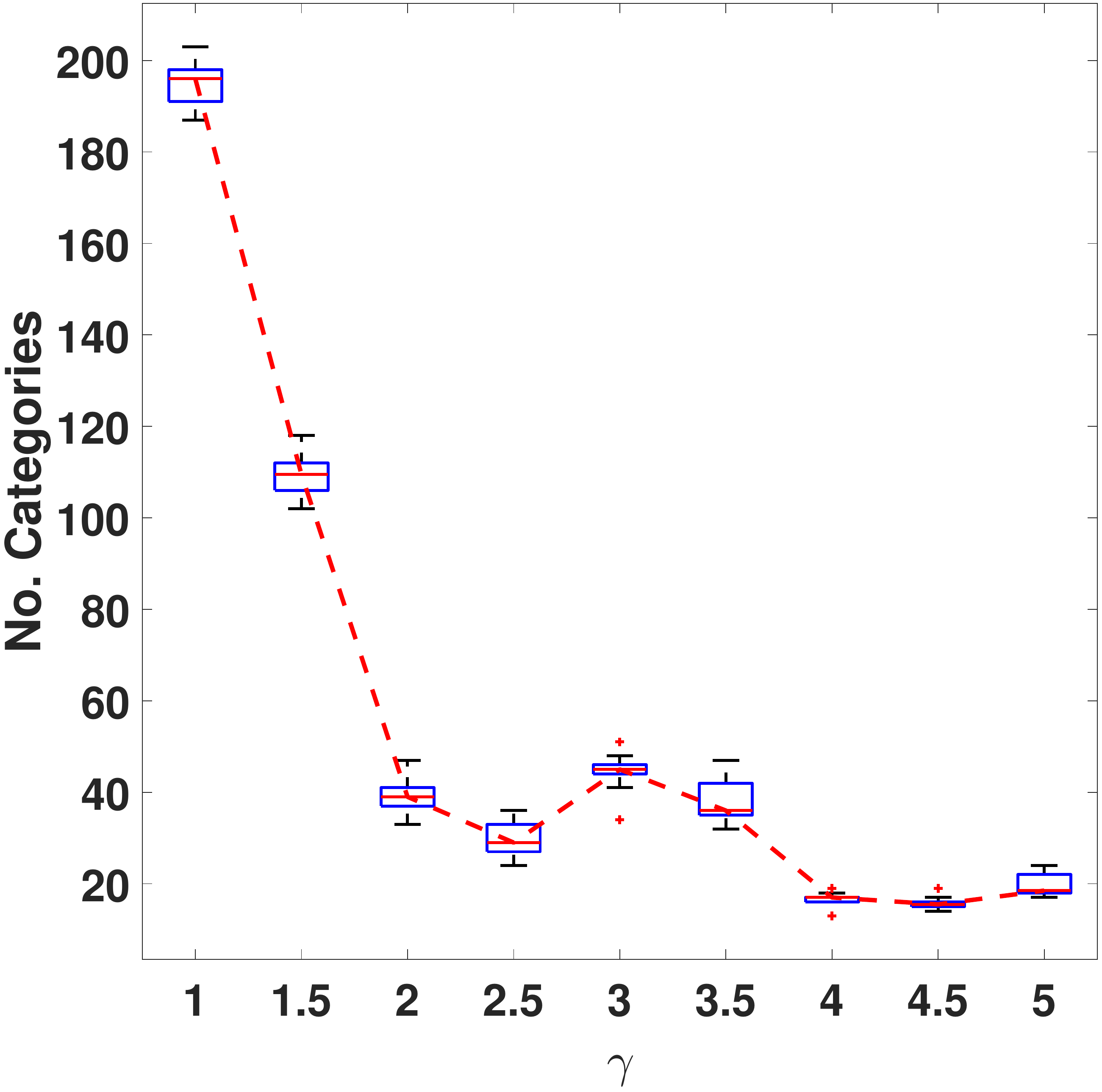} }
\subfloat[Moon]{\includegraphics[width=\gammaScale\textwidth]{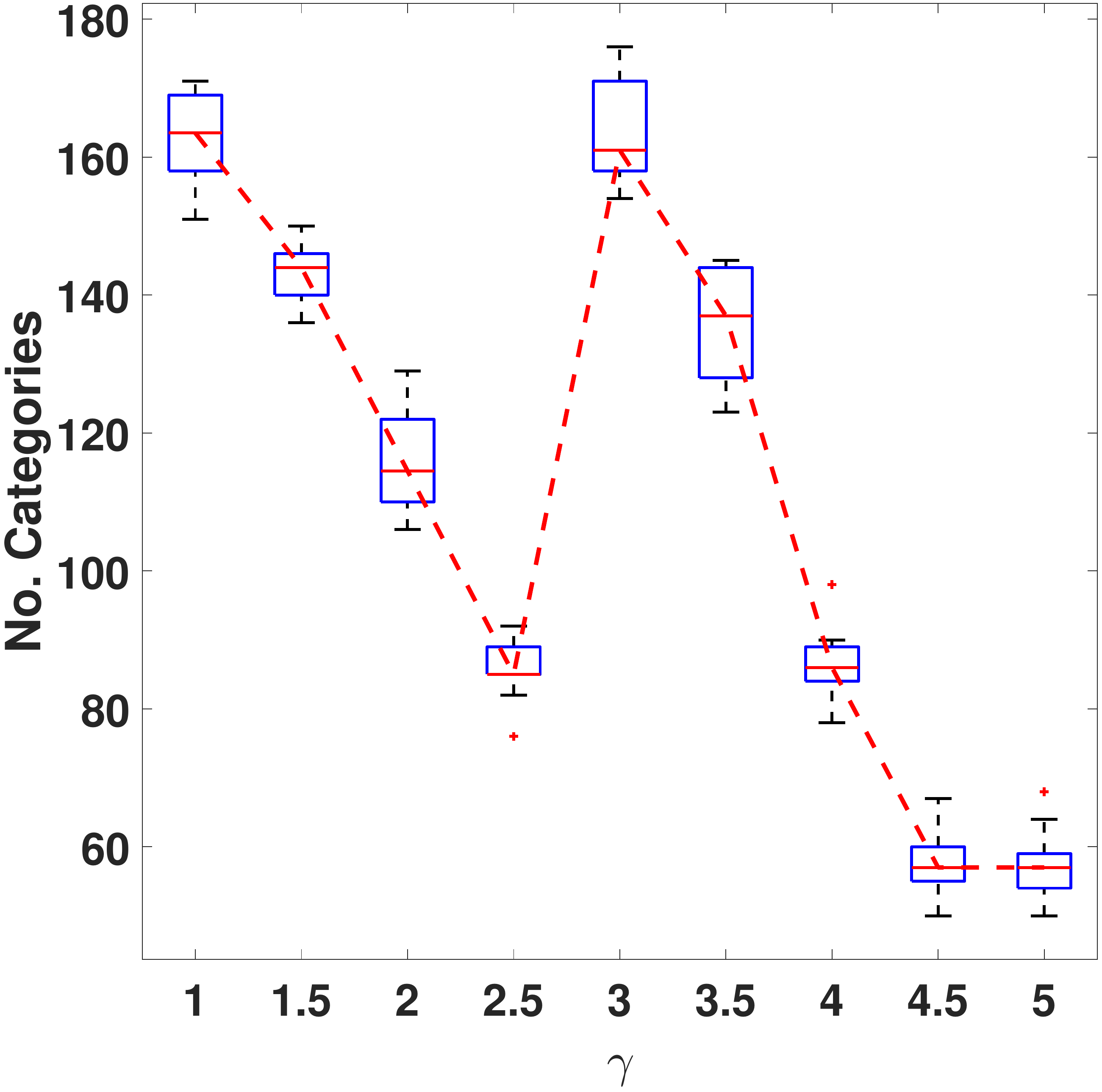} }
}
\caption{The behavior of the DDVFA system with respect to parameter $\gamma$ using the \textit{Seeds}, \textit{Wine}, \textit{Target}, \textit{Tetra}, \textit{Lsun}, and \textit{Moon} data sets: (a)-(f) peak average performance ($AR$), (g)-(l) number of clusters, and (m)-(r) total number of categories created. Both the number of clusters and categories are taken with respect to the most compact model that yields the depicted peak average performance (i.e., dual vigilance parameterization is \textit{not} held constant while varying parameter $\gamma$).}
\label{Fig:gamma_behavior_02}
\end{figure*}

\begin{figure*}[!ht]
\centerline{
\subfloat[Seeds]{\includegraphics[width=\gammaScale\textwidth]{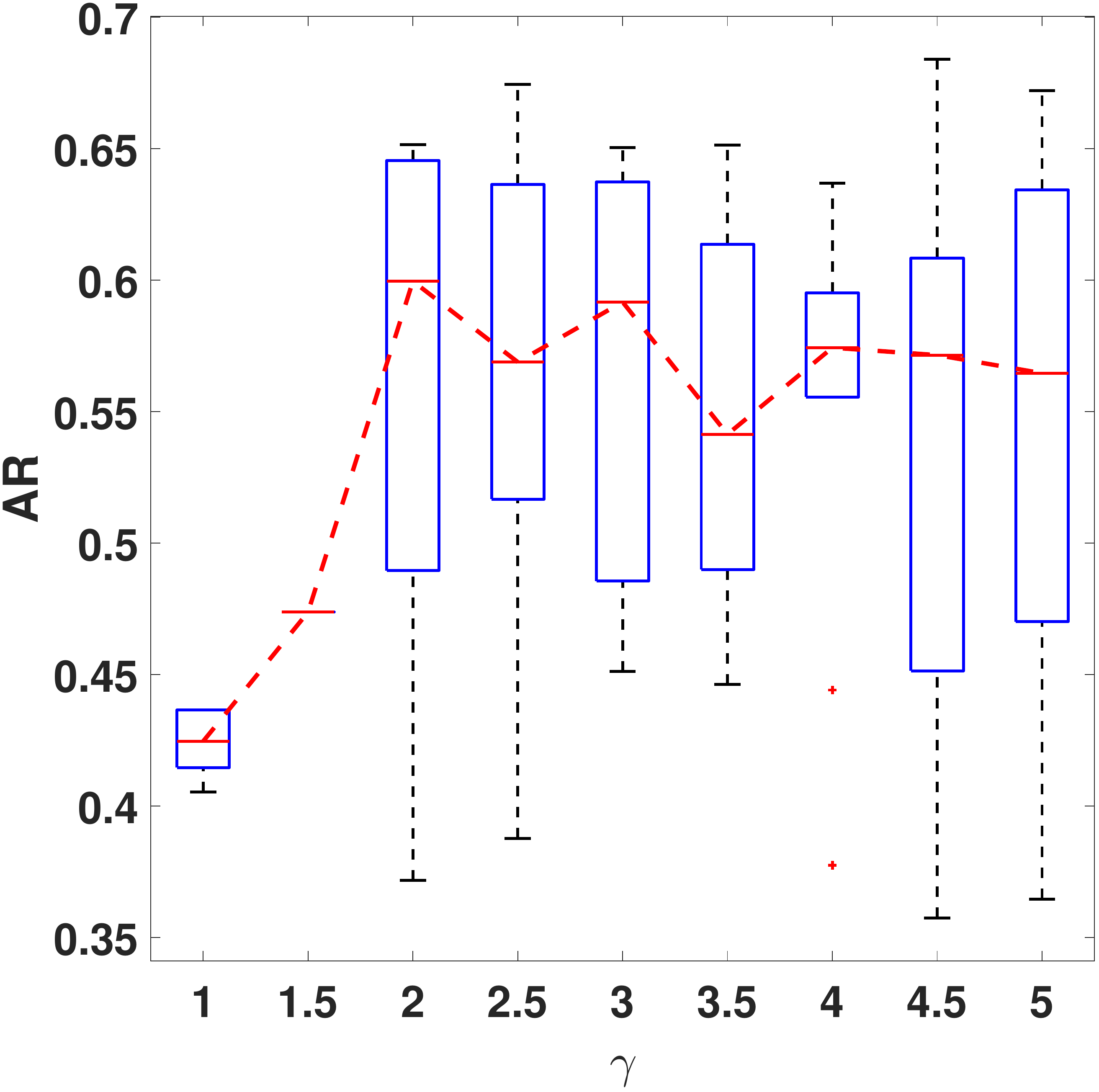} }
\subfloat[Wine]{\includegraphics[width=\gammaScale\textwidth]{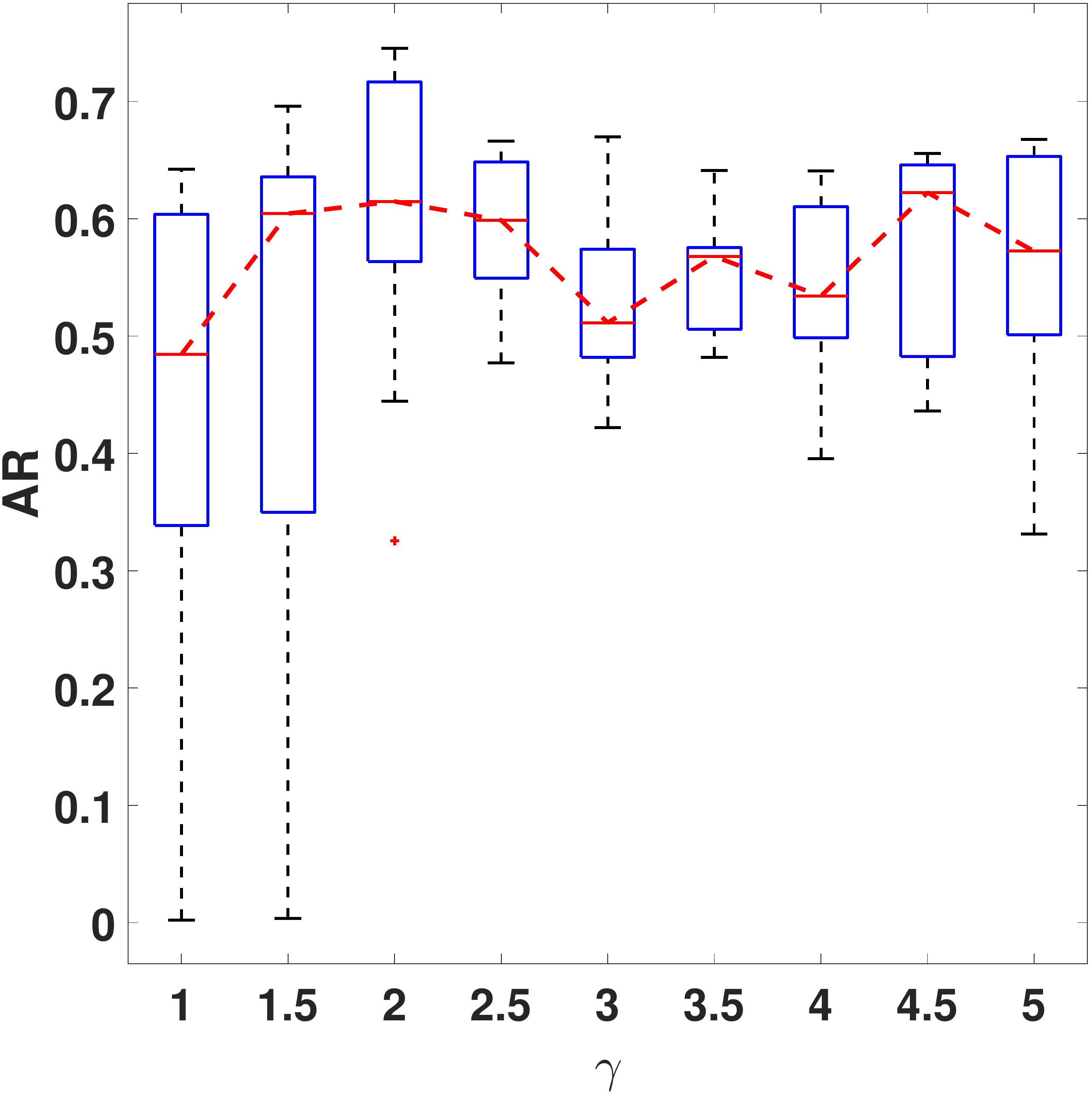} }
\subfloat[Target]{\includegraphics[width=\gammaScale\textwidth]{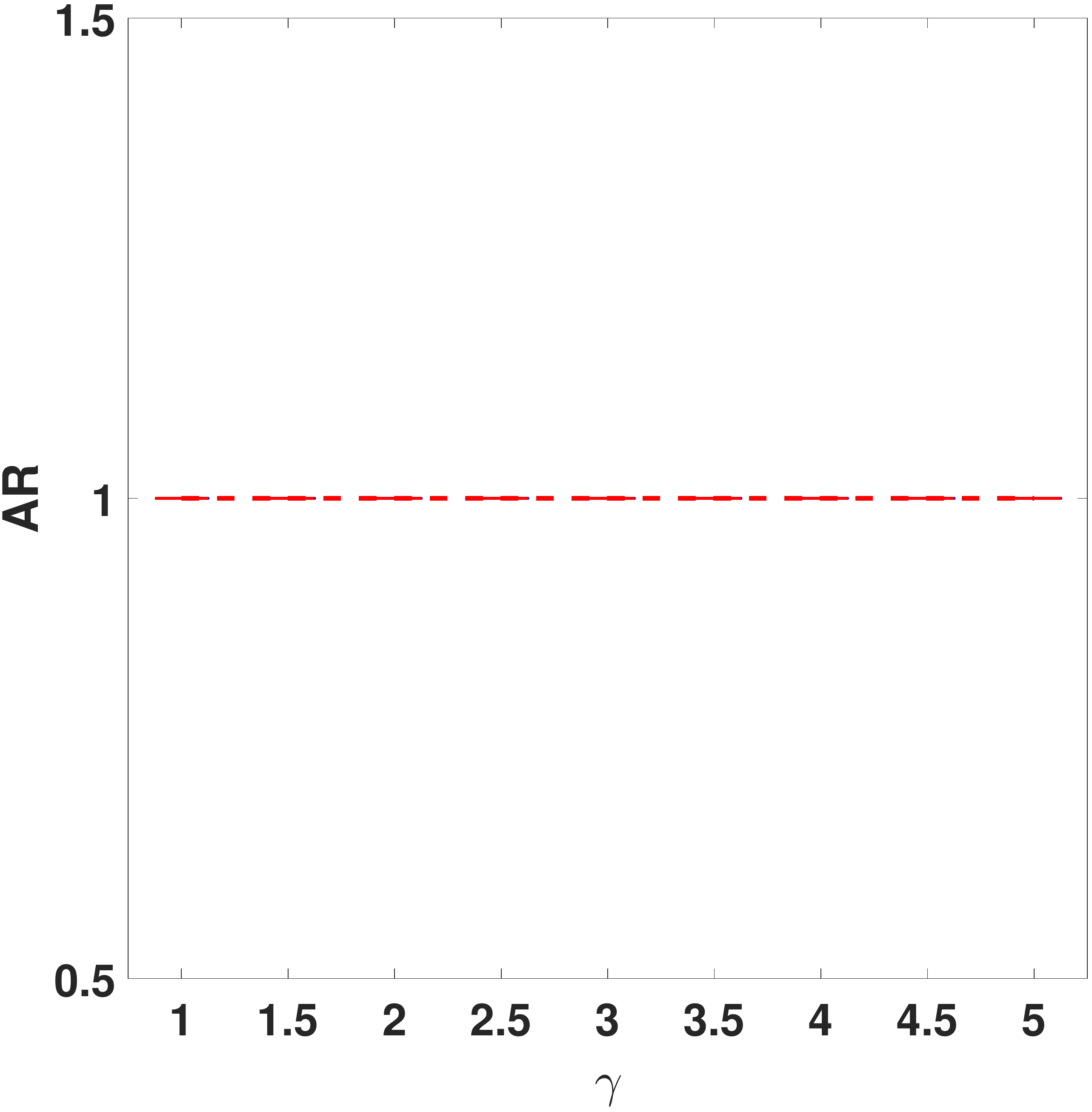} }
\subfloat[Tetra]{\includegraphics[width=\gammaScale\textwidth]{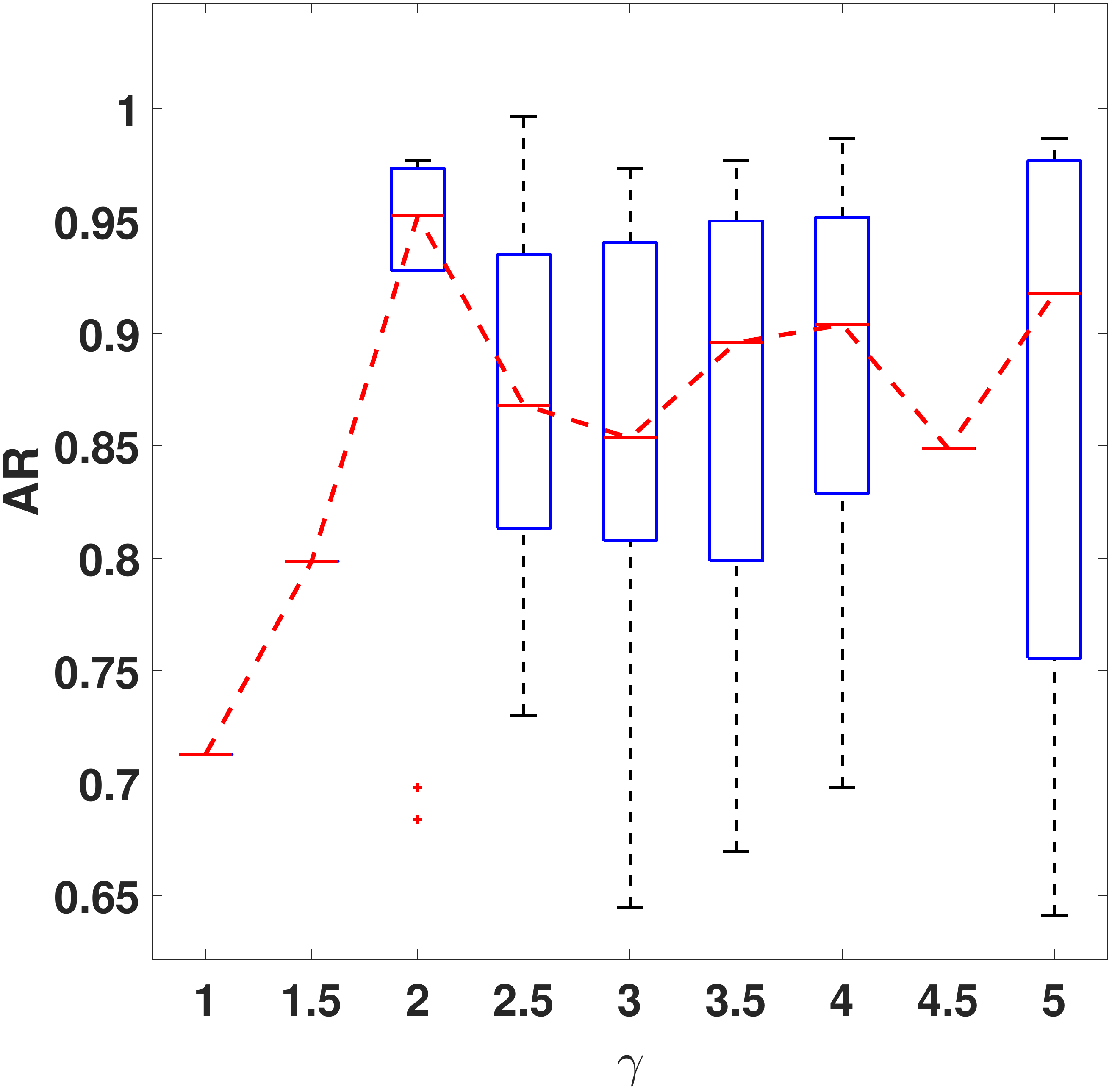} }
\subfloat[Lsun]{\includegraphics[width=\gammaScale\textwidth]{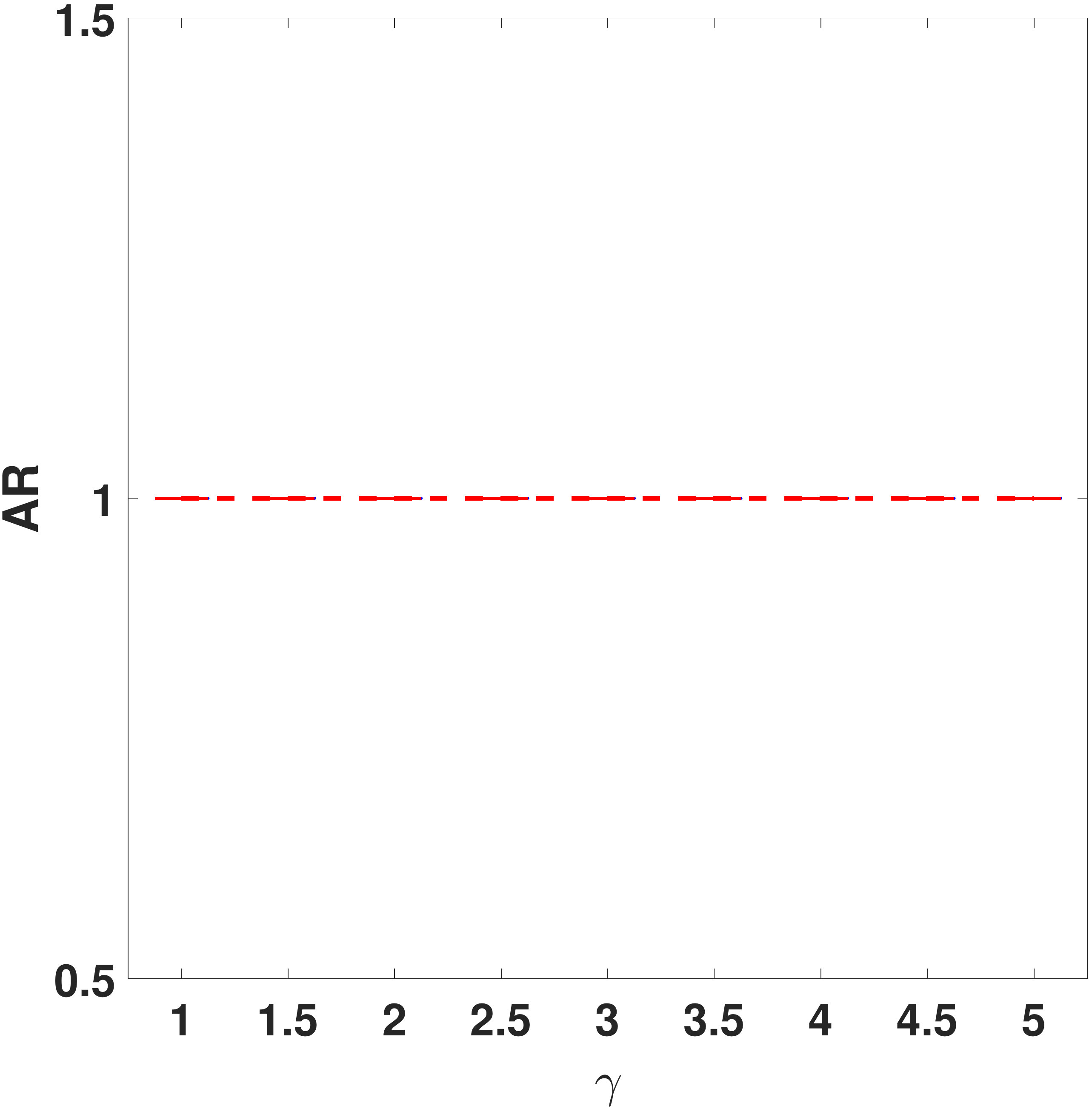} }
\subfloat[Moon]{\includegraphics[width=\gammaScale\textwidth]{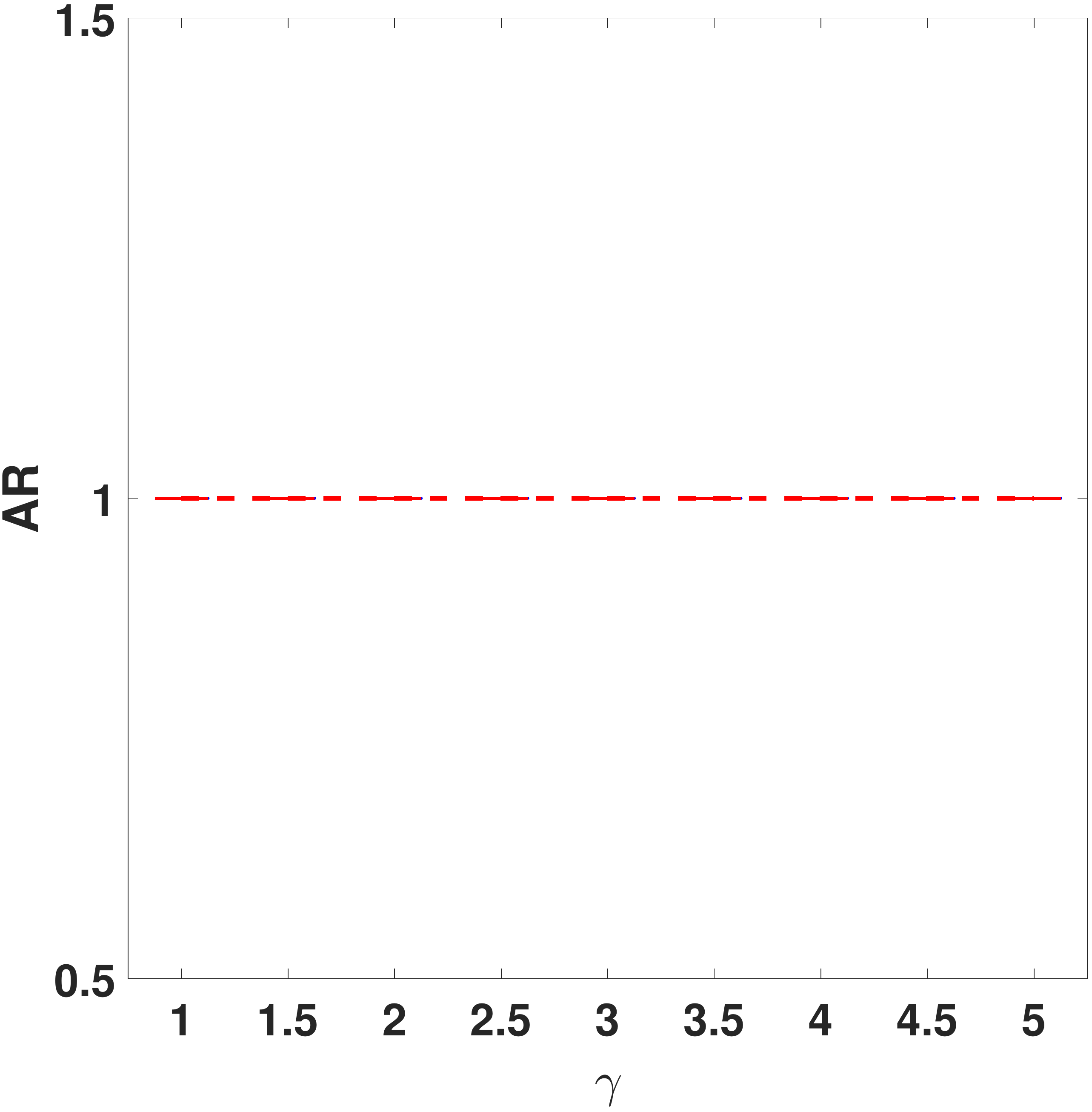} }
}
\vspace{-0.5\baselineskip}
\centerline{
\subfloat[Seeds]{\includegraphics[width=\gammaScale\textwidth]{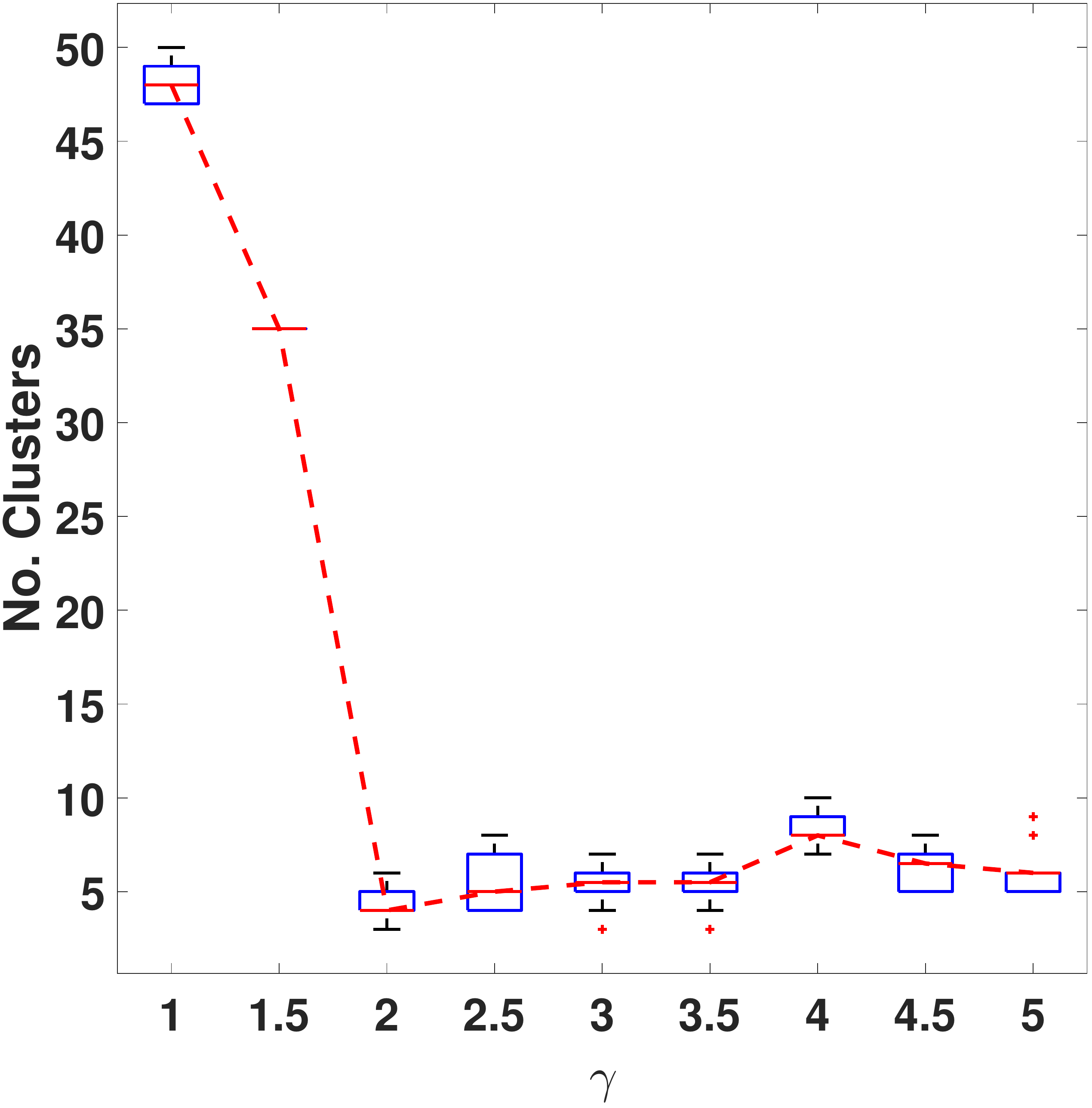} }
\subfloat[Wine]{\includegraphics[width=\gammaScale\textwidth]{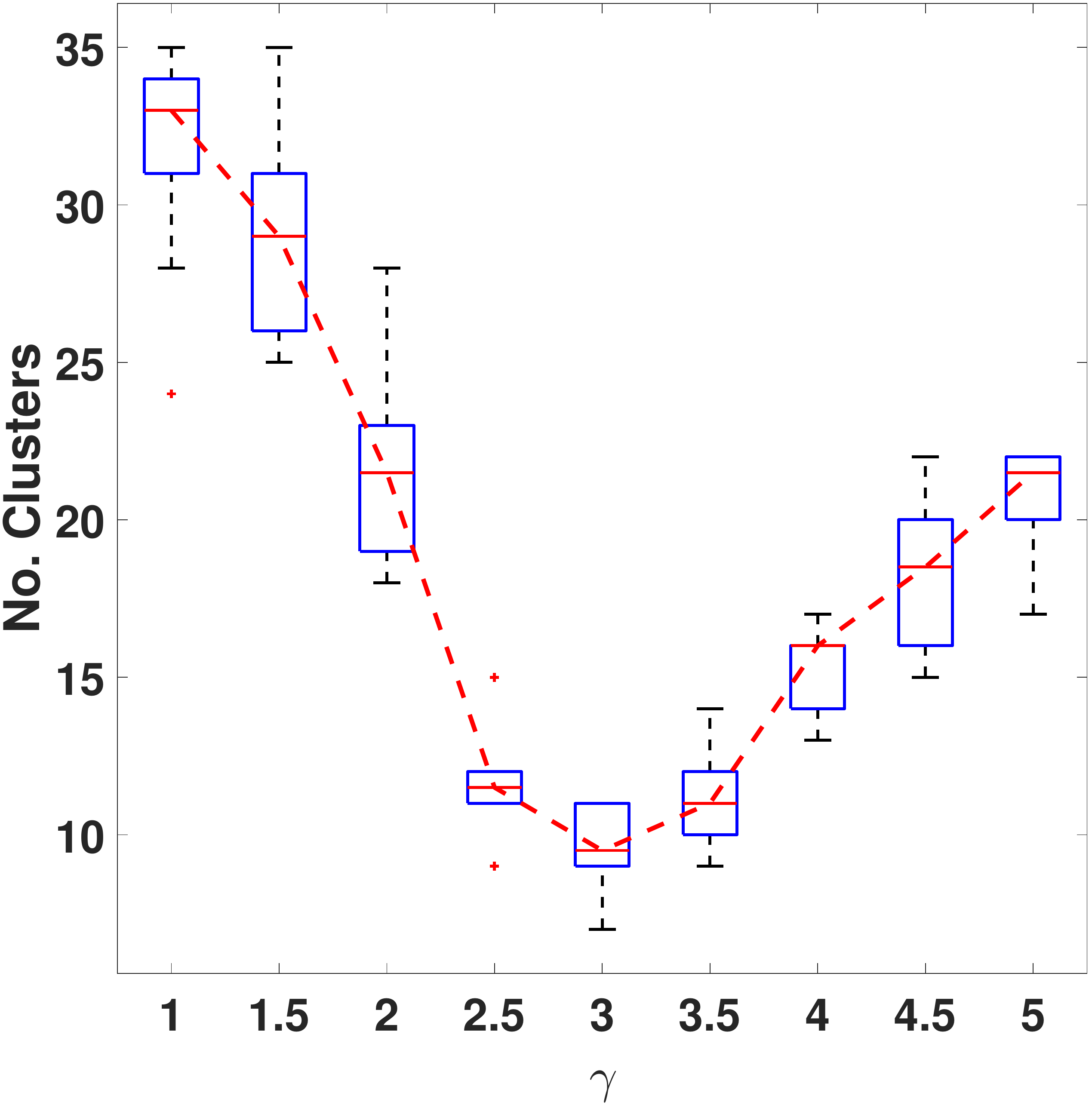} }
\subfloat[Target]{\includegraphics[width=\gammaScale\textwidth]{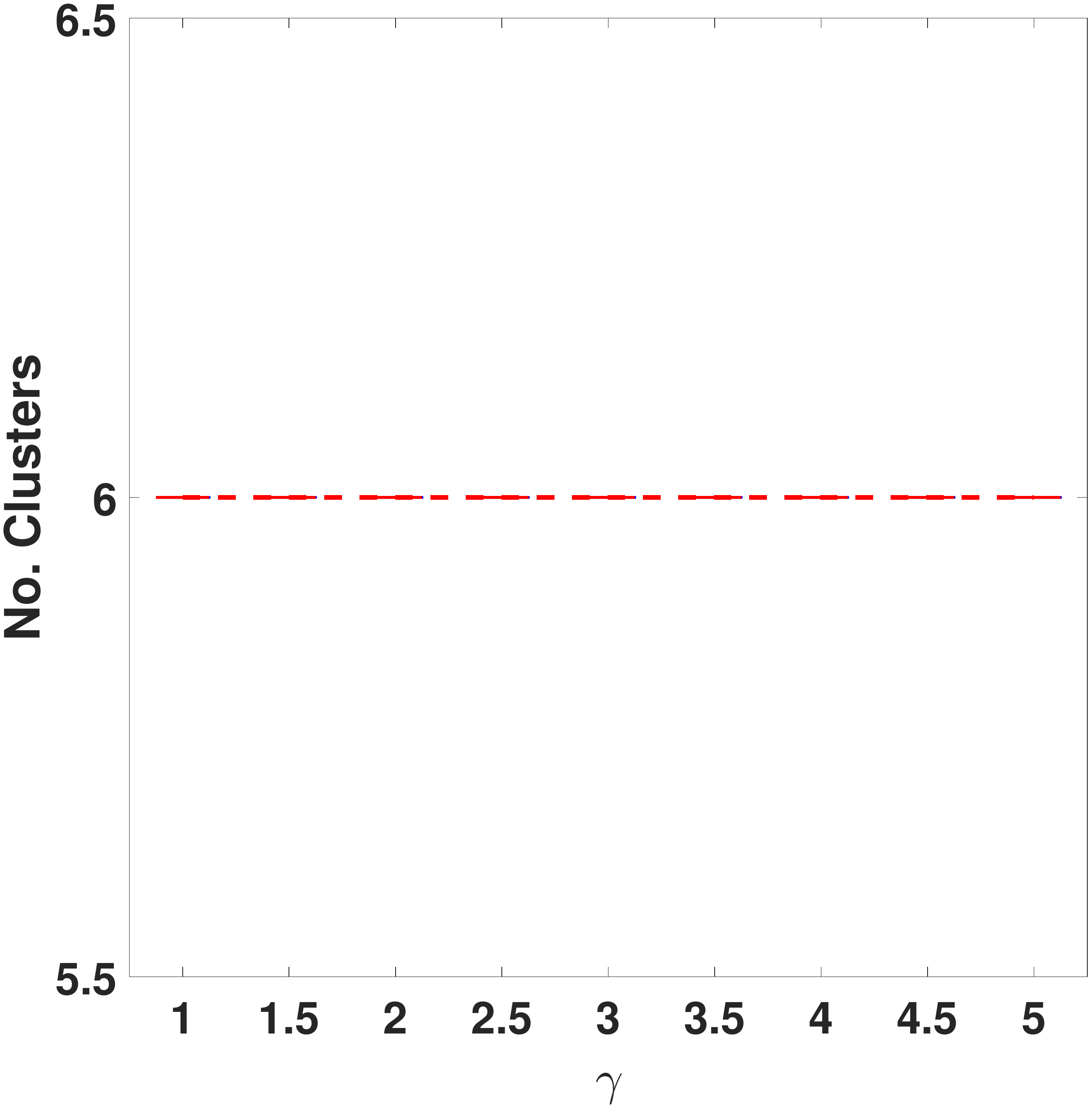} }
\subfloat[Tetra]{\includegraphics[width=\gammaScale\textwidth]{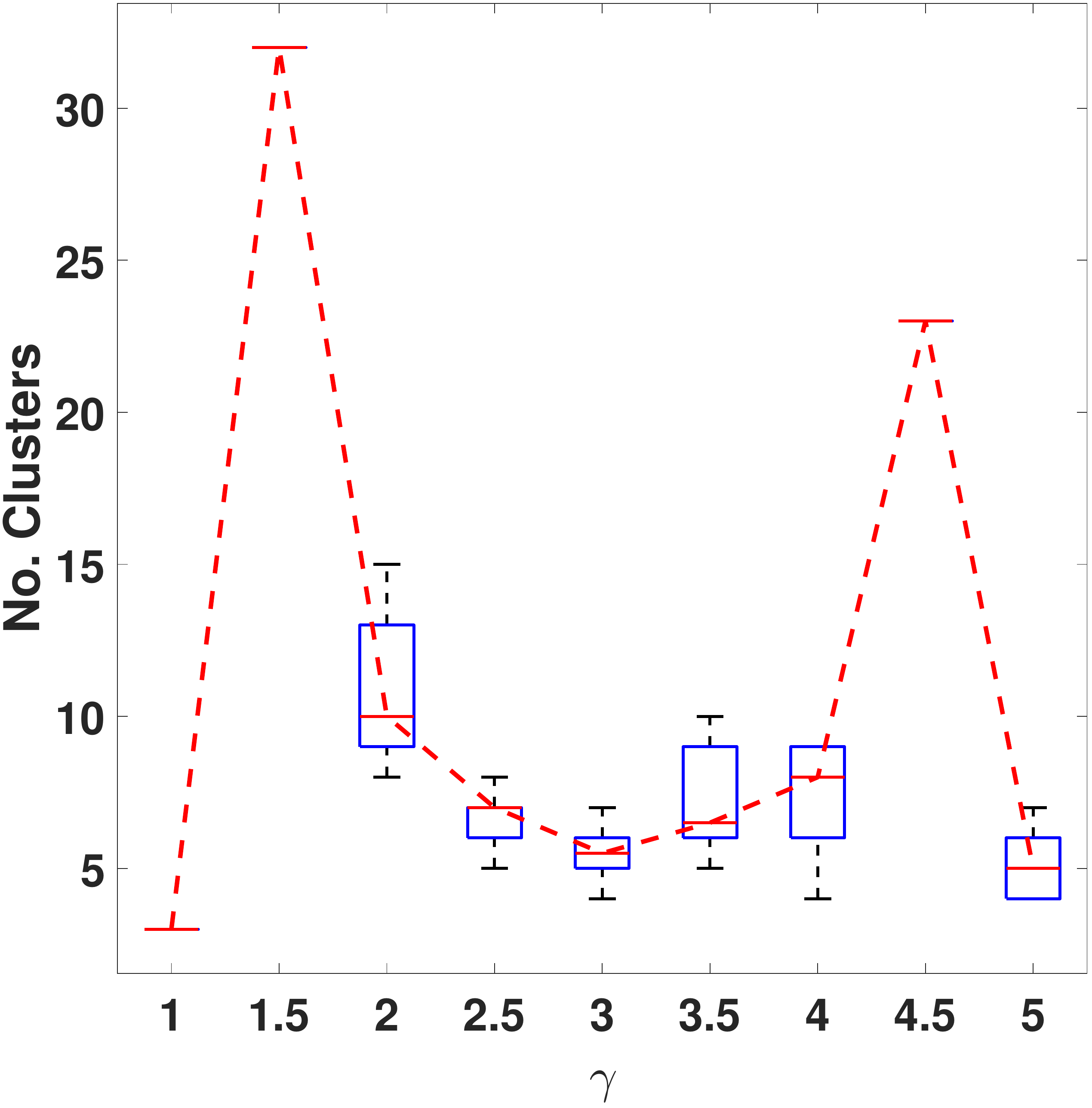} }
\subfloat[Lsun]{\includegraphics[width=\gammaScale\textwidth]{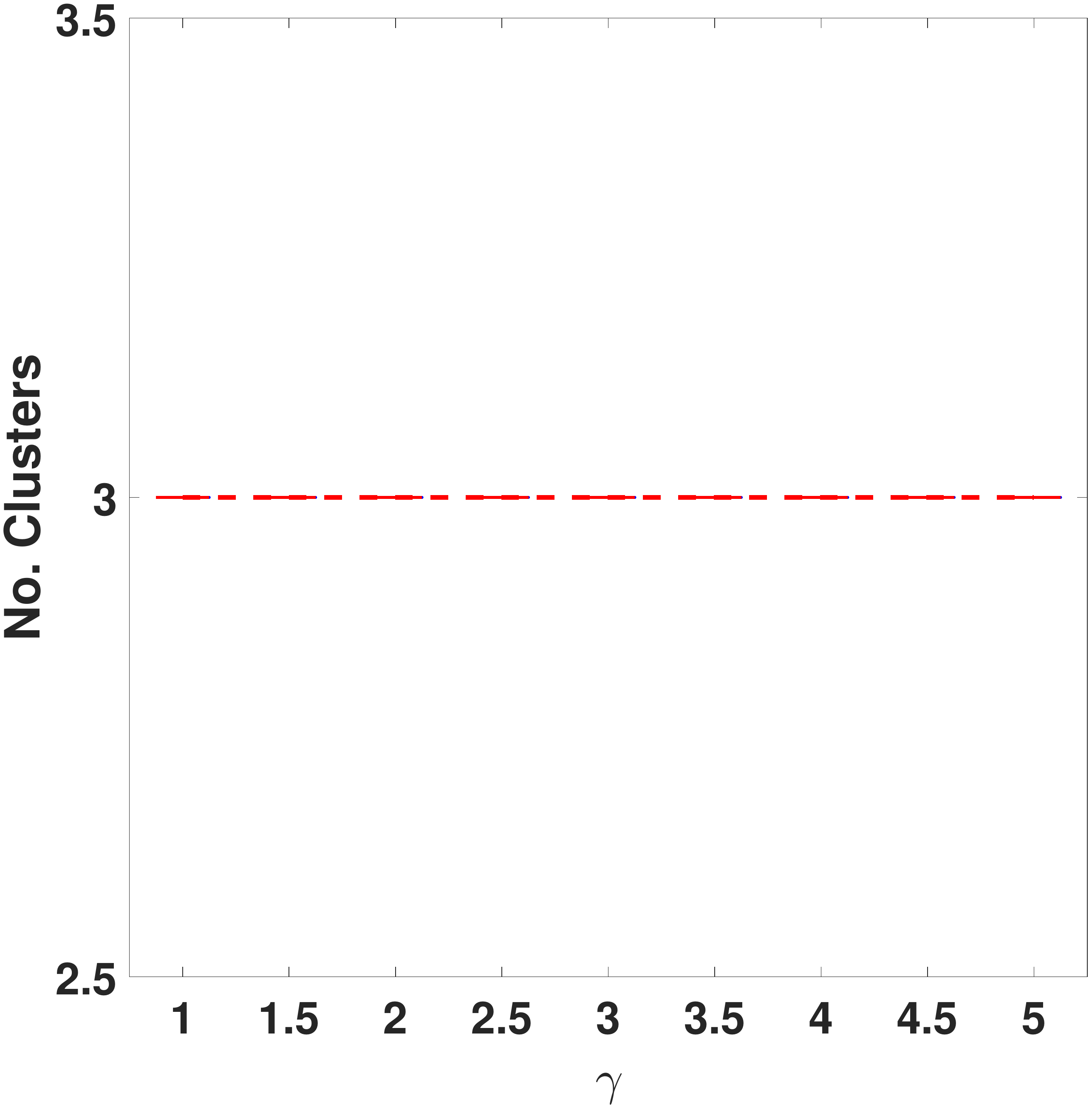} }
\subfloat[Moon]{\includegraphics[width=\gammaScale\textwidth]{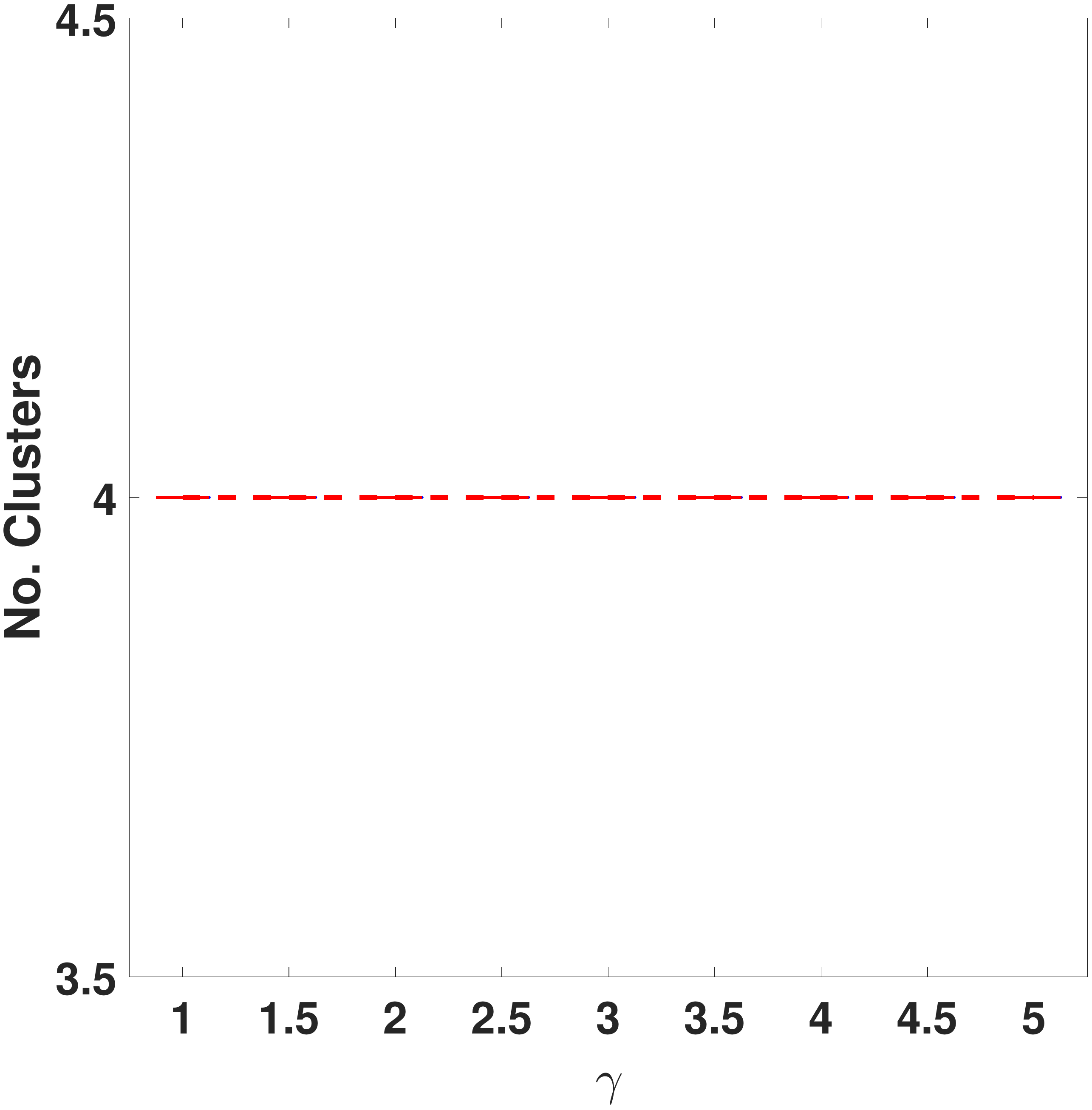} }
}
\vspace{-0.5\baselineskip}
\centerline{
\subfloat[Seeds]{\includegraphics[width=\gammaScale\textwidth]{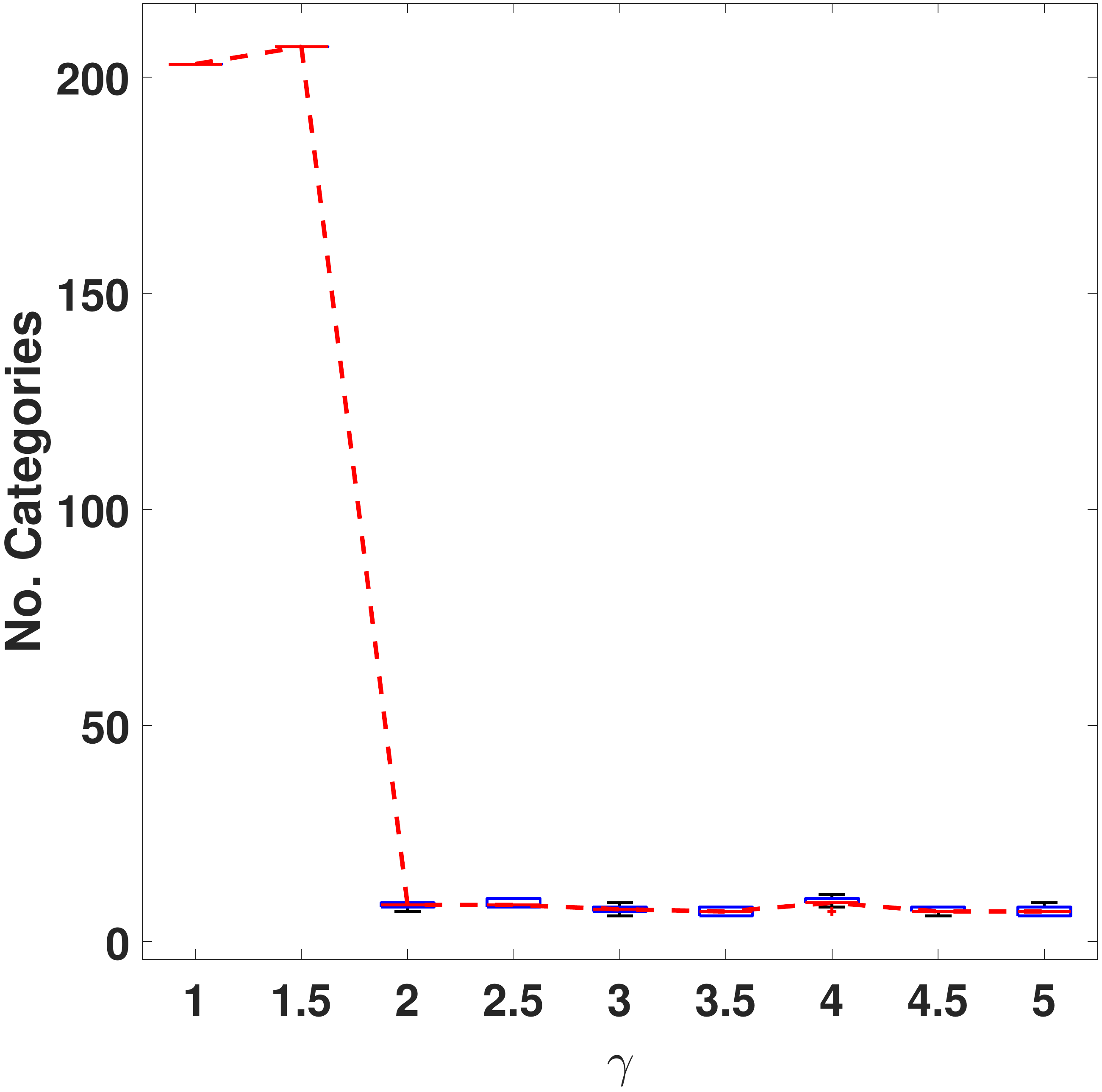} }
\subfloat[Wine]{\includegraphics[width=\gammaScale\textwidth]{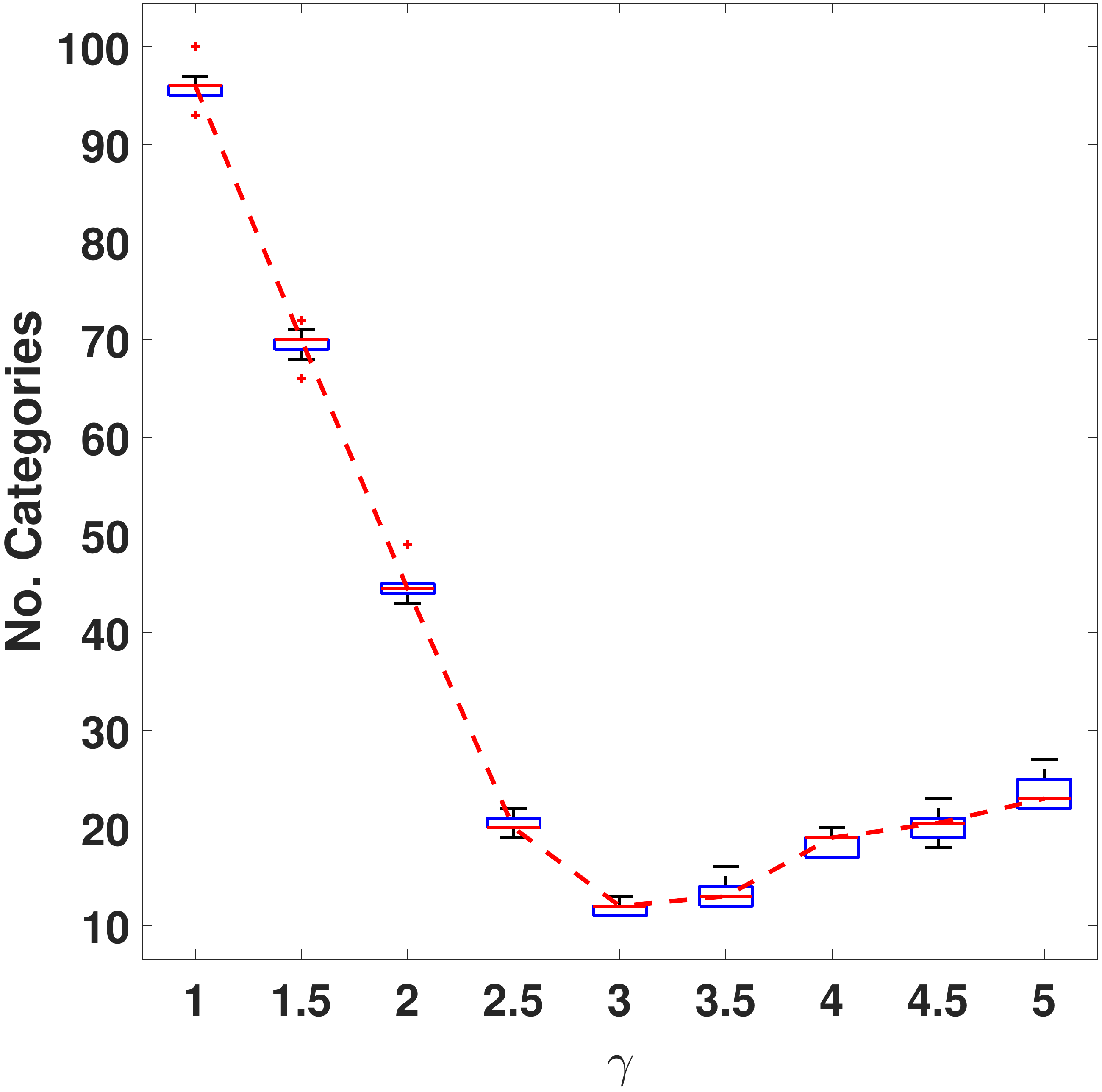} }
\subfloat[Target]{\includegraphics[width=\gammaScale\textwidth]{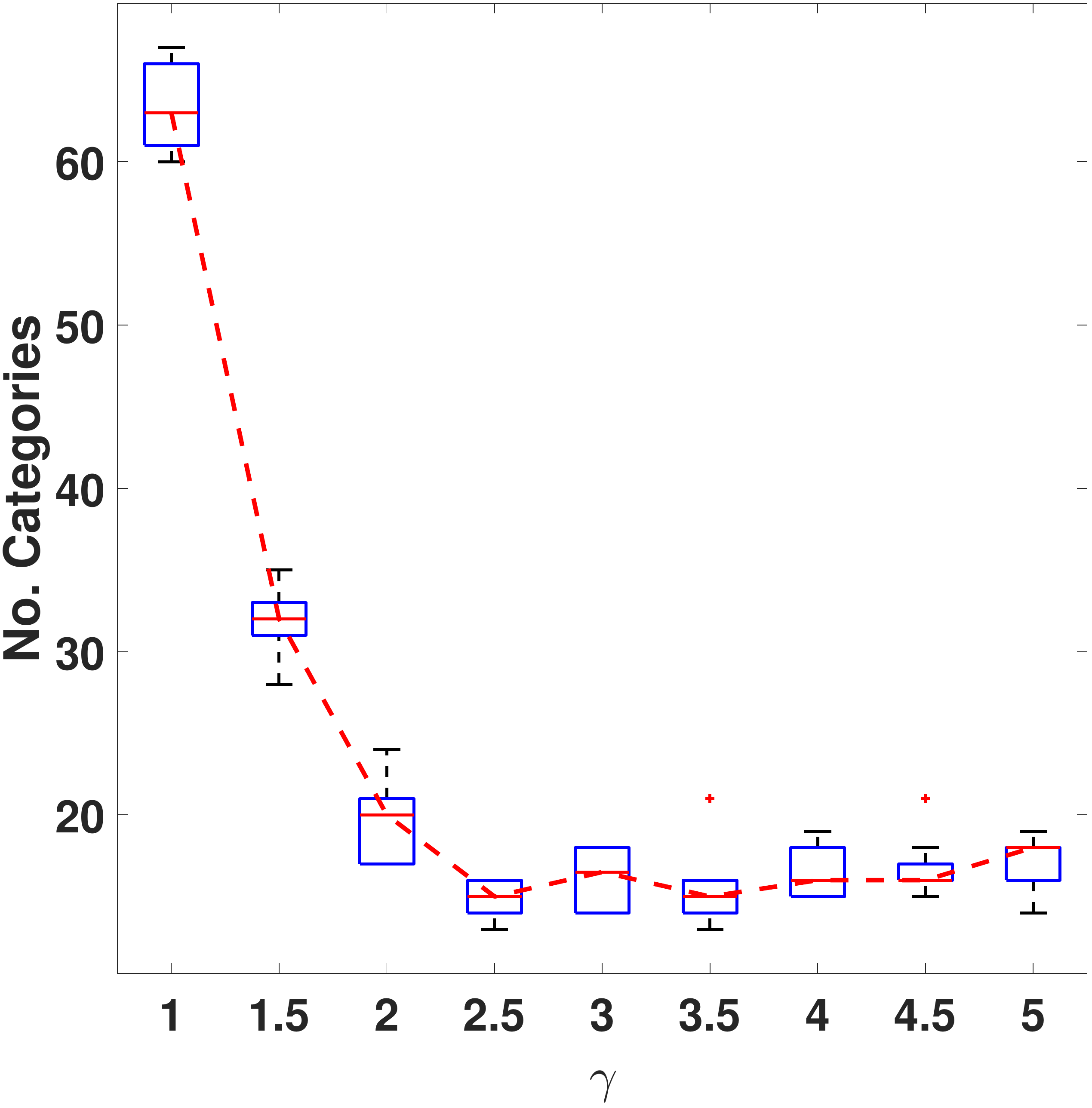} }
\subfloat[Tetra]{\includegraphics[width=\gammaScale\textwidth]{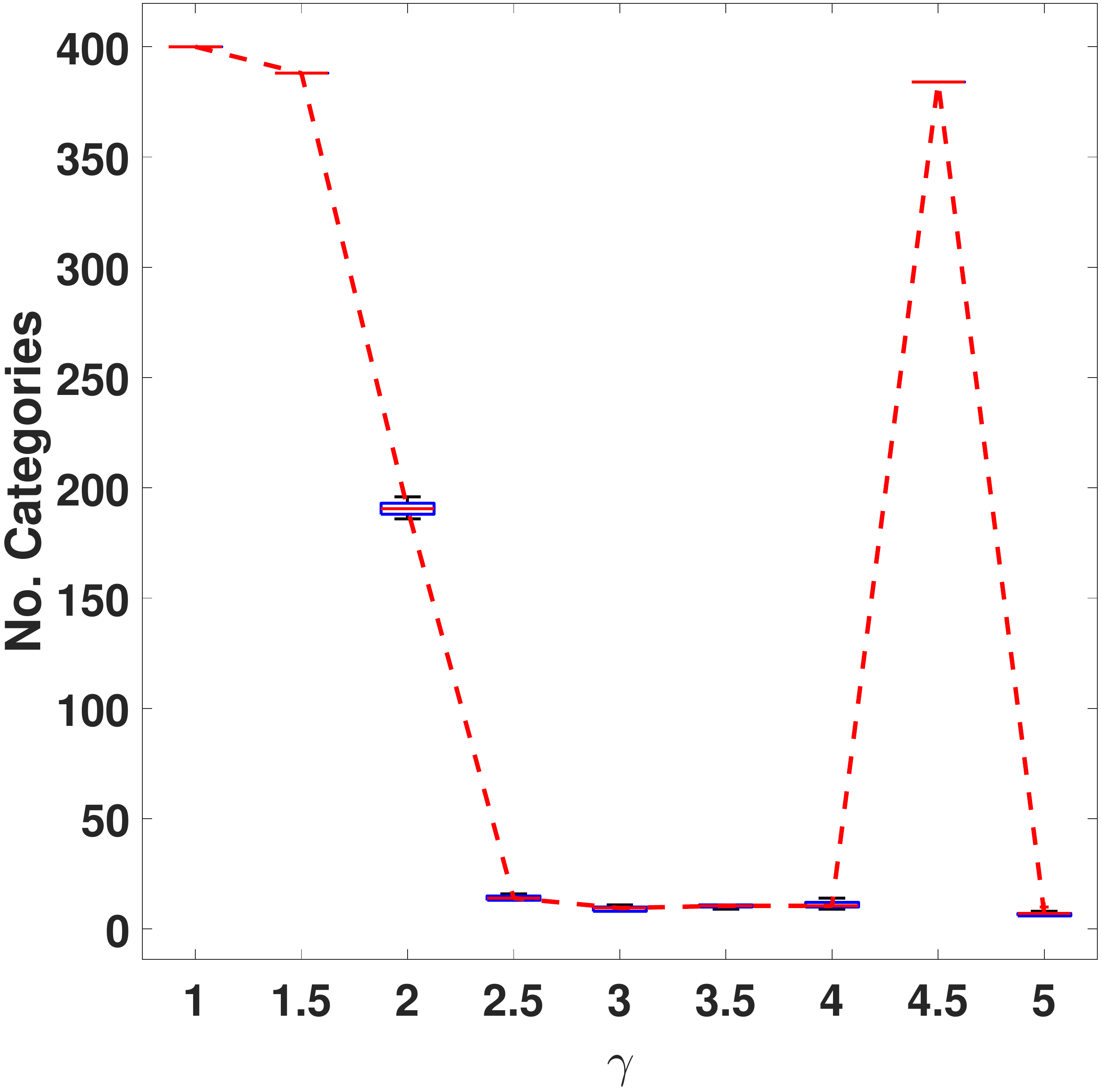} }
\subfloat[Lsun]{\includegraphics[width=\gammaScale\textwidth]{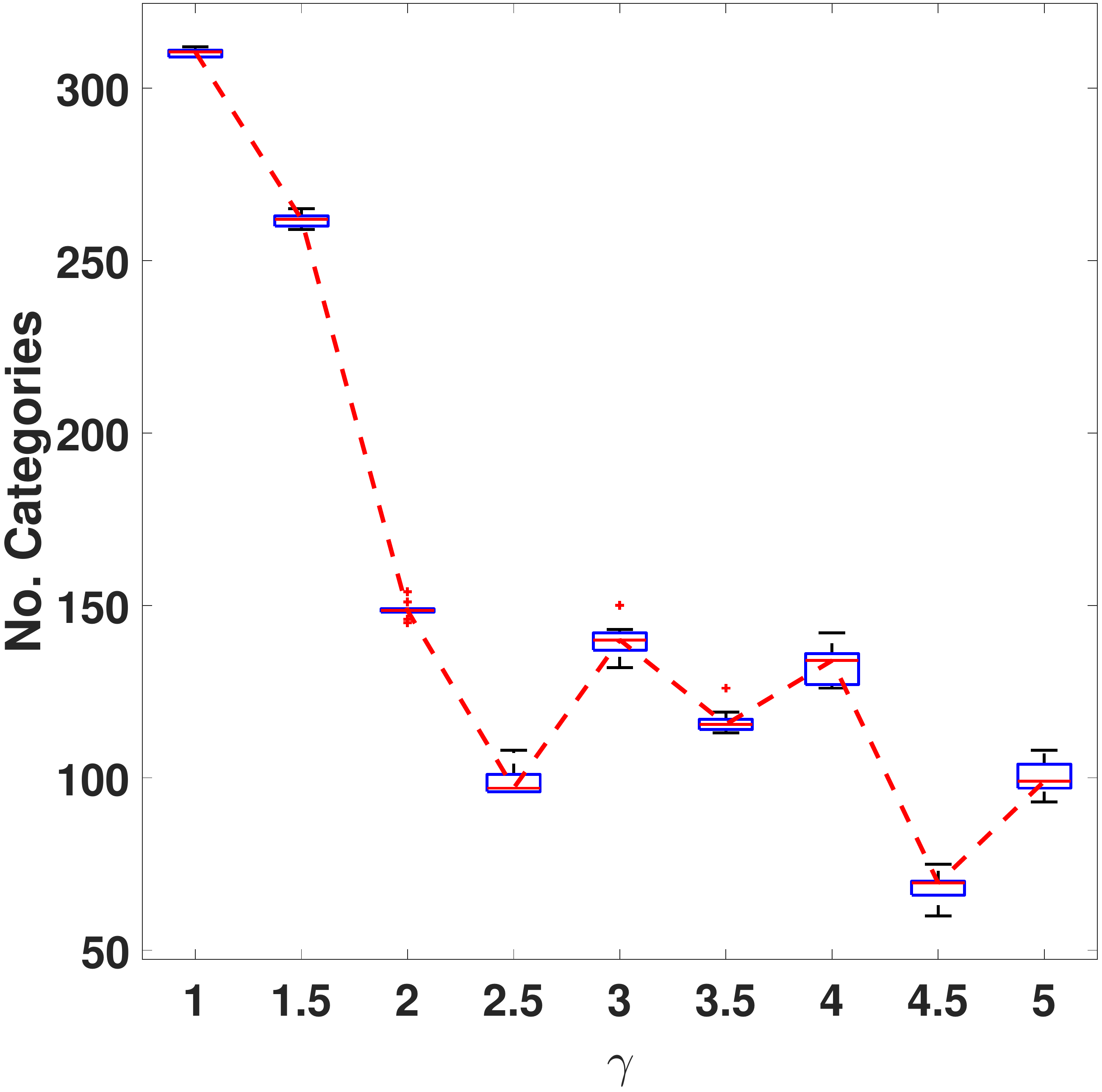} }
\subfloat[Moon]{\includegraphics[width=\gammaScale\textwidth]{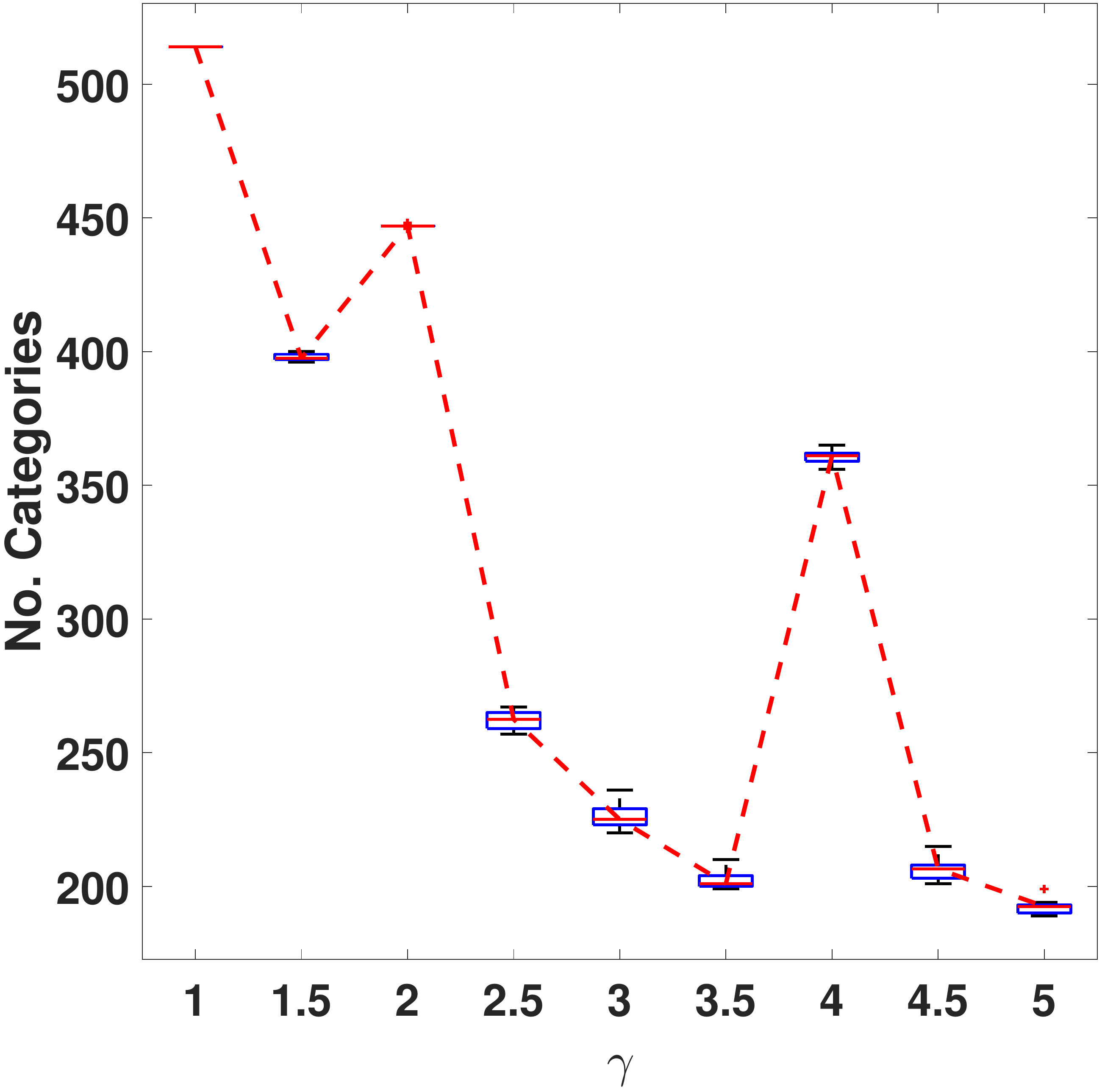} }
}
\caption{The behavior of the DDVFA + Merge ART system with respect to parameter $\gamma$ using the \textit{Seeds}, \textit{Wine}, \textit{Target}, \textit{Tetra}, \textit{Lsun}, and \textit{Moon} data sets: (a)-(f) peak average performance ($AR$), (g)-(l) number of clusters, and (m)-(r) total number of categories created. Both the number of clusters and categories are taken with respect to the most compact model that yields the depicted peak average performance (i.e., dual vigilance parameterization is \textit{not} held constant while varying parameter $\gamma$).}
\label{Fig:gamma_behavior_03}
\end{figure*}

To examine the behavior of the DDVFA systems with respect to parameter $\gamma$, $\gamma=1$ and $\gamma=3$ were arbitrarily set, and Wilcoxon's signed-ranks tests~\cite{Wilcoxon1945} were conducted to compare the performance and compactness of the best dual vigilance parameter combination (peak average performances over 30 runs). The results are reported in Table~\ref{Tab:gamma_tests}.

\begin{table}[!ht]
\centering
\caption{A statistical comparison of $\gamma=1$ versus $\gamma=3$: Wilcoxon p-values.}
\resizebox{\textwidth}{!}{
\begin{threeparttable}
\begingroup\setlength{\fboxsep}{0pt}
\colorbox{lightgray}{
\begin{tabular}{lllllll}
\toprule
\multirow{2}[4]{*}{Systems} & \multicolumn{6}{c}{Methods} \\
\cmidrule{2-7}          & \multicolumn{1}{l}{Average} & \multicolumn{1}{l}{Centroid} & \multicolumn{1}{l}{Complete} & \multicolumn{1}{l}{Median} & \multicolumn{1}{l}{Single} & \multicolumn{1}{l}{Weighted} \\
\midrule
\midrule
\multicolumn{7}{l}{Performance} \\
\midrule
VAT + DDVFA & \textbf{2.1228E-02} & \textbf{4.8300E-03} & 1.0200E-01 & \textbf{3.9650E-02} & \textbf{3.1506E-02} & 1.6480E-01 \\
DDVFA & 1.3591E-01 & \textbf{1.8254E-06} & \textbf{3.7323E-04} & \textbf{5.2872E-04} & \textbf{1.9209E-06} & 3.4935E-01 \\
DDVFA  +Merge ART & 2.2101E-01 & \textbf{3.3445E-06} & \textbf{4.0355E-04} & \textbf{1.7515E-02} & \textbf{2.6539E-03} & 1.5884E-01 \\
\midrule
\multicolumn{7}{l}{Compactness} \\
\midrule
VAT + DDVFA & 7.1864E-01 & 1.8663E-01 & 5.6445E-01 & 3.2279E-01 & \textbf{1.7982E-02} & 7.9707E-01 \\
DDVFA & \textbf{6.8344E-03} & \textbf{1.7697E-03} & \textbf{7.1966E-05} & \textbf{7.9639E-03} & \textbf{6.6540E-06} & \textbf{3.0581E-03} \\
DDVFA + Merge ART & 1.0000E+00 & \textbf{4.5022E-06} & \textbf{1.5649E-05} & \textbf{3.1513E-02} & \textbf{8.0045E-04} & 2.0223E-01 \\
\bottomrule
\end{tabular}
} \endgroup
\begin{tablenotes}[normal,flushleft]
\item[] Bold values indicate statistically significant results.
\end{tablenotes}
\end{threeparttable}
}
\label{Tab:gamma_tests}
\end{table}

Regarding the HAC-based activation/match functions, a significant statistical difference for both performance and compactness was observed for (a)~the single variant, (b)~most systems using centroid and median, and (c)~the complete variant but to a lesser extent. Average and weighted variants do not appear to be very affected by changing parameter $\gamma$ between these two values. With respect to the three DDVFA systems, performance and compactness are affected by parameter $\gamma$, except for the compactness of the VAT + DDVFA system which remains mostly unaffected.  

Due to these statistical analysis results, the DDVFA systems' behavior was further investigated using single-linkage HAC activation/match functions with respect to parameter~$\gamma$. The study is performed by varying~$\gamma$ in the interval $[0,5]$ with a step size of $0.5$ and observing the following aspects: peak average performance ($AR$), number of clusters, and number of categories created. The last two quantities were examined since DDVFA belongs to the class of multi-prototype-based clustering methods, i.e., each cluster may be represented by multiple categories. Such behaviors are illustrated in Figs.~\ref{Fig:gamma_behavior_01} through~\ref{Fig:gamma_behavior_03}. For clarity, and according to the recommendations outlined in Section~\ref{Sec:ResultsA}, only the behavior with respect to the data sets \textit{Seeds}, \textit{Wine}, \textit{Target}, \textit{Tetra}, \textit{Lsun}, and \textit{Moon} is reported.  

For each value of $\gamma$, the vigilance parameter combination corresponding to the best average performance over $10$ different input permutation orders is selected. Following Occam's razor and the principle of parsimony~\cite{duda2000}, among all models that yield the best performance, the one with the simplest clustering structure is selected, i.e., the one that requires the smaller number of categories to encode its clustering partition. Thus, the depicted box-plots relate to the simplest model that achieved the peak average performance for each value of $\gamma$.

\textbf{Remark 4.} Note that the vigilance parameter combinations that yield each box-plot in Figs.~\ref{Fig:gamma_behavior_01} through~\ref{Fig:gamma_behavior_03} are not held constant across the different values of~$\gamma$; therefore, they may not be necessarily the same. For instance, Fig.~\ref{Fig:gamma_behavior_01} shows that, for the VAT + DDVFA system, given a value of~$\gamma$, there is a vigilance parameter combination that can find the correct partitions ($AR=1$) with similar compactness levels (number of categories) across $\gamma$ values for the \textit{Target}, \textit{Tetra}, \textit{Lsun}, and \textit{Moon} data sets. Analogously, given a value~$\gamma$, there is a vigilance parameter combination for the DDVFA + Merge ART system that yields maximum $AR$ for the \textit{Target}, \textit{Lsun}, and \textit{Moon} data sets; however, the number of categories fluctuates when the samples are randomly presented. If the dual vigilance parameter combination is held constant, e.g., by setting it to the best combination associated with $\gamma=1$, then, for other $\gamma$ values, the behaviors with respect to performance, number of clusters and categories may change for both systems, as shown in Fig.~\ref{Fig:gamma_vigilance_fixed} for the \textit{Target} data set. Note the increase in the number of categories due to the increase of~$\gamma$: the smallest dual vigilance parameter values required to achieve the best performance for $\gamma=1$ are somewhat large, and the same values coupled with a more selective kernel (larger~$\gamma$) result in more categories being created.

\newcommand{\gammafixedvig}{0.33}
\begin{figure}[!hp]
\centerline{
\subfloat[$AR$]{\includegraphics[width=\gammafixedvig\columnwidth]{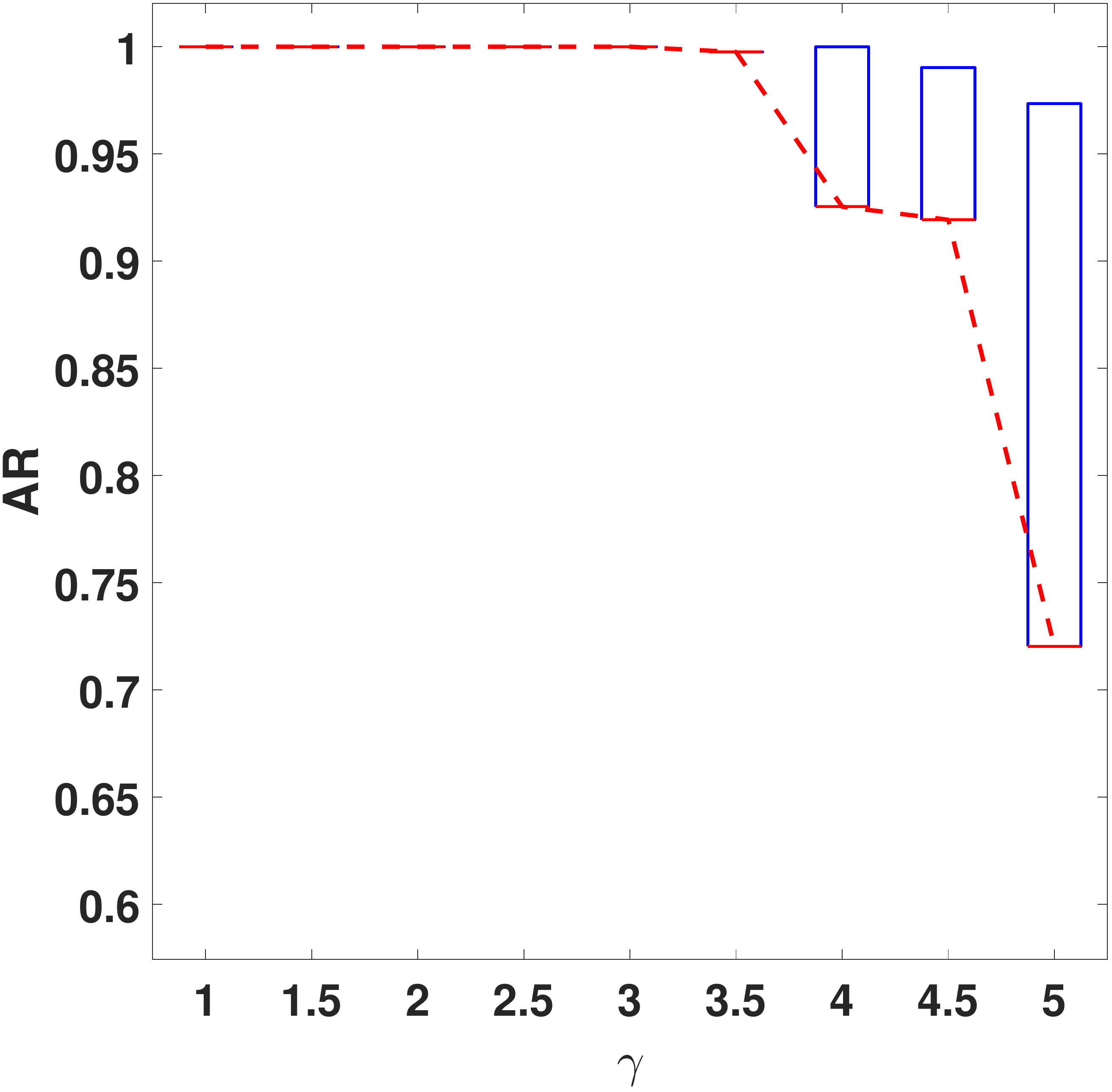} }
\subfloat[No. Clusters]{\includegraphics[width=\gammafixedvig\columnwidth]{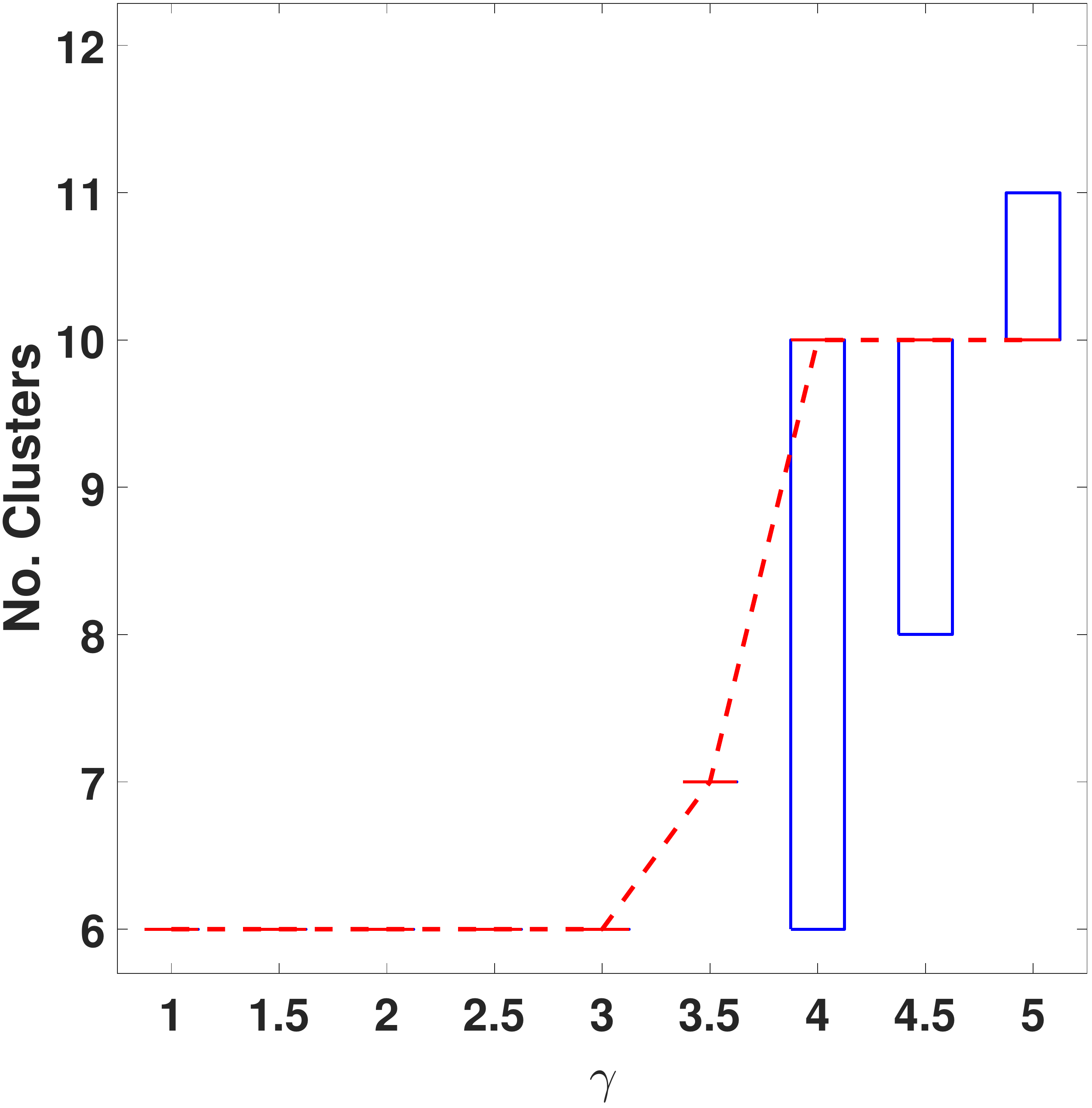} }
\subfloat[No. Categories]{\includegraphics[width=\gammafixedvig\columnwidth]{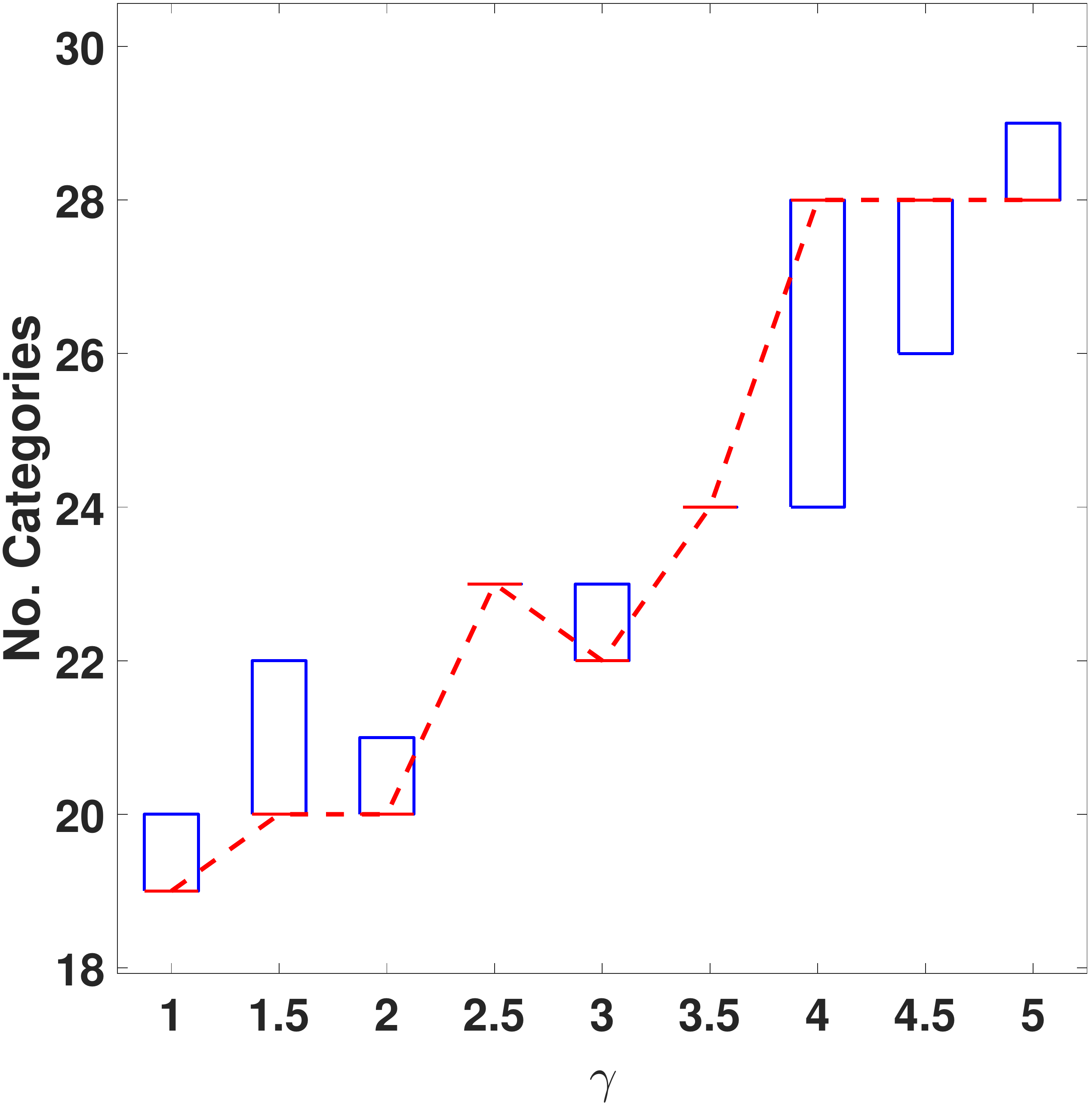} }
}
\centerline{
\subfloat[$AR$]{\includegraphics[width=\gammafixedvig\columnwidth]{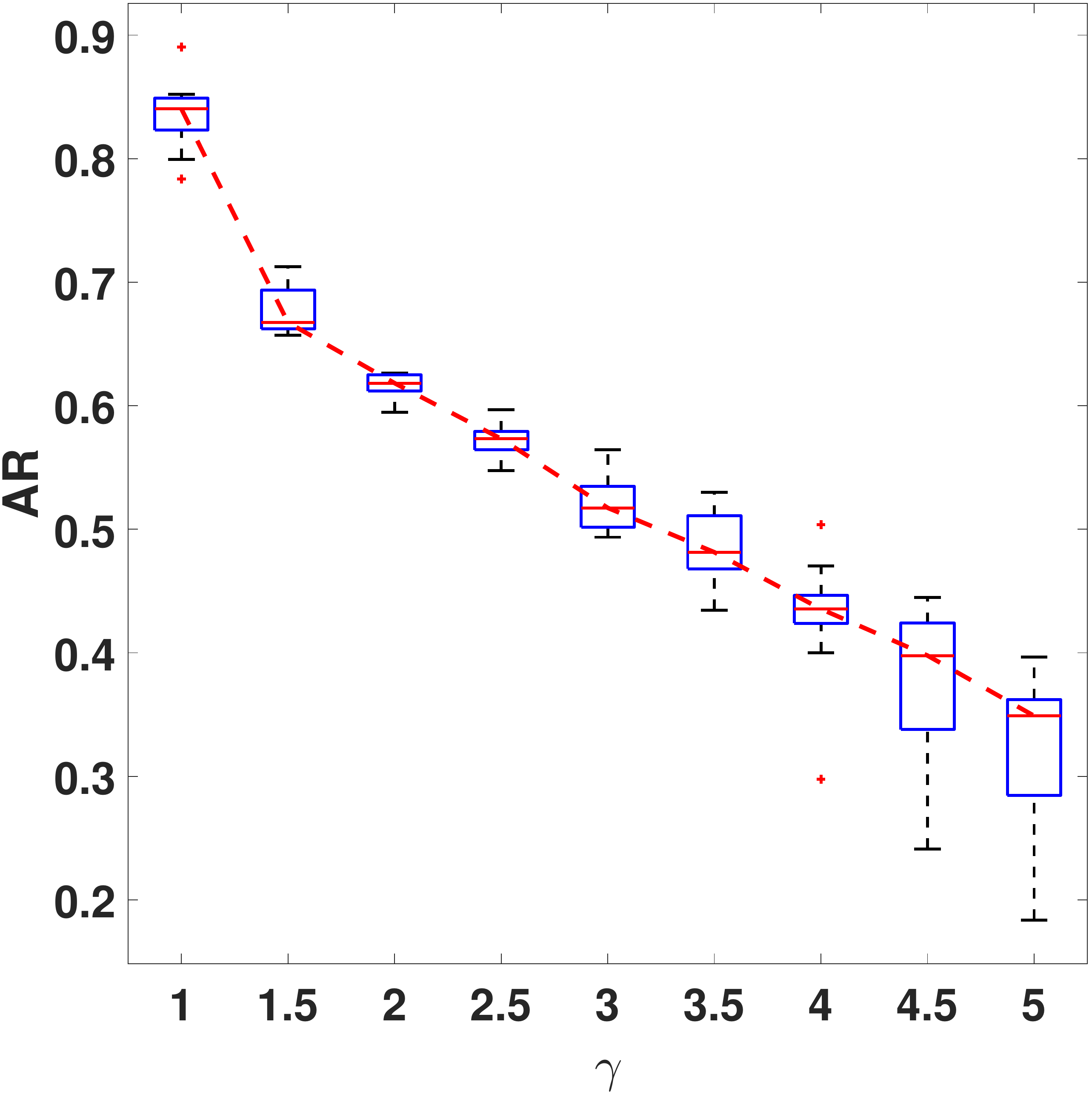} }
\subfloat[No. Clusters]{\includegraphics[width=\gammafixedvig\columnwidth]{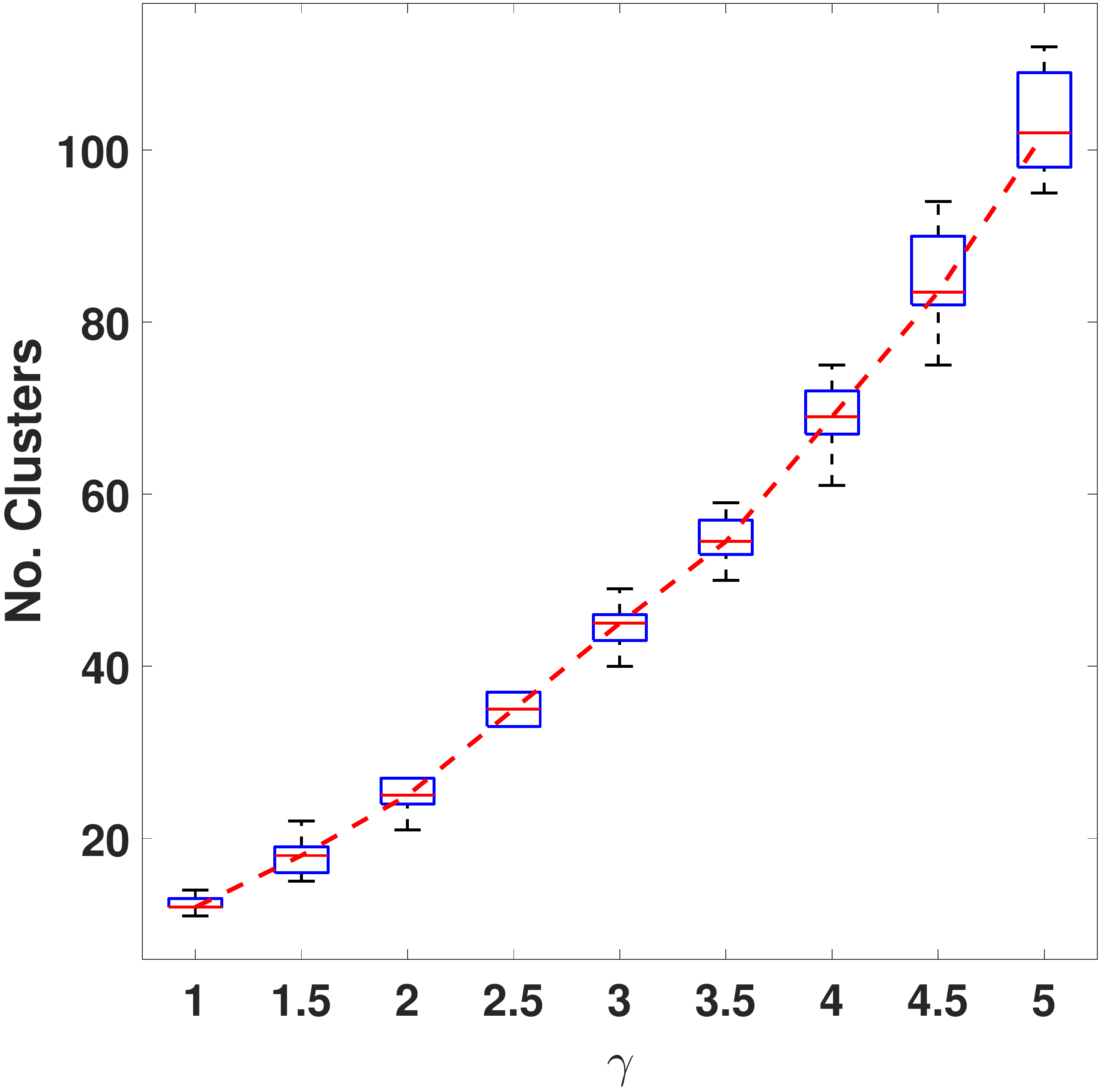} }
\subfloat[No. Categories]{\includegraphics[width=\gammafixedvig\columnwidth]{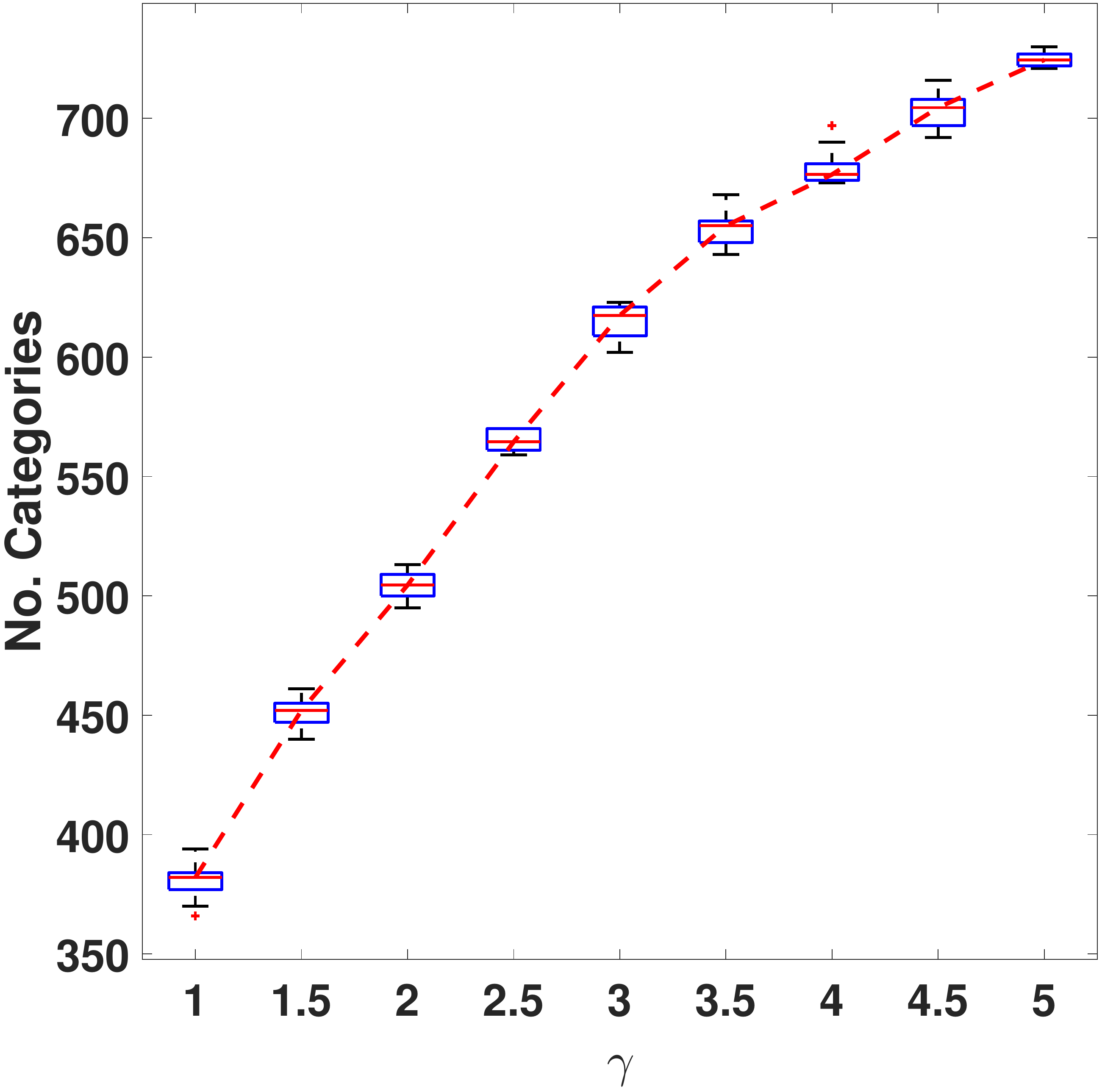} }
}
\centerline{
\subfloat[$AR$]{\includegraphics[width=\gammafixedvig\columnwidth]{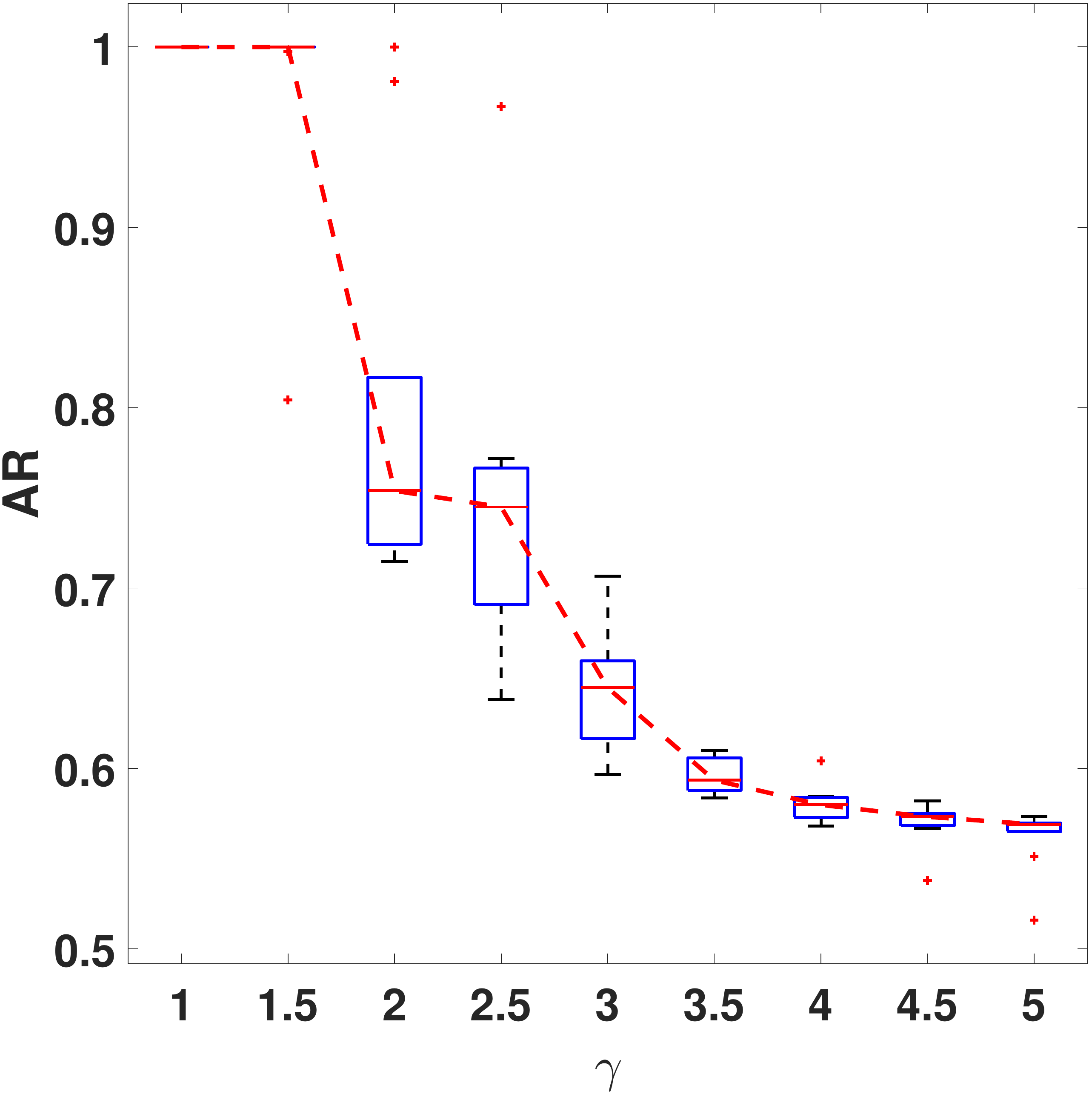} }
\subfloat[No. Clusters]{\includegraphics[width=\gammafixedvig\columnwidth]{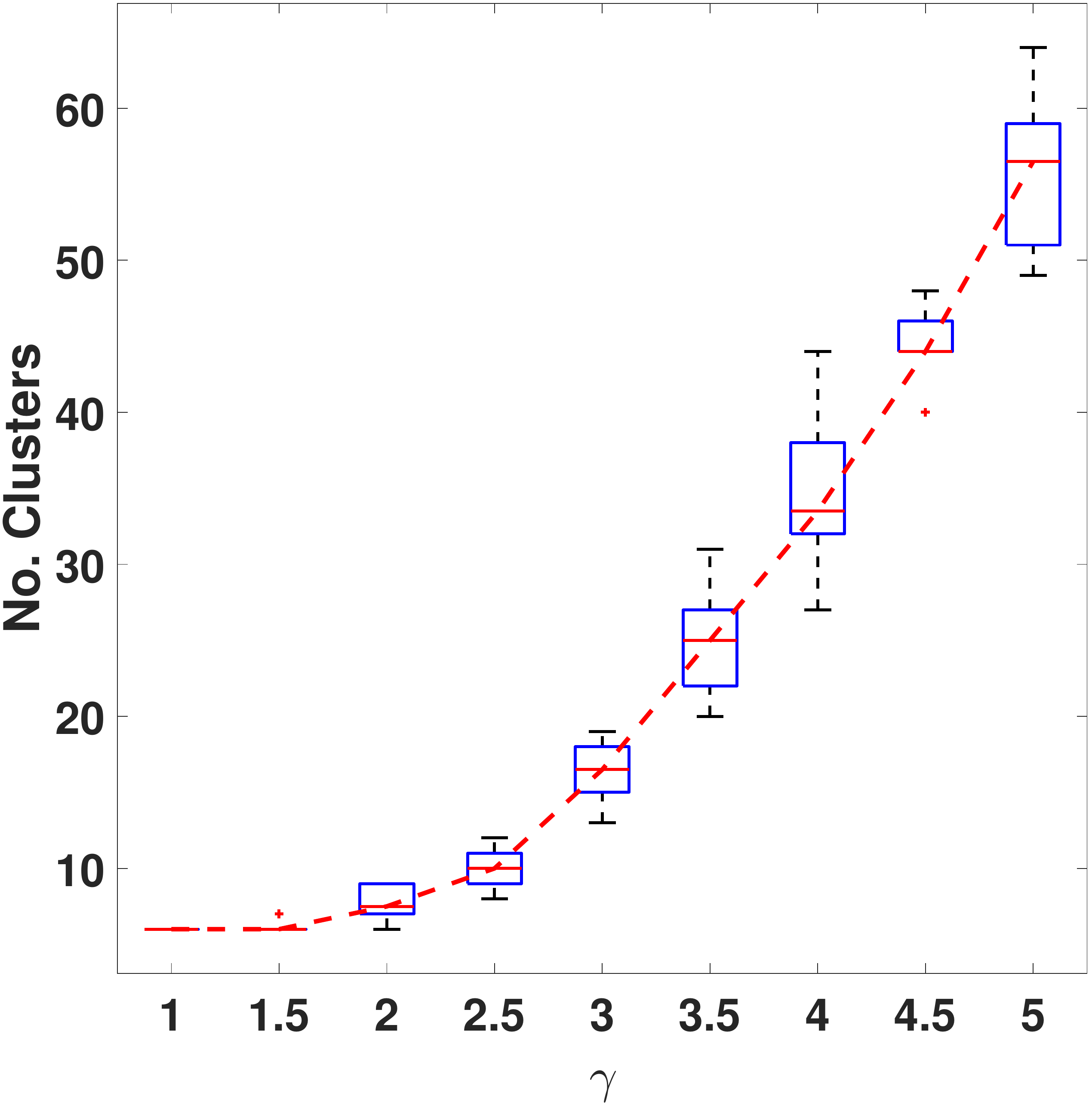} }
\subfloat[No. Categories]{\includegraphics[width=\gammafixedvig\columnwidth]{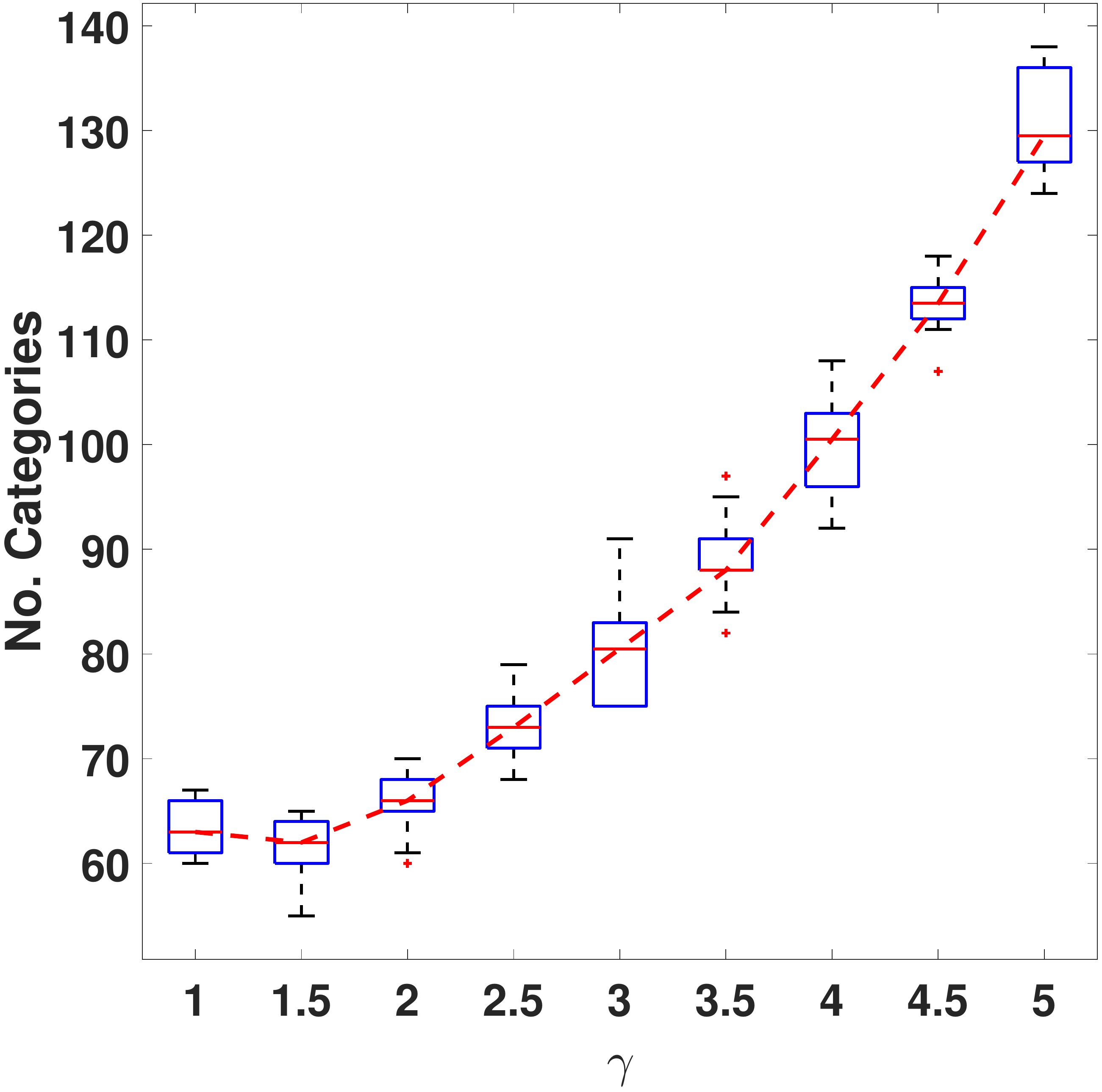} }
}
\caption{The behavior of the (a)-(c) VAT + DDVFA, (d)-(f) DDVFA, and (g)-(i) DDVFA + Merge ART systems for different values of parameter~$\gamma$ while holding the dual vigilance parameters constant. Single linkage HAC-based activation and match functions are used.}
\label{Fig:gamma_vigilance_fixed}
\end{figure}

Naturally, the behavior of the DDVFA systems with respect to $\gamma$ is data- and system-dependent. Although some $AR$ performance fluctuation exists across the values of $\gamma$ for some data sets, it generally seems to be fairly robust to this parameter. The number of categories, i.e., the compression level, often drastically changes with $\gamma$. For example, setting $\gamma=1$ (i.e., using standard fuzzy ART building blocks) versus $\gamma=2$ already yields noticeable changes in many data sets as shown in Figs.~\ref{Fig:gamma_behavior_01} through~\ref{Fig:gamma_behavior_03}, especially for the DDVFA + Merge ART system. Furthermore, the number of categories appears to decrease by increasing $\gamma$ as this tendency was observed in many of the data sets in Figs.~\ref{Fig:gamma_behavior_01} through~\ref{Fig:gamma_behavior_03}. Specifically, Fig.~\ref{Fig:Target_eg1} illustrates this effect in the \textit{Target} data set. These experimental results are consistent with previous findings in related work, in which improved memory compression is achieved when using power rules coupled with distributed learning in ART-systems~\cite{Carpenter1997b, Carpenter1998a}. Another important aspect refers to the region of the dual vigilance parameter space which correlates with better performance; such a region seems to increase with the value of $\gamma$ for some data sets (e.g., the \textit{Target} data set in Fig.~\ref{Fig:gamma_behavior_04}), usually at the expense of the network's compactness.

\newcommand{\egtargetcats}{0.24}
\begin{figure}[!t]
\centerline{
\subfloat[$(1, 19)$]{\includegraphics[width=\egtargetcats\columnwidth]{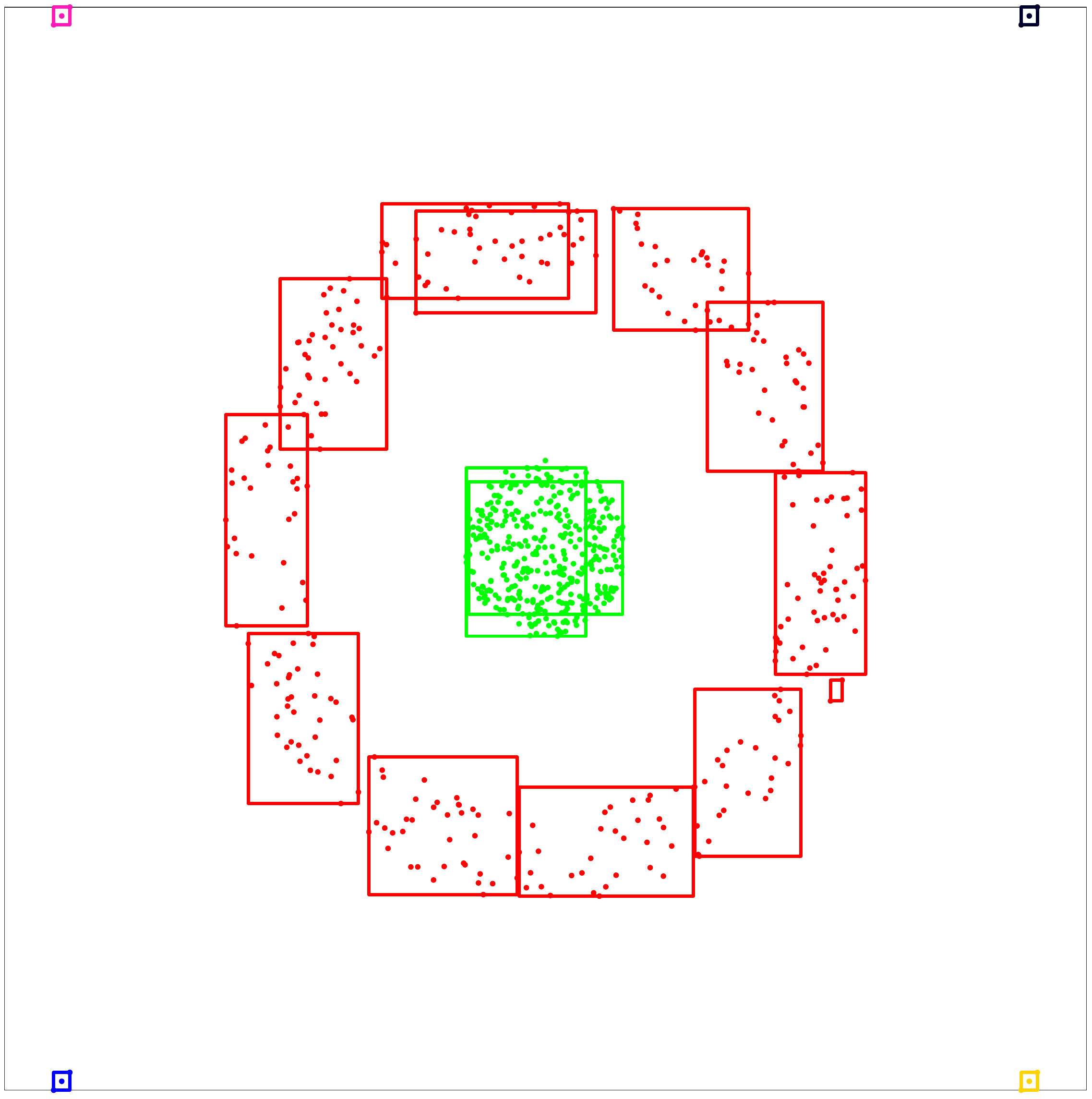}}
\hfil
\subfloat[$(3, 12)$]{\includegraphics[width=\egtargetcats\columnwidth]{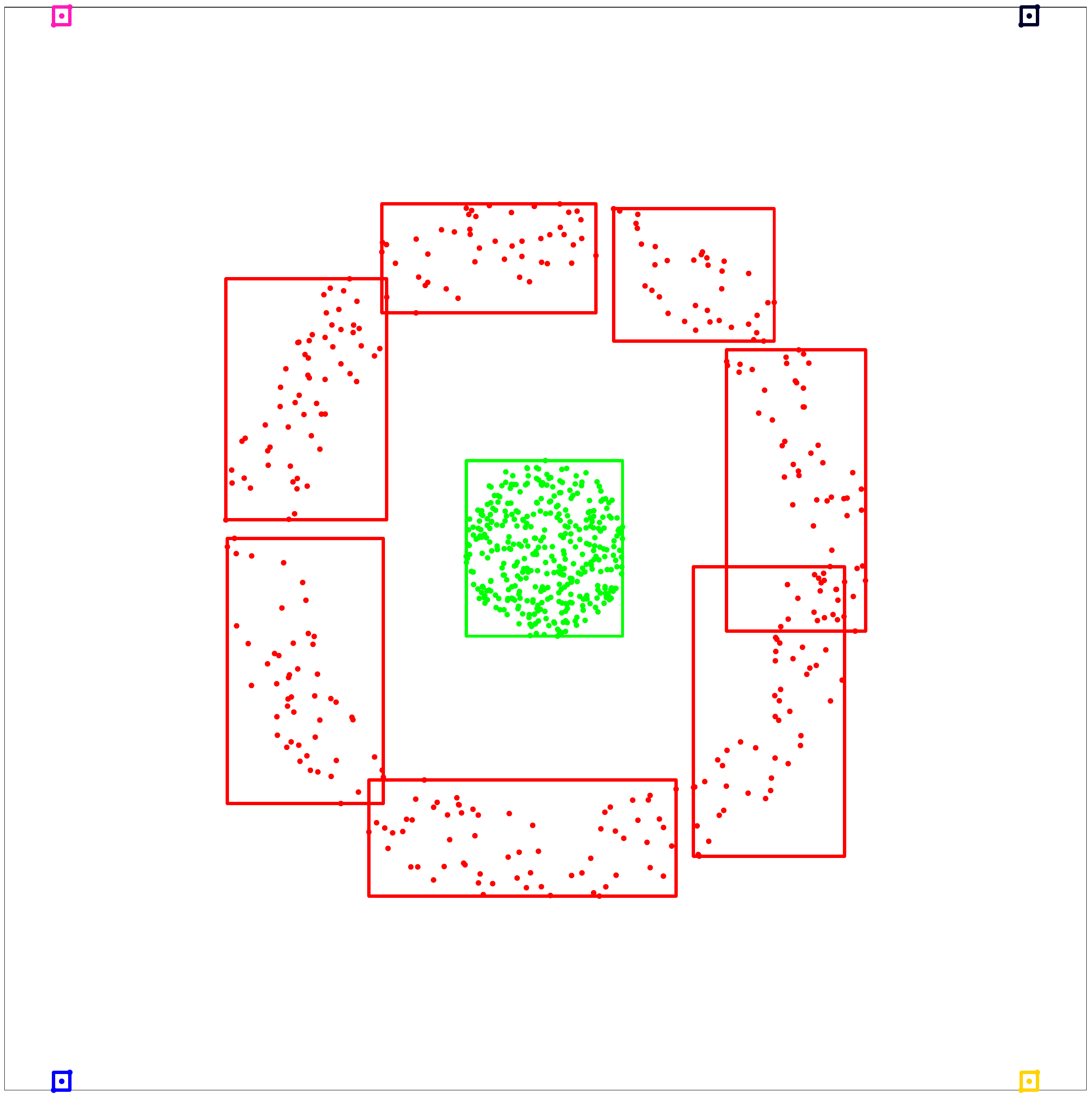}}
\hfil
\subfloat[$(1, 78)$]{\includegraphics[width=\egtargetcats\columnwidth]{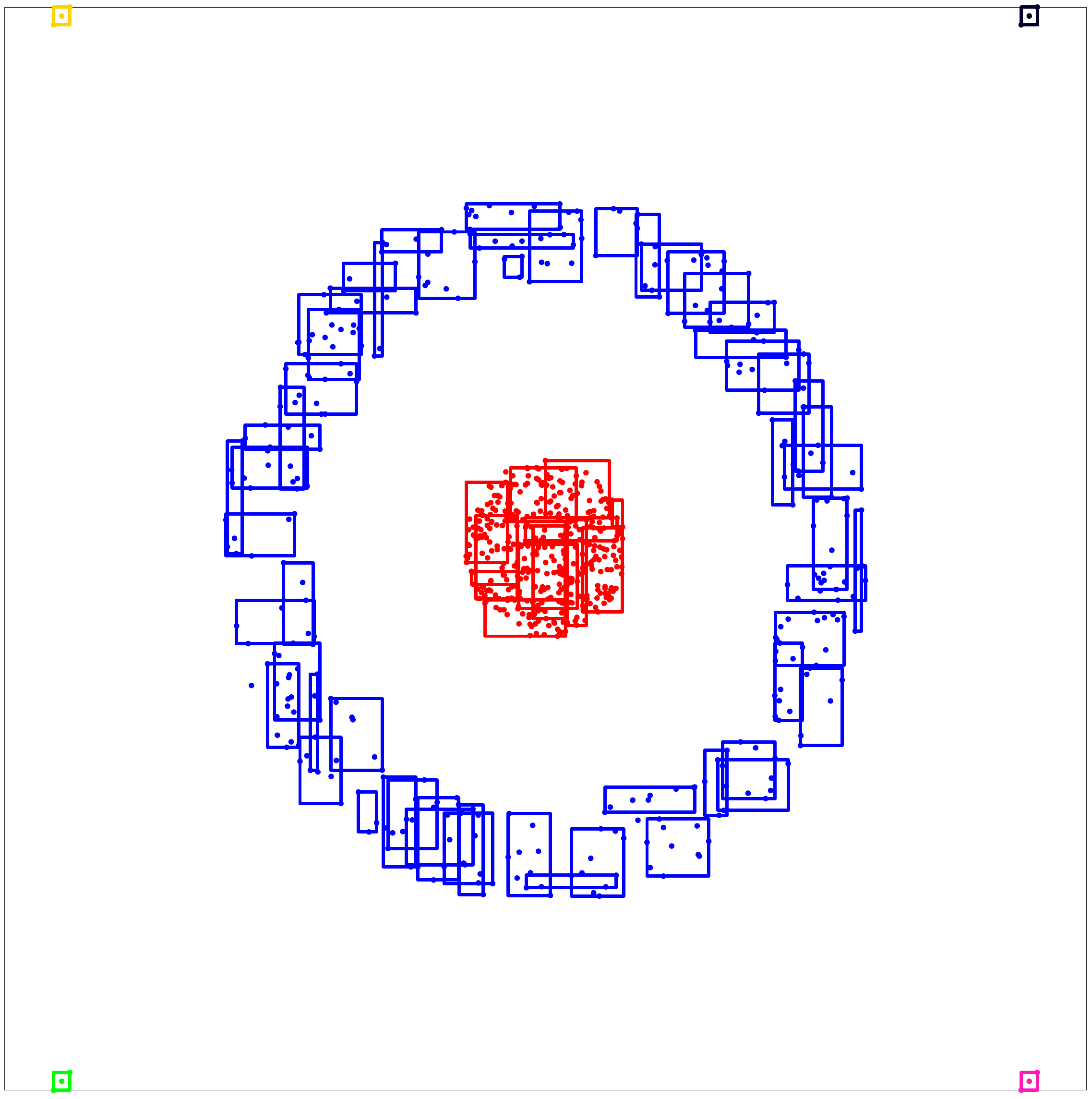}}
\hfil
\subfloat[$(3, 17)$]{\includegraphics[width=\egtargetcats\columnwidth]{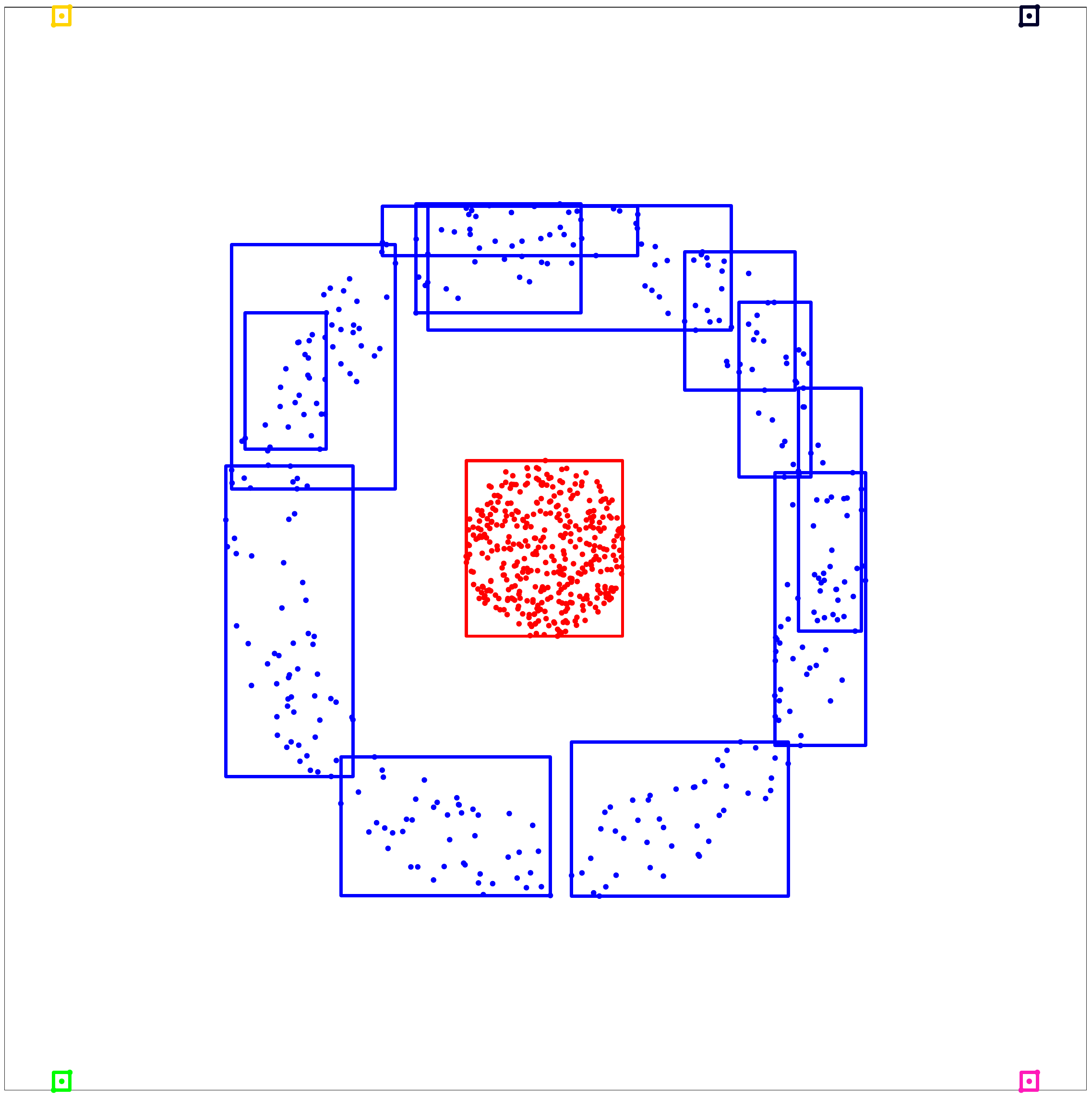}}
}
\caption{The best and most compact output partitions for the \textit{Target} data set using the (a)-(b) VAT + DDVFA and (c)-(d) DDVFA + Merge ART systems. The ordered pairs correspond to ($\gamma$, total number of categories). (a) and (d) correspond to fuzzy ART and are subject to category proliferation, whereas (b) and (d) correspond to DDVFA and represent the same data with fewer categories.}
\label{Fig:Target_eg1}
\end{figure}

\newcommand{\gammaMat}{0.33}
\begin{figure}[!ht]
\centerline{
\subfloat[$\gamma=1$]{\includegraphics[width=\gammaMat\columnwidth]{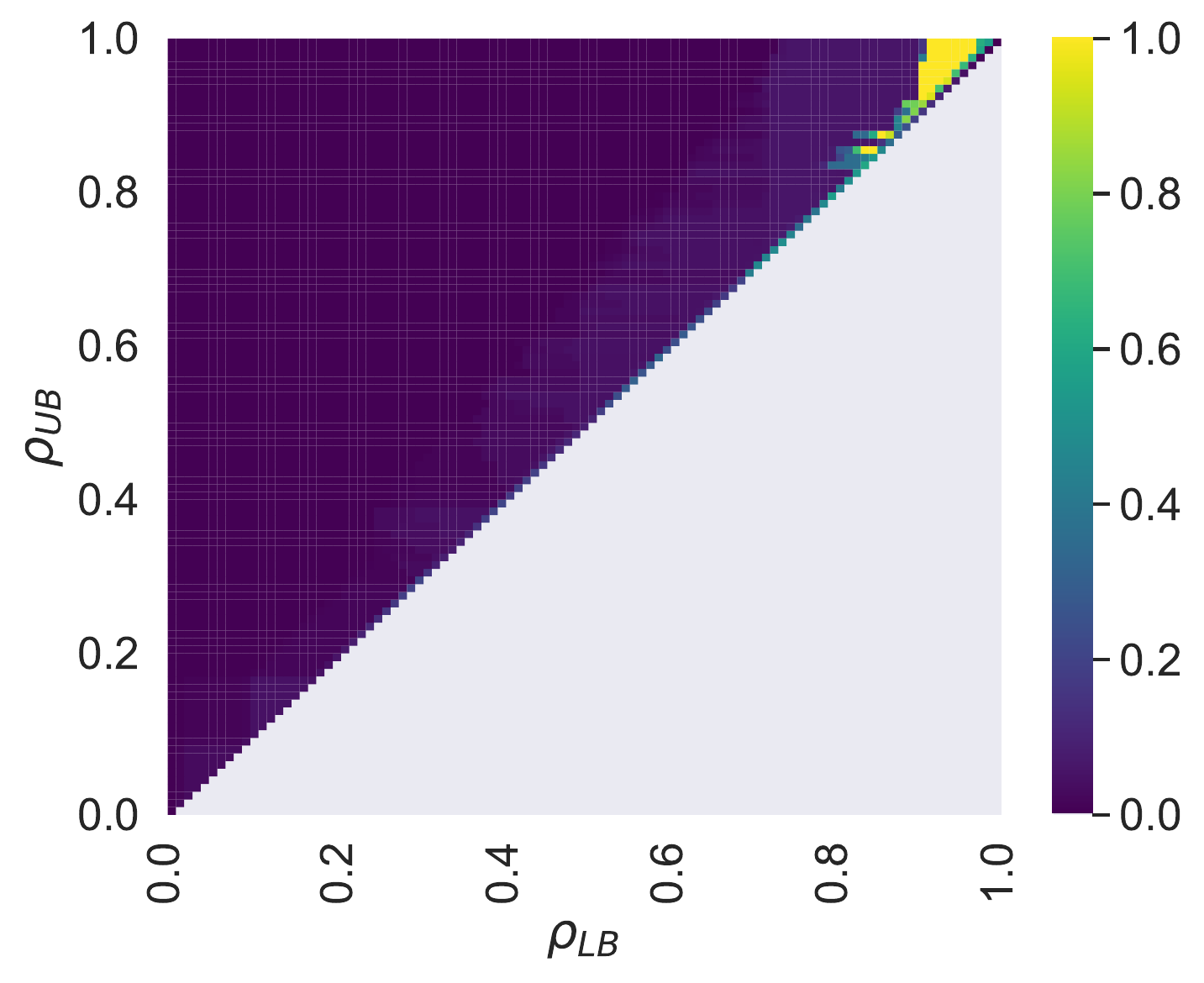} }
\subfloat[$\gamma=3$]{\includegraphics[width=\gammaMat\columnwidth]{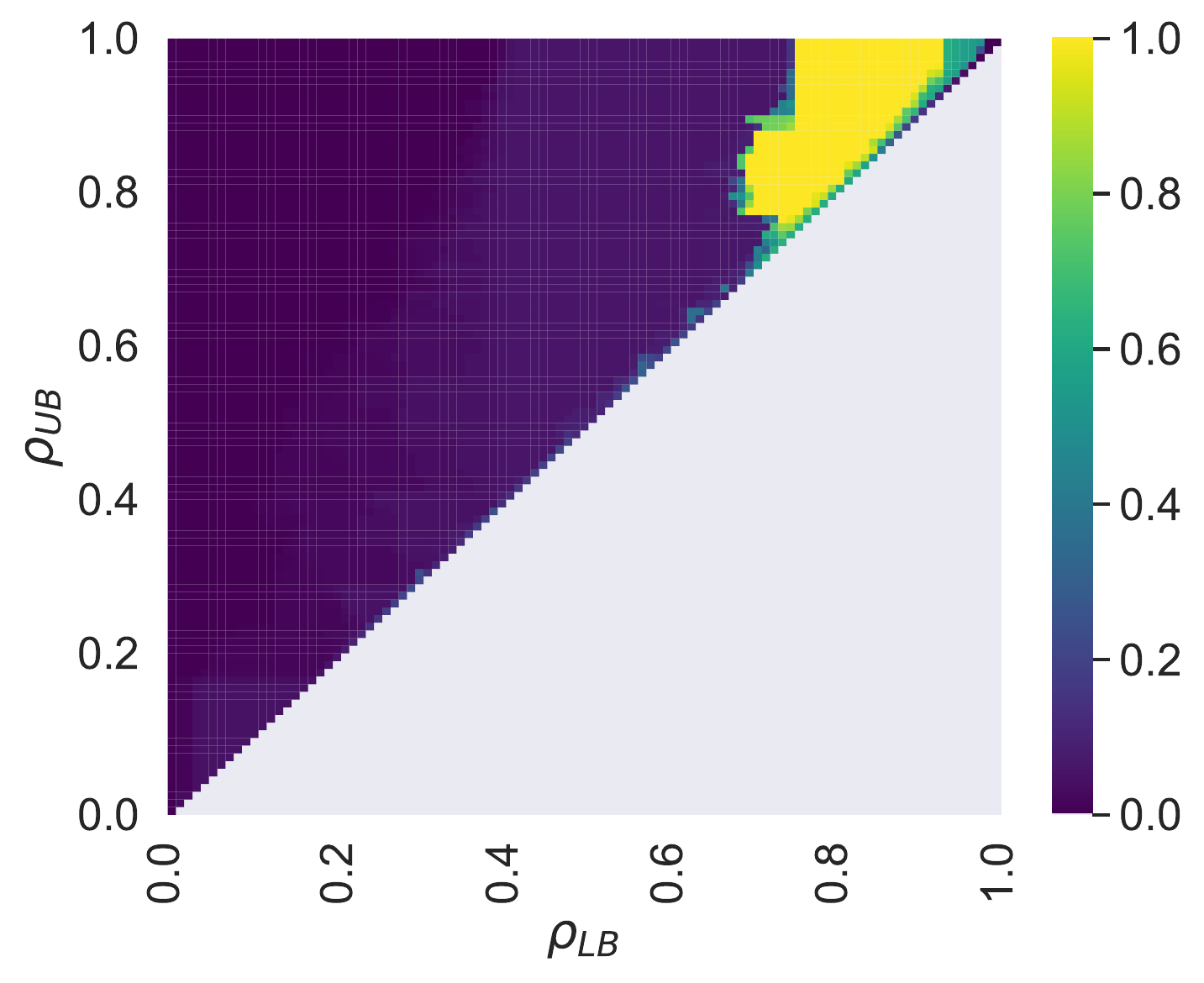} }
\subfloat[$\gamma=5$]{\includegraphics[width=\gammaMat\columnwidth]{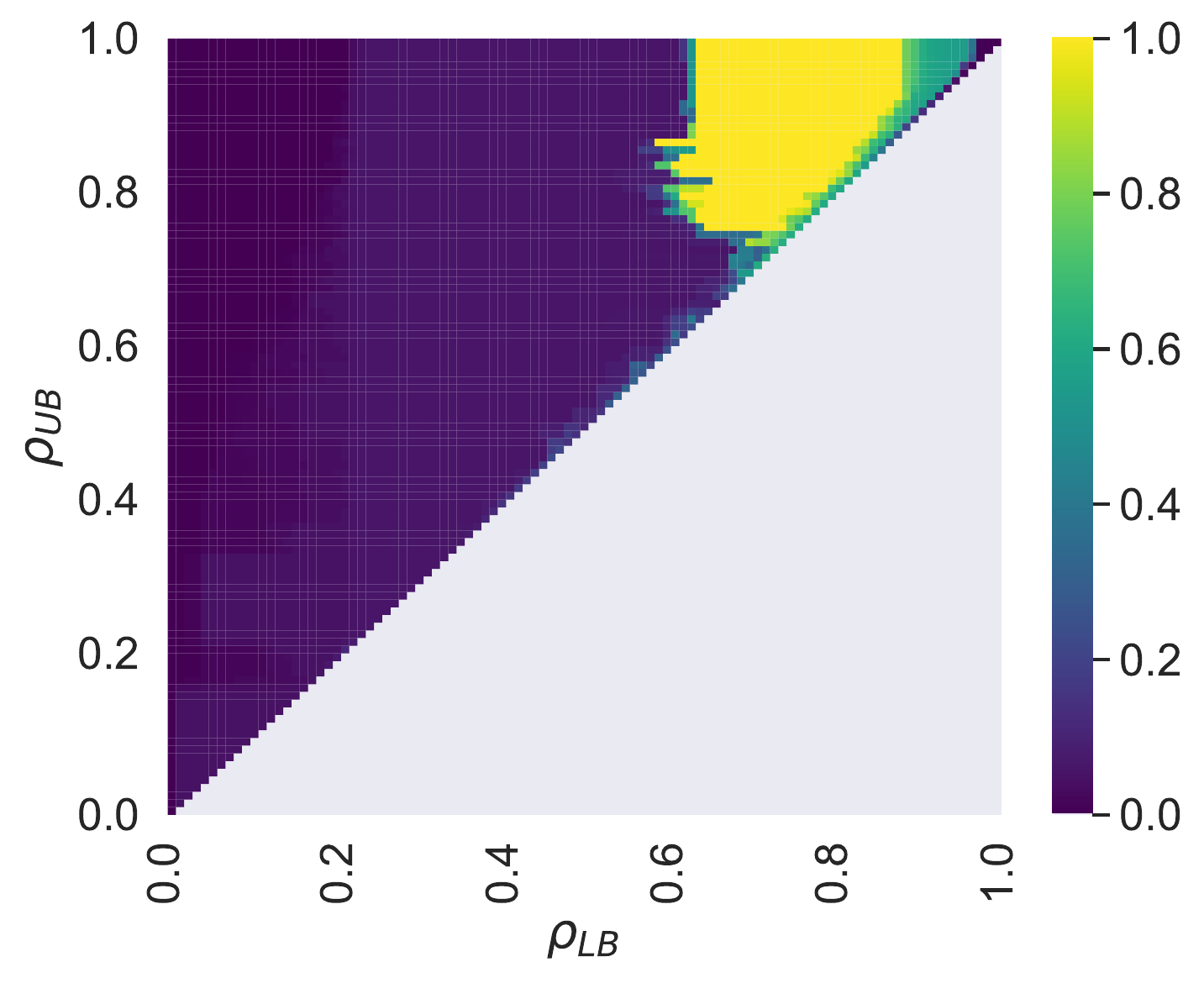} }
}
\centerline{
\subfloat[$\gamma=1$]{\includegraphics[width=\gammaMat\columnwidth]{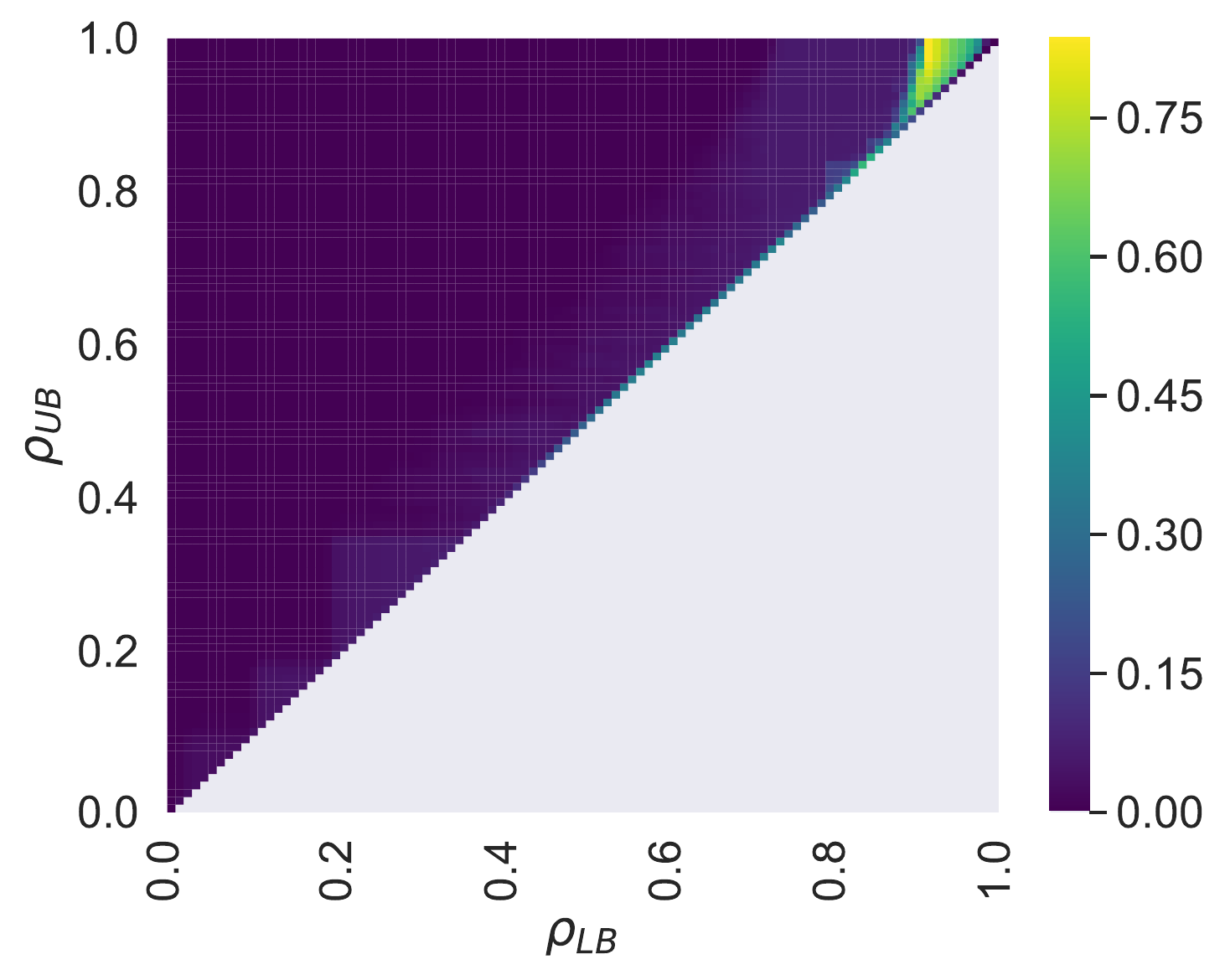} }
\subfloat[$\gamma=3$]{\includegraphics[width=\gammaMat\columnwidth]{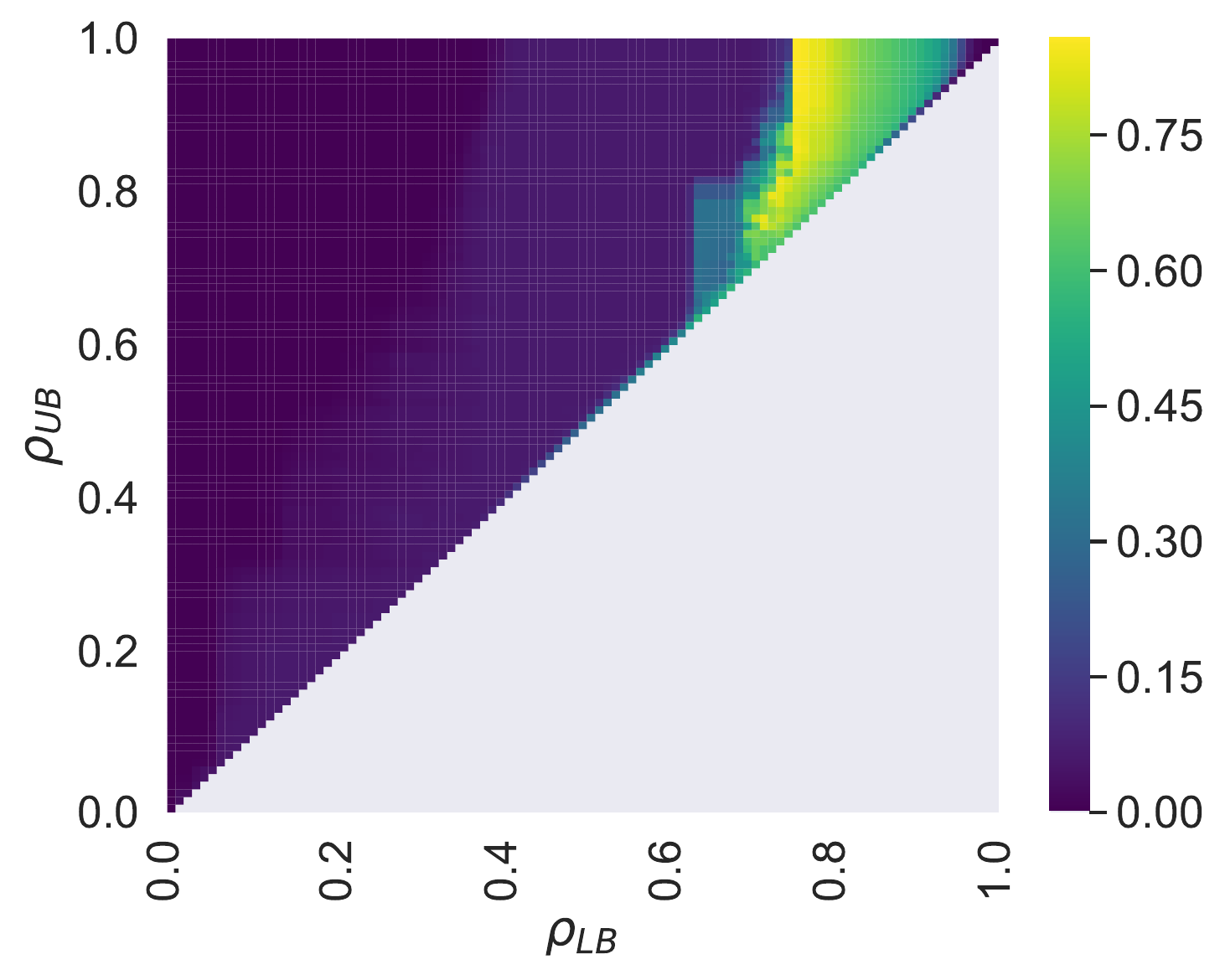} }
\subfloat[$\gamma=5$]{\includegraphics[width=\gammaMat\columnwidth]{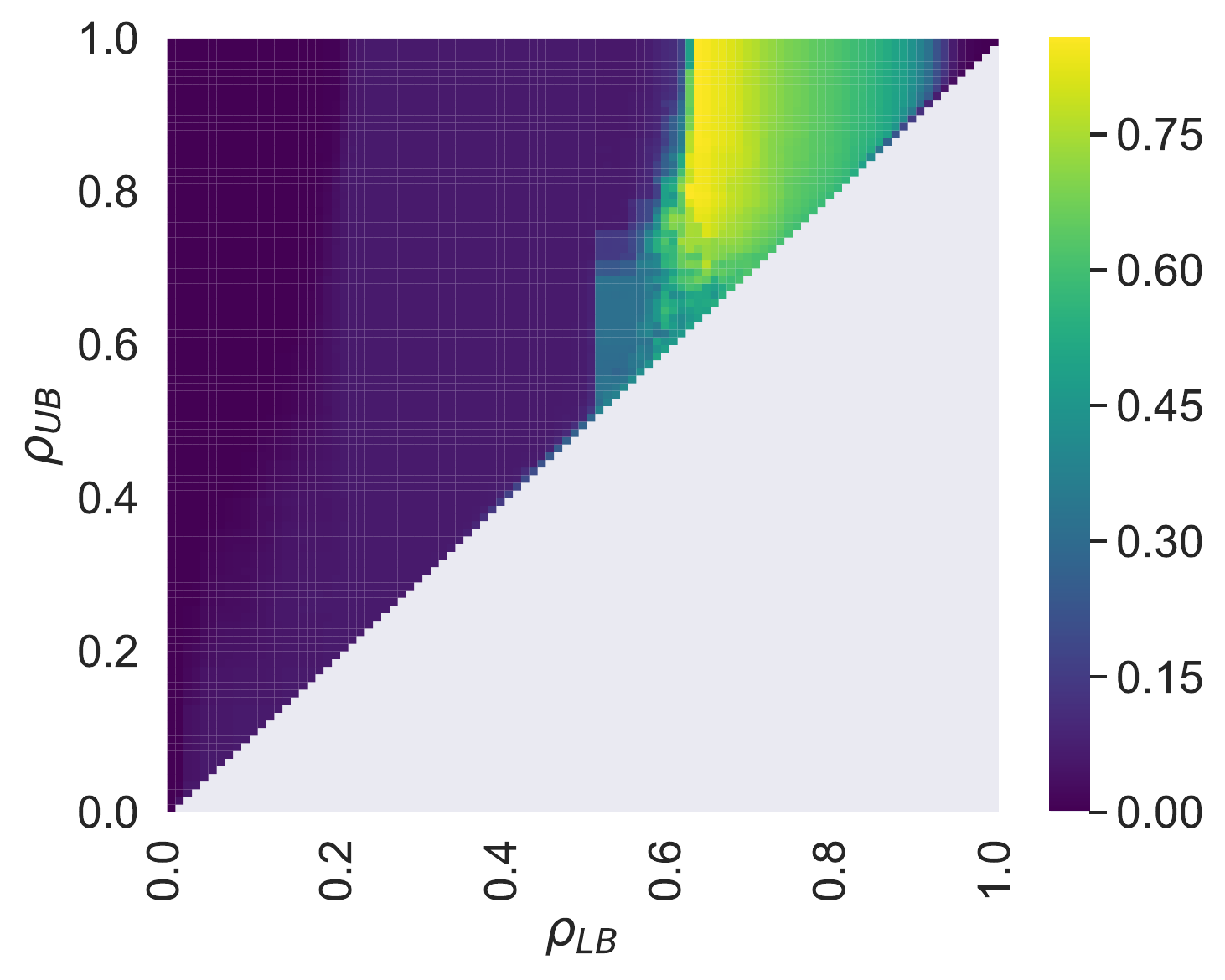} }
}
\centerline{
\subfloat[$\gamma=1$]{\includegraphics[width=\gammaMat\columnwidth]{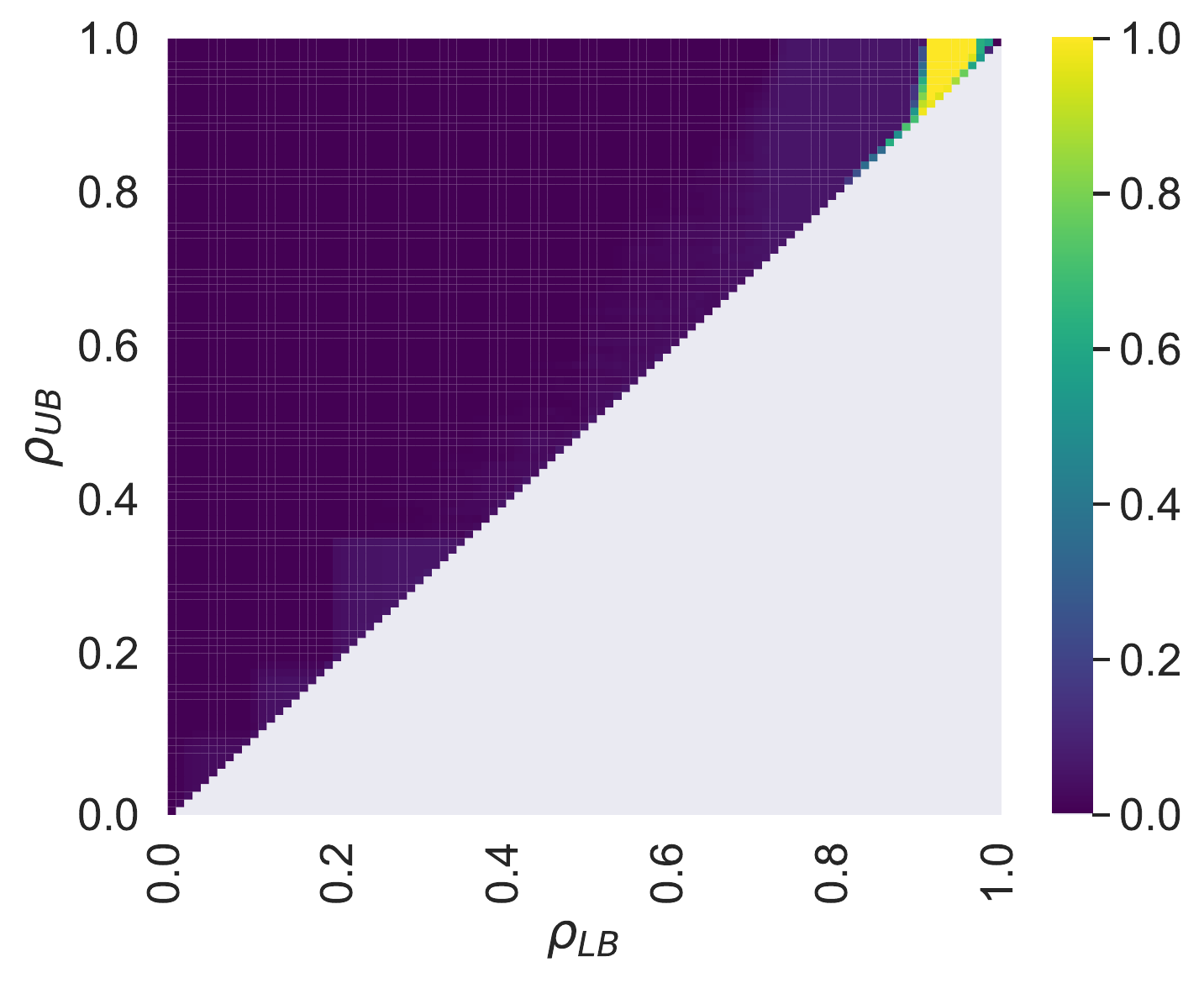} }
\subfloat[$\gamma=3$]{\includegraphics[width=\gammaMat\columnwidth]{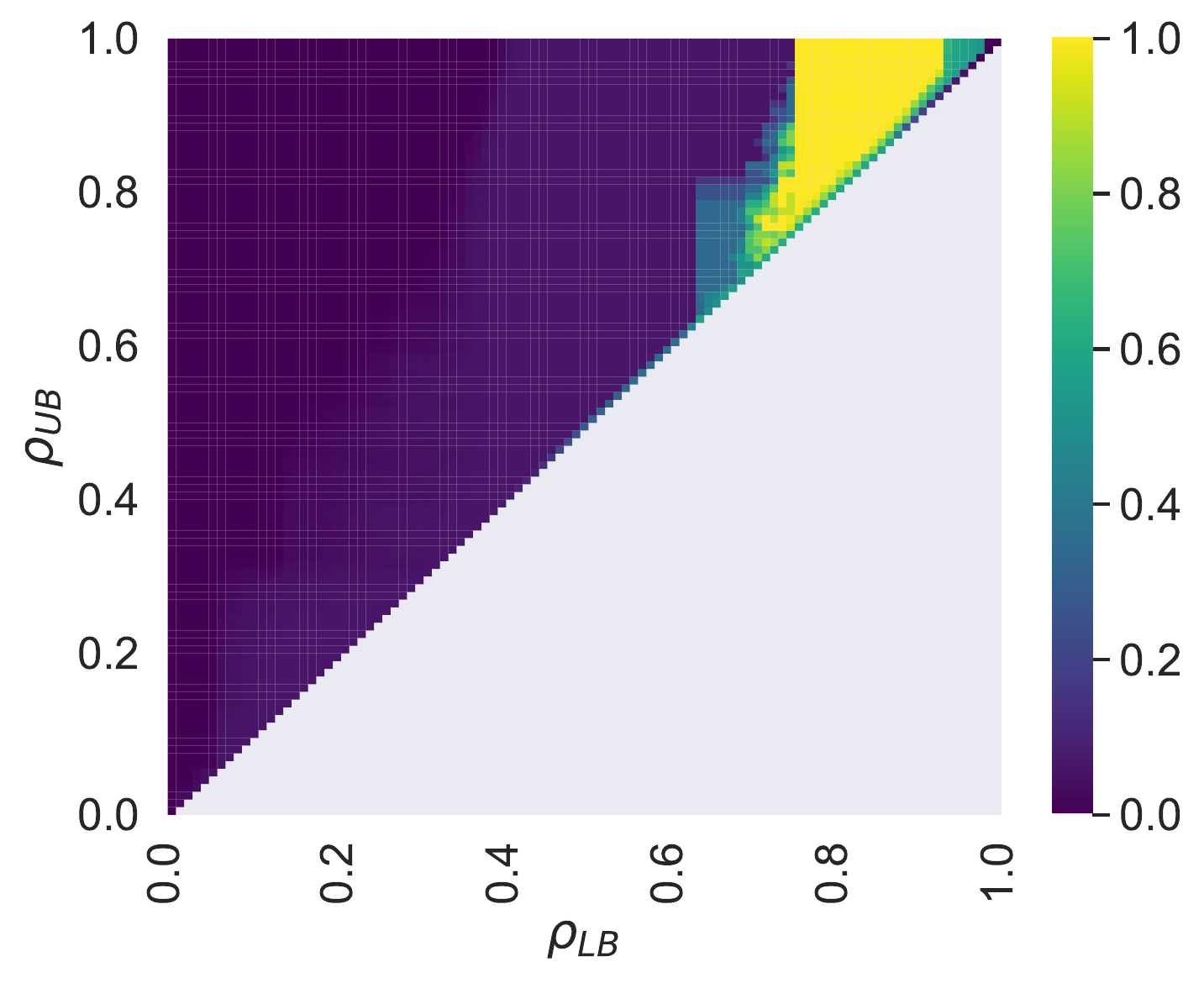} }
\subfloat[$\gamma=5$]{\includegraphics[width=\gammaMat\columnwidth]{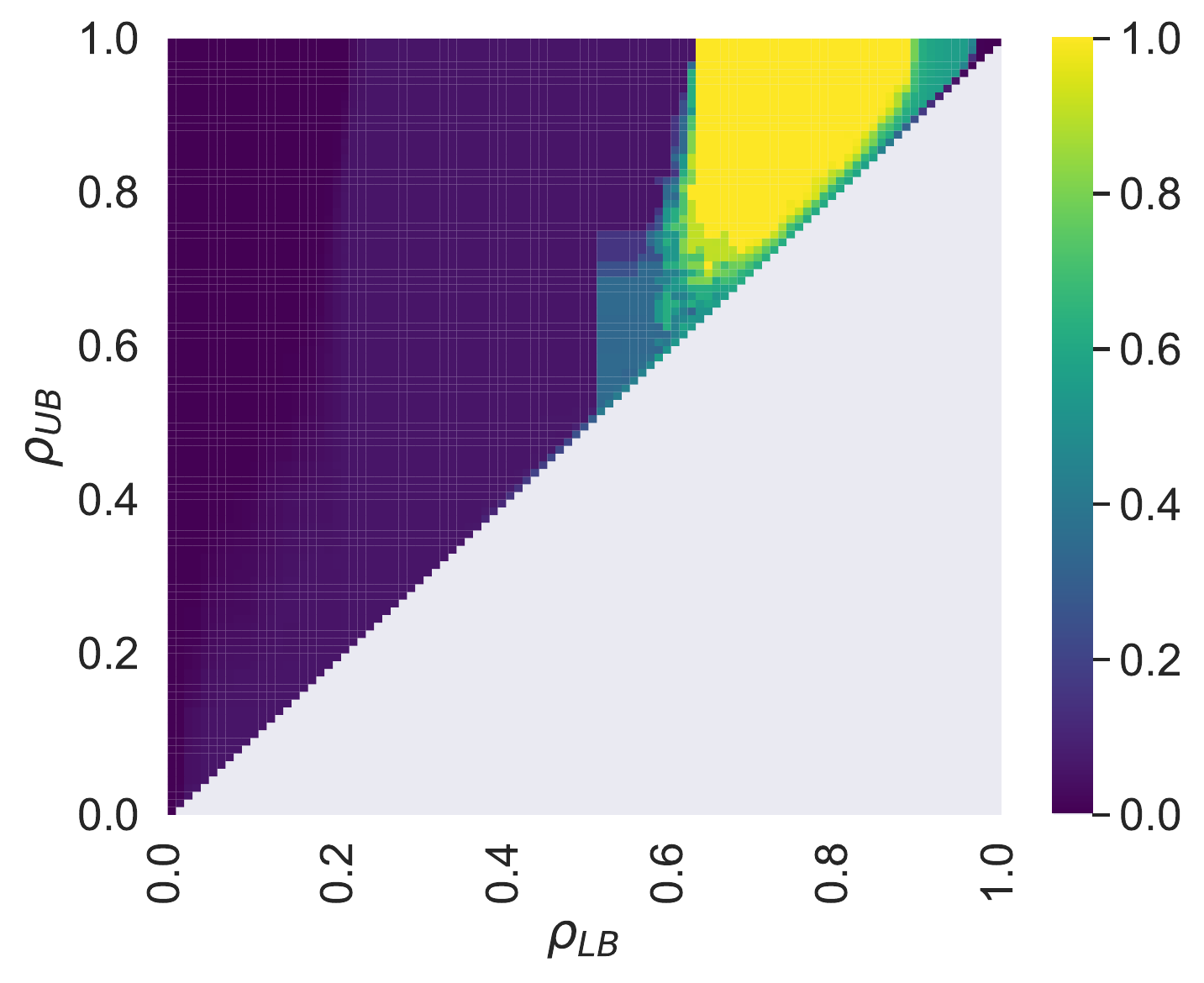} }
}
\caption{Heat maps corresponding to the average performance ($AR$) of (a)-(c) VAT + DDVFA, (d)-(f) DDVFA, and (g)-(i) DDVFA + Merge ART, for the \textit{Target} data set when varying parameter~$\gamma$. More yellow is better, implying a broader range of good parameter values. Sub-figures (a), (d), and (g) correspond to fuzzy ART building blocks, whereas the other portions of the figure correspond to contributions from this paper.}
\label{Fig:gamma_behavior_04}
\end{figure}

\section{Conclusion} \label{Sec:conclusion}

This paper presented DDVFA, a novel, modular, hierarchically self-consistent ART-based architecture for incremental, unsupervised learning. DDVFA features a number of innovations that differ from other ART-based systems. It relies on dual vigilance parameters to handle data quantization (local scale) and cluster similarity (global scale), features multi-prototype representations, and higher-order distributed activation and match functions. DDVFA consists of a global ART network whose nodes are local ART modules. The learning mechanism of the former is triggered by the feedback from the latter, thus enabling the system to capture arbitrary data distributions when using appropriate activation/match functions. DDVFA enables both one- and multi-category representations of clusters (i.e., one-to-one and one-to-many mappings of categories to clusters) according to the setting of the upper and lower vigilance parameter values. 

Like fuzzy ART and DVFA, DDVFA is sensitive to input order presentation. This work also introduces a compatible Merge ART module that yields improved performance in the online mode where samples arrive in a random order and pre-processing cannot be employed. Experiments were conducted with random and VAT ordered samples. As expected, the latter approach yields better average performance ranks, and thus it is recommended in applications where the offline learning mode is available. Otherwise, for online incremental learning, the usage of a Merge ART module cascaded with DDVFA is recommended, given that the latter showed superior performance and less sensitivity to input presentation order.The VAT + DDVFA and DDVFA + Merge ART systems were found to be statistically equivalent in this papers' experiments. Naturally, the type of distributed activation/match functions used for the similarity definition is data-dependent; the single-linkage-based ones typically yielded the best and second best average performance rank when cascading Merge ART and pre-processing with VAT, respectively. Conversely, weighted-based activation/match functions yielded the best average performance rank when solely using DDVFA. Naturally, as with other ART algorithms, the dual vigilance parameters must be carefully tuned.

The combination of DDVFA + Merge ART significantly outperformed fuzzy ART, DVFA, and topoART in most of the data sets with randomly presented samples, where a statistical difference was observed.  Conversely, when pre-processing with VAT, no statistical difference was observed, except for in standard fuzzy ART. The compactness (i.e., number of categories created) of the networks generated by the multi-prototype ART-based architectures were also compared, and again, no statistical difference was observed. Furthermore, the clustering performance of these best performing DDVFA systems were compared with single-linkage HAC, DBSCAN, k-means and affinity propagation. The results indicated that these DDVFA systems are statistically equivalent to the first three clustering algorithms mentioned, and they all perform statistically better than affinity propagation. This is noteworthy since DDVFA-based systems are based on incremental learning, whereas all the other non-ART-based algorithms used batch learning.

Finally, this work investigated the effect of the parameter~$\gamma$ in the behavior of DDVFA. The performance was robust toward this parameter, and with appropriate selection it can potentially increase the compactness (or equivalently, reduce the model complexity) of the DDVFA systems. This memory compression characteristic is consistent with findings from previous related work (distributed ART and ARTMAP systems), which combines power rules and distributed learning. Moreover, it was observed that~$\gamma$ can extend the subspace of dual vigilance parameter combinations that yield effective performance.

\appendix
\section{Derivation of the match function in DDVFA} \label{AppendixA}
This section contains the derivation of Eq.~(\ref{Eq:M3}). Let \mbox{$M_\gamma=M^{ART^{(1)}_i}_j$} be the activation function of category $j$ of $ART^{(1)}_i$ using $\gamma$ and $M_{\gamma^{*}}$ the activation function of the same category using $\gamma^{*}$. Then, the normalized version of $M_\gamma$ with respect to $M_{\gamma^{*}}$ ($M_{\gamma}^n$) is defined as

\begin{equation}
\begin{split}
M_{\gamma}^n  =  \left( \max(M_{\gamma^{*}}) - \min(M_{\gamma^{*}}) \right) \left( \frac{M_\gamma - \min(M_\gamma)}{\max(M_\gamma) - \min(M_\gamma)} \right) + \min(M_{\gamma^{*}}).  
\end{split}
\label{Eq:A1a}
\end{equation}

The values of $\max(M_{\gamma^{*}})$ and $\max(M_{\gamma})$ are easily obtainable, since any point inside the hyperrectangular category representation would have this value, particularly the weight \mbox{$\bm{w} = \bm{w}_j^{ART^{(1)}_i}$} of category $j$ itself. Furthermore, when using complement coding, $|\bm{x}|=d$ is a constant. The values $\min(M_{\gamma^{*}})$ and $\min(M_{\gamma})$ must be located at some corner of the d-dimensional unit hyperbox data space $[0,1]^{d}$. These values can also be easily calculated for data sets with small dimensionalities. However, as the dimension increases, searching $2^{d}$ points quickly becomes impractical. Therefore, since a match function $M$ satisfies \mbox{$0\leq M \leq 1$} by definition, a design decision was made to set \mbox{$\min(M_{\gamma^{*}}) = \min(M_{\gamma}) = 0$} in the normalization procedure. Hence,

\begin{equation}
\begin{split}
M_{\gamma}^n  &=  \max(M_{\gamma^{*}})) \left( \frac{M_\gamma}{\max(M_\gamma)} \right) =  \left( \frac{|\bm{w} \wedge \bm{w}|}{|\bm{x}|}\right)^{\gamma^{*}} \left( \frac{\frac{|\bm{x} \wedge \bm{w}|}{|\bm{x}|}}{\frac{|\bm{w} \wedge \bm{w}|}{|\bm{x}|}} \right)^\gamma \\
&= \left( \frac{|\bm{w}|}{|\bm{x}|} \right)^{\gamma^{*}} \left( \frac{|\bm{x} \wedge \bm{w}|}{c + |\bm{w}|}\right)^\gamma, 
\end{split}
\label{Eq:A1b}
\end{equation}

\noindent where the constant $c$ is inserted to safeguard against divisions by zero (since \mbox{$0 \leq \rho d \leq |\bm{w}| \leq d$}). This parameter implies that $\bm{w}=\bm{x}$ no longer yields a match function value equal to $1$. By making $c$ equal to the choice parameter $\alpha$, then Eq.~(\ref{Eq:A1b}) becomes

\begin{equation}
M_{\gamma}^n = \left(\frac{|\bm{w}|}{|\bm{x}|}\right)^{\gamma^{*}}T_\gamma,
\label{Eq:A1c}
\end{equation}

\noindent where $T_\gamma=T^{ART^{(1)}_i}_j$ is the activation function of category~$j$ of $ART^{(1)}_i$ using $\gamma$ (Eq.~(\ref{Eq:T2})). Naturally, if $\gamma^{*}=0$ then $M_{\gamma}^n = T_\gamma$, and for $\alpha \ll |\bm{w}|$, if $\gamma = \gamma^{*}$ then $M_{\gamma}^n \approx M_{\gamma^{*}}$ (Eq.~(\ref{Eq:M2})). 

\section*{Acknowledgment}  \label{Sec:thanks}

This research was sponsored by the Missouri University of Science and Technology Mary K. Finley Endowment and Intelligent Systems Center; the Coordena\c{c}\~{a}o de Aperfei\c{c}oamento de Pessoal de N\'{i}vel Superior - Brazil (CAPES) - Finance code BEX 13494/13-9; and the Army Research Laboratory (ARL), and it was accomplished under Cooperative Agreement Number W911NF-18-2-0260. The views and conclusions contained in this document are those of the authors and should not be interpreted as representing the official policies, either expressed or implied, of the Army Research Laboratory or the U.S. Government. The U.S. Government is authorized to reproduce and distribute reprints for Government purposes notwithstanding any copyright notation herein.

\biboptions{numbers,sort&compress}
\bibliography{bib/references}

\end{document}